\documentclass[11pt,a4paper]{report}

\usepackage[utf8]{inputenc}

\usepackage{feupteses} 

\graphicspath{{figures/}}

\usepackage{amsmath}
\usepackage{amssymb}
\usepackage{bm}  
\usepackage{algorithm}
\usepackage{algorithmic}
\usepackage{float}
\usepackage{booktabs}
\usepackage{listings}
\usepackage{makecell}
\usepackage{enumitem}
\usepackage{xurl}
\usepackage{subcaption}
\usepackage{graphicx}
\usepackage{rotating} 
\usepackage{longtable}

\begin{document}

\title{Video-based Music Generation}
\author{Serkan Sulun}

\degree{Doctoral Program in Electrical and Computer Engineering}

\thesisdate{May 23, 2025}

\copyrightnotice{Serkan Sulun, 2025}

\supervisor{Supervisor}{Dr. Paula Viana}
\supervisor{Supervisor}{Dr. Matthew E. P. Davies}


\committeetext{Approved in oral examination by the committee:}
\committeemember{President}{Dr.\ João Paulo Trigueiros da Silva Cunha\\ \textit{Faculty of Engineering, University of Porto}}
\committeemember{Member}{Dr.\ Rui Pedro Pinto de Carvalho e Paiva\\ \textit{Faculty of Science and Technology, University of Coimbra}}
\committeemember{Member}{Dr.\ Tom Collins\\ \textit{Frost School of Music, University of Miami}}
\committeemember{Member}{Dr.\ Sofia Carmen Faria Maia Cavaco\\ \textit{Faculty of Science and Technology, NOVA University Lisbon}}
\committeemember{Member}{Dr.\ Maria Teresa Magalhães da Silva Pinto de Andrade\\ \textit{Faculty of Engineering, University of Porto}}
\committeemember{Member (Supervisor)}{Dr.\ Paula Maria Marques de Moura Gomes Viana\\ \textit{ISEP, Polytechnic of Porto}}

\signature

\logo{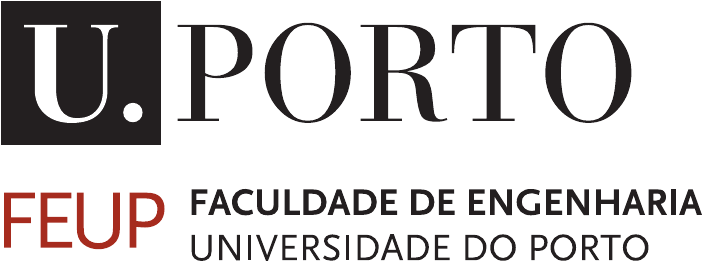}



\begin{Prolog}

\chapter*{Abstract}
As the volume of video content on the internet grows rapidly, finding a suitable soundtrack remains a significant challenge. This thesis presents EMSYNC (EMotion and SYNChronization), a fast, free, and automatic solution that generates music tailored to the input video, enabling content creators to enhance their productions without composing or licensing music, streamlining creativity and production. Our model creates music that is emotionally and rhythmically synchronized with the video, offering an adaptive and expressive solution for automatic soundtrack generation.

A core component of EMSYNC is a novel video emotion classifier. To achieve accurate and efficient video classification, we intelligently fuse pretrained models. We additionally address the data-centric challenges in video classification through cinematic trailer genre classification experiments using a large-scale dataset. By leveraging pretrained deep neural networks for feature extraction and keeping them frozen while training only fusion layers, we reduce computational complexity while improving accuracy. We show the generalization abilities of our method by obtaining state-of-the-art results on Ekman-6 and MovieNet, the largest video datasets for emotion and cinematic genre classification, respectively. 

Another key contribution is a large-scale, emotion-labeled MIDI dataset for affective music generation. Using annotations from online resources, we build the largest MIDI dataset with valence-arousal labels. We additionally analyze the emotional content of song lyrics within the MIDI files. We then present an emotion-based MIDI generator, the first to condition on continuous emotional values rather than discrete categories, enabling nuanced music generation aligned with complex emotional content.

To enhance temporal synchronization, we introduce a novel temporal boundary conditioning method, called "boundary offset encodings," aligning musical chords with scene changes. Integrated into EMSYNC, this method ensures music naturally follows the video's pacing and rhythm, improving the overall user experience.

We also explore audio synthesis, focusing on audio bandwidth enhancement due to the scarcity of paired MIDI-audio data. We present a proof-of-concept to highlight and address the challenges in audio synthesis, emphasizing generalization. For the first time, we identify the problem of "\mbox{filter} overfitting," where models trained on specific low-pass filters fail to generalize to real-world scenarios. To address this, we propose a data augmentation strategy that outperforms standard regularization methods, marking the first step toward developing robust audio enhancement models for real-world use.

Combining video emotion classification, emotion-based music generation, and temporal boundary conditioning, EMSYNC emerges as a fully automatic video-based music generator. User studies show that it consistently outperforms existing methods in terms of music richness, emotional alignment, temporal synchronization, and overall preference. As a result, EMSYNC sets a new state-of-the-art in video-based music generation, creating music that is both emotionally and rhythmically aligned with the video.

\chapter*{Resumo}

A disponibilização de conteúdos vídeo na internet tem crescido de forma exponencial colocando um acervo significativo, e de valor elevado, disponível à população em geral. No entanto, para os produtores destes conteúdos, encontrar uma banda sonora adequada continua a ser um desafio considerável. Esta tese apresenta o EMSYNC (EMotion and SYNChronization), uma solução rápida, gratuita e automática que gera música adaptada ao vídeo de entrada, permitindo aos criadores de conteúdo melhorar as suas produções sem necessidade de compor ou licenciar música, facilitando assim a criatividade e a produção. O nosso modelo cria música emocional e ritmicamente sincronizada com o vídeo, oferecendo uma solução adaptativa e expressiva para geração automática de bandas sonoras.
Um dos componentes centrais do EMSYNC é um novo classificador de emoções em vídeo. Para alcançar uma classificação precisa e eficiente, propomos uma fusão inteligente de modelos previamente treinados. De forma a ultrapassar os desafios associados à escassez de dados adequados a esta tarefa, e atendendo aos objetivos similares, tiramos partido das grandes coleções de dados associados a tarefas de classificação de géneros de trailers cinematográficos. Ao explorar redes neuronais profundas pré-treinadas para extração de características e mantendo-as congeladas enquanto treinamos apenas as camadas de fusão, reduzimos a complexidade computacional e melhoramos a precisão. Demonstramos as capacidades de generalização do nosso método ao ultrapassar os atuais resultados estado da arte nos conjuntos de dados Ekman-6 e MovieNet, os maiores conjuntos de dados de vídeo para classificação emocional e de género cinematográfico, respetivamente.

Outra contribuição relevante desta tese é um conjunto de dados MIDI anotado com emoções. Combinando informações recolhidas de recursos online e a análise do conteúdo emocional das letras das músicas, construímos o maior conjunto de dados MIDI com anotações de valência-excitação, disponibilizando assim um recurso poderoso para treinar modelos para geração de música baseados em emoções.
Um modelo capaz de gerar MIDI baseado em emoções, o primeiro a ser condicionado por valores emocionais contínuos em vez de categorias discretas, é outra das contribuições fundamentais da tese. Esta abordagem permite a geração de música alinhada com conteúdos emocionais complexos. Para melhorar a sincronização temporal entre os conteúdos de áudio e de vídeo, introduzimos um novo método de condicionamento de fronteiras temporais, denominado "codificação de desfasamento de fronteira", que alinha acordes musicais com mudanças de cena no vídeo. Integrado no EMSYNC, este método garante que a música segue naturalmente o ritmo e o andamento do vídeo, melhorando a experiência do utilizador.

A tese explora ainda a síntese de áudio. Dada a escassez de associação de dados MIDI-áudio multi-instrumento, focamo-nos em aspetos relacionados com a melhoria da largura de banda. Apresentamos uma prova de conceito que permite identificar e responder aos principais desafios associados à síntese áudio, com especial foco na capacidade de generalização. Identificamos o problema de "overfitting de filtros", em que modelos treinados com filtros passa-baixo específicos não generalizam bem para cenários do mundo real. Para ultrapassar este problema, propomos uma estratégia de aumento de dados que supera os métodos de regularização padrão, representando o primeiro passo para o desenvolvimento de modelos robustos de melhoria de áudio para uso real.

Combinando classificação emocional de vídeo, geração musical baseada em emoções e condicionamento temporal, o EMSYNC afirma-se como um gerador musical automático baseado em vídeo. Estudos com utilizadores demonstram que supera consistentemente os métodos existentes em termos de riqueza musical, alinhamento emocional, sincronização temporal e preferência geral. Assim, o EMSYNC estabelece um novo estado da arte na geração musical baseada em vídeo, criando música que se alinha emocional e ritmicamente com o vídeo.
  \chapter*{Acknowledgements}

I would like to begin by expressing my sincere gratitude to my supervisors, Paula Viana and Matthew Davies, for their continual support and guidance. I am also forever grateful to my family—Ali, Lale, and Zeynep—for their unconditional love and support. They have provided me with a solid foundation, allowing me to build an independent life without the weight of childhood traumas on top of the usual pains of adulthood.

People often say that relationships during a PhD are difficult, but having an incredible girlfriend has made this period much easier for me. I can never thank my love, companion, and best friend Chloé enough. In this foreign country, she has given me a new home.

A special thanks goes to Bianca, not only for being an amazing friend but also for setting me up with teaching jobs and, most importantly, for tolerating my complaints. I also want to thank my good friends and fellow academics in Porto—Göksu, Nathalie, and Nabila—for their unwavering support, friendship, and valuable advice. I furthermore thank you all for welcoming me into the "Porto Crew."

I am incredibly lucky to have my great friend Doğan, with whom I've maintained a close friendship for the past 20 years, despite living in different countries. I am grateful to my colleagues and office mates—Inês, Luís, Eduardo, Américo, and Tiago—for their friendship and support, helping me navigate my time in Portugal.

Finally, I want to express my thanks to my fellow PhDs from the la Caixa Fellowship cohort—Gizem, Marta, Inês, Arturo, Javi, Chris, Jacob, Paula, Jaime, and Ludovica—for their friendship and organizing some unforgettable holidays.

I am also grateful to have secured two fellowships, namely the la Caixa Doctoral Fellowship (INPhINIT) (LCF/BQ/DI19/11730032) and the FCT (Fundacao para a Ciencia e a Tecnologia) Doctoral Fellowship (2022.09594.BD), that enabled me to pursue this doctoral research. Additional support was provided by funding from the NEXUS-2 Project (12407/BI-M-ED\_B2/2025).

\vspace{10mm}
\flushleft{Serkan Sulun}
  \cleardoublepage
  \pdfbookmark[0]{Table of Contents}{contents}
  \tableofcontents
  \cleardoublepage
  \pdfbookmark[0]{List of Figures}{figures}
  \listoffigures
  \cleardoublepage
  \pdfbookmark[0]{List of Tables}{tables}
  \listoftables
  \chapter*{List of Acronyms}
\chaptermark{LIST OF ACRONYMS}

\begin{flushleft}
	\renewcommand{\arraystretch}{1.2} 
	\begin{longtable}{l p{0.8\linewidth}}
		
		AAC & Affective Algorithmic Composition \\
		AEC & Acoustic Echo Cancellation\\
		API & Application Programming Interface\\
		ASR & Automatic Speech Recognition\\
		AST & Audio Spectrogram Transformer\\
		BERT & Bidirectional Encoder Representations from Transformers\\
		BEATs & Bidirectional Encoder representation from Audio Transformers \\
		BN & Batch Normalization\\
		BPM & Beats Per Minute\\
		CLIP & Contrastive Language-Image Pretraining\\
		CMT & Controllable Music Transformer\\
		CNN & Convolutional Neural Network\\
		CTRL & Conditional TRansformer Language\\
		DA & Data Augmentation\\
		DAW & Digital Audio Workstations\\
		DSD & Demixing Secrets Dataset\\
		DNN & Deep Neural Network\\
		DO & DropOut \\
		EMSYNC & EMotion and SYNChronization\\
		EMOPIA & EMOtion PIAno\\
		FAD & Fréchet Audio Distance\\
		FC & Fully-connected\\
		GPT & Generative Pre-Training\\
		GRU & Gated Recurrent Unit\\
		I3D & Two-Stream Inflated 3D ConvNets\\
		ICASSP & International Conference on Acoustics, Speech, and Signal Processing\\
		LLM & Large Language Model\\
		LMD & Lakh MIDI Dataset\\
		LPC & Linear Predictive Coding\\
		LPD & Lakh Pianoroll Dataset\\
		LSTM & Long Short-Term Memory\\
		MAC & Multiply-ACcumulate\\ 
		mAP & Mean Average Precision\\
		MAESTRO & MIDI and Audio Edited for Synchronous TRacks and Organization\\
		MIDI & Musical Instrument Digital Interface \\
		MLP & Multi Layer Perceptron\\
		MSD & Million Song Dataset\\
		MVED & Music Video Emotion Dataset\\
		NaN & Not A Number\\
		NLL & Negative Log-likelihood\\
		NLP & Natural Language Processing\\
		OCR & Optical Character Recognition\\
		P & Precision\\
		PERR & Pairwise Emotional Relationship Recognition\\
		PESQ & Perceptual Evaluation of Speech Quality\\
		R & Recall\\
		ResNet & Residual Network\\
		RNN & Recurrent Neural Network\\
		SNR & Signal-to-Noise Ratio\\
		SR & Super-Resolution\\
		TRN & Temporal Relation Network\\
		TSN & Temporal Segment Network\\
		VEMOCLAP & Video EMOtion CLassifier using Pretrained features\\
		VGMIDI & Video Game Musical Instrument Digital Interface\\
		VGG & Visual Geometry Group\\
		ViSQOL & Virtual Speech Quality Objective Listener\\
		ViT & Vision Transformer\\
		ViViT & Video Vision Transformer\\
		YOLO & You Only Look Once\\
		
	\end{longtable}
\end{flushleft}
\end{Prolog}

\StartBody

\chapter{Introduction} \label{chap:intro}

The exponential growth of user-generated multimedia content is driven by the increasing availability of affordable, high-quality recording equipment and video editing software~\parencite{multimedia}. A major challenge in content creation is selecting suitable soundtracks to enhance viewer engagement~\parencite{soundtrack}. However, the unauthorized use of commercially published music infringes copyright, restricting monetization on platforms like YouTube. Alternatives such as purchasing music, hiring composers, or searching for royalty-free tracks can be costly, time-consuming, or fail to ensure proper synchronization with the video. This thesis proposes a fast, cost-free, and royalty-free solution: a model that takes any user-provided video as input and automatically composes a soundtrack that matches the emotional content of the video.

Music can be represented in two primary formats: audio and symbolic. Audio, the standard musical format perceived by general listeners, represents sound as a waveform. Symbolic music, on the other hand, is primarily used by musicians and is exemplified by sheet music, which encodes pitches, durations, and includes performance guides such as crescendo or pianissimo. In the digital domain, symbolic music is represented as a discrete sequence, similar to text. The most widely used digital symbolic music format is MIDI (Musical Instrument Digital Interface). A MIDI file consists of a sequence of events, each encoding information about a note, such as its pitch, instrument (channel), and intensity (velocity/loudness). While a musical composition can be stored as a MIDI file, similar to sheet music, MIDI files must be synthesized into audio to be heard. Software like Fluidsynth\footnote{https://fluidsynth.com} assigns sounds to MIDI events using standard libraries, converting them into an audio waveform.

Existing approaches that generate music as audio waveforms lack editability~\parencite{musicgen1,musicgen3,musicgen2,musicgen4}. These methods compress all stages of music production---composition, performance, editing, mixing, and mastering---into a single process, restricting creative control for professionals. In contrast, generating music in a symbolic format such as MIDI provides greater flexibility. Since MIDI explicitly encodes musical elements like notes, timing, and dynamics, professionals can refine compositions, synthesize audio using digital audio workstations (DAWs), or record and adjust performances with ease.

Our work focuses on generating music in MIDI format from arbitrary videos using deep neural networks (DNNs). Since MIDI is our exclusive format, we use "MIDI" and "music" interchangeably. A key challenge in training video-to-MIDI DNNs is the lack of large-scale paired video-MIDI datasets. Existing datasets are either domain-specific, such as pianist hand videos~\parencite{sighttosound}, or contain only a limited number of samples (around 1k)~\parencite{di,zhuo}. While large-scale MIDI datasets like the Lakh MIDI Dataset exist, they are not paired with videos, making end-to-end learning infeasible~\parencite{lmd}. Our approach addresses this by identifying intermediate representations that link video and music. This leads to a two-stage framework: first, analysis models extract representations from the input video; then, these representations condition the music generation model.

\begin{figure}[t]
	\centering
	\includegraphics[width=0.5\linewidth]{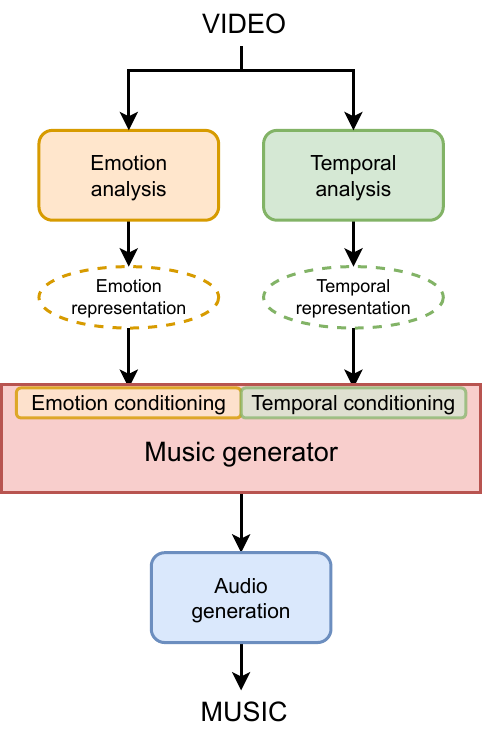}
	\caption{Our video-based music generation methodology.}
	\label{fig:flowchart}
\end{figure}

To realize an automatic soundtrack generator that appeals to human listeners, we must emulate the composition process of human soundtrack composers. Therefore, the intermediate representations should align with how composers create soundtracks. The question of how to compose music that fits a video is longstanding~\parencite{soundtrack}. However, it is an artistic rather than an engineering problem—there is no objective or definitive answer. Nevertheless, certain heuristics and general knowledge suggest that an effective soundtrack should: (1) align with the video's underlying emotions to enhance them \parencite{soundtrack_emotion2,soundtrack_emotion1}, and (2) match its temporal dynamics, ensuring synchronization to improve viewer engagement \parencite{soundtrack_scenes2,soundtrack_scenes1}. Based on this perspective, our method follows a two-branch approach, generating music using both emotional cues and temporal dynamics. Figure~\ref{fig:flowchart} illustrates our two-stage, two-branch methodology.

The user-provided input video is first analyzed, followed by symbolic music generation based on the analysis output, and finally, the audio that corresponds to the symbolic music is generated. However, in both our work and the structure of this thesis, we adopt a more strategic ordering of these three processes.

In both our research and the narrative of this thesis, we begin with music generation, followed by video analysis. This order ensures that the generated music remains perceptually coherent and musically meaningful at each stage. Designing the video analysis module first would be risky: if the music generator cannot effectively use the extracted representations, we would need to revisit and redesign the video analysis module. Furthermore, once a conditional music generator is in place, even if video feature extraction proves unsuccessful, we could fall back on a contingency plan by getting these features directly from the user.

After designing the music generation and video analysis modules, we focus on integrating them. For MIDI-to-audio generation, we use a standard approach with the dedicated Fluidsynth software and standard sound libraries. While experimental methods based on deep neural networks exist, they are limited to piano tracks and do not generalize to multi-instrument music \parencite{maestro,midi2audio2}. 

Nevertheless, we explore the use of DNNs for audio generation in a different context towards the end of this thesis. This exploration is separate from and not integrated into our main model. We focus specifically on the challenges associated with audio generation. As discussed later in Section~\ref{sec:intro_audio}, the lack of datasets containing paired audio and multi-instrumental MIDI forces us to rely on audio-only datasets, particularly for generating high-quality audio from low-quality sources. In this context, we address the problem of audio bandwidth extension, where we create input-target pairs by synthetically band-limiting high-quality audio.

In summary, here is the high-level order of topics we explore throughout this thesis:

\begin{enumerate}
	\item Symbolic music generation
	\item Video analysis
	\item Combining the video analysis and music generation modules (video-based music generation)
	\item Audio generation and bandwidth extension
\end{enumerate}

We begin our discussion with symbolic music generation.

\section{Symbolic music generation}

Symbolic music, represented as a sequence of notes, is widely used in machine learning models due to its compact and structured format. Large-scale raw MIDI datasets enable unsupervised training of DNNs for automatic symbolic music generation~\parencite{lmd}. The state-of-the-art architecture for sequence generation is the transformer model, which serves as the backbone of all large language models (LLMs) \parencite{transformer}. Similar to language modeling, transformers learn to predict the next token, i.e., the next note, and generate output autoregressively, one token at a time during inference. However, human composers do not simply construct music note by note in a sequential manner; their creative process involves high-level concepts such as motifs, themes, emotions, and temporal structures like rhythm~\parencite{composing_w_emotion}. Inspired by this, we present a \textit{conditional music generator} with two distinct mechanisms to independently incorporate temporal and emotional conditioning.

\subsection{Temporally-conditioned music generation}
\label{sec:intro_temporal}
We first address the challenge of generating music based on temporal dynamics. However, it is first necessary to define what "temporal dynamics" entail. Previous works on video-based music generation have used motion speed and motion saliency as temporal features~\parencite{di}. This approach produces a dense representation of temporal dynamics, resulting in continuous conditioning on musical note density. However, our preliminary listening tests of \textcite{di} revealed two key limitations. First, the continuously varying note density disrupts rhythmic consistency. Second, because the mapping is dense, synchronization between music and video is difficult to perceive unless there are strong contrasts in motion speed and note density across consecutive sections.

We pose the following research question: 

\begin{quote}
	\textit{Which type of temporal feature is the most useful for synchronizing video and music while maintaining rhythmic consistency? }
\end{quote}

Our hypotheses are as follows:

\begin{quote}
	\textit{Dense temporal representations, such as speed, can result in continuous modifications and disruptions of rhythm.}
	
	\textit{Matching sparse temporal features, such as temporal boundaries, would help preserve a steady rhythm while enhancing the perceived synchronization between video and music.}
\end{quote}

Temporal conditioning in MIDI-generating models presents unique challenges. While deep transformer models can process time-based data such as videos, their sequence dimension correlates linearly with time due to a fixed frame rate~\parencite{swin,vivit}. In contrast, MIDI is typically represented using an event-based encoding, where sequence and time dimensions are not linearly correlated~\parencite{event_encoding}.

In event-based MIDI encoding, two primary token types exist: "note" and "time shift." Note tokens represent pitch, while time shift tokens advance the time axis, capturing both note durations and silences. Each time shift token specifies a time increment, creating a one-dimensional sequence where position in the sequence does not directly correspond to position in time. The key advantage of this encoding is the absence of a fixed time grid, allowing for subtle, expressive timing variations that reflect human musicianship.

We further ask:

\begin{quote}
	\textit{How can we design methods of temporal synchronization that are independent of token positions, while preserving event-based MIDI encoding?}
\end{quote}

We hypothesize: 

\begin{quote}
	\textit{Auxiliary inputs in the form of temporal conditions can facilitate synchronization. }
\end{quote}

A particular challenge arises when the model is required to synchronize at a specific timestamp, as forced synchronization may disrupt the consistent rhythm of the generated music. To address this, we further investigate learned conditioning methods with minimal intervention, allowing the model to select the exact synchronization point that aligns harmoniously with the music it generates. Additionally, we hypothesize that providing the model with information about future boundaries can guide it in generating musical notes that precede the boundary in a coherent manner.

\subsection{Emotion-based music generation}

Music has long been a powerful medium for emotional expression and communication~\parencite{music_and_emotion2}. Its ability to evoke emotions has been studied across various disciplines, including psychology~\parencite{psychology}, musicology~\parencite{music_and_emotion}, and neuroscience~\parencite{neuroscience}. With the rise of deep learning, there is growing interest in machine learning algorithms capable of automatically analyzing and generating music to elicit specific emotional responses in listeners~\parencite{music_survey}.

We address the problem of generating music from emotions, aiming to achieve the highest possible performance. Our approach follows the deep learning trend, where larger models trained with more data perform better~\parencite{scaling}. However, current MIDI datasets that contain emotion labels are very small, containing fewer than 400 samples~\parencite{vgmidi,emo_lstm,emopia}. This limitation poses a significant challenge to training large and powerful DNNs for emotion-based MIDI generation.

To address the issue of data scarcity, we explore the following research questions:

\begin{quote}
\textit{How can we acquire large-scale datasets consisting of MIDI-emotion label pairs?}

\textit{Is it possible to label existing MIDI datasets with emotion labels without human input?}

\textit{Do other modalities, such as audio or text (e.g., song lyrics), contain information related to emotions?}

\textit{Can we leverage existing music databases and platforms to identify emotion labels and assign them to corresponding MIDI samples?}
\end{quote}

We hypothesize that: 
\begin{quote}
\textit{Emotions that are automatically extracted from other related modalities, particularly from the audio version and built-in song lyrics will align with the corresponding MIDI samples. }
\end{quote}
Due to the inherent differences between these modalities, i.e., audio vs. MIDI and text vs. MIDI, these labels will serve as "weak" labels. However, the large volume of available samples is expected to compensate for the weakness of the match between MIDI and emotion labels. Since most musical content available on the internet is in audio format, we aim to explore music platforms to identify metadata related to emotions. We additionally exploit language models to analyze the emotions of the lyrics within the MIDI files.

We now continue with exploring video analysis.

\section{Video analysis}
\label{sec:intro_video}
In video analysis, representations are typically categorized as high-level or low-level based on their abstraction and complexity. Low-level video features are extracted directly from video data without interpreting its content. Examples include pixel-based features such as RGB values and edge information~\parencite{canny}, motion features such as optical flow~\parencite{flow}, texture features like local binary patterns~\parencite{texture}, and transition features such as dissolves and scene cuts~\parencite{scenechange}.

High-level video features provide a more abstract interpretation of video content, often captured using machine learning methods. Examples include actions \parencite{action}, objects \parencite{object}, emotions \parencite{emotion_intro}, and genres \parencite{wehrmann}.

In this work, we utilize both low-level scene cuts and high-level semantic information. Since scene cut detection is a well-established problem, we directly apply existing methods using the FFMpeg software\footnote{\url{https://ffmpeg.org}}. Our research primarily focuses on semantic video classification, particularly emotion classification, as this is crucial for our emotion-based music generator. Since we aim to develop a music generator that works with any type of video, we specifically focus on the emotion classification of \textit{arbitrary} videos, rather than domain-specific videos, such as those containing dialogues \parencite{social_video_emotion} or human faces \parencite{caer}.

In semantic video classification, as in most video classification tasks, DNNs represent the state of the art. These models primarily use raw data, such as pixel values and audio waveforms, as inputs. Unlike older methods that rely on hand-crafted features, DNNs are designed to implicitly extract high-level features within their layers. However, these models often contain millions of parameters, requiring substantial computational resources, including time, memory, and large-scale training data.

Video classification presents additional challenges due to the sheer size of uncompressed video frame sequences, which further increases computational demands. To address this, prior works have processed only a fixed, limited number of frames or clips~\parencite{vaanet, wehrmann}. However, this approach inevitably leads to information loss, potentially compromising classification accuracy. The computational complexity of processing raw videos increases further with the adaptation of larger models---a trend that continues to expand over the years \parencite{larger2,larger1}. When the training data is limited, using larger input dimensionality and larger models also increases the chance of overfitting \parencite{avoiding_overfitting}. 

We ask the following research question:

\begin{quote}
\textit{How can we design efficient video analysis models with high representation power, potentially capable of handling an arbitrary number of input frames, while ensuring generalization?}
\end{quote}

Our hypotheses are:

\begin{quote}
\textit{Due to the large size of raw video inputs, training models end-to-end on such data is prohibitively expensive and likely inefficient. }

\textit{Models pretrained on various multimedia tasks can provide compact and meaningful inputs for video classification, leading to improved performance and efficiency.}
\end{quote}

Using pretrained models brings multiple advantages. Firstly, since the pretrained model is used in inference mode, its weights remain frozen, therefore reducing the time and memory complexity by requiring only a forward pass, without an additional backward pass. Secondly, it considerably reduces the size of the input fed to the subsequent model that is being trained. As an example, a very small video frame contains $224 \times 224 \times 3 = 163{,}968$ values, while the state-of-the-art image analysis model CLIP (Contrastive Language-Image Pre-Training) represents an image using only $512$ values \parencite{clip}. This size reduction becomes more important as the duration of the video increases. Thirdly, assuming that the pretrained features of multimedia analysis models present valuable and relevant information for the task of video classification, their use removes the need to train new models to extract similar features implicitly from raw data. This reduces the size of the subsequent model that needs to be trained. Finally, since it reduces the input size and the number of weights in training, it also reduces the chance of overfitting \parencite{avoiding_overfitting}.

There are several methods to utilize pretrained networks, which have significant overlaps. These methods are commonly referred to as transfer learning, fine-tuning, or using pretrained features \parencite{transfer1,transfer2}. These terminologies can be distinguished as follows: Transfer learning involves applying a model trained on one task to a different task \parencite{transfer_learning}. Fine-tuning entails further training the transferred model for the new task \parencite{finetuning}. In our work, we use pretrained features, which involves performing inference on the data using a pretrained model, and then feeding the resulting outputs or activations (features) into a new model that is trained~\parencite{pretrained_features}. These methods are frequently employed in video processing tasks such as human activity recognition \parencite{activity_e}, video summarization \parencite{summarization_e}, and video recommendation \parencite{recommendation_e}.

We continue with the following research questions: 

\begin{quote}
\textit{How does training a task-specific DNN end-to-end compare with utilizing frozen pretrained models?}

\textit{How can we efficiently use open-source pretrained networks and combine them for video classification?}
\end{quote}

Our hypothesis is as follows:
\begin{quote}
\textit{Powerful multimodal pretrained models can be used in inference mode, and their outputs can be fused with lightweight neural networks trained separately---enhancing both performance and efficiency.}
\end{quote}

We present architectures trained end-to-end and models that fuse pretrained networks, and then compare their performance. We employ pretrained models for image analysis, optical character recognition, automatic speech recognition, audio event classification, music classification, and facial emotion classification to incorporate information about visual scenery, objects, text, speech, music, audio, and human faces. These pretrained models are used in inference mode, meaning they are not fine-tuned, which results in significantly lower time and memory requirements. We then use trained neural networks to fuse the pretrained features and obtain the final classification results. Since pretrained models extract meaningful features from raw data, a shallow fusion network with just one or two layers is sufficient as the classifier, greatly reducing computational complexity, hardware requirements, and training time.

We also investigate various semantic video classification datasets in terms of quality, quantity, and their suitability for our overall objective. We explore the types of video datasets that offer the most variety to enable our models to generalize to arbitrary videos. We hypothesize that user-generated videos and cinematic videos are the two best candidates. User-generated videos, collected from platforms such as YouTube\footnote{\url{https://youtube.com}} and Flickr\footnote{\url{https://flickr.com}}, have minimal restrictions on the type of videos uploaded, aside from legal considerations. Similarly, cinema, as an art form, seeks to push the boundaries of creativity and, therefore, contains a wide range of visual and narrative variety.

Our preliminary investigations reveal a lack of high-quality, large-scale video datasets with emotion labels. Ekman-6, one of the largest video emotion classification benchmarks, only contains 1,637 videos and leads to overfitting~\parencite{ekman6}. Using a small, fixed number of input frames addresses overfitting but at the cost of information loss~\parencite{wehrmann,vaanet}. In this work, we aim to tackle these issues of data scarcity and information loss. We pose the following research questions: 

\begin{quote}
\textit{What other semantic video classification tasks can provide a large amount of training data to further mitigate overfitting? }

\textit{Given sufficient data, what is the relationship between the number of input frames and classification performance?}
\end{quote}

We further challenge the notion of using a small and fixed number of frames. We argue that:

\begin{quote}
	
\textit{Cinematic video datasets provide a vast amount of labeled video samples, thanks to the availability of online catalogs such as IMDb\footnote{\url{https://imdb.com}} and video hosting platforms like YouTube~\parencite{movienet,movielens}. }
	
\textit{By using compact inputs and a large amount of data, increasing the number of input frames will consistently improve classification performance.}
\end{quote}

To this end, we explore cinematic datasets that offer a substantial number of samples. By leveraging their size, we can address overfitting, train larger and more powerful models, and utilize more input frames to fully maximize the information provided to the model.

A cinematic counterpart to video emotion classification is movie trailer genre classification. Since both emotion and genre classification involve discrete categories, we can employ classifier models that predict probability distributions over a set of predefined labels. This similarity enables us to tackle both tasks using a shared model architecture.

We continue our discussion with the integration of the symbolic music generation and video analysis modules to realize a video-based music generator.

\section{Video-based music generation}

We finally integrate the previously built components—video emotion classifier, FFMpeg scene cut extractor, emotion-conditioning mechanism, and temporal boundary conditioning mechanism—to construct the video-based music generator.

A key challenge arises from the mismatch in representation between the video emotion classifier and the emotion-conditioning mechanism of the music generator. While our video emotion classifier outputs probabilities for discrete emotions, our music generator represents emotions using continuous valence-arousal values~\parencite{valence_arousal}. We ask:

\begin{quote}
\textit{How can we bridge categorical (discrete) and dimensional (continuous) emotion representations?}
\end{quote}

We hypothesize that:

\begin{quote}
\textit{Statistical data from psychological user studies may support the development of a mapping method between these two representations.}
\end{quote}

To bridge this gap, we introduce a simple mapping method based on psychological studies~\parencite{mapping}. These studies present participants with scenarios associated with discrete emotions and ask them to rate the felt emotion in terms of valence (pleasantness vs. unpleasantness) and arousal (intensity). Researchers then report the mean and standard deviation of valence and arousal for each discrete emotion category. We use these statistical values to construct a mixture of Gaussian distributions, where each component corresponds to a discrete emotion. The final valence-arousal value is then sampled from this mixture, weighted according to the probability distribution output by the video emotion classifier. Finally, we evaluate our complete model against existing video-based music generators. Extensive user studies demonstrate that our method outperforms prior approaches across all evaluated metrics.

While the audio samples in our user study are generated using off-the-shelf Fluidsynth software, we further explore the task of audio generation---focusing particularly on bandwidth extension and its associated challenges.

\section{Audio generation and bandwidth extension}
\label{sec:intro_audio}
Our overall video-based music generator outputs symbolic music that is compact, editable, and interpretable. To generate audio that can be heard by humans, the users have several alternatives, ranging from quick but low-quality methods to high-quality but time-consuming approaches. First, users can utilize computer software such as Fluidsynth and soundfont libraries to quickly synthesize MIDI into audio with minimal intervention. These soundfont libraries provide fixed sounds for specific pitches of particular instruments. This is the approach we use to generate our final samples, as our work is focused on music composition rather than sound design. However, as expected, the resulting audio is mechanical, unrealistic, and generally of lower quality compared to music produced by human musicians.

Another alternative is to use Digital Audio Workstations (DAWs) to generate audio from MIDI. In this approach, the end user can edit all aspects of the MIDI, including notes, tempo, instruments, and velocities, as well as the audio itself, using proprietary soundfonts, custom effects, and various mixing and mastering techniques. While this approach produces higher-quality audio, it is manual, requiring significant time, skills, and labor. Finally, human musicians can perform and record the compositions generated by our music generator. This method yields the highest-quality audio but also demands the most time, skills, and labor.

An alternative research-oriented solution is to use DNNs to synthesize audio from MIDI. However, this is an ill-defined problem, as the same music composition, i.e., a MIDI file, can have an infinite number of human performances. There are models that attempt to tackle this task, but they rely on paired datasets of MIDI and audio, which are typically piano-based~\parencite{maestro,midi2audio2}. Since our work focuses on multi-instrumental music generation, we face the challenge that there are no multi-instrument datasets with pairs of audio performed by humans and MIDI. One could train DNNs on synthetic datasets containing MIDI and the corresponding synthesized audio, but this would limit the DNN's performance to the quality of the synthesis method, rendering it somewhat redundant. Additionally, this could be viewed as a generalization problem, where the model only generates synthetic-sounding audio and fails to generalize to real-world samples created by musicians.

To investigate this problem further, we ask the following research questions:

\begin{quote}
\textit{What other audio generation tasks exhibit overfitting caused by synthetic data?}

\textit{How can we quantify and alleviate overfitting to synthetic data in audio generation tasks?}
\end{quote}

Our hypotheses are as follows:

\begin{quote}
	
\textit{Audio enhancement tasks also suffer from overfitting, due to their use of synthetically degraded audio as inputs.}

\textit{To systematically analyze overfitting, and to mitigate it, we can utilize a diverse set of degradation functions during the training and testing of DNNs.}
\end{quote}

Audio enhancement tasks involve using low-quality audio as input and its high-quality counterpart as output. Audio enhancement models are often trained with synthetic data, where high-quality audio is degraded using digital filters to create the low-quality counterparts~\parencite{kuleshov}. This task only requires a high-quality audio dataset, without the need for a corresponding sample from a different modality, such as MIDI. As audio enhancement tasks only use audio-only samples, the data is abundant. We aim to quantify generalization by applying multiple digital degradation filters and measuring whether the trained models generalize across different filters. Specifically, we focus on the task of audio bandwidth extension, which allows for both qualitative and quantitative analysis by observing and evaluating the reconstruction quality of the missing frequency bands. Furthermore, in audio bandwidth extension, we can use various low-pass filters by varying their types and orders.

As this problem stems from the use of synthetic data, we hypothesize the solution is to use data augmentation strategies, and not model-based methods. Our results show that using a variety of filters while creating the input samples alleviate overfitting. Our approach marks the first step towards achieving real-world audio bandwidth extension models.

We now summarize our contributions.

\section{Contributions}

Our first key contribution is the construction of a large-scale MIDI dataset with emotion labels. This dataset enables training models where emotions serve as conditional inputs and MIDI sequences act as ground-truth targets. Motivated by the well-established principle that larger models trained on larger datasets yield better results~\parencite{larger_models}, we avoid using existing emotion-labeled MIDI datasets, as they contain few samples~\parencite{vgmidi,emo_lstm,emopia}. Instead, we assign emotion labels to the Lakh MIDI dataset, which comprises 176,581 samples~\parencite{lmd}. By integrating multiple auxiliary datasets, we annotate 34,791 MIDI samples with valence and arousal values, creating the largest emotion-labeled MIDI dataset to date. Using this dataset, we successfully train large transformer models~\parencite{transformer} to generate multi-instrument symbolic music conditioned on emotion. To the best of our knowledge, this is the first music generation model capable of conditioning on both valence and arousal simultaneously, allowing control over any emotion within the widely used circumplex model of affect \parencite{valence_arousal}. We shared our findings in the following paper and open-source code.

\begin{quote}

\textit{\textbf{S. Sulun}, M. E. P. Davies, and P. Viana, "Symbolic Music Generation Conditioned on Continuous-Valued Emotions," IEEE Access, vol. 10, pp. 44617–44626, 2022.} \parencite{sulun_midi}

\textit{Emotion-based MIDI generator.} \url{https://github.com/serkansulun/midi-emotion}

\end{quote}

We additionally provide emotion labels for MIDI files by classifying the sentiments of lyrics contained within the MIDI files. To achieve this, we first train emotion classification models on GoEmotions, one of the largest text datasets with 28 fine-grained emotion labels~\parencite{goemotions}. Using these models, we infer emotion labels from the lyrics associated with the MIDI files. Although this approach did not succeed in improving emotion-based MIDI generation, our model achieved state-of-the-art performance on the GoEmotions test set while being only half the size of the baseline model. We presented our findings in the following paper and open-source code and dataset.

\begin{quote}

\sloppypar \textit{\textbf{S. Sulun}, P. Oliveira, and P. Viana, "Emotion4MIDI: A Lyrics-Based Emotion-Labeled Symbolic Music Dataset," in Progress in Artificial Intelligence, Cham: Springer Nature Switzerland, 2023, pp. 77–89.} \parencite{epia}

\textit{Emotion classification of MIDI lyrics. Includes pretrained emotion text classifier, and dataset with the labels for the MIDI files in Lakh and Reddit MIDI datasets.} \url{https://github.com/serkansulun/lyricsemotions}

\end{quote}

We also present a novel mechanism for the temporal conditioning of our music generator. Unlike state-of-the-art video-based music generators that rely on fixed, coarse time grids~\parencite{di,kang}, our method preserves event-based encoding while enabling fine-grained temporal control. We achieve this by encoding and feeding \textit{boundary offsets} into the model. These offsets inform the model of the remaining time until the next chord, allowing it to "anticipate" upcoming chords and generate preceding notes accordingly. As a result, the generated chords and surrounding notes remain rhythmically and  harmonically coherent. We introduced this method in our final paper that presents our overall model, along with an online demo.

\begin{quote}

\textit{\textbf{S. Sulun}, P. Viana, and M. E. P. Davies, "Video Soundtrack Generation by Aligning Emotions and Temporal Boundaries," IEEE Transactions on Multimedia, 2026, in print.} \parencite{emsync}

\textit{Video-based music generator.} \url{https://github.com/serkansulun/emsync}

\textit{Demo.} \url{https://colab.research.google.com/drive/1hq6WmZztJk-yKp0UX1UTon8rJ3xC-bV6}

\end{quote}

We then present our approach for semantic video classification. Our method maintains relatively low computational complexity by leveraging compact pretrained features instead of raw frames. In video emotion classification, we fuse the pretrained features intelligently using cross-attention layers \parencite{attention}. Our method outperforms state-of-the-art models in one of the largest video emotion classification datasets, Ekman-6~\parencite{ekman6}, by a significant margin. As a secondary contribution, we identify labeling errors within the Ekman-6 dataset. These errors stem from issues in the video search process, such as mislabeling videos of celebrities named "Joy" as expressing the emotion of joy. To address this, we manually review all videos in the Ekman-6 dataset and compile a blacklist of mislabeled samples to assist future researchers. We presented our findings in the following paper and online demo:

\begin{quote}

\textit{\textbf{S. Sulun}, P. Viana, and M. E. P. Davies, “VEMOCLAP: A video emotion classification web application,” International Symposium on Multimedia (ISM), IEEE Computer Society, pp. 137–140, 2024.} \parencite{vemoclap}

\textit{Video emotion classifier.} \url{https://github.com/serkansulun/video-emotion}

\textit{Demo.} \url{https://colab.research.google.com/drive/1S-Nsm968-KTErbuU0qHOp7mEEzh6xOVx}

\end{quote}

While the Ekman-6 dataset contains 1,637 videos averaging 10 seconds in length~\parencite{ekman6}, the cinematic dataset MovieNet consists of 33,000 movie trailers, each averaging 2 minutes~\parencite{movienet}. For trailer genre classification using MovieNet, we train more robust architectures by incorporating a greater number of input frames without risking overfitting. To fuse pretrained features extracted from both video frames and audio segments, we employ the transformer architecture—recognized as the state-of-the-art for sequence processing and the backbone of modern large language models~\parencite{transformer}. This architecture is well-suited for handling long input sequences and capturing long-term dependencies. Leveraging this capability, our genre classification model processes the full video and audio content: it first identifies scene cuts and then uses \textit{all} the resulting frames, alongside the complete audio. Consequently, our approach outperforms existing state-of-the-art models in genre classification on MovieNet, the largest cinematic dataset available. Additionally, we demonstrate empirically that classification performance improves consistently as more input frames are incorporated. We detail our findings in the following paper and provide the corresponding open-source code.

\begin{quote}

\textit{\textbf{S. Sulun}, P. Viana, and M. E. P. Davies, “Movie trailer genre classification using multimodal pretrained features,” Expert Systems with Applications, vol. 258, p. 125209, 2024.} \parencite{eswa}

\textit{Trailer genre classifier.} \url{https://github.com/serkansulun/trailer-genre-classification}

\end{quote}

We finally raise the issue of filter generalization for DNNs neural networks applied to musical audio bandwidth extension. Contrary to many problems for which deep learning is used, we do not find any evidence of overfitting to audio samples themselves (i.e. the training data), but rather, we observe a clear trend for state of the art DNNs to overfit to filter shapes. When these DNNs are presented with audio excerpts that have been preprocessed with low pass filters not included in the training, then no meaningful extension of the bandwidth can be obtained. Furthermore, the use of widely adopted regularization layers such as batch normalization and dropout fall short in alleviating this problem. To address the filter overfitting issue we propose a novel data augmentation approach, which uses multiple filters at the time of training. Our results demonstrate that without data augmentation, filter overfitting increases as training progresses, whereas including data augmentation is a promising step towards achieving filter generalization. Our findings are presented in the following paper and open-source code.

\begin{quote}

\textit{\textbf{S. Sulun} and M. E. P. Davies, "On filter generalization for music bandwidth extension using deep neural networks," IEEE Journal of Selected Topics in Signal Processing, vol. 15, no. 1, pp. 132–142, 2020.} \parencite{sulun_audio}

\textit{Audio bandwidth enhancer.} \url{https://github.com/serkansulun/deep-music-enhancer}

\end{quote}

The remainder of this thesis is structured as follows. Chapter~\ref{chap:background} introduces the background and related work on emotion representations, transformers, sequence processing (text and MIDI), video analysis, video-based music generation, and audio bandwidth extension. Chapter~\ref{chap:labeling} details our approach to labeling a large MIDI dataset with emotions. Chapter~\ref{chap:music} discusses our conditional music generation methods, focusing on emotion and temporal boundary conditioning. Chapter~\ref{chap:video} presents our video analysis models, specifically for emotion classification, genre classification, and scene cut extraction. Chapter~\ref{chap:video_music} integrates these components into a complete model for video-based music generation. Chapter~\ref{chap:audio} presents our exploration on the task of audio bandwidth extension. Finally, Chapter~\ref{chap:conclusion} summarizes our conclusions and outlines future directions. 
\chapter{Background and related work} \label{chap:background}

In this chapter, we begin by introducing the key concepts related to our goal of video-based music generation. Our work involves multiple types of data—video, image, audio, text, symbolic music, and emotion. While video, image, audio, and text have well-known and standardized digital formats, symbolic music and emotion do not have a single common way of being represented. We therefore start by presenting the symbolic music and emotion representations, and the different ways that they are encoded. 

Next, we introduce the transformer architecture, which is the state-of-the-art approach for processing sequences~\parencite{transformer}. Since symbolic music is a sequence of notes, video is a sequence of frames, and text is a sequence of words or subwords, we rely heavily on transformers across the various subtasks in our work. Later in the thesis, we present significant modifications to the standard transformer to implement novel conditioning mechanisms as well as for multimodal processing. For this reason, it is important to first provide a theoretical background on how transformers work.

After explaining aforementioned key concepts, we present a literature review on specific applications. We first describe the generic task of sequence processing and then its particular subtasks such as text emotion classification, conditional generic sequence generation, and conditional symbolic music generation. 

We then cover video analysis in a dedicated section, separate from generic sequence processing. Although video can be described as a sequence of frames, its structured pixel layout—spanning width, height, and time—distinguishes it from tokenized sequences like MIDI and text. In this section, we also discuss video emotion classification and trailer genre classification.

Next, we review the literature on video-based symbolic music generation, which represents the current state-of-the-art in our overall task. We finally present the works on audio bandwidth extension, which is a subtask of audio generation.

\section{Symbolic music}
\label{sec:midi_background}

Symbolic music formats, such as MIDI (Musical Instrument Digital Interface), are used to represent musical performances and compositions in the digital domain. These files contain only the musical information, such as notes, tempo, and dynamics, without the actual sound, making them analogous to a "digital music sheet." Compared to audio formats, symbolic music files are much smaller in size and dimensionality, which makes them more manageable and suitable for modeling with deep neural networks~\parencite{music_survey}. The two prevailing music formats are MIDI (Musical Instrument Digital Interface) and pianoroll, which can seamlessly be converted into each other.

Unlike audio formats (e.g., WAV, MP3), MIDI does not store actual sound. Instead, it encodes musical instructions, such as pitch, velocity, duration, and instrument choice, which can be interpreted by MIDI-compatible devices (e.g., synthesizers, virtual instruments). MIDI files consist of \textit{events}, where each note is encoded with a delta-time value (denoting the time shift), pitch (denoting the note), channel (denoting the instrument), and velocity (denoting the loudness). This results in a 1-dimensional sequence representation.

\begin{figure}[htbp]
	\centering
	\includegraphics[width=1\linewidth]{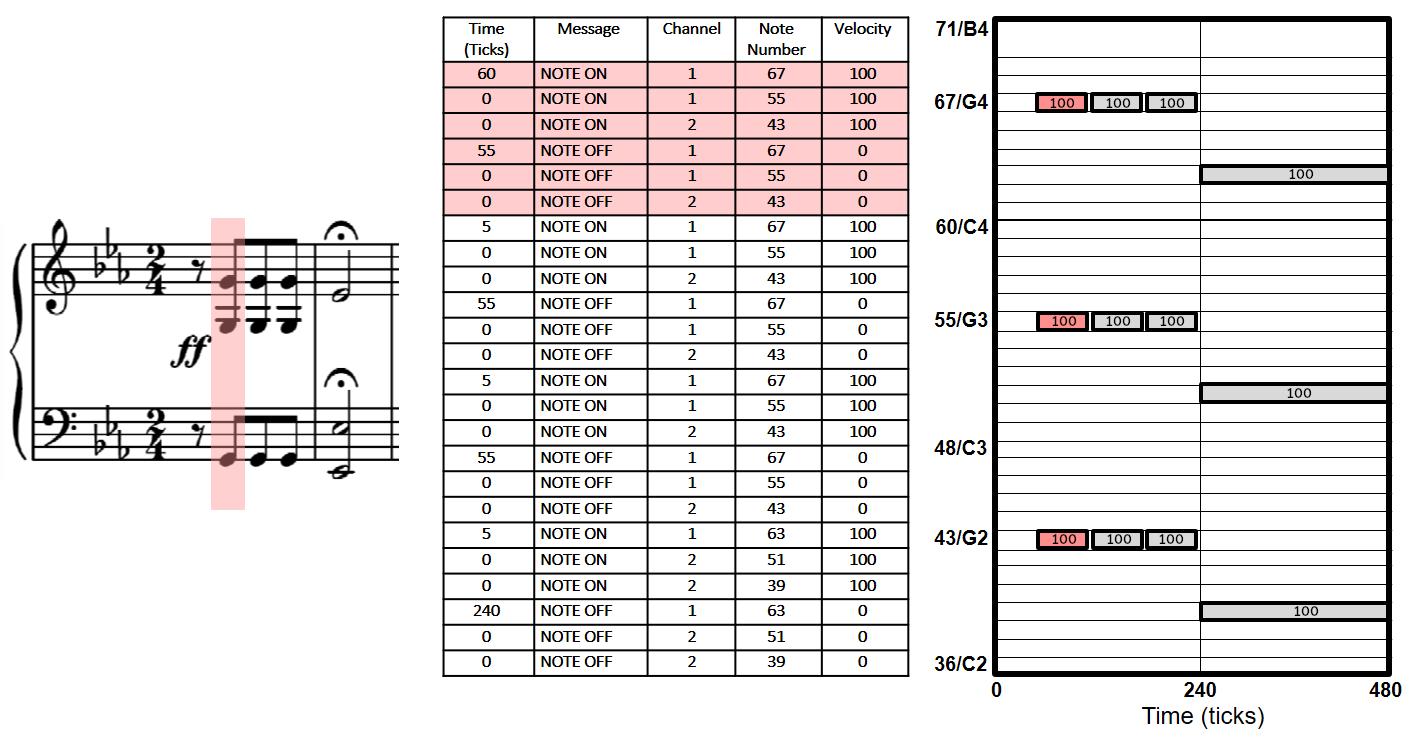}
	\caption{Sheet music (left), MIDI (middle), and pianoroll (right) representations. Taken from \textcite{fundamentals_music_processing}.}
	\label{fig:symbolic_music}
\end{figure}

In contrast, the pianoroll format organizes the notes in a 2-dimensional representation: the horizontal axis represents time, the vertical axis represents pitch, and the values correspond to velocity, resulting in a 2-dimensional matrix. For multi-instrument music, multiple 2-dimensional matrices are used—one per instrument. Figure~\ref{fig:symbolic_music} displays the first twelve notes of Beethoven's Fifth Symphony, represented in sheet music, MIDI, and pianoroll formats. In this figure, the sheet music and the MIDI represents the left and right hand notes of the piano using different staves and channels, respectively. The pianoroll collects them into a single matrix since they belong to the same instrument.

The \textit{event-based} symbolic music representation is an adaptation of MIDI for neural network use~\parencite{event_encoding}. It splits some of the MIDI properties into separate events before tokenization, creating a dictionary of manageable size. In particular, it separates the delta-time information to create \texttt{TIME-SHIFT} events and the velocity information to create \texttt{SET-VELOCITY} events. \texttt{TIME-SHIFT} tokens are used to move along the time axis, representing both note durations and the silences between them. Each token specifies a time increment in milliseconds. For example, an 800-millisecond shift is encoded as \texttt{TIME-SHIFT<800ms>}. Both velocities and time shifts can be quantized to reduce the vocabulary size. Typically, time shifts are quantized at 8 milliseconds, with a maximum time shift of 1000 milliseconds~\parencite{event_encoding}. Timeshifts longer than 1000 milliseconds can be represented by multiple consecutive time shift tokens.

\begin{figure}[h]
	\centering
	\includegraphics[width=1\textwidth]{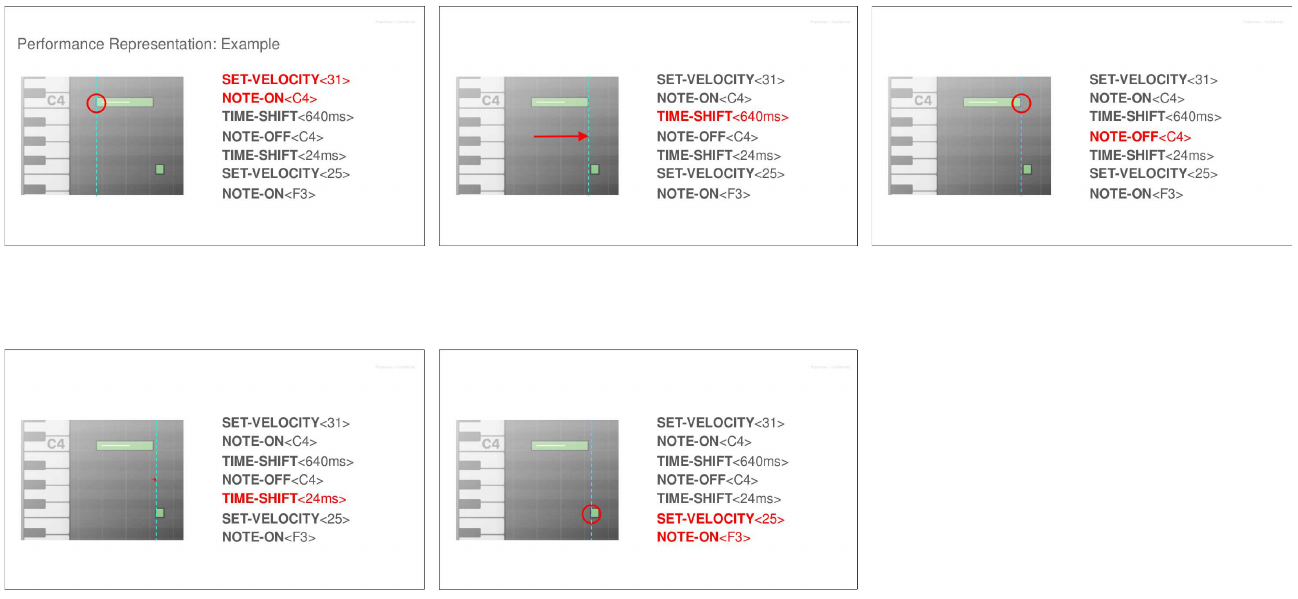}
	\caption{Conversion of pianoroll into event-based representation. Taken from \textcite{event_encoding}. License: CC 4.0 BY}
	\label{fig:event}
\end{figure}

Figure~\ref{fig:event} displays the conversion of symbolic music into an event-based representation. Pianoroll is used in this figure as it is a more visual representation compared to MIDI, as seen in~\ref{fig:symbolic_music}.  However, the same conversion process applies to MIDI as well. First, a C4 note with a duration of 640 milliseconds and a velocity of 31 is played, followed by a 24-millisecond silence. After that, an F3 note is played with a velocity of 25. When processing the first note, assuming the previous velocity was different, we set the velocity to 31 using a \texttt{SET-VELOCITY<31>} token. Then, we mark the beginning of the first note and its velocity using a \texttt{NOTE-ON<C4>} token. Next, using the \texttt{TIME-SHIFT<640ms>}, we move forward in time by 640 milliseconds, until we arrive at the next note boundary (i.e., a \texttt{NOTE-ON} or a \texttt{NOTE-OFF} token). Since the velocity of the next note differs from the currently set velocity of 31, we set the new velocity to 25 using a \texttt{SET-VELOCITY<25>} token. Finally, we mark the beginning and the pitch of the next note using a \texttt{NOTE-ON<F3>} token.

Event-based encoding was originally developed for single-instrument music, particularly piano music~\parencite{event_encoding}. However, it is straightforward to encode instrument information by injecting it into the \texttt{NOTE-ON} and \texttt{NOTE-OFF} tokens~\parencite{lakhnes}. This modification allows for the representation of multi-instrument music, though it increases the vocabulary size. As a result, we obtain tokens such as \texttt{NOTE-ON-PIANO<C4>} and \texttt{NOTE-OFF-GUITAR<F3>}.

We now turn to the next modality relevant to our work: emotions.

\section{Emotion representations}

Emotion is a central concept supporting multiple subtasks in this thesis. It forms one of the two branches of our video-based music generator, serving as the bridge that links video content to musical output. We employ emotions as targets and outputs while classifying videos and as input conditions while generating music. However, unlike modalities such as video, which is always represented as an array of pixels, emotions can be modeled in multiple ways, making their computational modeling less standardized.

Emotion representation is a fundamental concept in affective computing and psychology, forming the basis for analyzing and modeling human emotions~\parencite{emotionrepresentation1,emotionrepresentation2,emotionrepresentation3}. There are two primary approaches to representing emotions: categorical and dimensional models.

\textcite{ekman} presented a categorical approach to classify facial expressions and physiological responses using discrete categories, such as happiness, sadness, anger, fear, surprise, and disgust. The dimensional approach represents emotions as points within a continuous space, typically defined by valence (ranging from unpleasant to pleasant) and arousal (ranging from calm to excited)~\parencite{valence_arousal}. Both approaches offer distinct strengths and limitations. Categorical models align more naturally with how humans perceive and label emotions, offering simplicity and clarity. Yet, their reliance on a fixed set of discrete categories results in a coarse representation that may overlook the fluid and overlapping nature of emotional experiences. In contrast, dimensional models provide a nuanced and flexible representation of emotional states, capturing subtle variations through continuous numerical coordinates. However, this complexity can make them less intuitive and harder for the general public to interpret.

\subsection{Categorical emotion representation}
\label{sec:categorical_emotion}
\textcite{ekman} conducted pioneering research on facial expressions and emotions, identifying six universal facial emotions that are present across cultures: happiness, sadness, anger, fear, surprise, and disgust. These categories continue to serve as labels for modern emotion-related datasets~\parencite{ekman6}. Expanding on the role of emotions in early development and psychological functioning, \textcite{humanemotions} proposed a broader set of ten basic emotions: joy, interest, sadness, anger, disgust, contempt, fear, shame, guilt, and surprise.

\begin{figure}[htbp]
   	\centering
   	\includegraphics[width=0.7\linewidth]{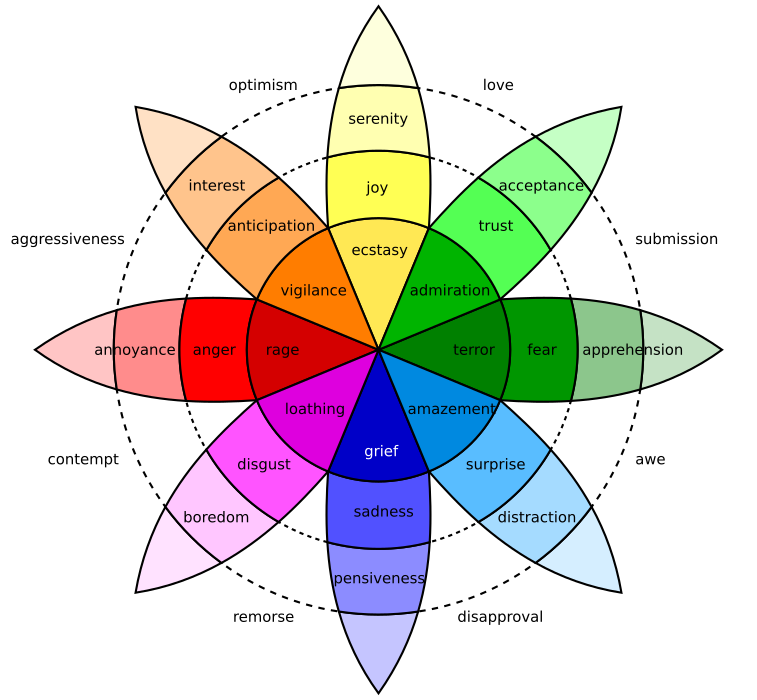}
   	\caption{Emotion wheel of \textcite{emotionwheel}. Author/Copyright holder: Machine Elf 1735. Copyright terms and licence: Public Domain.}
   	\label{fig:emotion_wheel}
\end{figure}

\textcite{emotionwheel} introduced an alternative framework by organizing eight primary emotions---joy, trust, fear, surprise, sadness, disgust, anger, and anticipation---into a circular structure, pairing them as opposites to form an emotion wheel. Each primary emotion is further divided into high- and low-intensity counterparts, and secondary emotions emerge from blending two primary emotions. His representation of the emotion wheel is illustrated in Figure~\ref{fig:emotion_wheel}. Notably, his concept of emotion intensities aligns with dimensional models, where arousal serves as a key dimension for defining emotional intensity.

\begin{figure}[hpbt]
	\centering
	\includegraphics[width=0.7\linewidth]{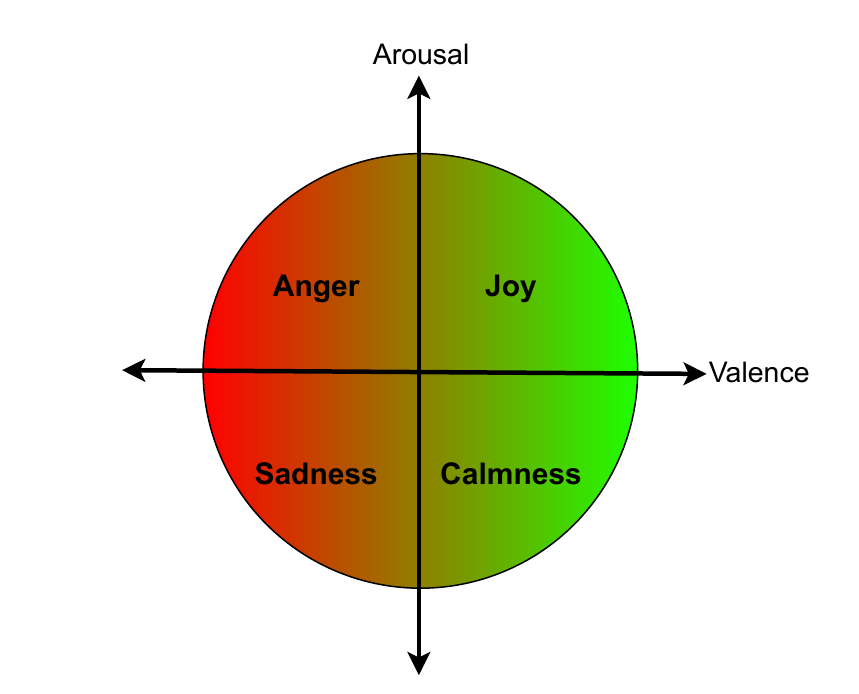}
	\caption{Circumplex model of emotions \parencite{valence_arousal}.}
	\label{fig:valence_arousal}
\end{figure}

\subsection{Dimensional emotion representation}
\label{sec:valence_arousal}
Dimensional models of emotion represent affect as points in a continuous space instead of fixed categories. These models use numerical values along different axes to describe emotions, allowing for a more flexible and detailed representation of emotional states. \textcite{valence_arousal} famously introduced the circumplex model of affect, which maps emotions onto a two-dimensional circular plane. The horizontal axis (valence) indicates how pleasant or unpleasant an emotion is (e.g., joy is positive, while sadness is negative). The vertical axis (arousal) represents the level of activation or energy (e.g., anger is positive, while calmness is negative).

Due to its simplicity in representing emotions with just two continuous values, the circumplex model is widely applied in emotion analysis tasks, including emotion prediction from images \parencite{facialemotion}, physiological signals \parencite{physiologicalemotion}, and music \parencite{musicemotion}. Figure~\ref{fig:valence_arousal} illustrates the circumplex model, while showing how categorical emotions are distributed within its quadrants.

Having covered the key data modalities---MIDI and emotion---we now introduce the architecture central to our work: the transformer.

\section{Transformers}

In this thesis, we tackle several sequence processing tasks, which we describe in detail in later chapters. MIDI is treated as a sequence of notes, and transformers are used to model and generate these sequences. In video classification, features extracted from individual frames form a sequence that can be processed by transformers to predict emotions or cinematic genres. We also work with raw audio, which is often too lengthy to handle in a single pass. To manage this, we split the audio into chunks, extract features from each chunk, and then process the resulting sequence using transformers. Similarly, in text-based tasks, sequences of words or subwords are fed into transformers to perform emotion classification.

Transformers have revolutionized deep learning, particularly in natural language processing (NLP). They form the backbone of all large language models, including ChatGPT\footnote{\url{https://chatgpt.com}}, Llama\footnote{\url{https://llama.com}}, and Gemini\footnote{\url{https://gemini.google.com}}. Unlike recurrent models, transformers process entire sequences in parallel, eliminating the need for recurrence. By leveraging attention mechanisms and parallel computation, transformers have set new performance benchmarks in NLP tasks such as machine translation~\parencite{translation}, text generation~\parencite{textgeneration}, and question answering~\parencite{questionanswering}. However, their superiority extends beyond NLP. Modern architectures for automatic speech recognition~\parencite{whisper}, image processing~\parencite{vit}, and MIDI generation \parencite{musictransformer} also rely on transformers.

\begin{figure}[h]
	\centering
	\includegraphics[width=0.6\textwidth]{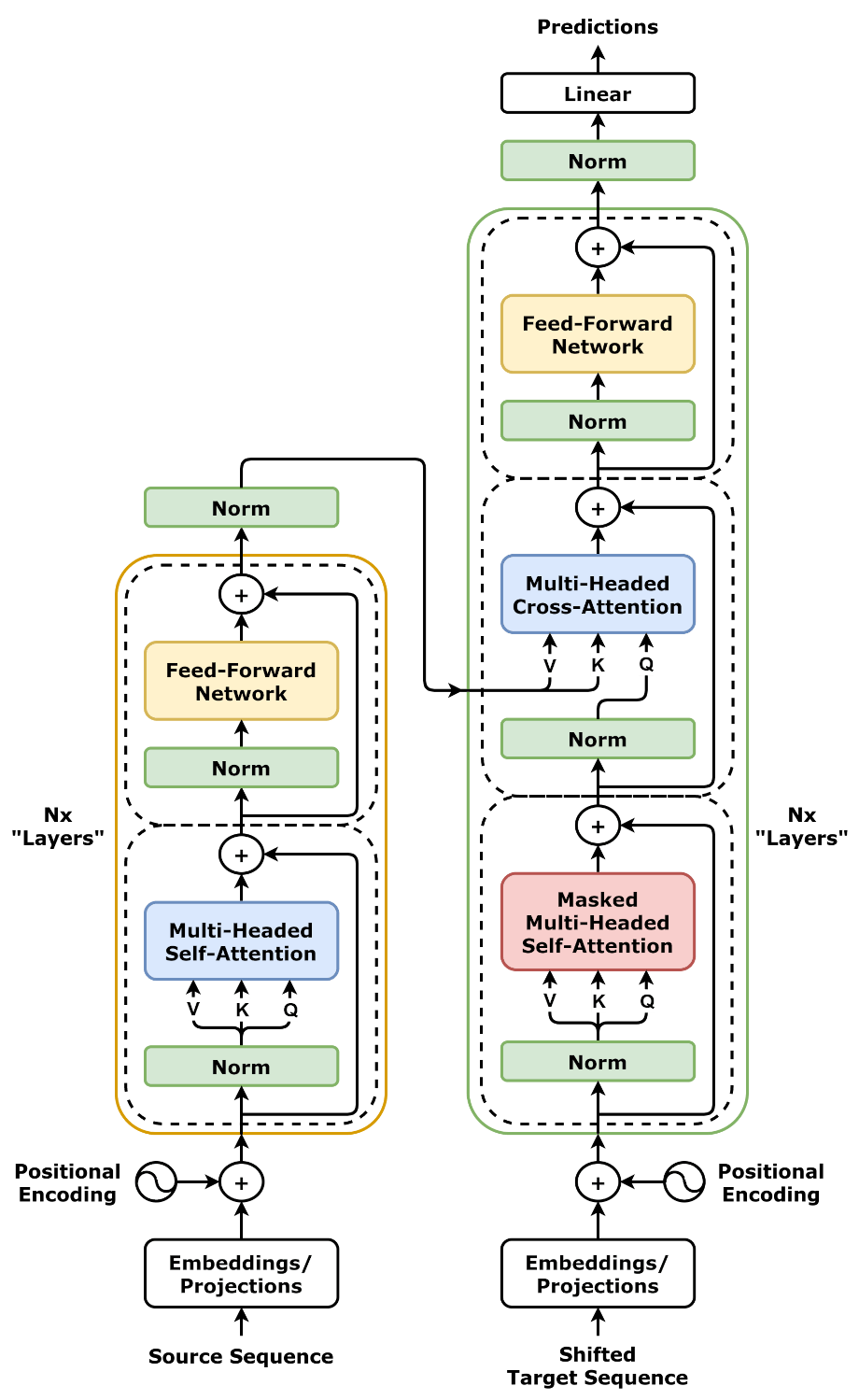}
	\caption{The transformer model, with the encoder on the left and the decoder on the right. - \url{https://github.com/dvgodoy/dl-visuals/?tab=readme-ov-file}, CC BY 4.0}
	\label{fig:transformer}
\end{figure}

We now explain the theory behind transformers, beginning with the block diagram shown in Figure~\ref{fig:transformer}. The full architecture consists of an encoder (left) and a decoder (right). This encoder-decoder setup is commonly used in tasks like machine translation, where the encoder processes the source sequence and the decoder generates the target sequence. In purely generative tasks, such as text or MIDI generation, only the decoder is used. In this configuration, the target sequence is a shifted version of the input sequence, meaning that for an input token at position $n$, the target token is at position $n+1$. In simpler terms, given an input sequence, the model predicts the next token.

Tokens are discrete elements of a sequence. In MIDI processing, tokens correspond to individual MIDI messages, such as triggering a note with a specific pitch. In text processing, tokens typically represent words, though subwords are often used to capture semantic relationships between word variations~\parencite{subwords}. This prevents redundant tokenization of words with shared roots, improving efficiency.


After each token is extracted, unique tokens define the \textit{vocabulary}. Each unique token can be represented with a numerical index within the vocabulary, allowing the input sequence to be represented as a sequence of discrete numbers. Embedding refers to the process of projecting these discrete numbers into vectors with continuous values. The embedding layer is simply a lookup table, defined as a weight matrix with size $V \times D$, where $V$ is the vocabulary length and $D$ is the model dimensionality. The embedding layer maps each token's index to the corresponding row in this matrix, extracting a vector of length $D$. These vectors are then concatenated along the sequence dimension. For an input sequence of length $S$, the embedding layer produces a matrix of continuous values with size $S \times D$.

\paragraph{Positional encoding}
\label{sec:position_background}
Transformers, unlike Recurrent Neural Networks (RNNs)~\parencite{rnn}, do not process sequences sequentially. Instead, they process the entire input sequence in parallel, significantly improving efficiency. However, this parallelization introduces a key challenge: transformers lack an inherent sense of word order because self-attention treats all input tokens simultaneously. 

The example below shows the importance of positional order. Without positional information, the following two sentences would be interpreted as identical:

1- "I only said that she could help him."

2- "I said that only she could help him."

These two sentences are different because of the \textit{position} of the word "only," which changes the emphasis and meaning. In the first sentence, "only" modifies the verb "said." This means that the speaker's action was limited to just saying that she could help him, and not doing anything more. It suggests that the speaker didn't imply anything else beyond saying that she could help. In the second sentence, "only" modifies "she." This means that the speaker is emphasizing that she is the exclusive person who can help him. No one else, except for her, can help.

In order to incorporate positional information, transformers utilize \textit{positional encodings}, which can be defined in various ways. The original transformer employs sinusoidal encodings:

\begin{equation}
PE_{(pos, 2i)} = \sin \left(\frac{pos}{10000^{2i/D}}\right)
\end{equation}

\begin{equation}
PE_{(pos, 2i+1)} = \cos \left(\frac{pos}{10000^{2i/D}}\right)
\end{equation}

\begin{figure}[htbp]
	\centering
	\includegraphics[width=1\linewidth]{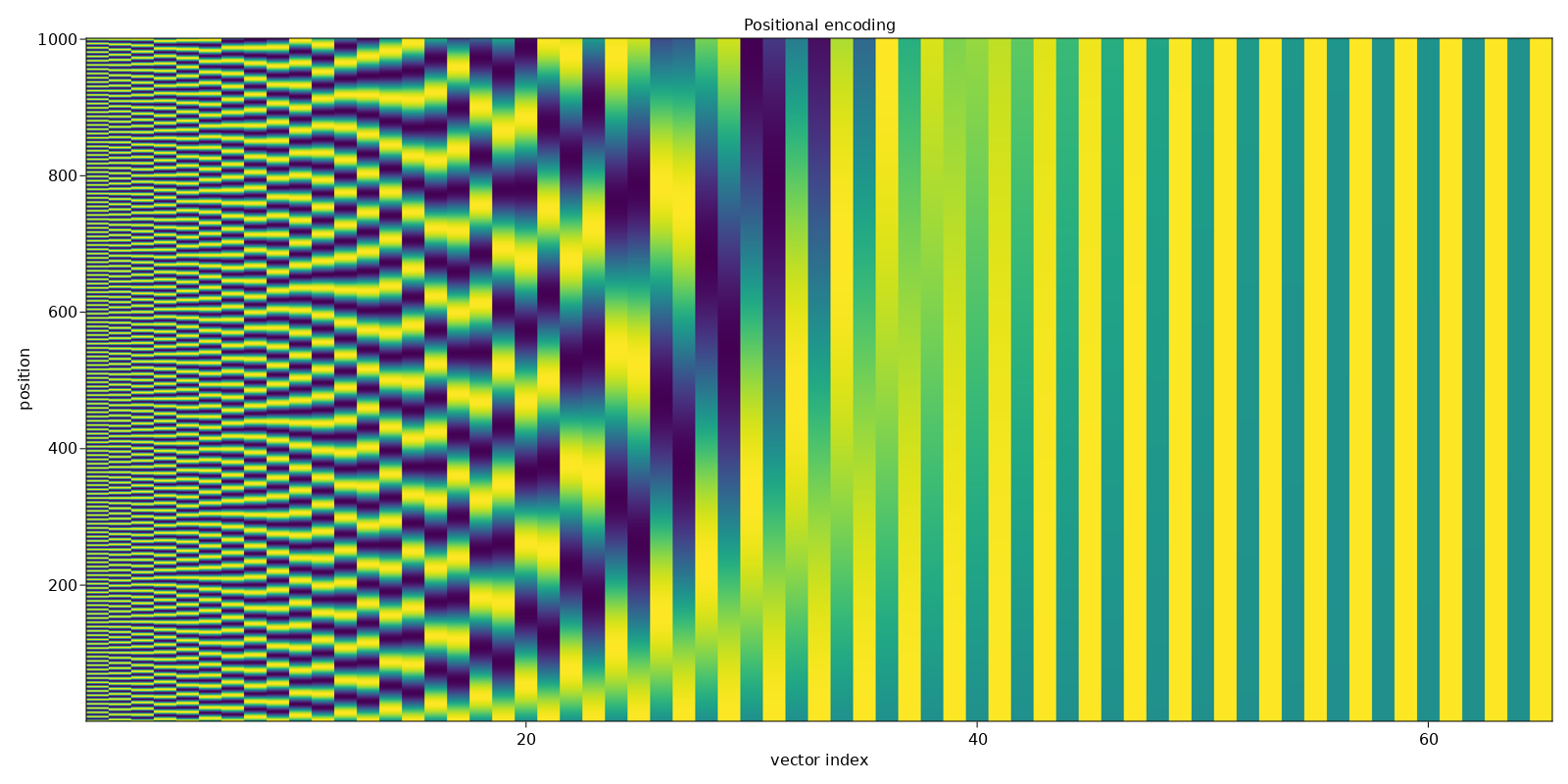}
	\caption{Sinusoidal positional encoding. By Cosmia Nebula - Own work, CC BY-SA 4.0, \url{https://commons.wikimedia.org/w/index.php?curid=119948133}}
	\label{fig:position}
\end{figure}

This results in an interleaved pattern of sine and cosine waves, as illustrated in Figure~\ref{fig:position}.

This position information is injected into the model by simple addition to the embedded sequence. More recent models use learned positional encoding instead of fixed sinusoidal waves~\parencite{learned_position}. This learned encoding is represented as an $S_{max} \times D$ matrix, where $S_{max}$ is the maximum input sequence length.

\textcite{musictransformer} introduced the \textit{music transformer}, where relative position encoding~\parencite{relative_position} is intelligently integrated into the attention mechanism. Unlike absolute positional encoding, relative positions encode the distances between tokens, rather than their absolute positions. In music, rhythm and harmony are defined by how notes interact with one another, not by their fixed time locations. Therefore relative positional encodings are particularly well-suited for music generation, as they reflect the relationships between musical events, rather than their absolute positions in time. 

\paragraph{Normalization}

Normalization of internal activations is crucial for stable training in any neural network. While traditional neural networks typically use batch normalization, which normalizes activations along the batch dimension~\parencite{batchnorm}, transformers use layer normalization, where activations are normalized across the feature dimension~\parencite{layernorm}.

\paragraph{Multi-head attention}

At the core of the transformer is the attention mechanism, which computes a weighted sum of input values based on their relevance. This enables transformers to process all words in a sentence simultaneously and determine the importance of each word relative to the others.

Consider the example of translating a sentence from a source language to a target language. In self-attention, the model focuses on understanding relationships between words within the target sentence. In cross-attention, the model still focuses on the target sentence, but also considers the influence of the source sentence. While the mechanism behind self- and cross-attention is the same, the key difference lies in the input.

In multi-head attention, the word embeddings are first projected into queries, keys, and values using separate projection matrices. In self-attention, queries, keys, and values are derived from the same sentence. In contrast, in cross-attention, the source sequence forms the keys and values, while the target sequence forms the queries.

\begin{equation}
Q = X_Q W_Q, \quad K = X_K W_K, \quad V = X_V W_V
\end{equation}

In the above formulas, $X_Q$, $X_K$, and $X_V$ represent the input word embeddings for query, key, and values, respectively. $W_Q$, $W_K$, and $W_V$ represent the learned projection matrices.

The attention mechanism then calculates the interaction between the keys and queries to assign a weight (importance) to each value.

\begin{equation}
\text{Attention}(Q, K, V) = \text{softmax} \left( \frac{QK^T}{\sqrt{D_k}} \right) V
\end{equation}

The query ($Q$), key ($K$), and value ($V$) matrices determine the attention output. The dot product $QK^T$ computes similarity, while $\sqrt{D_k}$, the square-root fo the dimensionality of the keys, scales the values to prevent large gradients. The softmax function converts the scores into probabilities, which weight the values in $V$.

Attention works by comparing the importance of words to one another through queries ($Q$), keys ($K$), and values ($V$). These are learned representations of the input sequence. The query ($Q$) represents what we are looking for, such as a word that needs context. The key ($K$) serves as the reference against which we compare, such as other words in the sequence. The value ($V$) holds the actual information that is processed, such as the word to which we are assigning importance.

Instead of computing attention once, multi-head attention applies it multiple times in parallel:

\begin{equation}
\text{head}^i = \text{Attention}(W_Q^i, W_K^i, W_V^i)
\end{equation}

\begin{equation}
\text{MultiHead}(Q, K, V) = \text{Concat} ( \text{head}_1, \dots, \text{head}_h ) W_O
\end{equation}

Each attention head ($\text{head}_i$) projects the inputs using different weights. The outputs from all heads are concatenated and linearly transformed by $W_O$. Using multiple heads allows the model to capture different relationships in parallel.

\paragraph{Feed-forward Network}

The feed-forward network, represented by the yellow blocks in Figure \ref{fig:transformer} processes each word individually, without considering interactions between them. It typically consists of two linear layers with a rectified linear unit (ReLU) activation in between~\parencite{relu}. The first linear layer projects the input to a higher dimensionality, while the second linear layer reduces it back to the model's original dimensionality. This expansion in dimensionality enhances the network’s representation capacity.

To stabilize training, residual connections are used throughout the model~\parencite{resnet}. The final linear layer projects the activations back to a dimensionality corresponding to the vocabulary size. The output vector thus represents the probability of each word in the vocabulary.

Having introduced the key concepts, we now turn to a review of related literature on the tasks addressed in this work.

\section{Sequence processing}
\label{sequence_background}
Symbolic music can be represented as sequential data, similar to text. Hence, the same models can, in principle, be used for both natural language processing (NLP) and symbolic music processing. One of the oldest neural network architectures for sequence modeling is the recurrent neural network (RNN), where a single input sample (token) is processed at each timestep~\parencite{rnn}. The network is trained by calculating the gradient of the error across each timestep, using the algorithm named \textit{backpropagation through time}.

However, RNNs aren't very successful in modeling long sequences because, as the sequence grows longer, the backpropagated gradients can approach zero. This problem is known as the \textit{vanishing gradient problem}. Long short-term memory (LSTM) networks alleviate this issue by using specialized gates~\parencite{lstm}. A similar variant, the gated recurrent unit (GRU), can achieve performance similar to LSTM but with a simpler architecture and fewer parameters~\parencite{gru}.

Despite these improvements, even LSTMs and GRUs struggle to process long sequences efficiently and tend to "forget" earlier input samples as the sequence grows. The attention mechanism addresses this issue by explicitly modeling dependencies between \textit{all} pairs of input samples~\parencite{attention}. Finally, the transformer model has set the current state-of-the-art in sequence processing by incorporating the attention mechanism in a multi-headed and multi-layered architecture~\parencite{transformer}.

The original transformer implementation consisted of an encoder and a decoder network, and it was tested on the task of machine translation. Encoder-decoder architectures are commonly used for machine translation, where the encoder processes the source text and the decoder generates the output~\parencite{encoder_decoder_rnn}. Since language modeling involves generating text from scratch, it can be seen as analogous to music generation, meaning both tasks can be categorized under the task of \textit{sequence generation}.

In sequence generation, there are no separate source and target sequences. As a result, state-of-the-art language models consist solely of a decoder~\parencite{gpt3}. Sequence-generating neural networks are trained with input and target sequences from the same domain. Specifically, the target sequence is a shifted version of the input sequence, so for each input token, the network predicts the next token.

During the training of a sequence generation model, the training loss is calculated based on the prediction performance of all tokens, even though the model only considers the previous tokens. To ensure that the model does not attend to future tokens, a triangular mask is added to the softmax output of the attention module. An example of a triangular mask for a sequence length of 5 is shown below.

\[
mask =
\begin{bmatrix}
0 & -\infty & -\infty & -\infty & -\infty \\
0 & 0 & -\infty & -\infty & -\infty \\
0 & 0 & 0 & -\infty & -\infty \\
0 & 0 & 0 & 0 & -\infty \\
0 & 0 & 0 & 0 & 0
\end{bmatrix}
\]

This ensures that the attention outputs related to future tokens are masked. For example, while the model is predicting the 3rd token, it will only use the 1st and 2nd tokens. This generation task can be viewed as a classification task for all tokens. The most commonly used loss function for this task is cross-entropy loss:

\begin{equation}
\mathcal{L} = - \sum_{s=1}^{S} \sum_{v=1}^{V} y_{s,i} \log \hat{y}_{s,i}
\end{equation}

This loss function measures the difference between the predicted token probabilities and the true labels. The index $s$ represents the position in the sequence, ranging from $1$ to $S$, where $S$ is the total number of tokens. The index $i$ corresponds to the vocabulary word indices, ranging from $1$ to $V$, where $V$ is the vocabulary size. The term $y_{s,v}$ is a one-hot encoded label, meaning it is $1$ if the correct word at position $s$ is word $v$, and $0$ otherwise. The term $\hat{y}_{s,v}$ is the predicted probability of word $i$ at position $s$, which is obtained from the softmax output of the transformer model. The logarithm $\log \hat{y}_{s,v}$ ensures that incorrect predictions are heavily penalized, as lower predicted probabilities lead to larger negative values. Since only one word is correct at each position, only one term in the summation contributes to the loss at each timestep. The model optimizes this function by minimizing $\mathcal{L}$, encouraging higher probabilities for the correct words while reducing the likelihood of incorrect predictions.

Sequence generation models operate autoregressively during inference, meaning that each new token is generated one at a time. To generate each token, a forward pass is performed. After generating a token, it is added to the sequence, and this updated sequence is then fed back into the model to predict the next token. For instance, if we have already generated a sequence consisting of tokens 1 and 2, we use this sequence as input to predict token 3. Once token 3 is generated, it is appended to the sequence, forming the sequence [1, 2, 3], which is then used to predict token 4, and so on. Often, a special \texttt{START} token is used as the first token to initiate inference from scratch. Alternatively, inference can be performed using a predefined sequence, known as a \textit{primer} or \textit{priming sequence}. In this case, the model continues from the sequence provided by the user, generating tokens that extend the given sequence.

In contrast to generative tasks, classification tasks predict a single set of probabilities for the entire sequence, rather than for each token. To facilitate this, the input sequence is prepended with a special token called the \texttt{CLS} (classification) token, and only the output corresponding to this token is considered. In classification tasks, no triangular mask is applied, so each token can attend to every other token. As a result, the output corresponding to the \texttt{CLS} token utilizes the entire sequence.

In the following sections, we discuss sequence classification and sequence generation tasks that are relevant to our work. We begin with a review of text emotion classification, followed by conditional sequence generation, and specifically conditional music generation.

\subsection{Text emotion classification}

Emotion classification from text---or sentiment analysis, as used interchangeably in the machine learning literature---allows us to automatically identify and/or quantify the emotion expressed in a piece of text, such as a review, social media post, or customer feedback \parencite{sentiment_survey}. Identifying the underlying emotion in text is useful in various fields such as customer service \parencite{customer}, finance \parencite{finance}, politics \parencite{politics}, and entertainment \parencite{emotionpv}.

We now present the most recent and comprehensive text emotion classification datasets, along with the methods.

\subsubsection{Datasets}

The SemEval (Semantic Evaluation) workshops regularly host a variety of challenges focused on semantic text analysis~\parencite{semeval}. Several of these challenges target text-based emotion classification. Task 11 of the 2025 edition, titled "Bridging the Gap in Text-Based Emotion Detection", compiles samples from diverse sources including social media, speeches, literature, and news~\parencite{semeval25}. The dataset includes over 100,000 samples across more than 30 languages, all manually labeled with Ekman’s six basic emotions—joy, sadness, anger, fear, surprise, and disgust—along with a neutral category~\parencite{ekman}. Of these, 5,651 samples are in English. In the 2019 edition, Task 3—"EmoContext: Contextual Emotion Detection in Text"—challenges participants to determine the emotion of the final utterance in a short dialogue~\parencite{semeval19}. This dataset contains 38,424 samples labeled with four categories: happy, sad, angry, and others. Task 1 of the 2018 edition, titled "Affect in Tweets", focuses on emotional content in Twitter data\footnote{\url{https://twitter.com}}~\parencite{semeval18}. It includes multiple subtasks: emotion intensity regression, ordinal classification of intensity, valence regression, valence ordinal classification, and emotion classification. The challenge supports three languages---English, Spanish, and Arabic---with 10,983 English samples specifically available for the emotion classification task.

The EmoBank dataset consists of 10,062 English sentences collected from sources such as news articles, blogs, fiction, and travel guides. These samples are annotated through crowdsourcing using a dimensional emotion representation scheme \parencite{emobank}. The Multimodal EmotionLines Dataset (MELD) is designed for emotion recognition in multi-party conversations \parencite{meld}. It includes text, video, and audio data. The text component features 13,000 English utterances labeled with Ekman's six basic emotions—joy, sadness, anger, fear, surprise, and disgust—along with a neutral category \parencite{ekman}. The Cleaned Balanced Emotional Tweets (CBET) dataset provides 81,163 tweets annotated with nine emotion labels: joy, anger, sorrow, love, contempt, surprise, fear, guilt, and thankfulness \parencite{cbet}. The GoEmotions dataset contains 58,000 English text samples sourced from Reddit\footnote{\url{https://www.reddit.com}} comments \parencite{goemotions}. It uses a broad range of 27 categorical emotion labels, including admiration, amusement, approval, annoyance, anger, curiosity, confusion, caring, desire, disgust, disapproval, disappointment, embarrassment, excitement, fear, joy, grief, gratitude, love, sadness, nervousness, optimism, pride, remorse, relief, realization, and surprise.

Table \ref{table:text_emotion} provides a summary of the text emotion datasets mentioned above. It outlines the source or context of each dataset, the number of English-language samples, and the type of emotion representation used.

\begin{table}[h]
	\caption{Text emotion classification datasets}
	\label{table:text_emotion}
	\centering
	\begin{tabular}{c|ccc}
		Dataset & Context & Sample size & Emotion representation \\ \hline \hline
		
		\makecell{CBET\\\parencite{cbet}} & Tweets & 81,163 & Categorical \\ \hline
		
		\makecell{EmoBank\\\parencite{emobank}} & \makecell{News, blogs,\\fiction, travel guides}  & 10,062 & Dimensional \\ \hline
	
		\makecell{SemEval-2018\\\parencite{semeval18}} & Tweets & 10,983 & Categorical \\ \hline
		
		\makecell{SemEval-2019\\\parencite{semeval19}} & Dialogues & 38,424 & Categorical \\ \hline
		
		\makecell{MELD\\\parencite{meld}} & Dialogues & 13,000 & Categorical \\ \hline
		
		\makecell{GoEmotions\\\parencite{goemotions}} & Reddit comments & 58,000 & Categorical \\ \hline
		
		\makecell{SemEval-2025\\\parencite{semeval25}} & \makecell{Social media, speeches,\\literature, news}  & 5651 & Categorical 

	\end{tabular}
\end{table}

\subsubsection{Methods}

Machine learning methods have significantly advanced the state of the art in text emotion classification for the past two decades. However, the earliest works in this field relied on hand-crafted features, such as frequently used n-grams \parencite{ngrams}, or adjectives and adverbs that are associated with particular emotions \parencite{appraisal}. Nonetheless, the advent of deep learning has made it computationally feasible to process raw inputs without extracting features manually, leading to better performance \parencite{alexnet}. Recurrent Neural Networks and their improved variants such as Long Short-Term Memory were initially used \parencite{rnn} but were later replaced by the transformer model \parencite{transformer}, which is the current state of the art in text classification tasks. The 2025 SemEval task on text emotion classification, titled "Bridging the Gap in Text-Based Emotion Detection", highlights that current state-of-the-art approaches rely on pretrained transformers with fine-tuning, or large language models (LLMs) enhanced through instruction tuning, adapter modules, or data augmentation techniques~\parencite{semeval25}.

Fine-tuning pretrained transformers for specific tasks leverages the knowledge acquired during their initial training phase. The GPT (Generative Pretraining Transformer) model, for instance, is a large transformer pretrained on next-token prediction and later fine-tuned for various NLP tasks, achieving state-of-the-art performance~\parencite{gpt}. BERT (Bidirectional Encoder Representations from Transformers) enhanced this approach by introducing masked token prediction during pretraining, which allowed the model to learn bidirectional context~\parencite{bert}. Building on BERT, the RoBERTa (Robustly Optimized BERT Approach) model further improved performance by employing extensive hyperparameter tuning, training on more data, using larger batch sizes, and extending the pretraining duration~\parencite{roberta}.

Instruction-tuning involves the supervised fine-tuning of large language models (LLMs) using instructions, typically in the form of prompt-output pairs~\parencite{instruction_tuning}. Since fine-tuning an entire LLM can be resource-intensive, many approaches reduce the number of trainable parameters by using adapters~\parencite{adapters}. Adapters are small modules inserted within a neural network, enabling fine-tuning of the model while keeping the large network’s weights frozen. This allows the adapter modules to be trained while maintaining computational efficiency. Advanced adapters, like Low-Rank Adapters (LoRA), use low-rank matrix decomposition to represent weight updates with fewer parameters than the original weight matrix~\parencite{lora}. \textcite{uob} utilized adapters for cross-lingual text emotion classification, leveraging datasets from multiple languages. \textcite{lotus} employed LLMs for data augmentation in fine-tuning pretrained transformers. They used the LLaMa-3 model\footnote{\url{https://llama.com}} to generate explanations for text in their training dataset, enhancing the performance of a pretrained RoBERTa model on text emotion classification tasks.

We continue by providing background on sequence processing, focusing on conditional sequence generation.

\subsection{Conditional generic sequence generation}

Although it is a loosely used term, \textit{conditioning} refers to controlling a model's output by providing auxiliary inputs, i.e., conditions. These conditions can belong to the same domain as the input and target, enabling training with unlabeled data. Alternatively, conditions can belong to different domains, such as the labels of a labeled dataset. 

\sloppypar Even the earliest neural networks for natural language processing utilized conditioning. \textcite{conditionalLM} developed a language model conditioned on factors such as topic and genre, where a conditioning vector was created using a linear layer and concatenated with the hidden state of the recurrent neural network (RNN). \textcite{conditional_translator} developed an encoder-decoder RNN model for translation from English to German, conditioned on politeness. They used control tokens to specify whether the user preferred a formal or informal translation.

The Conditional Transformer Language (CTRL) model feeds control tokens that denote domain, style, topics, and more into a large transformer, achieving state-of-the-art results in conditional language modeling~\parencite{ctrl}. \textcite{paraphrase} performed style transfer by generating paraphrases and demonstrated that training separate models for each style outperforms training a single model that uses style-specific control tokens.

\textcite{bias} identified triggers—subsequences that generate biased text when used as inputs—and employed them as primers to induce or balance bias in language modeling. \textcite{dialogue} investigated controlling the style of dialogue generation by comparing three methods: retrieve-and-refine~\parencite{retrieve}, inference-time iterative refinement \parencite{iterative}, and conditional generation using control tokens~\parencite{ctrl}. They showed that conditional generation with control tokens outperformed the other methods.

Most works in the literature use categorical variables, such as control tokens, to guide language modeling. In contrast, image captioning can be framed as a text generation task based on images, which are non-categorical variables. In this context, the input image is typically processed by a convolutional neural network (CNN), and the resulting features are used to condition a separate language model. Earlier works used recurrent neural networks (RNNs) as the language model for this task~\parencite{image_caption_rnn}, but state-of-the-art models have replaced RNNs with transformers~\parencite{image_caption_transformer}. \textcite{image_caption_transformer} compared different conditioning methods and observed similar performance across them. These methods include feeding the spatial image features into the cross-attention layer of the decoder~\parencite{image_caption_attention}, combining the image features with each word embedding, and feeding the image features before the word embeddings~\parencite{image_caption_rnn}.

Next, we move on to the task most closely related to our overall work: conditional music generation.

\subsection{Conditional symbolic music generation}

While we maintain our focus on the conditional generation of symbolic music, it is essential to understand non-conditional generation of symbolic music as well. We proceed by presenting the relevant datasets and methods.

\subsubsection{Datasets}
\label{sec:midi_datasets}

The largest symbolic music dataset is the Lakh MIDI Dataset (LMD)~\parencite{lmd}, which includes 176,581 unlabeled, multi-instrument MIDI files. Of these, 45,129 files are matched to 31,034 tracks in the Million Song Dataset (MSD)~\parencite{msd}. The mismatch in numbers arises because multiple MIDI files can correspond to the same MSD track, and a single MIDI file can also map to multiple tracks.

The MAESTRO dataset (MIDI and Audio Edited for Synchronous TRacks and Organization) is a large, high-quality dataset designed for music research, especially in areas like transcription, generation, and synthesis~\parencite{maestro}. It includes over 200 hours of professional piano performances, with precisely aligned audio and MIDI recordings. The MIDI data captures not just the notes and timing but also note velocities, which represent the intensity or dynamics of each keystroke---crucial for modeling expressive piano performance.

\textcite{zhuo} introduced the Symbolic Music Videos (SymMV) dataset, which contains video-MIDI pairs. They sourced MIDI data from piano tutorial videos, automatically transcribing audio using the Onsets and Frames model~\parencite{onset_and_frames}, resulting in 1,140 samples. They also provide low-level features, such as color histograms and RGB frame differences for motion, as well as high-level CLIP features extracted from the videos.

To the best of our knowledge, there are only four publicly available symbolic music datasets with emotion labels, though their sample sizes are quite small. The VGMIDI (Video Game MIDI) dataset consists of $204$ video game soundtracks played on piano, with continuous-valued labels for valence and arousal annotated by 30 human subjects~\parencite{vgmidi}. 
\textcite{mirexlike} created the MIREX-like dataset with discrete emotion labels sourced from the AllMusic database\footnote{https://allmusic.com}. The dataset primarily consists of audio recordings, but 193 of the samples are also available as MIDI files, featuring a varying number of instruments.
The EMOPIA (EMOtion PIAno) dataset consists of paired audio and MIDI clips featuring solo piano, extracted from 387 songs~\parencite{emopia}. Emotional labels were manually annotated by the authors using discrete categories based on the four quadrants of the two-dimensional circumplex model of affect~\parencite{valence_arousal}. Table~\ref{fig:midi_datasets} presents the MIDI datasets, the total durations in hours, instrumentations and the types of emotion labels.

Unfortunately, due to their small sizes, these datasets are insufficient for training deep neural networks with millions of parameters, as they tend to cause overfitting and poor generalization. Overall, there are no large-scale, multi-instrument MIDI datasets annotated with emotion labels.

\begin{table}[h]
	\caption{MIDI datasets}
	\label{table:data comparison}
	\centering
	\begin{tabular}{c|ccc}
		Dataset & Duration (hours) & Instrumentation & Emotion label \\ \hline \hline
		\makecell{MIREX-like\\\parencite{mirexlike}} & 11.6 & Multi-instrument & Categorical  \\ \hline
		\makecell{LMD\\\parencite{lmd}} & 9382.1 & Multi-instrument & None \\ \hline
		\makecell{MAESTRO\\\parencite{maestro}} & 200.0 & Piano-only & None \\ \hline
		\makecell{VGMIDI\\\parencite{vgmidi}} & 6.4  & Piano-only  & Dimensional  \\ \hline
		
		\makecell{EMOPIA\\\parencite{emopia}} & 11.0  & Piano-only & Categorical   \\ \hline
		\makecell{SymMV\\\parencite{zhuo}} & 76.5  & Piano-only & None 
	\end{tabular}
	\label{fig:midi_datasets}
\end{table}

\subsubsection{Methods}

Early works using neural networks for music generation employed recurrent neural networks (RNNs) \parencite{blues_lstm}. However, the recent advent of the transformer model has enabled the modeling of much longer dependencies. The music transformer builds upon the transformer model by incorporating relative positional information, providing a better representation of the relationships between notes~\parencite{musictransformer}.

The majority of existing literature on symbolic music generation relies on a non-conditional approach. These methods are trained on raw MIDI data without explicit labels, enabling the generation of new music similar to the examples in the training dataset~\parencite{musictransformer}. However, some approaches leverage low-level features within the data to generate music in a conditional manner~\parencite{coconet}. For example, they may use short melodies, chords, or single-instrument tracks as input to generate corresponding melodies. Using short melodies as input is particularly practical because they are directly fed into the model, requiring no changes to the architecture. In sequence processing, these short input sequences are referred to as \textit{primers}. The model then predicts the melody that follows the primer melody. In this case, both the condition and the target belong to the same domain, allowing the model to be trained using unlabeled data.

Other symbolic music generation tasks that utilize same-domain conditioning include accompaniment generation~\parencite{coconet,lpd}, interpolation~\parencite{musicvae}, inpainting~\parencite{sketchnet, music_machine}, and style transfer~\parencite{musical_style, wang}. MidiNet can generate melodies conditioned on chords by training on a private dataset that includes chord information~\parencite{midinet}. OpenAI's MuseNet is trained on a combination of datasets, and generation can be conditioned on specific artist names, genres, or styles using primer control tokens~\parencite{musenet}.

It is also possible to use low-level symbolic music features for conditioning~\parencite{tonal_tension, pati, yang}. Features such as tempo, note density, pitch range, and tonal tension can be automatically calculated, eliminating the need for a labeled dataset. \textcite{fadernets} aimed to compensate for the small size of the labeled VGMIDI dataset by augmenting it with the unlabeled MAESTRO (MIDI and Audio Edited for Synchronous TRacks and Organization) dataset~\parencite{maestro}---detailed in the next section. They used low-level rhythm and note density features to infer the high-level arousal feature. Similarly, \textcite{mmtbert} incorporated chord information using additional tokens and further enhanced output quality through the use of Generative Adversarial Networks~\parencite{gan}.

\textbf{Emotion-based music generation}, or equivalently Affective Algorithmic Composition (AAC) methods focus on the automatic composition of music based on specific emotions \parencite{aac}. Use cases for AAC include composing soundtracks for videos and video games \parencite{game_soundtrack}, neurofeedback training for medical applications \parencite{medicinal}, and developing brain-computer music interfaces \parencite{bci}.

Although the relationship between music and emotion is well-studied~\parencite{music_and_emotion}, choosing the "optimal" emotion model to explore this relationship remains a debated topic~\parencite{emotion_models_for_music}. Existing AAC work primarily employs the two main categories of emotion models: categorical and dimensional~\parencite{aac}.

As discussed in Section~\ref{sec:categorical_emotion}, categorical emotion models use discrete labels such as happy, sad, angry, and surprised~\parencite{discrete_emotions}. However, studies on dimensional approaches argue that categorical models fail to capture the complexities and subtleties of human emotions. Instead, they propose using continuous-valued coordinates to represent emotions in a low-dimensional space~\parencite{emotion_survey_continuous}.

Early works on AAC used various melodic, harmonic, and rhythmic features to target specific emotions (see the overview of~\textcite{aac}). However, this is an indirect approach, and the correspondence between emotions and musical features is only approximate~\parencite{music_emotion_features, music_emotion_features2}. The advent of deep learning allowed for the use of complex models trained on large labeled datasets, utilizing the emotion labels directly and eliminating the need for intermediate features~\parencite{alexnet}. 

More recent AAC models are trained in an end-to-end fashion on datasets containing symbolic music and emotion labels. The creators of the VGMIDI dataset developed a method for symbolic music generation conditioned on emotion~\parencite{vgmidi}. Using a genetic algorithm, they fine-tuned the weights of a pretrained LSTM, separately for positive and negative valence conditions, resulting in two models.

Both \textcite{emo_lstm} and \textcite{emopia} generated symbolic music conditioned on four categorical emotions, each corresponding to one of the four quadrants of the valence-arousal plane. \textcite{emo_lstm} labeled the classical piano music obtained from piano-midi.de, using categorical emotion labels and trained a biaxial LSTM~\parencite{balstm} on this labeled dataset. \textcite{emopia}, the creators of the EMOPIA dataset, trained a transformer model conditioned with control tokens~\parencite{ctrl}. However, these works were limited by the small number of categorical emotion labels they could use, likely due to the small sizes of their training datasets.

We reserve Section~\ref{sec:intro_video_based_music} to discuss \textit{video-based symbolic music generation}, as this task requires more complex models with video analysis modules. We now proceed by discussing the task of video analysis.

\section{Video analysis}\label{sec:video-processing}

Video-based music generation models typically consist of a video analysis module followed by a music generator~\parencite{di,kang}. As explained in Chapter \ref{chap:intro} our work involves both low-level video analysis, specifically scene cut detection, and high-level video analysis, namely semantic video classification. 

Scene cut detection is built into standard video processing libraries, such as FFMpeg\footnote{https://ffmpeg.com}. FFmpeg's algorithm computes the average pixel difference between consecutive frames and marks a scene cut when this difference exceeds a specified threshold. As scene cut detection is a solved problem, we center our discussion on semantic video classification, with a particular focus on emotion and genre classification.

Unlike the classification of other signals, video classification is a \textit{multimodal} task. Video classification models utilize both video frames—capturing scenes, objects, actions, faces, and text—and audio, which includes speech, music, effects, and audio events. Additionally, since videos are sequences, both short- and long-term dependencies must be considered. Furthermore, the nature of the video content significantly influences which modalities are most relevant. For instance, when classifying emotions based on facial expressions, visual information from video frames is essential. In contrast, for detecting emotions in social interactions, analyzing speech becomes equally or even more important.

\subsection{Video emotion classification}

Video emotion classification is the task of automatically detecting and categorizing emotions expressed in videos using computational methods. Video classification models take video frames as input, and often include audio as well, producing output in the form of probabilities for predefined emotion categories or continuous values such as valence and arousal. While our work focuses on the classification of arbitrary videos, we also list the following video emotion datasets that include both arbitrary and specific video types.

\subsubsection{Datasets}

\textcite{social_video_emotion} present the Pairwise Emotional Relationship Recognition
(PERR) dataset which includes 31,182 videos from TV drama series, focusing on social interactions between humans. The Context Aware Emotion Recognition (CAER) dataset similarly collects 13,201 clips from 79 TV series, but it specifically selects clips containing human faces~\parencite{caer}. \textcite{musicvideodataset} label the emotions in 3,438 music videos, creating the Music Video Emotion Dataset (MVED).

There are limited datasets for the classification of arbitrary videos, i.e., those that include a diverse set of videos. The VideoEmotion-8 dataset consists of 1,001 user-generated videos collected from YouTube and Flickr, labeled with 8 categorical emotions: anger, anticipation, disgust, fear, joy, sadness, surprise, and trust~\parencite{videoemotion8}. The Ekman-6 dataset~\parencite{ekman6} is an extended version of VideoEmotion-8, containing 1,637 videos labeled with the six basic emotions defined by \textcite{ekman}: anger, disgust, fear, joy, sadness, and surprise. Table~\ref{table:video_emotion} summarizes the video emotion datasets, all of which use categorical emotion representations.

\begin{table}[h]
	\caption{Video emotion datasets.}
	\label{table:video_emotion}
	\centering
	\begin{tabular}{c|ccc}
		
		Dataset & Context & Sample size  \\ \hline 
		
		VideoEmotion-8 \parencite{videoemotion8} & User-generated & 1,001 \\ 

		Ekman-6 \parencite{ekman6} & User-generated & 1,637 \\ 
		
		CAER \parencite{caer} & Human faces  & 13,201 \\ 

		PERR \parencite{social_video_emotion}& Social interactions & 31,182 \\ 

		MVED \parencite{musicvideodataset} & Music videos & 3,438  \\  

	\end{tabular}
\end{table}

We now proceed to discuss methods for video emotion classification.

\subsubsection{Methods}

All modern video classification models rely on deep neural networks. \textcite{ekman6} employed cross-domain knowledge transfer, particularly from image emotion classification to video emotion classification. \textcite{cfn} first classify scenes, objects, and events in the video and then fuse this information to classify the emotion. \textcite{vaanet} used 3D convolutional neural networks to process video frames and 2D convolutional neural networks to process audio spectrograms. They also used modified loss functions that apply a higher penalty for incorrect classifications when the prediction and the target share similar emotional polarities.

\textcite{cten} introduced inter- and intra-attention modules to analyze dependencies within and between video frames and audio. Based on the softmax outputs from the attention modules, they identify dominant features and then erase them, encouraging the model to focus on secondary information.

Due to the long length of video frame sequences, many methods extract and operate on \textit{keyframes}. A straightforward way to extract keyframes is by identifying scene boundaries, but \textcite{keyframe} further filter these based on their emotional saliency. They calculate emotional saliency by feeding frames into an image emotion classifier and using the probability of the highest class. Essentially, they select video frames that produce more pronounced emotion classification results, and use those frames in their final analysis. \textcite{lrcanet} applied data augmentation techniques for video emotion classification by blending frames through linear interpolation or cutting and pasting different sections. A majority of the aforementioned methods rely on data-oriented approaches, such as data selection (e.g., identifying key frames) or data augmentation (e.g., generating synthetic frames).

We now continue our discussion of semantic video classification by turning to an alternative task: cinematic trailer genre classification. Thanks to online film databases like IMDb and video platforms such as YouTube, cinematic trailer datasets tend to offer larger sample sizes and longer video durations. This abundance of data allows researchers to shift focus from data-oriented strategies to model-oriented approaches.

\subsection{Trailer genre classification}

In trailer genre classification, models process cinematic trailers to predict the movie's genres. This is usually a multi-label classification task since movies are usually associated with multiple genres. We continue our discussion with cinematic datasets and trailer genre classification methods.

\subsubsection{Datasets}

MovieLens is one of the earliest cinematic datasets, containing movie ratings and metadata such as genre and title~\parencite{movielens}. While the latest version includes 86k movies, it does not provide direct links to the movie trailers. These trailers can only be obtained by crawling the Internet Movie Database (IMDb) website\footnote{https://www.imdb.com}, which conflicts with their terms of use.

The LMTD (Large Movie Trailer Dataset) is a collection of features and metadata from 3,500 movie trailers, although the project is currently discontinued and the data is no longer available~\parencite{lmtd}. The MM-IMDb (Multimodal IMDb) dataset merges movie posters with metadata from MovieLens, resulting in 26k movies with posters, plots, genres, and other metadata, but without trailers or videos~\parencite{mmimdb}.

The MMTF-14K (Multimodal Movie Trailer Features dataset) provides multimodal features extracted from 14k movie trailers along with metadata such as user reviews and genre~\parencite{mmtf}. MovieScope is a comprehensive dataset for multi-modal movie analysis, including data such as movie trailers, posters, plot synopses, user reviews, and visual-auditory features, covering 5k distinct movies~\parencite{moviescope}. 

The Condensed Movies dataset includes full individual scenes, rather than trailers, along with plot details and character information, from 4k movies~\parencite{condensed}. MovieNet is a large-scale, holistic dataset providing movie, trailer, poster, subtitle, plot, tags, and genre metadata~\parencite{movienet}. While it contains metadata for 375k movies, trailers are available via YouTube links for 33k movies, making MovieNet the largest cinematic dataset in terms of labeled trailers.

In the context of trailer genre classification, earlier datasets either do not include trailers or require web scraping, which may conflict with the terms of use of online platforms. More recent efforts provide YouTube links to trailers associated with the movies in their datasets, but these are still limited in size. Among available options, MovieNet stands out as the largest dataset offering direct access to movie trailers, making it the most suitable candidate for training DNNs for trailer genre classification.

\subsubsection{Methods}

One of the earliest works on movie genre classification used scene detectors to obtain keyframes and extracted hand-crafted visual features, such as roughness, ruggedness, and openness, from a privately collected dataset~\parencite{zhou}. Genre classification was achieved by comparing the distance between the feature vectors from the training and testing sets.

One of the first works to use neural networks for movie genre classification also leveraged visual pretrained features~\parencite{wehrmann}, using the LMTD dataset~\parencite{lmtd}. They utilized pretrained features from classification models trained on the ImageNet~\parencite{imagenet} and Places365~\parencite{places} datasets, along with audio spectrograms. The features from individual frames were fused using a convolution-through-time module, which operates as a standard convolutional neural network (CNN) where the kernel length matches the input feature length of each keyframe. The kernel traverses features from subsequent frames along the temporal dimension~\parencite{wehrmann}. The kernel length along the temporal dimension is 3, meaning the model exploits the correspondence between only 3 frames. Additionally, since the input and output of the CNN have varying numbers of frames, the model takes the maximum values along the temporal dimension to create a fixed-size vector for the final classification layer. This approach inevitably leads to a loss of information. Another work explored the correspondence between facial emotions and cinematic genres by first extracting human faces from trailer videos, classifying their emotions, and then mapping these emotions to cinematic genres~\parencite{yadav}.

The work presenting the MovieNet dataset also introduces a model for movie genre classification, along with results from other state-of-the-art video classification models~\parencite{movienet}. While the model they introduce surpasses the performance of existing models, it only uses 8 clips, each containing 3 frames from the entire trailer. During inference, predictions are made for each individual clip, and these are averaged to generate the final prediction.  The MovieCLIP model first trains a scene classification model and then uses its final activations as input to a genre classification model, specifically using the Moviescope dataset~\parencite{movieclip}.
        
Some recent works have used transformers for movie genre classification. \textcite{trailer_metadata_classification} used transformers to individually process pretrained frame features, then fused the resulting activations with metadata and movie posters. \textcite{trailer_poster_classification_2} and \textcite{trailer_poster_classification} used pretrained transformers to process movie posters and plots, excluding any use of video data, for genre classification on the MM-IMDb dataset~\parencite{mmimdb}.

All of the aforementioned methods place limitations on input data—either by restricting the number of input frames to a fixed, small set or by relying on pretrained frame features derived from standard image classification models such as VGG (Visual Geometry Group)~\parencite{vgg}. The former fails to capture the long-term dependencies present in trailers, while the latter limits the model to features extracted from a single context, overlooking other informative cues such as human faces or on-screen text. These limitations highlight the need for improved methods that leverage a large and variable number of frames and incorporate pretrained features from a diverse range of tasks.

        
Next, we discuss works that combine video analysis with symbolic music generation, resulting in video-based symbolic music generators.

\section{Video-based symbolic music generation}
\label{sec:intro_video_based_music}

Video-based symbolic music generators take an input video and compose a suitable soundtrack in symbolic format. Several studies focus on generating symbolic music for specific types of videos, such as those featuring human movements like dancing or instrumental performances. The Foley Music model~\parencite{foley} generates MIDI from videos of musicians by processing body keypoint movements using a Graph Convolutional Network~\parencite{gcn} and a Transformer~\parencite{transformer}. Similarly, \textcite{sighttosound} and \textcite{audeo} use deep neural networks with residual connections to generate symbolic music from videos of finger movements on piano keyboards.

Due to their specialized nature, these approaches rely on datasets containing video-MIDI pairs, though these datasets typically contain fewer than 1k samples~\parencite{audeo,video_midi_dataset,music_dataset,sighttosound}. The RhythmicNet model employs a multi-stage process to generate music from dance videos by predicting beats and style, generating a drum track, and subsequently creating multitrack music~\parencite{rhythmicnet}. We now present works that address the more general task of generating symbolic music for arbitrary videos.

The Controllable Music Transformer (CMT) generates music based on video features such as motion speed, motion saliency, and timing~\parencite{di}. Our preliminary explorations indicate that these features are temporally dense and continuously influence music generation, which leads to an inconsistent tempo. CMT employs the Lakh Pianoroll Dataset~\parencite{lpd} and processes music using an extended compound-word representation~\parencite{compound_words}. In this representation, each token encodes type, beat/bar marking, note strength, note density, instrument, pitch, and duration. While this reduces the sequence length, it significantly increases the input dimensionality. Moreover, note events in CMT are strictly aligned with beat subdivisions, whereas human musicians often introduce expressive timing deviations---an essential element for conveying emotion in musical performance. In a follow-up work to CMT, \textcite{zhuo} trained three generation models sequentially: one for generating chord sequences, another for generating the melody, and a third for generating the accompaniment.

The Video2Music model utilizes both low-level video features and high-level emotional conditioning~\parencite{kang}. 
The authors compiled the MuVi-Sync dataset, which consists of 748 music videos labeled with musical attributes such as note density, loudness, chords, and key---but does not include symbolic music data.
Their encoder-decoder transformer takes low- and high-level video features, along with a user-provided primer chord and key, to generate chord sequences. These sequences are then arpeggiated using fixed patterns to create the final MIDI output. However, the model's reliance on a fixed time grid eliminates the subtle expressive timing found in human performances. Additionally, its use of fixed arpeggiation patterns limits musical diversity, and the requirement for user-defined chord and key inputs restricts accessibility for non-musicians.

Evaluation of video-based music generators is non-trivial. For objective evaluation, most works measure the difference between generated and ground-truth samples using metrics such as Pitch Class Histogram Entropy, Grooving Pattern Similarity, and Structureness Indicator~\parencite{metrics_music}, and Number of Statistically-Different Bins~\parencite{ndb}, or functions like contrastive loss~\parencite{contrastive}. When the dataset contains paired video and MIDI samples, this method is straightforward~\parencite{foley,zhuo}. For unpaired video and MIDI samples, objectively evaluating a model that generates MIDI from arbitrary videos is not feasible. 
\textcite{rhythmicnet} evaluate each module in their multi-stage architecture using ground-truth data from the training datasets corresponding to individual stages—namely, beat and style prediction, drum track creation, and ultimately, music generation.

The authors of the Controllable Music Transformer, using unpaired datasets, evaluate output MIDIs generated unconditionally, without input video~\parencite{di}. 
They empirically show that unconditionally generated MIDI samples exhibit metrics closer to the unpaired MIDI dataset compared to video-conditioned output samples. This occurs because constraining the model with video structure causes deviations from intrinsic MIDI structures, and the best way to match a specific MIDI dataset’s metrics is to train exclusively on that dataset.

Overall, existing approaches to video-based music generation exhibit several limitations. Some models are designed for narrow contexts, such as dance or instrument performance videos, limiting their generalizability. Others generate only piano music, restricting the expressive and instrumental range. Methods that rely on continuous temporal conditioning often disrupt the natural sense of tempo, while nearly all approaches use a fixed and coarse time grid, failing to capture the nuanced timing of human musical performance.

Evaluation practices also have notable shortcomings. Some studies assess only individual components of multi-stage models without evaluating the full pipeline. Others evaluate MIDI generation in isolation, ignoring its alignment with the video content. Subjective evaluation through user studies remains the most common practical approach for assessing video-based MIDI generation~\parencite{di,kang,zhuo,foley,rhythmicnet}.

We continue discussing audio generation and its challenges, specifically focusing on the task of audio bandwidth extension.

\section{Audio bandwidth extension}

Modern recording techniques provide music signals with extremely high audio quality. By contrast, the listening experience of archive recordings, such as jazz, pop, folk, and blues recorded before the 1960s is arguably limited by the recording techniques of the time as well as the degradation of physical media. Furthermore, modern recordings can also suffer from diminished audio quality, due to the use of lossy compression, downsampling, packet loss, or clipping. In the broadest sense, audio enhancement aims to restore a degraded signal to improve its sound quality \parencite{restoration}. As such, audio enhancement may target the removal of noise, the suppression of cracks or pops (e.g. from an old vinyl record), signal completion to fill in gaps (so-called "audio inpainting" \parencite{audioinpainting,perraudin2018inpainting}), or the extension of the bandwidth from a band-limited signal.

To transmit audio signals through internet streams, or for the ease of storing, common operations such as compression, bandwidth reduction, and low-pass filtering all result in the removal of at least part of the high-frequency audio content. While this process can be understood as a relatively straightforward mapping from a \textit{full-bandwidth}, or \textit{wideband} signal to a \textit{band-limited} or \textit{narrowband} signal, the corresponding inverse problem, namely \textit{bandwidth extension}, seeks to reconstructing missing high-frequency content and is thus non-trivial. Nevertheless, bandwidth extension is crucial for increasing the fidelity of audio, especially for speech and music signals. 

We move on to discuss the datasets and the methods employed in audio bandwidth extension.

\subsection{Datasets}

Audio bandwidth extension models take a band-limited audio input signal and create their full-band counterparts. Band-limited audio refers to audio signals whose frequency content is restricted to a certain range, typically lower than the full audible spectrum (20 Hz - 20 kHz). Since signals are converted from analog to digital using sampling, the highest representable frequency is directly related to the digital signal's sampling rate. The Nyquist theorem statest that the minimum sampling rate required to accurately reconstruct an analog signal is twice the highest frequency component present. Therefore, in audio processing, full-band signals are commonly sampled at 44.1 or 48 kHz.

To study the task of audio bandwidth extension, existing models typically use full-band (44.1+ kHz) audio as ground truth, with their band-limited, i.e., low-pass filtered counterparts serving as input. Several music source separation datasets include full-band audio and are well-suited for this task. Although these datasets provide isolated stems for individual instruments, the availability of full-band mixtures makes them particularly useful for audio bandwidth extension research. The DSD100 (Demixing Secrets Dataset) dataset contains 100 music tracks spanning various styles, along with isolated stems for drums, bass, vocals, and other sources~\parencite{dsd100}. Similarly, the MUSDB18 dataset includes 150 full-band tracks from diverse genres, each with stems for drums, bass, vocals, and other instruments~\parencite{musdb18hq}. The MedleyDB dataset offers 254 full-band songs and their individual tracks, although the stems are categorized with finer and more variable labels, such as audience noise from live performances or individual drum components like kick drum~\parencite{medleydb}.


The Open Dataset of Audio Quality (ODAQ) features 240 audio samples that are sampled at 44.1 or 48 kHz \parencite{odaq}. It features streaming audio provided by Netflix\footnote{\url{https://netflix.com}} and Fraunhofer IIS\footnote{\url{https://iis.fraunhofer.de}}. This dataset also includes quality-degraded versions of each sample, aimed at audio enhancement research. The ICASSP (International Conference on Acoustics, Speech, and Signal Processing) 2023 Acoustic Echo Cancellation (AEC) Challenge introduces a full-band speech dataset with 10,000 samples and versions with synthetic echo effect added \parencite{aec}. The Synthetic Polyphonic Ambient Sound Source (SPASS) Dataset presents 25,000 audio samples with a sampling rate of 44.1 kHz. This dataset mixes short audio events with longer environmental sounds, and is aimed at audio event detection~\parencite{spass}.

Table~\ref{table:audio} summarizes the full-band audio datasets, their intended tasks, and sample sizes.

\begin{table}[h]
	\caption{Full-band audio datasets}
	\label{table:audio}
	\centering
	\begin{tabular}{c|ccc}
		
		Dataset & Task & Sample size  \\ \hline 
		
		MedleyDB \parencite{medleydb} & Music source separation & 254 \\
		
		DSD100 \parencite{dsd100} & Music source separation & 100 \\ 
		
		MUSDB18 \parencite{musdb18} & Music source separation  & 150 \\ 

		SPASS \parencite{spass}& Audio event detection & 25,000 \\ 
		
		AEC \parencite{aec} & Acoustic echo cancellation & 10,000 \\
		
		ODAQ \parencite{odaq} & Streaming audio enhancement & 240\\ 

	\end{tabular}
\end{table}

\subsection{Methods}

The first applications of audio bandwidth extension addressed speech signals only, due to the practical problems arising from the low bandwidth of the telephone systems. One of the earliest works uses a statistical approach in which narrowband and wideband spectral envelopes which are assumed to be generated by a mixture of narrowband and wideband sources \parencite{cheng1994}. The probability of each source contributing to the speech signal is parameterized and then estimated using the expectation maximization (EM) algorithm. Mapping-based techniques aim at learning the associations between features belonging to the narrowband and wideband speech where features including linear predictive coding (LPC) coefficients and line spectral frequencies (LSFs). Codebook mapping-based methods make use of two learned codebooks, belonging to the narrowband and wideband signals, containing spectral envelope features, where a one-to-one mapping exists between their entries \parencite{yoshida1994,epps1999}. For each frame of the input signal, the best matching entry in the narrowband codebook is found and then mapped to the corresponding entry in the wideband codebook. A similar approach is used in linear mapping-based methods, where narrowband features are mapped to wideband features using a linear transformation, where the transformation matrix is learned using methods such as least-squares \parencite{nakatoh1997,chennoukh2001}.

Later methods sought to learn to model the wideband signal directly, rather than the mapping between predefined features. Gaussian mixture models (GMMs) have been used to estimate the joint probability density of narrowband and wideband signals \parencite{park2000,nour2008}. The parameters of the model, namely prior probabilities, mean vectors, and covariance matrices are learned from a training dataset. More sophisticated approaches include the use of hidden Markov models (HMMs), where each state of the model represents the wideband extension of its narrowband input \parencite{jax2003,bauer2008,song2009}. Due to its recursive mechanism, HMMs can leverage information from the past input frames. Methods based on non-negative matrix factorization (NMF) model the speech signals as a combination of learned non-negative bases \parencite{bansal2005,sun2013}. In the testing stage, low-frequency base components of the input can be used to estimate how to combine the high-frequency base components to create the wideband signal. Finally, the first works using neural networks for speech bandwidth extension employ multilayer perceptrons (MLPs) to estimate LPC coefficients of the wideband speech signal \parencite{iser2003}, or to find a shaping function which transforms the spectral magnitude \parencite{kontio2007}. We note that these early works used very small neural networks, in which the total number of parameters was around $100$.

More recent approaches to audio bandwidth extension use DNNs, with many more layers and far greater representation power than their older counterparts. DNNs also eliminate the need for hand-crafted features, as they can use raw audio or time-frequency transforms as input, and then learn appropriate intermediate representations. One of the earliest works using DNNs for audio bandwidth extension works with the frequency spectrogram \parencite{li}. A much deeper model employs the popular \textit{U-Net} architecture \parencite{unet} and works in the raw audio domain, performing experiments on both speech and single instrument music \parencite{kuleshov}. \textcite{tfnet} combines the two aforementioned approaches, resulting in a dual network, which operates separately in the time and frequency domains, and creates the final output using a fusion layer. To increase the qualitative performance, namely, the clarity of the produced audio, generative adversarial networks \parencite{gan} are also employed in DNN-based audio bandwidth extension \parencite{bwe_gan, bwe_gan2}.

While the enhancement of old music recordings can be partially framed in the context of bandwidth extension, certain risks arise when considering the data that DNNs are given for training. Even though trained DNNs can perform well on samples from the training data, they may not exhibit the same performance on unseen samples from testing data. This phenomenon is named \textit{sample overfitting} and even though it is an important concern, especially for classification tasks, its existence in generative tasks, such as image super-resolution, audio bandwidth extension, and adversarial generation, is debated. Recent studies show that sample overfitting is not observed for both discriminators and generators of generative adversarial networks \parencite{gan_overfitting1, gan_overfitting2}, and supervised generative networks for video frame generation \parencite{masters,frame_prediction_sulun}. Furthermore, state-of-the-art image super-resolution networks do not include any regularization layers, such as batch normalization \parencite{batchnorm} and dropout \parencite{dropout}, to avoid overfitting \parencite{imagesr1,imagesr2,imagesr3}. 

\section{Overview of the following chapters}

This concludes our discussion of the background and transitions us into presenting our novel contributions and overall methodology. We begin by developing a standalone conditional music generator, independent of video input. Initially, the model is conditioned on emotional and temporal signals. However, emotion-based music generation requires symbolic music data annotated with emotion labels. As discussed in Section~\ref{sec:midi_datasets}, existing emotion-labeled MIDI datasets are too small to effectively train deep neural networks. Therefore, in Chapter~\ref{chap:labeling}, we address this limitation by creating a large-scale emotion-labeled MIDI dataset.

With this dataset in place, we focus on conditional music generation in Chapter~\ref{chap:music}. After establishing a standalone music generator, we shift our attention to video analysis in Chapter~\ref{chap:video}. In Chapter~\ref{chap:video_music}, we describe how we integrate the conditional music generator with the video analysis module. We conclude our methodological discussion in Chapter~\ref{chap:audio}, where we explore the challenges of audio generation, specifically audio bandwidth extension. Finally, in Chapter~\ref{chap:conclusion}, we present our conclusions and outline directions for future research.
\chapter{Emotion labeling of MIDI} 
\label{chap:labeling}

Conditional deep neural networks are trained to generate outputs based on specific input conditions. Training such models requires a dataset that includes both the target outputs and their corresponding conditions. During training, the condition is provided as input to the DNN, which then generates an output. The model is optimized by minimizing a loss function that measures the difference between the generated output and the ground-truth target. During inference, only the condition is needed, as the trained DNN can produce an output that corresponds to the given condition.

A core component of our video-based music generator is the conditional music generator. Although the conditioning inputs are ultimately derived from video in later stages of our work, we begin by designing and testing the music generator with manually provided conditions. 

Our ultimate goal is to establish a correspondence between input video and output MIDI by leveraging both high-level emotional features and low-level temporal features. Low-level temporal features can be directly extracted from the MIDI data itself, so this stage does not require any additional labels. In contrast, MIDI files do not contain high-level emotion information. Therefore, to train an emotion-conditioned MIDI generator, we require MIDI datasets annotated with emotion labels. Later in Chapter~\ref{chap:music}, we describe how these labels are utilized during MIDI generation.

As our goal is to train deep neural networks for improved music generation, we require large datasets. However, as discussed in Section~\ref{sec:midi_datasets}, existing emotion-labeled MIDI datasets are too limited in size to support such training. To address this, we provide emotion annotations for large-scale MIDI-only datasets. Given the scale of these datasets, manual labeling is impractical. Instead, we propose an automatic labeling approach that leverages additional modalities such as audio and text. Specifically, we first utilize the alignment between MIDI and audio versions of songs. Then, we explore potential emotional correlations between a song’s MIDI representation and its lyrics.

Spotify\footnote{\url{https://spotify.com}} provides both low- and high-level features for the songs in its catalog. Our analysis indicates that features can be effective for representing musical emotion. Furthermore, since some MIDI files include embedded lyrics, we can also leverage text processing models to extract emotional information from the lyrical content. Next, we detail these two approaches.

\section{Using Spotify features}
\label{sec:lakh_spotify}

The Spotify for Developers API (Application Programming Interface)\footnote{\url{https://developer.spotify.com}} provides automated access to audio features of songs. Through this API, we retrieve various musical attributes, including danceability, energy, key, loudness, mode, speechiness, acousticness, instrumentalness, liveness, valence, and tempo. Among these, \textit{valence} is especially relevant for representing musical emotion. Although the exact methods used to compute these features are not publicly disclosed, detailed descriptions are available in the official documentation\footnote{\url{https://developer.spotify.com/documentation/web-api/reference}}.

Using this API, our goal is to label the Lakh MIDI Dataset (LMD), the largest available collection of MIDI files. To do so, we must first identify corresponding entries in Spotify's catalog. Fortunately, LMD includes a subset called \textit{LMD-matched}, which aligns a portion of its MIDI files with entries from the Million Song Dataset (MSD)~\parencite{msd}. Each MSD entry provides metadata such as song title, artist name, and identifiers that link to another music database, The Echo Nest\footnote{\url{https://en.wikipedia.org/wiki/The_Echo_Nest}}. The Million Song Dataset Echo Nest mapping archive\footnote{\url{https://labs.acousticbrainz.org/million-song-dataset-echonest-archive}} extends this connection by providing Spotify track IDs for the Echo Nest. In summary, we follow this chain of mappings---starting from LMD-matched, passing through MSD and Echo Nest---to ultimately associate MIDI files with their corresponding Spotify tracks.

However, not all entries in the MSD have an associated Spotify track ID. To expand our labeled dataset, we adopt an alternative strategy: we use the artist names and song titles provided in the MSD to perform direct searches in Spotify’s database.

The presence of the \textit{valence} feature within Spotify's audio attributes allows us to directly integrate it into the circumplex model of affect (Section~\ref{sec:valence_arousal}, Figure~\ref{fig:valence_arousal})~\parencite{valence_arousal}, which uses valence and arousal to represent the emotion of a MIDI track. This simplifies our task to modeling the arousal feature. However, our initial analysis of the Spotify features suggests that the \textit{energy} feature is not an ideal fit for modeling arousal. We hypothesize that this discrepancy arises from the timbral differences between audio and symbolic music, meaning that the same composition can be performed with varying energy levels (e.g., volume, intonation, effects) depending on the recording.

To model the arousal dimension, we draw on existing literature and use low-level features that can be directly extracted from MIDI~\parencite{aac, fadernets}. Our experiments show that tempo and note density are particularly effective in representing arousal.

We estimate the tempo using the built-in metadata of the MIDI file. The tick scale indicates the temporal resolution of the MIDI file. For instance, a tick scale of 0.001 means that MIDI events can be separated by at least 1 millisecond. The resolution refers to the number of ticks per musical beat. In music, a beat serves as the basic unit of time, often represented as a rhythmic pulse that defines the tempo and structure of a piece. By multiplying the tick scale by the resolution, we obtain the duration of a beat in seconds. To estimate the tempo in beats per minute (BPM), we divide the duration of one minute by the beat duration. The following equation summarizes this approach:

\begin{equation}
\text{Estimated tempo} = \frac{60.0}{\text{Tick scale} \times \text{Resolution}}
\end{equation}

We manually reviewed a set of samples to estimate their tempo by ear and compared these estimates with our computed results. We found that the estimation aligns well when the MIDI file includes a drum track, which is arguably the most influential instrument for establishing rhythm. However, our method occasionally fails in the absence of a drum track, particularly in piano-only pieces. Later in our emotion-based music generation research we experiment with using estimated tempo as arousal only when a drum track is present.

We also extract the note density, which is defined as the average number of notes played by all instruments per second. Additionally, we calculate the number of instruments, and the average note density, which is obtained by dividing the note density by the number of instruments. Using average note density as the arousal feature also works even when there are no drum tracks. However, a manual analysis of the outputs of conditional music generation shows that using average density as arousal is less reliable compared to using the estimated tempo.

\begin{figure}[h]
\centering
\includegraphics[width=0.9\linewidth]{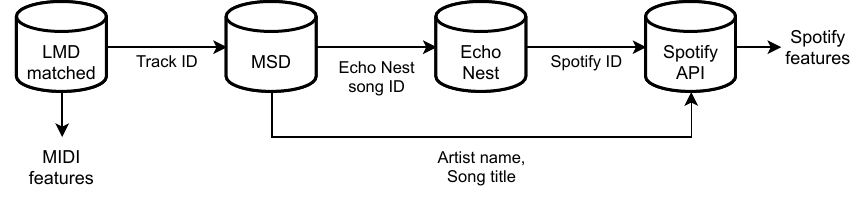}
\caption{Dataset creation pipeline.}
\label{fig:data}
\end{figure}

Our entire dataset creation pipeline is illustrated in Figure \ref{fig:data}.

For completeness and to foster future research, we derive the low-level MIDI features for the entire Lakh MIDI dataset, labeled as \textit{LMD-full}, where not all samples are necessarily mapped to the entries in the MSD. The LMD-full dataset consists of 178,561 MIDI files. After investigation, we find that 174,270 of these were valid, and we discard the remaining corrupt or empty files. The Lakh MIDI dataset was constructed by downloading MIDI files from publicly available sources on the internet and then retaining unique files based on their hash values. However, upon further examination, we find that MIDI files with different hash values can still have the same musical content, possibly due to differences in metadata. To further filter the data and retain only MIDI files with unique musical content, we first convert the MIDI files to pianorolls using the pretty\_midi package\footnote{\url{https://craffel.github.io/pretty-midi}}. We then concatenate the 2-dimensional piano roll matrices of all instruments alphabetically to create a single 2-dimensional matrix that describes only the musical content. After recalculating the hash values using these matrices, we select the files with unique hashes. This process resulted in 152,968 MIDI files with unique musical content. These are the files we ultimately use to train our music generator model.

The matched split of the Lakh MIDI dataset, LMD-matched, consists of 31,034 tracks from the MSD matched with 116,189 MIDI files from the LMD. Since multiple MIDI files can be matched to the same track, and multiple tracks to the same MIDI file, we applied filtering to retain only MIDI files with unique musical content, just as we did for the LMD-full split. Additionally, we kept only the best-matching track from the MSD for each MIDI file, based on the matching scores provided in the LMD-matched subset, ensuring that each MIDI file has a single set of labels. After filtering for valid MIDI files with unique content, we obtained 36,545 MIDI files matched with MSD entries. Using the metadata from the MSD, we queried Spotify’s dataset and successfully retrieved audio features for 34,791 of these MIDI files.

In its complete form, our dataset contains high-level labels such as danceability, energy, key, loudness, mode, speechiness, acousticness, instrumentalness, liveness, and valence; as well as low-level MIDI features such as note density, tempo and the number of instruments. A comparison between our dataset and existing MIDI datasets with emotion labels is shown in Table \ref{table:data_comparison}. This dataset forms our training dataset for emotion-based MIDI generator, which is discussed in Section \ref{sec:emotion_based_midi_generation}. We show a sample entry and the included features in Listing \ref{lst:dataset}.

\begin{table}[h]
	\caption{Emotion-labeled MIDI datasets}
	\label{table:data_comparison}
	\centering
	\begin{tabular}{c|cccc}
		Dataset & \makecell{Number of\\songs} & \makecell{Duration\\(hours)} &  Instrumentation & Emotion label \\ \hline \hline 
		\makecell{MIREX-like\\\parencite{mirexlike}}	& 193	 & 11.6 & Multi-instrument & Categorical  \\ \hline

		\makecell{VGMIDI\\\parencite{vgmidi}} & 204 & 6.4  & Piano-only  & Dimensional  \\ \hline
		
		\makecell{EMOPIA\\\parencite{emopia}} & 387 & 11.0  & Piano-only & Categorical   \\ \hline
		
		Ours & 34791 & 2161.9  & Multi-instrument & Dimensional 
	\end{tabular}
\end{table}


\begin{minipage}[t]{0.9\linewidth}
	
	\label{lst:dataset}
   	\begin{lstlisting}[frame=single,basicstyle=\small,caption=A sample entry from proposed dataset.\vspace{4mm},label=dataset]
   	"cc992d0d8e82d09b7fe2466cf851497a": {
   	
   	"midi_features": {
   	"note_density": 30.364985431879415,
   	"tempo": 84.000084000084,
   	"n_instruments": 10
   	},
   	"matched_features": {
   	"track_id": "TRUHHPK12903CBA84F",
   	"match_score": 0.7362919446232072,
   	"song_id": "SOSYWZT12AB0187E45",
   	"title": "In The Summertime",
   	"artist": "Mungo Jerry",
   	"release": "Uber 30 - das rockt!",
   	"spotify_id": "5VPOrzHyuULaiCKnwQNNCN",
   	"spotify_title": "In The Summertime",
   	"spotify_artist": "Mungo Jerry",
   	"spotify_album": "Uber 30 - das rockt!",
   	"spotify_audio_features": {
   	"danceability": 0.681,
   	"energy": 0.509,
   	"key": 4,
   	"loudness": -8.504,
   	"mode": 1,
   	"speechiness": 0.0461,
   	"acousticness": 0.497,
   	"instrumentalness": 2.72e-06,
   	"liveness": 0.188,
   	"valence": 0.963,
   	"tempo": 82.614,
   	"type": "audio_features",
   	"id": "5VPOrzHyuULaiCKnwQNNCN",
   	"uri": 
   	"spotify:track:5VPOrzHyuULaiCKnwQNNCN",
   	"track_href": "https://api.spotify.com/v1/
		       tracks/5VPOrzHyuULaiCKnwQNNCN",
	"analysis_url":"https://api.spotify.com/v1/
			audio-analysis/5VPOrzHyuULaiCKnwQNNCN",
   	"duration_ms": 210387,
   	"time_signature": 4
   	}
   	}
   	}
   	\end{lstlisting}
\end{minipage}

\section{Emotion classification of song lyrics}
\label{sec:lyrics}
Our second method for obtaining emotion labels for MIDI samples involves analyzing the emotions in the song lyrics included in the MIDI files. While MIDI files are primarily designed to contain note events, they can optionally include text-based metadata such as track name, copyright notice, and lyrics.
    
Our approach takes advantage of the natural connection between lyrics and music, both of which are often tied to emotions~\parencite{music_lyrics_emotion}. Similar to the Spotify music features, we note that the emotional content of lyrics can also provide weak features for the MIDI samples, though there are instances where the lyrics and music convey conflicting emotions. A notable example is the song \textit{Every Breath You Take} by The Police, where the music conveys a loving and caring tone, while the lyrics are written from the perspective of a stalker, reflecting unhealthy fixation, jealousy, and control\footnote{\url{https://www.bbc.co.uk/radio2/soldonsong/songlibrary/indepth/everybreathyoutake.shtml}}.

While the Spotify Developers API directly provides valence values, emotional features from lyrics must be extracted through text analysis. Our approach is to use a text emotion classifier to analyze the song lyrics and assign the predicted emotions to the MIDI samples. To this end, we first train models for emotion classification on the GoEmotions dataset, one of the largest text datasets with 28 fine-grained emotion labels~\parencite{goemotions}.

\subsection{Training}
The first step toward our aim of building an emotion-labeled symbolic music dataset is training a model to perform multi-label emotion classification based on text input.

\subsubsection{Dataset}

We train our model using the GoEmotions dataset \parencite{goemotions}. This dataset consists of English comments from Reddit\footnote{https://www.reddit.com}, which are manually annotated to identify the underlying emotions. It is a multi-label dataset, meaning each sample can have more than one emotion label. The dataset includes 27 emotions and a "neutral" label. \textcite{goemotions} further grouped the labels into 7 categories, including the six basic emotions identified by \textcite{ekman} (joy, anger, fear, sadness, disgust, and surprise) as well as the "neutral" label, as shown in Table~\ref{tab:ekman-mapping} . The dataset contains a total of 58k samples, which are split into training, validation, and testing sets in a ratio of 80\%, 10\%, and 10\%, respectively. Given the number of labels and its size, GoEmotions is one of the largest emotion classification datasets and has the highest number of discrete emotion labels \parencite{emotion_review}.

\begin{table}[]
	\centering
	\caption{Mapping of GoEmotions labels to Ekman categories}
	\renewcommand{\arraystretch}{1.2}
	\begin{tabular}{@{}c|c@{}}

		\textbf{GoEmotions labels} & \textbf{Ekman categories} \\
		\hline \hline
		anger, annoyance, disapproval & anger \\ \hline
		disgust & disgust \\ \hline
		fear, nervousness & fear \\ \hline
		admiration, amusement, approval, caring, desire, & joy \\ 
		excitement, gratitude, joy, love, optimism, pride, relief & \\ \hline
		sadness, disappointment, embarrassment, grief, remorse & sadness \\ \hline
		confusion, curiosity, realization, surprise & surprise \\ \hline
		neutral & neutral \\ 

	\end{tabular}

	\label{tab:ekman-mapping}
\end{table}

\subsubsection{Implementation details}

We employ DistilBERT as the backbone of our model \parencite{distilbert}, which is a condensed and compressed variant of the BERT (Bidirectional Encoder Representations from Transformers) model \parencite{bert}, achieved through knowledge distillation \parencite{distillation_2006,distillation_2015}. DistilBERT utilizes fewer layers than BERT and learns from BERT's outputs to mimic its behavior. Our model consists of $6$ layers, with each layer containing $12$ attention heads and a dimensionality of $768$, yielding a total of 67M parameters. To facilitate multi-label classification, we customize the output layer while adding a sigmoid activation layer at the end, as opposed to using a softmax layer for single label classification. The output layer's size is determined by the number of labels present in the training dataset, which can be either $7$ or $28$.

We train our models using binary cross-entropy loss. For evaluation, we use precision, recall, and F1-score, with macro averaging. We search for the optimal decision cutoff to maximize the F1-score. The decision cutoff is set at $0.3$, meaning that predictions with a value of $0.3$ or greater are considered positive predictions and others negative. 

We train two models to classify a given text into $7$ and $28$ labels. We use a dropout rate of $0.1$ and a gradient clipping norm of $1$. The batch size was set to $16$ for the model with $7$ output labels and to $32$ for the model with $28$ output labels. We apply a learning rate of 5e-5 for the former and 3e-5 for the latter. We use early stopping considering the F1-score on the validation dataset, which corresponded to training for $10$ epochs for both models. We implement the models using Huggingface library \parencite{huggingface} with Pytorch backend \parencite{pytorch} and train them using a single Nvidia GeForce GTX 1080 Ti GPU.

\subsection{Inference}

After training the models for text-based emotion classification, we use it in inference mode, using the song lyrics from the MIDI files as inputs. This allowed us to create a MIDI dataset labeled with emotions. Similar to using audio-related labels in Section~\ref{sec:lakh_spotify}, we consider these as "weak labels" because they are derived from text rather than MIDI, and generated by a trained model rather than annotated by humans.

We use two MIDI datasets that are publicly available and were created by gathering MIDI files from various online sources: the Lakh MIDI dataset consisting of $176k$ samples \parencite{lmd} and the Reddit MIDI dataset containing $130k$ samples\footnote{https://archive.org/details/themagicofmidiv1}. We filter the datasets by selecting MIDI files that contain lyrics in the English language with at least $50$ words. This filtering process resulted in a total of $12509$ files, consisting of $8386$ files from the Lakh MIDI dataset and $4123$ files from the Reddit MIDI dataset. During inference, we utilized the two pretrained models, feeding the entire song's lyrics, using a truncation length of $512$ tokens.

 \subsection{Results}
 \label{sec:labeling_results}
 In this section, we first present the emotion classification performance of our trained models. Then, we introduce the emotion-labeled MIDI dataset, which we created by analyzing the sentiment of the song lyrics using our trained models.
 
 \subsubsection{Emotion classification on the GoEmotions dataset}
 
 We evaluate the performance of our trained models on the test split of the GoEmotions dataset and compared our results with the baseline presented in the original paper \parencite{goemotions}. Similar to the original paper, we report our results for scenarios using two sets of labels, with $7$ and $28$ emotions. For each label, we report the precision, recall, and F1-scores along with the macro-averages. It is important to mention that, as the dataset is imbalanced, macro-averaging is more appropriate than micro-averaging, as it was also used in the original paper. We note that the baseline model is BERT and has twice the size of our model \parencite{bert}. 
 
 The trade-off between precision and recall is determined by the cutoff value. Therefore, we emphasize higher F1-scores because they provide a more balanced perspective by taking the harmonic mean of precision and recall, and are much less sensitive to the cutoff value. Although the original paper does not state the cutoff value, we achieve the best F1-score and similar performance to the original paper on the $7$-label dataset using a cutoff value of $0.3$. For consistency, we use the same value for the $28$-label dataset. We present our results on the dataset with $7$ and $28$ labels in Tables \ref{table:7labels} and \ref{table:28labels}, respectively. Higher values indicate better performance, and the best results are highlighted in bold.
 
 
 \begin{table}[]
    	\centering
    	\caption{7-label classification results}
    	\begin{tabular}{lcc|cc|cc}
    		\\
    		& \multicolumn{2}{c}{Precision} & \multicolumn{2}{c}{Recall} & \multicolumn{2}{c}{F1-score} \\
    		\hline 
    		& Baseline & Ours & Baseline & Ours & Baseline & Ours \\
    		\hline \hline 
    		anger & 0.50  & 0.50  & 0.65  & 0.67  & \textbf{0.57}  & \textbf{0.57} \\
    		disgust & 0.52  & 0.57  & 0.53  & 0.49  & \textbf{0.53}  & 0.52 \\
    		fear  & 0.61  & 0.57  & 0.76  & 0.73  & \textbf{0.68}  & 0.64 \\
    		joy   & 0.77  & 0.75  & 0.88  & 0.89  & \textbf{0.82}  & \textbf{0.82} \\
    		neutral & 0.66  & 0.63  & 0.67  & 0.75  & 0.66  & \textbf{0.68} \\
    		sadness & 0.56  & 0.57  & 0.62  & 0.67  & 0.59  & \textbf{0.61} \\
    		surprise & 0.53  & 0.59  & 0.70  & 0.62  & \textbf{0.61}  & \textbf{0.61}  \\ \hline
    		
    		macro-average & 0.59  & 0.60 & 0.69  & 0.69  & \textbf{0.64}  & \textbf{0.64} \\
    		
    	\end{tabular}
    	\label{table:7labels}
 \end{table}
 
 \begin{table}[]
    	\centering
    	\caption{28-label classification results}
    	\begin{tabular}{lcc|cc|cc}
    		\\
    		& \multicolumn{2}{c}{Precision} & \multicolumn{2}{c}{Recall} & \multicolumn{2}{c}{F1-score}        \\ \hline
    		& Baseline        & Ours        & Baseline       & Ours      & Baseline      & Ours          \\ \hline \hline
    		admiration     & 0.53            & 0.65        & 0.83           & 0.75      & 0.65          & \textbf{0.70} \\
    		amusement      & 0.70            & 0.72        & 0.94           & 0.91      & 0.80          & \textbf{0.81} \\
    		anger          & 0.36            & 0.53        & 0.66           & 0.49      & 0.47          & \textbf{0.51} \\
    		annoyance      & 0.24            & 0.40        & 0.63           & 0.31      & 0.34          & \textbf{0.35} \\
    		approval       & 0.26            & 0.39        & 0.57           & 0.38      & 0.36          & \textbf{0.39} \\
    		caring         & 0.30            & 0.37        & 0.56           & 0.46      & 0.39          & \textbf{0.41} \\
    		confusion      & 0.24            & 0.52        & 0.76           & 0.42      & 0.37          & \textbf{0.47} \\
    		curiosity      & 0.40            & 0.47        & 0.84           & 0.62      & \textbf{0.54} & 0.53          \\
    		desire         & 0.43            & 0.66        & 0.59           & 0.42      & 0.49          & \textbf{0.51} \\
    		disappointment & 0.19            & 0.39        & 0.52           & 0.22      & \textbf{0.28} & \textbf{0.28} \\
    		disapproval    & 0.29            & 0.39        & 0.61           & 0.41      & 0.39          & \textbf{0.40} \\
    		disgust        & 0.34            & 0.64        & 0.66           & 0.39      & 0.45          & \textbf{0.48} \\
    		embarrassment  & 0.39            & 0.72        & 0.49           & 0.35      & 0.43          & \textbf{0.47} \\
    		excitement     & 0.26            & 0.43        & 0.52           & 0.47      & 0.34          & \textbf{0.45} \\
    		fear           & 0.46            & 0.60        & 0.85           & 0.76      & 0.60          & \textbf{0.67} \\
    		gratitude      & 0.79            & 0.88        & 0.95           & 0.92      & 0.86          & \textbf{0.90} \\
    		grief          & 0.00            & 0.00        & 0.00           & 0.00      & \textbf{0.00} & \textbf{0.00} \\
    		joy            & 0.39            & 0.59        & 0.73           & 0.61      & 0.51          & \textbf{0.60} \\
    		love           & 0.68            & 0.78        & 0.92           & 0.85      & 0.78          & \textbf{0.81} \\
    		nervousness    & 0.28            & 0.45        & 0.48           & 0.43      & 0.35          & \textbf{0.44} \\
    		neutral        & 0.56            & 0.61        & 0.84           & 0.76      & \textbf{0.68} & \textbf{0.68} \\
    		optimism       & 0.41            & 0.56        & 0.69           & 0.52      & 0.51          & \textbf{0.54} \\
    		pride          & 0.67            & 0.83        & 0.25           & 0.31      & 0.36          & \textbf{0.45} \\
    		realization    & 0.16            & 0.39        & 0.29           & 0.14      & \textbf{0.21} & \textbf{0.21} \\
    		relief         & 0.50            & 0.00        & 0.09           & 0.00      & \textbf{0.15} & 0.00          \\
    		remorse        & 0.53            & 0.59        & 0.88           & 0.86      & 0.66          & \textbf{0.70} \\
    		sadness        & 0.38            & 0.57        & 0.71           & 0.60      & 0.49          & \textbf{0.59} \\
    		surprise       & 0.40            & 0.56        & 0.66           & 0.50      & 0.50          & \textbf{0.53} \\ \hline
    		macro-average  & 0.40            & 0.53        & 0.63           & 0.50      & 0.46          & \textbf{0.50}
    	\end{tabular}
    	\label{table:28labels}
 \end{table}
 
 Based on the F1-scores, our model performs comparably to the baseline on the $7$-label dataset. Specifically, our model has a better performance on $2$ labels, worse on $2$ labels, and the same on $3$ labels, as well as for the macro-average. On the $28$-label dataset, our model surpasses the baseline with only a lower performance on $2$ labels, equal performance on $4$ labels, and better performance on the remaining $22$ labels. Furthermore, our model demonstrates an improvement of $0.04$ in terms of the macro-average.
 
 We hypothesize that a smaller model, such as ours (DistilBERT), may perform better than a larger baseline model (BERT) in certain settings, such as when there are a limited number of training samples or a high output/target dimensionality, as in the case of the $28$-label dataset. In these scenarios, models are more prone to overfitting, as has been previously observed \parencite{overfitting}. Additionally, the original paper \parencite{distilbert} demonstrates that the DistilBERT model outperforms BERT on the Winograd Natural Language Inference (WNLI) dataset \parencite{wnli}.
 
 \subsubsection{Inference on MIDI datasets}
 
 We use our trained models to analyze the song lyrics of the Lakh and Reddit MIDI datasets, resulting in an augmented dataset that contains the file paths to $12509$ MIDI files and their corresponding predicted probabilities for emotion labels. To provide more flexibility to the users, we do not apply a threshold to the predicted probabilities and provide the raw classification outputs, allowing the entire dataset to be used as is. We generate two CSV (comma-separated values) files containing the $7$ and $28$ emotion labels as columns, with the $12509$ MIDI file paths as rows. 
 
 For demonstration purposes, we provide transposed versions of the tables, using only $3$ samples, shown in Tables \ref{table:sample_7labels} and \ref{table:sample_28labels}.  For further demonstration and ease of analysis, we provide excerpts from the lyrics of each of the three sample songs in Listing \ref{lyrics}, along with the emotions having predicted probabilities higher than $0.1$ in descending order. It is noteworthy that having a dataset with $28$ emotion labels allows for a more nuanced representation of emotions. For instance, when we examine this dataset, the song "Imagine" is predicted to have ``optimism" as its top emotion, whereas "Take a Chance on Me" is predicted to have "caring" as its top emotion. However, both songs are predicted to have "joy" as their top emotion in the dataset with only seven labels.
 
\begin{figure}[htbp]
    	\includegraphics[width=\textwidth]{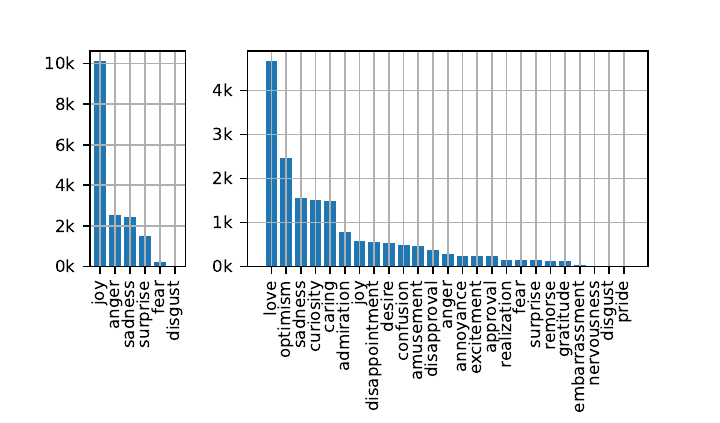}
    	\caption{The number of samples containing each emotion in our $7$-label (left) and $28$-label (right) datasets. 
    	} 
    	\label{fig:counts}
 \end{figure}

 We also present the number of samples containing each emotion in our datasets in Figure \ref{fig:counts}. In these figures, we exclude the "neutral" label. For the sake of demonstration, we also consider emotions with a prediction value higher than $0.1$ as positive labels, meaning that those emotions are present for a given sample. Due to space limitations, the file paths are replaced with the artist and song names and are as the following: John Lennon - Imagine: "\texttt{\path{lakh/5/58c076b72d5115486c09a7d9e6df1029.mid}}" (artist and title obtained using Million Song Dataset \parencite{msd}), ABBA - Take a Chance on Me: "\texttt{\path{reddit/A/ABBA.Take a chance on me K.mid}}", Elvis Presley - Are You Lonesome Tonight: "\texttt{\path{reddit/P/PRESLEY.Are you lonesome tonight K.mid}}"


 \begin{table}[]
	\centering
	\caption{Sample entries from the $7$-label dataset.}
	\label{table:sample_7labels}
	
	\begin{tabular}{l@{\hskip 6mm}c@{\hskip 9mm}c@{\hskip 3mm}c}
		& \begin{tabular}{@{}c@{}}John Lennon \\ - Imagine \end{tabular} & \begin{tabular}{@{}c@{}}ABBA - Take \\ a Chance on Me\end{tabular} & \begin{tabular}{@{}c@{}}Elvis Presley - Are \\ You Lonesome Tonight\end{tabular}  \\ \hline \hline
		anger    & 0.0051                                      & 0.0146                                  & 0.0272                                          \\
		disgust  & 0.0003                                      & 0.0009                                  & 0.0045                                          \\
		fear     & 0.0005                                      & 0.0024                                  & 0.0131                                          \\
		joy      & \textbf{0.8072}                                      & \textbf{0.8948}                                  & 0.0477                                          \\
		neutral  & 0.1953                                      & 0.1420                                  & 0.0782                                          \\
		sadness  & 0.0013                                      & 0.0069                                  & \textbf{0.7372}                                          \\
		surprise & 0.0754                                      & 0.0053                                  & 0.5465                                         
	\end{tabular}
\end{table}

 \begin{table}[]
    	
    	\centering
    	\caption{Sample entries from the $28$-label dataset.} 
    	\vspace{3mm}
    	\label{table:sample_28labels}
    	
    	\begin{tabular}{l@{\hskip 6mm}c@{\hskip 9mm}c@{\hskip 3mm}c}
    		& \begin{tabular}{@{}c@{}}John Lennon \\ - Imagine \end{tabular} & \begin{tabular}{@{}c@{}}ABBA - Take \\ a Chance on Me\end{tabular} & \begin{tabular}{@{}c@{}}Elvis Presley - Are \\ You Lonesome Tonight\end{tabular}  \\ \hline \hline
    		admiration     & 0.0021                                      & 0.0091                                  & 0.0048                                          \\
    		amusement      & 0.0051                                      & 0.0012                                  & 0.0027                                          \\
    		anger          & 0.0025                                      & 0.0018                                  & 0.0053                                          \\
    		annoyance      & 0.0024                                      & 0.0020                                   & 0.0075                                          \\
    		approval       & 0.0026                                      & 0.0809                                  & 0.0072                                          \\
    		caring         & 0.0067                                      & \textbf{0.6169}                                  & 0.0601                                          \\
    		confusion      & 0.0070                                       & 0.0035                                  & 0.1029                                          \\
    		curiosity      & 0.0332                                      & 0.0141                                  & \textbf{0.6502}                                          \\
    		desire         & 0.0482                                      & 0.0472                                  & 0.0055                                          \\
    		disappointment & 0.0044                                      & 0.0016                                  & 0.0199                                          \\
    		disapproval    & 0.0019                                      & 0.0030                                   & 0.0048                                          \\
    		disgust        & 0.0007                                      & 0.0003                                  & 0.0009                                          \\
    		embarrassment  & 0.0006                                      & 0.0002                                  & 0.0045                                          \\
    		excitement     & 0.0130                                       & 0.0049                                  & 0.0011                                          \\
    		fear           & 0.0026                                      & 0.0026                                  & 0.0035                                          \\
    		gratitude      & 0.0007                                      & 0.0017                                  & 0.0059                                          \\
    		grief          & 0.0008                                      & 0.0016                                  & 0.0085                                          \\
    		joy            & 0.0025                                      & 0.0040                                   & 0.0018                                          \\
    		love           & 0.0021                                      & 0.1079                                  & 0.0193                                          \\
    		nervousness    & 0.0007                                      & 0.0017                                  & 0.0094                                          \\
    		neutral        & 0.2954                                      & 0.4288                                  & 0.0757                                          \\
    		optimism       & \textbf{0.7554}                                      & 0.1423                                  & 0.0060                                           \\
    		pride          & 0.0010                                       & 0.0013                                  & 0.0006                                          \\
    		realization    & 0.0023                                      & 0.0040                                   & 0.0045                                          \\
    		relief         & 0.0004                                      & 0.0033                                  & 0.0011                                          \\
    		remorse        & 0.0005                                      & 0.0012                                  & 0.1491                                          \\
    		sadness        & 0.0011                                      & 0.0027                                  & 0.1767                                          \\
    		surprise       & 0.0107                                      & 0.0005                                  & 0.0020                                          
    	\end{tabular}
 \end{table}
 
\begin{lstlisting}[float,floatplacement=htbp,basicstyle=\ttfamily\fontsize{9.0}{8}\selectfont,caption={Sample entries with excerpts from lyrics, and emotions with a predicted value higher than 0.1.},label=lyrics]
File path: lakh/5/58c076b72d5115486c09a7d9e6df1029.mid
Artist - Title: John Lennon - Imagine
Lyrics: 
Imagine there's no heaven. 
It's easy if you try. 
No hell below us. 
Above us, only sky. 
Imagine all the people. 
Livin' for today.
7-label predictions:
joy:            0.8072
neutral:        0.1953
28-label predictions:
optimism:       0.7554
neutral:        0.2954

File path: reddit/A/ABBA.Take a chance on me K.mid
Artist - Title: ABBA - Take a Chance on Me
Lyrics: 
If you change your mind, I'm the first in line.
Honey, I'm still free. 
Take a chance on me.
If you need me, let me know, gonna be around.
If you've got no place to go, if you're feeling down.
7-label predictions:
joy:            0.8948
neutral:        0.1420
28-label predictions:
caring:         0.6169
neutral:        0.4288
optimism:       0.1423
love:           0.1079

File path: reddit/P/PRESLEY.Are you lonesome tonight K.mid
Artist - Title: Elvis Presley - Are You Lonesome Tonight
Lyrics: 
Are you lonesome tonight? 
Do you miss me tonight? 
Are you sorry we drifted apart? 
Does your memory stray to a bright summer day, 
When I kissed you and called you sweetheart?
7-label predictions:
sadness:        0.7372
surprise:       0.5465
28-label predictions:
curiosity:      0.6502
sadness:        0.1767
remorse:        0.1491    
confusion:      0.1029
\end{lstlisting}

\clearpage

\section{Discussion}

In our exploration of automatic emotion labeling for symbolic music, we investigated two primary approaches: utilizing Spotify’s audio-related features and analyzing text-based song lyrics embedded within MIDI files. Our objective was to augment large-scale MIDI datasets with emotional metadata to facilitate the training of conditional music generators. Our preliminary evaluations revealed that emotion labels inferred from Spotify features were more reliable than those derived from lyrics. As a result, we ultimately relied on Spotify-derived labels for emotion-conditioned MIDI generation.

Spotify offers a set of high-level musical features, with valence in particular aligning closely with established emotion models such as the circumplex model of affect. We leveraged the alignment between the Lakh MIDI Dataset and Spotify’s catalog through the Million Song Dataset and Echo Nest IDs. This process yielded emotional metadata for tens of thousands of MIDI files, offering a practical foundation for emotion-based MIDI generation.

While we also explored using lyrics to infer emotional content, this approach proved less effective in our setting. Despite training a multi-label emotion classifier using the GoEmotions dataset and a DistilBERT backbone, the labels generated from lyrics lacked consistency and failed to produce perceptually coherent results in downstream generation tasks. This limitation may stem from the ambiguity of lyrics, mismatches between lyrical and musical emotion, and the domain gap between internet text and song lyrics. As such, lyric-based annotations were excluded from the final generation pipeline.

Our quantitative results on text-based emotion classification highlights the efficiency of DistilBERT for emotion classification, especially in cases where the number of output labels is high, as in the 28-emotion label set. Notably, our model outperformed the baseline on the 28-label dataset, demonstrating that smaller models can prevent overfitting and still achieve strong results in settings with limited training data and many classes. The ability to classify text into 28 fine-grained emotion categories enabled us to capture more nuanced emotional variations in songs, as shown by the our demonstrative output samples in Listing~\ref{lyrics}.

Although we exclude the lyric-based labels from our music generation pipeline, emotion classification from lyrics remains a promising research direction. Lyrics can convey the songwriter’s emotional intent and provide a semantic layer beyond what audio features reveal. With improved domain-specific models or multi-modal strategies that combine lyrics with audio and symbolic data, more nuanced and accurate emotional labeling could be achieved. Our dataset and models are included to support these avenues, particularly for tasks such as mood-based music retrieval or affect-aware lyric composition.

Together, these labeling approaches lay the groundwork for emotion-conditioned music generation, where symbolic music can be composed or guided based on target emotional characteristics. In the next chapter, we describe how these labels are used to train generative models that respond to emotional conditions.

\chapter{Conditional music generation} \label{chap:music}

In this chapter, we present our approach to conditional music generation, which is the most important component of our video-based music generation system, as this module is responsible for producing the final musical output. While the input conditions are manually provided in this chapter, in Chapter~\ref{chap:video_music} we replace them with results obtained from video analysis. As previously discussed, we chose to focus on conditional music generation first, and video analysis later in Chapter~\ref{chap:video}, to ensure musically meaningful results at every stage of our research.

As explained in Chapter~\ref{chap:intro}, our goal is to achieve both high- and low-level alignment between the input video and the generated music. We use emotions as the high-level and temporal boundaries as the low-level intermediary features. This enables the generated soundtrack to reflect the video's emotional tone while maintaining temporal synchronization.

We first outline our method for temporally-conditioned music generation.

\section{Music generation based on temporal boundaries}
\label{sec:boundary_conditioning}

Temporal alignment of the video and music is essential to create a "synchronized" feel~\parencite{soundtrack_scenes1}. Our approach seeks to overcome the limitations of the dense temporal conditioning mechanisms used by state-of-the-art methods~\parencite{di}, as discussed in Section~\ref{sec:intro_temporal}. To address this, we propose utilizing sparse temporal features, specifically temporal boundaries. In this approach, we first define what constitutes a temporal boundary in both video and music.

Since scenes (or equivalently, shots) are the fundamental units of a video, we use scene cuts to mark temporal boundaries~\parencite{scenecut1,scenecut2}. In the musical domain, we adopt chord locations as temporal boundaries, as chords frequently function as structural anchors in compositions~\parencite{chords}. By aligning these two modalities, our goal is to train a music generator that places chords near the scene boundaries of the input video, while maintaining overall musical coherence throughout the generated melody.

To enable temporally-conditioned music generation, we first extract chord locations to use as input during training. We label chords in our training dataset—the Lakh Pianoroll Dataset-5~\parencite{lpd}, a version of the Lakh MIDI Dataset~\parencite{lmd} with instruments merged into five categories: bass, drums, guitar, piano, and strings. We focus on guitar and piano chords containing at least three simultaneous notes lasting a minimum of two beats. We refer to these as \textit{strong chords}, noting that this terminology differs from any potential uses of the term in harmonic contexts. To label these strong chords, we extend our token vocabulary by adding the \texttt{CHORD} token. Each strong chord is marked by inserting a \texttt{CHORD} token before its first \texttt{ON} (i.e., \texttt{NOTE-ON}) token. To allow the model to generate chords not only at video scene cut locations but also independently where musically appropriate, we randomly remove 20\% of \texttt{CHORD} tokens during training.

Chords are integral to a melody, providing harmonic and rhythmic support to surrounding notes, both preceding and following~\parencite{chords}. During inference, forcing a \texttt{CHORD} token into the sequence at a specific location may cause the chord to sound off-beat or overly abrupt. This occurs because the model, having no prior knowledge of the upcoming chord, may generate preceding notes that do not align naturally with it. To address this, we propose a method that enables the model to "anticipate" upcoming chords and generate preceding notes and time shifts accordingly. Additionally, we train the model to generate the \texttt{CHORD} token itself, ensuring rhythmic consistency in the generated music. To achieve this, we define \textit{boundary offsets} for each input token, representing the remaining time until the next boundary. These offsets are capped at a maximum value and, since our music generator is autoregressive, they are computed based on future chords rather than past ones. Furthermore, we amplify the loss of the \texttt{CHORD} token by a factor of 10, as it influences multiple preceding chord notes and plays a critical role in structuring the generated music.

\begin{figure}[b]
	\centering
	\includegraphics[width=1\textwidth]{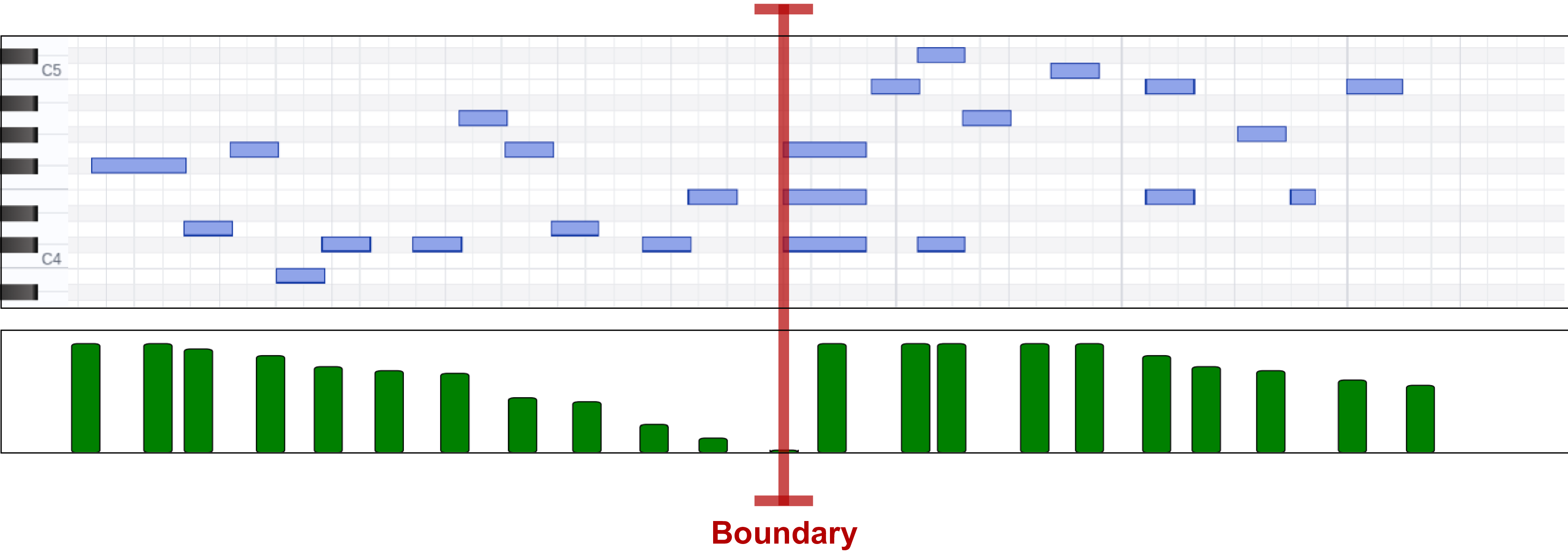}
	\caption{Graphical illustration of a musical boundary and its offsets.}
	\label{fig:chord_boundaries}
\end{figure}

In Figure~\ref{fig:chord_boundaries}, we illustrate the concept of boundary offsets relative to the strong chord. The top part of the figure shows the symbolic music, where a chord containing three or more simultaneous notes and lasting longer than a certain threshold defines a musical boundary. The bottom part of the figure displays the boundary offsets, which represent the temporal distance to the next boundary. It is important to note that these figures are illustrative; the offsets do not perfectly align with the music, except at the defined boundary.

We process boundary offsets using a feed-forward network to generate boundary offset encodings, which are then concatenated with learned positional encodings (defined in Section~\ref{sec:position_background})~\parencite{learned_position} along the feature dimension. This architectural choice is motivated by several key factors. First, we inject boundary offset encodings at the input level rather than within the transformer body, ensuring that the core model remains unchanged. This allows the model to process inputs with or without boundary offsets, enabling seamless fine-tuning by repeating the learned positional encodings along the feature dimension when boundary offsets are absent. Second, we avoid adding boundary offset encodings directly to learned positional encodings to maintain a distinction between the two. Finally, recent studies suggest that decoder-only transformers can implicitly learn positional encodings through their internal weights, even without explicitly adding them~\parencite{position1,position2}. Based on this insight, we halve the feature length of learned positional encodings and allocate the remaining feature space to boundary offset encodings. The resulting vector sequence consists of positional encodings augmented with boundary offset encodings. Following the standard transformer model, we add this sequence to the token embeddings~\parencite{transformer} before passing it to the transformer body with relative global attention~\parencite{musictransformer}.

\begin{figure}[hbp]
	\centering
	\includegraphics[width=0.6\textwidth]{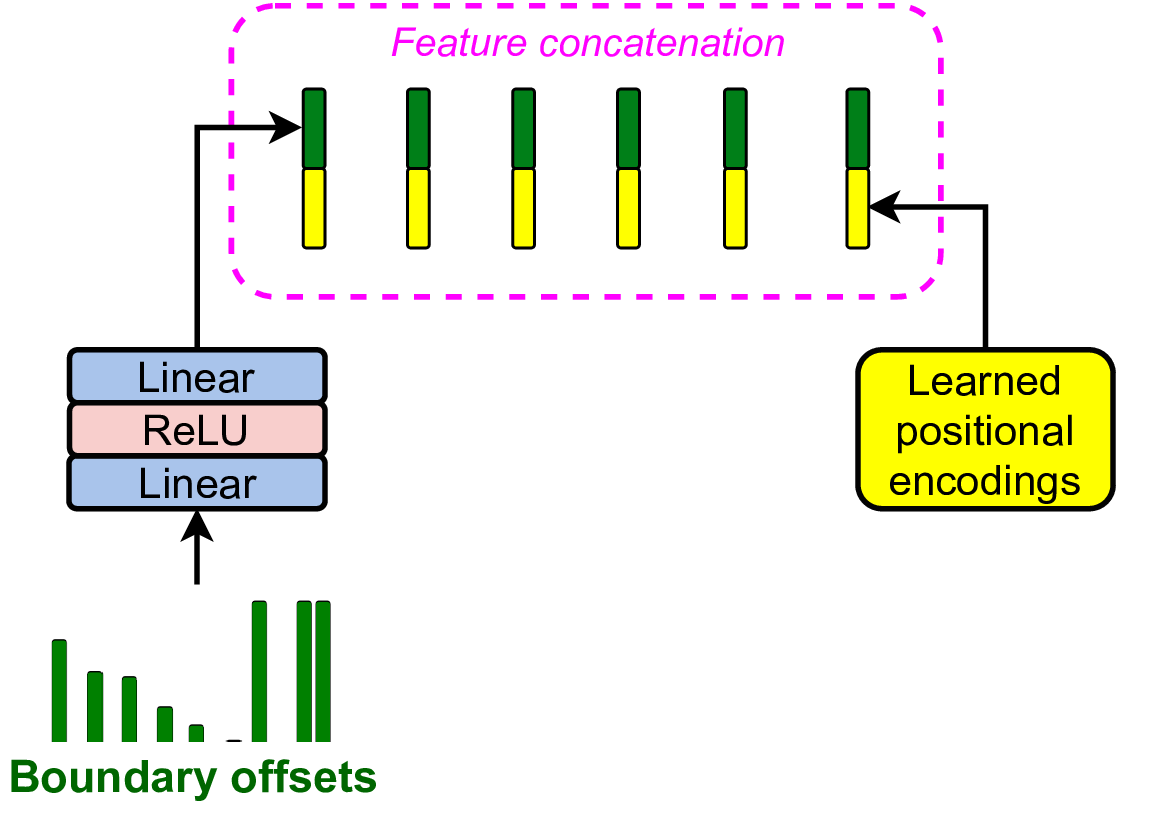}
	\caption{Conditioning mechanism based on boundary offsets.}
	\label{fig:boundary_conditioning}
\end{figure}

Algorithm \ref{alg:offset} outlines the process of computing boundary offsets during inference, i.e., music generation. The model generates music autoregressively, producing one token per forward pass. We maintain a time cursor by tracking the generated \texttt{TIMESHIFT} tokens and compute the boundary offset, i.e., the time remaining until the next boundary, for each generated token. If the model generates a \texttt{CHORD} token, we calculate the absolute difference between the time cursor and each input boundary. The boundary with a difference smaller than the predefined sensitivity threshold of 1\,s is considered successfully generated. We then remove these boundaries from future offset calculations by replacing them with infinity. Next, we compute the boundary offset for the generated token, regardless of its type. 
This offset represents the distance to the next closest boundary. We compute it by subtracting the time cursor from each input boundary and replacing any negative values (corresponding to past boundaries) with infinity to ignore them. We then take the minimum of the resulting values as the offset to the nearest boundary and clamp it to a maximum value if necessary. The resulting boundary offset and generated token are appended to their respective lists to be used in the next timestep.
For simplicity, Algorithm \ref{alg:offset} is presented for a single sample, but in practice, this operation is performed in minibatches. During training, we preprocess the entire input sequence at once by constructing a time grid instead of a time cursor, allowing us to calculate boundary offsets for all tokens simultaneously. For clarity, the initial list of generated tokens is shown as empty; however, in practice, we begin with a \texttt{START} token.

\begin{algorithm}[t]
	\caption{Creating boundary offsets during music generation in inference.}
	\label{alg:offset}
	\begin{flushleft}
	\textbf{Input}: List of input boundaries (in seconds), $\mathbf{b}$; video duration, $d$; valence, $x_v$; arousal $x_a$\\
	\textbf{Parameter}: Sensitivity, $\xi$; distances to all boundaries, $\boldsymbol{\delta}_b$; maximum offset, $\delta_{max}$; generated boundary mask, $\mathbf{m_b}$; token type $t_t$; token value, $t_v$; time cursor, $c$; generation function including forward-pass and sampling, g\\
	\textbf{Output}: List of generated tokens $\mathbf{t}$; list of boundary offsets, $\boldsymbol{\delta}$
\end{flushleft}
	\begin{algorithmic}
		\STATE Let $c=0$; $\boldsymbol{\delta}=\left[ \hspace{2pt} \right]$; $\mathbf{t}=\left[ \hspace{2pt} \right]$.
		\WHILE{$c < d$}
		\STATE \texttt{\# generate token as (type, value):} 
		\STATE $(t_t, t_v) = g(\mathbf{t}, \boldsymbol{\delta}, x_v, x_a)$
		\IF {$t_t$ == \texttt{TIMESHIFT}}
		\STATE $c = c + t_v $
		\ELSIF {$t_t$ == \texttt{CHORD}}
		\STATE $\mathbf{m_b} = \left| c - \mathbf{b} \right| < \xi$.
		\STATE $\mathbf{b}\left[\mathbf{m_b}\right] = +\infty$.
		\ENDIF
		
		\STATE $\boldsymbol{\delta}_b = \mathbf{b} - c $
		\STATE $\boldsymbol{\delta}_b[\boldsymbol{\delta}_b < 0] = +\infty $
		
		\STATE $\boldsymbol{\delta}$.append$(\min (\min (\boldsymbol{\delta}_b), \delta_{max}))$
		
		\STATE $\mathbf{t}$.append$((t_t, t_v))$
		
		\ENDWHILE
	\end{algorithmic}
\end{algorithm}

%

The \texttt{ON} tokens that appear after a \texttt{CHORD} token and before the next \texttt{TIME-SHIFT} token are considered notes of the generated chord. To make the generated chord more distinctive, we increase the velocity of these notes. 

While in training, the temporal locations of chords from the ground-truth MIDI serve as input boundaries. As we later detail in Chapter~\ref{chap:video_music}, during video-based inference, we replace these boundaries with video scene cut locations.

We continue our discussion with conditioning music generation on emotions.

\section{Music generation based on emotions}
\label{sec:emotion_based_midi_generation}
\label{sec:music_models}


Emotion-based music generation is the second module of our conditional music generation model. Compared to temporal conditioning, emotion-based conditioning aims to establish a higher-level connection between the video and music, focusing on the emotional tone.

To enable emotion-based music generation, we previously obtained emotion labels for MIDI datasets, as described in Chapter~\ref{chap:labeling}. These labels serve as conditional inputs during model development. Next, we describe the methods we developed for emotion-based music generation.

The backbone of our models is the music transformer \parencite{musictransformer}, which is a decoder-only transformer using relative position embeddings. We first design a \textit{vanilla} model that does not use any conditioning. It takes musical tokens, projects them into vector space through an embedding layer, and feeds the output into the music transformer.

We explore several methods for conditioning the music generation process on emotion features, which we refer to as \textit{discrete-token}, \textit{continuous-token}, and \textit{continuous-concatenated}. The \textit{discrete-token} approach represents the state-of-the-art in conditional sequence generation~\parencite{musenet, emopia, ctrl}, where a discrete control token is prepended to the original sequence---in our case, a tokenized MIDI sequence. However, since our emotion conditions are continuous rather than discrete, we design alternative methods to incorporate these continuous conditions, enabling finer-grained control over the generated music.

In the \textit{discrete-token} approach, we convert the continuous emotion conditions into discrete values by placing them into bins. Specifically, we quantize the condition values using 5 equal-sized bins, with the central bin indexed as 0. The number of bins is chosen to reflect typical verbal quantifiers, such as \textit{very low}, \textit{low}, \textit{moderate}, \textit{high}, and \textit{very high}. The control tokens for valence and arousal are prepended to the music tokens, i.e., they are concatenated in the sequence dimension, before being fed into the transformer model. A key limitation of this method is the potential loss of information due to the binning of continuous values.

In the next approach, called \textit{continuous-token}, we utilize the condition values in their original continuous form. The valence and arousal values are passed through separate linear layers to create condition vectors, which match the feature length of the music token embeddings. These condition vectors are then concatenated with the music token embeddings in the sequence dimension and fed into the transformer. Unlike the \textit{discrete-token} approach, this method avoids the information loss associated with quantization by preserving the continuous nature of the emotion conditions.

\begin{figure}[t]
	\centering
	\includegraphics[width=1\textwidth]{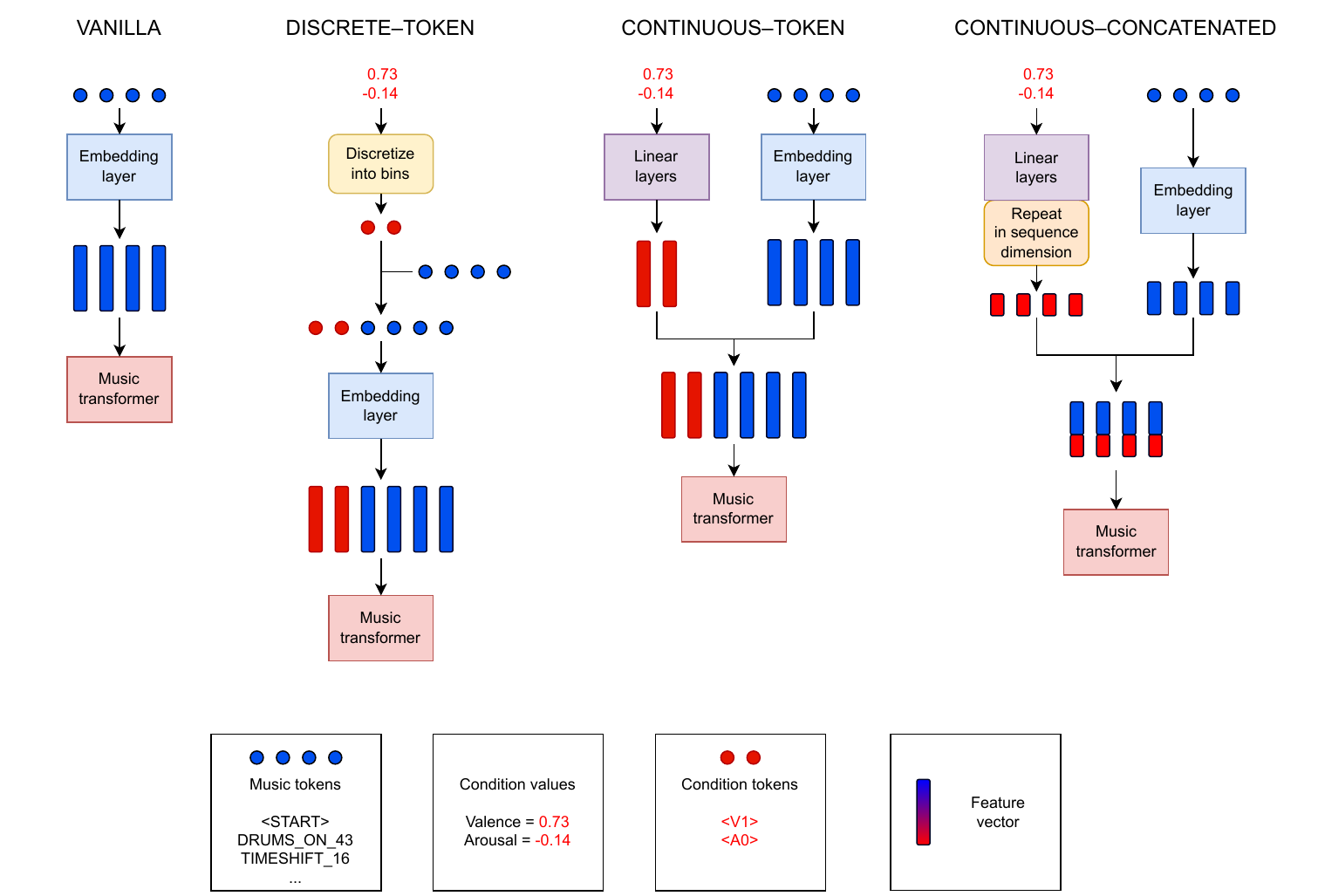}
	\caption{Emotion-based music generation models.}
	\label{fig:models}
\end{figure}

Our final approach, \textit{continuous-concatenated}, combines the two continuous condition values into a single vector, which is then repeated across the sequence dimension and concatenated with each music token embedding. This design addresses a limitation of the \textit{continuous-token} method, where condition vectors are treated with equal importance as individual music tokens. We argue that emotional conditions influence the musical sequence globally, unlike individual notes that contribute locally. By embedding the condition vector into every token, we ensure that the emotional context is present throughout the entire sequence, reinforcing its influence at every step. The representations of the models can be seen in Figure \ref{fig:models}.

For a given sample, if a condition label is not present, i.e. is NaN (Not a Number), we are not able to quantize it into a bin, nor project it into a vector space. In this case, we employ learned embedding instead of projections. As any discrete token is projected into vector space using an embedding layer, this approach is equivalent to having special discrete tokens such as \texttt{NO\_VALENCE} and \texttt{NO\_AROUSAL} in model's vocabulary. This approach not only accommodates missing data but also enables the model to switch between conditional and non-conditional generation within a unified framework.

\section{Dataset and preprocessing}
\label{sec:music_dataset}

We train our music generator on the Lakh Pianoroll Dataset (LPD)~\parencite{lpd}, which contains 174,154 pianorolls derived from the Lakh MIDI Dataset~\parencite{lmd}. We tokenize the pianorolls using an event-based symbolic music representation~\parencite{event_encoding}. Specifically, an \texttt{ON} (note on) token marks the start of a note, and an \texttt{OFF} (note off) token marks its end. These tokens also encode pitch and instrument information. For example, a piano note with a MIDI pitch of 60 (C4) is denoted as \texttt{PIANO\_ON\_60}. We filter out MIDI note-on and note-off events that have a pitch outside the range of the piano, i.e., lower than $21$ (A0) and higher than $108$ (C8), since these notes aren't audible using standard MIDI soundfonts. Since a larger number of instruments increases vocabulary size, we use the Lakh Pianoroll Dataset-5 variant, where all instrument tracks are merged into five predefined categories: bass, drums, guitar, piano, and strings~\parencite{lpd}. However, our method is adaptable to datasets with different instrument groupings.

We use a temporal resolution of 8 milliseconds with a maximum shift of 1000 milliseconds, as done by \textcite{event_encoding}. Longer durations are represented using multiple consecutive \texttt{TIME-SHIFT} tokens. 
We also use the \texttt{START} tokens to mark the beginning of the songs, the \texttt{BAR} tokens to indicate the musical bars, and the \texttt{PAD} tokens to standardize input sequence lengths in minibatches. As mentioned earlier, strong chords are marked by inserting a \texttt{CHORD} token before their first \texttt{ON} token.

Since we aim to generate multi-instrument compositions, we prioritize pieces with three or more instruments. However, the dataset contains many songs with only one or two instruments. Rather than filtering them out, we prepend special tokens: \texttt{FEWER\_INSTRUMENTS} for songs with two or fewer instruments and \texttt{MORE\_INSTRUMENTS} for those with three or more. These tokens allow users to specify instrumentation preferences at inference time while leveraging the entire dataset during training.

As emotion labels, we use our previously constructed Lakh-Spotify dataset, described in Section~\ref{sec:lakh_spotify}. Emotions are represented using the dimensional circumplex model of affect~\parencite{valence_arousal}, with a slight modification: both valence and arousal are constrained to the range $[-1, 1]$, forming a square valence-arousal space rather than a circular one. Since the original valence values in the Lakh-Spotify dataset span $[0, 1]$, a shift and scale transformation is required. However, we observed an unexpected peak at the 0.0 value in the original valence distribution. Upon examining the corresponding musical samples, this value did not consistently reflect low-valence characteristics, suggesting that missing or undefined values might have been marked as zero. We therefore replace these values with NaN (Not A Number) values, resulting in 23,653 samples with valid valence labels.

To model arousal, we use low-level MIDI features such as note density and tempo, following the approach suggested by \textcite{aac}, and our qualitative viewing of the output showed musically coherent results similar to \textcite{fadernets}. We initially experimented with average note density and observed moderate success; however, we ultimately adopted estimated tempo exclusively for samples containing a drum track, as this produced more consistent qualitative results, as explained in Section~\ref{sec:lakh_spotify}. To ensure a robust generation process, we exclude extreme tempo values by setting empirical bounds of 50 and 150, determined through trial and error and manual inspection. Samples falling outside these limits or lacking a drum track are assigned NaN values for arousal. This results in 103,735 samples with valid arousal values. Both valence and arousal values are shifted and scaled to lie within the range $[-1, 1]$. Finally, to create the testing data split, we order the file names from the LPD-matched subset alphabetically, and reserve the last 5\%.

\section{Implementation details}

The training input sequences are fixed-sized chunks extracted from the MIDI sequences. In our experiments, we use an input length of 1216 tokens to maximize memory usage. With a probability of 0.05, the input chunk is selected from the beginning of the song, which helps the model learn how songs typically start and better understand the \texttt{START} token. With the remaining 0.95 probability, the chunk is taken from a random location in the song, which increases the variability of the inputs. 

For data augmentation, we transpose the pitches of all instruments, except for drums, by a randomly chosen integer value between -3 and 3 inclusive. This relatively narrow range ensures that the emotional character of the songs remains intact. By applying these two methods, i.e., random starting points and pitch transposition, the effective training data size becomes significantly larger than the number of songs in the training data split.

Our models have 20 layers and a feature dimension of 768. Each layer has 16 heads and a feed-forward layer with a dimension of 3072. In the \textit{continuous-concat} model, the dimensionalities of the conditioning vectors and token embeddings are 192 and 576, respectively, ensuring that the total dimensionality of the transformer input remains constant at 768 across models. Overall, our models have around 145 million parameters. 

We implement our models using the Pytorch library \parencite{pytorch} and train them on a single NVIDIA Quadro RTX 6000 GPU. We use the Adam optimizer \parencite{adam} with a learning rate of 2e-5. Our preliminary experiments confirm the findings of \textcite{lakhnes}, that common learning rates for language modeling tasks, about 2e-4, are too high for MIDI generation tasks. We reduced the learning rate to 2e-6 when the training loss plateaued and kept training until convergence. We use gradient clipping at a norm of 1, with a dropout rate of 0.1, a batch size of 4, and an input length of 1216 tokens. We use a triangular autoregressive mask to prevent the model from attending to future tokens.

At inference, and before the generation starts, the input sequence only consists of the \texttt{[START, BAR]} tokens. Alternatively, to generate a piece of music with a more abrupt beginning, we can omit the \texttt{START} token and use only the \texttt{BAR} token as the priming sequence. This approach typically results in an output that resembles the middle portion of a song. Additionally, users can choose any MIDI sequence as a primer and have the model continue the given melody.

We generate the output autoregressively, where the generated token is appended to the input sequence, forming the input sequence for the next timestep. When the generated sequence starts to exceed the maximum length of 1216 tokens, we use only the last 1216 tokens of the generated sequence as input, ensuring the length limit is not exceeded. For token generation, we use nucleus sampling with a probability of $p=0.7$ from a temperature-adjusted softmax distribution \parencite{nucleus}, where the temperature is set to 1.2. To prevent excessive repetition, if the number of tokens in the nucleus from the previous step is fewer than 3, we slightly increase the temperature. These hyperparameters are selected through grid search, where different combinations are tested and the results are manually evaluated by listening to the generated music.

We now discuss the evaluation methods for conditional music generation.

\section{Evaluation}

Quantitatively evaluating our temporal conditioning mechanism is challenging. As mentioned earlier, forcing the model to place a chord \textit{exactly} at the conditioning boundary could disrupt the rhythm of the generated music. Instead, our goal is to guide the model to place a chord in the \textit{vicinity} of the temporal boundary, ensuring it fits rhythmically within the rest of the generated sequence. Moreover, we also want to allow the model to ignore a particular conditioning boundary if it is impossible to place a chord near it without disrupting the rhythm. Therefore, these temporal conditions are soft rather than hard constraints. We do not aim to maximize an accuracy metric; instead, our goal is to achieve a subjective sense of synchronization in the context of video-based music generation. To assess this, we opt for a subjective evaluation where we present the model's outputs overlaid with various input videos. This evaluation is further detailed in Section~\ref{sec:video_music_eval}, where we use a user survey to gather feedback. One of the survey questions specifically addresses temporal alignment, asking, "How well does the music align with the video in terms of rhythm and timing?" This question helps evaluate the effectiveness of our temporal boundary conditioning method.

We evaluate our emotion-conditioning methods using both qualitative and quantitative approaches. As later detailed in Section~\ref{sec:video_music_eval}, our video-based user study includes the question: "How well does the music match the video in terms of emotion?" For the quantitative evaluation, we isolate the emotion-conditioning mechanism by disabling temporal conditioning. We evaluate our emotion-conditioning methods using the metrics negative log-likelihood (NLL), top-1, and top-5 accuracies. While measuring top-$n$ accuracy, for each token, the model's output is considered accurate if the ground-truth class is within the top $n$ probabilities of the model's output. The evaluation configuration is the same as the training, namely using chunks with a length of 1216 tokens, and calculating the loss for every single token in the target sequence. This is much more challenging than only predicting the next token given the full sequence, since, at the extreme, the model tries to predict the first note of a song, only given the \texttt{START} and \texttt{BAR} tokens. We ensure that the entirety of the test split is used by sequentially taking non-overlapping chunks, resulting in 1836 chunks overall.

\begin{figure}[htbp]
	\centering
	\includegraphics[width=0.7\textwidth]{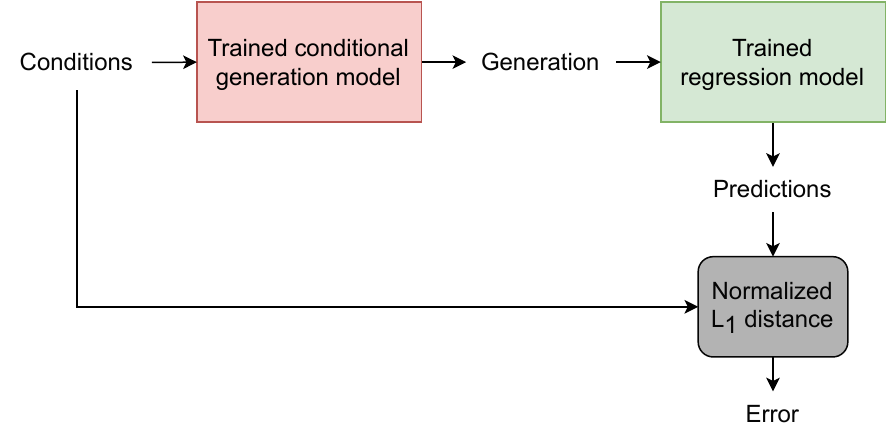}
	\caption{Pipeline for evaluation of emotion representation.}
	\label{fig:eval_inference}
\end{figure}

We additionally perform a quantitative evaluation on samples generated by our conditional models, by analyzing their emotional content, as done by \textcite{emopia}. To this end, we first train a regression model to predict the emotion values of the samples from the training data split. The architecture of the regression model is a music transformer with 8 layers, and the final layer outputs two continuous values, namely the valence and the arousal. Then using the trained conditional generation models, we perform inference using a collection of conditions, and later predict the emotional content of the generated samples using the trained regression model. As the error metric, we use the normalized $L_1$-distance between the predictions of the regression model and the conditions that were fed during inference. To make a fair comparison against the \textit{discrete-token} model, the condition values are chosen as the midpoints of the bins used by the discrete condition tokens, namely -0.8, -0.4, 0, 0.4, and 0.8. Using a combination of 5 values for valence, and 5 values for arousal, we end up with a collection of 25 condition value pairs. For each model and each condition, we generate 8 samples without "cherry-picking" and report the average error. Each sample has 4096 tokens. The regression model takes inputs with a length of 1216 tokens, similar to the generator. Samples are fed into the regression model using a sliding window with 50\% overlap, and outputs are averaged. The overall scheme for evaluating generated samples is visualized in Figure \ref{fig:eval_inference}.

We also make the generated samples available online\footnote{\url{https://www.serkansulun.com/midi}}. As explained previously, using the conditional models, we generate 200 samples for each. Since the generated melodies have a fixed number of tokens, their durations in time are inversely proportional to their tempo. The melodies generated with the lowest arousal conditions results in the slowest tempo, and are on average 159.6 seconds long. We also generate 200 more samples using the vanilla model, with no conditioning. Overall, we present 800 samples. The midi files are rendered into mp3 format using the Fluidsynth software and FluidR3\_GM soundfont\footnote{\url{https://archive.org/details/fluidr3-gm-gs}}.

\section{Results and discussion}
\label{sec:music_results}

We now present the results and discussion of our emotion-conditioned music generation methods. As explained earlier, the results for the temporal-conditioning mechanism are discussed later in Section~\ref{sec:results_video_music}.

The performance of the emotion-conditioning methods according to the prediction accuracy-based evaluation can be seen in Table \ref{table:eval}. NLL refers to negative log-likelihood, where lower is better. Top-1 and Top-5 refer to the accuracy, where higher is better. The \textit{continuous-concatenated} method outperforms other method, including the state-of-the-art \textit{discrete-token} method, across all metrics.

\begin{table}[H]
	\caption{Performance of our methods during evaluation.}
	\label{table:eval}
	\centering
		\begin{tabular}{l|ccc}
			Model                    & NLL    & Top-1  & Top-5  \\  \hline 
			vanilla                     & 0.7445 & 0.7784 & 0.9513 \\ 
			discrete-token              & 0.7375 & 0.7885 & 0.9536 \\ 
			continuous-token            & 0.7122 & 0.7895 & 0.9545 \\ 
			continuous-concatenated     & \textbf{0.7075} & \textbf{0.7913} & \textbf{0.9548} 
		\end{tabular}
\end{table}

In Table \ref{table:infer} we show the results for the regression-based evaluation. Error refers to the normalized $L_1$ distance between conditions fed during inference and the output of the trained regression which consumes the generated samples. We demonstrate that \textit{continuous-concatenated} method outperforms others in terms of its ability to convey emotion. Note, here the vanilla approach is not included since it is not conditioned on any emotion information.

\begin{table}[htbp]
	\caption{Performance of our methods during inference.}
	\label{table:infer}
	\centering
	\begin{tabular}{l|c}
		Model             & Error           \\ \hline 
		discrete-token    & 0.2164          \\ 
		continuous-token  & 0.1951          \\ 
		continuous-concatenated & \textbf{0.1948} 
	\end{tabular}
\end{table}

\sloppypar
When comparing the performance of the presented methods, we speculate that the main shortcoming of the \textit{discrete-token} and \textit{continuous-token} methods, in contrast to the \textit{continuous-concatenated} method, is their equal treatment of condition values and sequence tokens. While each token in the sequence is primarily useful for making local predictions (i.e., predicting tokens nearby), the condition values have a global impact, as they directly influence the entire generated sample. Our proposed \textit{continuous-concatenated} method maximizes the utility of condition information by incorporating it into each embedding of the input sequence for the transformer. Both of our proposed methods can also use continuous-valued conditions, offering finer control over the generation process. 

%
For reproducibility and to help future research, we open-source our dataset and the code that we used to prepare it\footnote{\url{https://github.com/serkansulun/midi-emotion}}. In the NLP community, training large transformers from scratch is a rare practice that is typically replaced by transfer learning, namely by fine-tuning open-source pre-trained models. However, a similar phenomenon does not exist in the field of symbolic music generation. Thus, we additionally open-source our trained models to allow other researchers to cut down on the time and resources for training, with transfer learning. To the best of our knowledge, ours are the largest open-source symbolic music generation models, that are trained on the largest multi-instrument symbolic music dataset, in the literature.

The conditions used in this chapter were either derived from the MIDI data or manually provided during inference. In Chapter~\ref{chap:video_music}, we show how these same conditions are automatically extracted from input videos and then passed into our conditional music generator. Before introducing the full video-based system, however, we first describe how these conditions are obtained from video. We now turn to our video analysis methods.

\chapter{Video analysis} \label{chap:video}

In this chapter, we present our video analysis methods, which produce the inputs for the conditional music generator described earlier. We remind that our overall approach follows a backward design: we first develop the music generation component and then construct the video analysis models to ensure musically meaningful outputs at every stage.  Specifically, we analyze videos from two complementary perspectives: low-level temporal boundaries and high-level semantics. For the temporal analysis we detect scene cut boundaries, while the semantic analysis involves emotion and genre classification. In the final stage of our research, these video-derived features will enable the generation of music that aligns strong chord boundaries with scene cuts, while also reflecting the video's emotional tone.

As explained in Section~\ref{sec:video-processing}, scene cut extraction is a solved problem. Therefore, we concentrate our research on the more challenging task of semantic video classification. Our video-based music generator relies on a video emotion classifier; however, as noted in Sections~\ref{sec:intro_video} and~\ref{sec:video-processing}, the limited size of video emotion datasets constrains model complexity and often necessitates using a fixed and small number of input frames. To address this, we first design an emotion classifier suited for these constraints. We then apply the same architectures to trailer genre classification—a task with significantly more data—allowing us to explore the model's capacity beyond the limitations imposed by emotion datasets.

\section{Scene cut extraction}
\label{sec:scenecut}

In our video-based music generator, we use video boundaries in the form of scene cuts, along with high-level emotions. Since scene cut detection is a well-established problem, we use the FFmpeg software for this task. We identify a scene cut when the average pixel difference between consecutive frames exceeds 27\%.

To generate musical chords near the scene boundaries, we aim to avoid an excessive number of close boundaries. Therefore, we apply a difference filter to the extracted scene cuts and remove those occurring less than 4 seconds apart. We determined these parameters through trial and error, as well as manual inspection.

We continue with our methodology for semantic video classification.

\section{Semantic video classification}

Both video emotion classification and trailer genre classification fall under the category of semantic video classification. Since both tasks involve using audiovisual data for predictions, we employ similar architectures for both. The video emotion classifier, being integrated into the final video-based music generator, incorporates the most recent updates and adjustments compared to the trailer genre classifier. We highlight these differences where they occur.

Most existing video classification methods use raw pixels and audio as inputs, but this high-dimensional input space can lead to overfitting when data is limited, as discussed in Sections~\ref{sec:intro_video} and~\ref{sec:video-processing}. Moreover, videos are inherently multimodal, involving elements like objects, scenes, speech, music, sound events, and on-screen text—all of which can contribute to semantic understanding. However, capturing all of these modalities with a single model is a major challenge.

To address this, we use learned features extracted from a variety of pretrained deep neural networks (DNNs). These features are compact, helping to reduce input dimensionality and mitigate overfitting. Because they are derived from models already trained on large datasets, they contain rich, semantically useful information for video analysis. Importantly, we use these pretrained models in forward mode only—without any fine-tuning—significantly lowering memory and computational requirements. We then train lightweight models on top of these features to make predictions, keeping training efficient while still leveraging high-level representations. We now introduce the pretrained models which we use for feature extraction.

\subsection{Feature extraction}

We use the following pretrained models for feature extraction, taking activations from the layer just before the final classification layer, except in the case of text. This approach provides richer and more informative representations than using final predictions alone. For text, we first extract raw output and then process it through additional text models, again using the penultimate layer activations for feature representation.

We note that some of the pretrained models we use also analyze on-screen music. Although our primary task is music generation, we incorporate music analysis models to enhance video classification performance and remain competitive with state-of-the-art methods, while also enabling our classifiers to be applied to tasks that don't involve music generation. Furthermore, our video emotion classifier does not rely on a model dedicated exclusively to music analysis; instead, it uses a unified, state-of-the-art model for general audio analysis. We do not attempt to separate music from other audio sources, as this remains an open research challenge~\parencite{sound_separation}.

\paragraph{CLIP}
Contrastive Language-Image Pretraining (CLIP) is a state-of-the-art model for image understanding, pretrained using contrastive learning with a large collection of images and their associated captions from the internet \parencite{clip}. The pretrained model first resizes the image so that the longer side has 224 pixels, followed by a 224×224 center crop. Then, the model encodes each frame into a 512-dimensional vector.

\paragraph{Audiotag}
To extract audio features, we use a model pretrained for audio event tagging \parencite{audiotag}. This model processes 3-second audio chunks and outputs a probability vector for 527 different labels. We extract the activations before the classification layer, resulting in 128-dimensional vectors for each audio chunk.

\paragraph{Music}
Since cinematic trailers often feature significant soundtracks, it is essential to investigate the musical aspect further. To capture this, we extract a musical feature using a model pretrained for music genre classification \parencite{musicnet}. This model processes 22-second audio chunks and outputs a probability vector for 50 different labels and 64-dimensional activations for each audio chunk. 

We used the Audiotag and Music models only for genre classification and later replaced them with the following improved and unified audio analysis model for emotion classification.

\paragraph{BEATs} Bidirectional Encoder representation from Audio Transformers (BEATs) is an audio classification model that combines an acoustic tokenizer and a classifier, which are trained iteratively \parencite{beats}. This model is capable of classifying both audio events and music genres.

\paragraph{Face detector} Human faces are detected and extracted using the model from the Ultralytics group\footnote{\url{https://github.com/ultralytics/ultralytics}}, which is based on YOLO (You Only Look Once) \parencite{yolo}. YOLO is a real-time object detection algorithm that divides an image into a grid and predicts bounding boxes and class probabilities for each grid cell using convolutional neural networks. This model was later included for emotion classification and was not used for genre classification.

\paragraph{Expression classifier} The cropped human faces are then fed into a facial emotion classifier\footnote{\url{https://huggingface.co/trpakov/vit-face-expression}}. This model, a Vision Transformer (ViT) \parencite{vit}, was trained on the FER-2013 (Facial Emotion Recognition) dataset \parencite{fer}. The model takes a facial image and predicts the facial expression as angry, disgusted, fearful, happy, sad, surprised, or neutral. This model was later included for emotion classification and was not used for genre classification.

\paragraph{Automatic Speech Recognition (ASR)} OpenAI's Whisper model processes audio input, detects speech patterns, and outputs text, as done in the genre classification task \parencite{whisper}. We use the larger variant of the model which also translates non-English text into English.

\paragraph{Optical Character Recognition (OCR)}
We also performed optical character recognition (OCR) on each frame using the PaddleOCR model\footnote{\url{https://paddlepaddle.github.io/PaddleOCR/main/en}}. We additionally used a pretrained spell correction model on the produced output\footnote{\url{https://huggingface.co/oliverguhr/spelling-correction-english-base}}. The overall output of this stage is in text format. We improve the OCR output by incorporating the following additional post-processing models.

\paragraph{Language identification and translation} 
We translate non-English OCR text into English as the final text encoding models are trained on English text. We first use a language identification model\footnote{\url{https://huggingface.co/papluca/xlm-roberta-base-language-detection}} to detect if translation is necessary. This model, based on XLM-RoBERTa \parencite{xlmroberta}, is trained on multiple language identification datasets \parencite{language_identification1,language_identification2,language_identification3}.

The non-English OCR text is then translated into English using Facebook's NLLB-200 (No Language Left Behind) model \parencite{translation}. This model uses a combination of transformer \parencite{transformer} and Sparsely Gated Mixture of Experts \parencite{dynamic} layers, trained on internet-sourced text data, to translate between 200 languages. While the NLLB-200 model requires the source language to be specified, the language identification model automatically detects it. As a large language model, NLLB-200 can translate long texts and yield error-free outputs even if the input contains spelling errors. However, for English OCR output, we use the following correction models.

\paragraph{Spell correction} The Symspell package provides tools for spell checking\footnote{\url{https://github.com/wolfgarbe/SymSpell}}. Among these tools, the word segmenter separates words in sentences where spaces are missing, which is useful for post-processing OCR outputs that lack spaces. We also use a spellcheck model to correct potential typos\footnote{\url{https://huggingface.co/ai-forever/T5-large-spell}}. This is a T5 transformer language model \parencite{t5} trained on a dataset containing synthetic spelling errors, allowing it to correct any English text. After obtaining the final text output from ASR and OCR, we encode them to obtain feature vectors, using pretrained language models.

\paragraph{DistilBERT}
For genre classification, we encode the output text of the OCR and ASR models using a pretrained language model, specifically DistilBERT \parencite{distilbert}, which is a condensed and compressed variant of the BERT (Bidirectional Encoder Representations from Transformers) model \parencite{bert}, achieved through knowledge distillation \parencite{distillation_2006,distillation_2015}. DistilBERT utilizes fewer layers than BERT and learns from BERT's outputs to mimic its behavior. This model converts the input text into tokens, including words and sub-words, and encodes each token as a 768-dimensional vector.

\paragraph{Text sentiment classifier} 

For emotion classification, we use another language model that is trained for sentiment analysis. We use a RoBERTa language model \parencite{roberta} that was trained on the TweetEval sentiment classification benchmark \parencite{sentiment}\footnote{\url{https://huggingface.co/cardiffnlp/twitter-roberta-base-sentiment}}. This model can predict the sentiment of a given text as positive, negative, or neutral. As with the previous features, we utilize the activations before the final classification layer.

After introducing our pretrained features, we now discuss the models used to process these features and make final predictions for the emotion and genre classification tasks.

\subsection{Emotion classification}
\label{sec:emotion_classifier}

In this task, we aim to classify input videos into emotion categories. We name our model VEMOCLAP, which stands for \textbf{V}ideo \textbf{EMO}tion \textbf{CLA}ssifier
using \textbf{P}retrained features. We begin by describing the dataset used and our approach for extracting pretrained features from it.

\subsubsection{Dataset and feature extraction}

We use the Ekman-6 dataset, as it is one of the largest emotion-labeled datasets that includes arbitrary user-generated videos \parencite{ekman6}. We avoid using datasets that feature more specific types of videos, such as those focused on human faces or social interactions, because we want our final video-based music generator to be applicable to any type of video. The Ekman-6 dataset contains 1,637 short videos, split equally for training and testing. Each video is labeled with one of the six basic emotions \parencite{ekman}: anger, disgust, fear, joy, sadness, and surprise.

We extract a fixed number, $n$, of frames from each video, along with the entire audio, which is resampled at 16 kHz and converted to mono. The features from the facial expression classifier and CLIP are sequences of vectors, as their inputs are sequences of frames. If multiple faces are detected in a single video frame, we average the features of the two largest faces. Similarly, the BEATs model processes sequences of 3-second audio chunks. We extract $n$ audio chunks to match the number of video frames. While CLIP and BEATs produce $n$ output vectors, the facial expression classifier may produce fewer vectors, depending on whether faces are present in each frame.

The ASR model generates a single block of text from the entire audio. If the source language is not English, it automatically translates the output text into English. 
The OCR model generates blocks of text for each video frame. The language identification model processes each block, and the text is translated into English if the identified language is not English. If the language is already in English, the text is passed through the word segmenter and then the spell corrector. The resulting text blocks from each frame are concatenated to form a single block of text. The texts resulting from ASR and OCR are fed into the sentiment classifier separately. Since the sentiment classifier predicts a single label for any length of text, a single vector is extracted as the feature.

After feature extraction, for a single video, we are left with $n$ CLIP, $n$ BEATs, $k \leq n$ facial expression, 1 OCR sentiment, and 1 ASR sentiment feature. Fusing these pretrained features presents several challenges. First, since they are extracted using different pretrained models, the lengths (dimensionalities) of the feature vectors vary. Second, when the feature vectors form a sequence, as in the case of facial expressions, CLIP, and BEATs, their temporal lengths can also differ. Finally, since these features belong to different modalities, the content of the feature vectors can be vastly different. We now present the method we use to fuse the pretrained features and produce the final emotion predictions.

\subsubsection{Classification method}

To handle the value range difference of the multimodal features, we first perform min-max normalization on each feature using statistics extracted from the collection of pretrained features. To handle differing dimensionalities, all sequential input features are first projected to queries, keys, and values with a common dimensionality~\parencite{transformer}. Next, the cross-attention modules~\parencite{attention} exploit correspondence between pairs of sequential features. The attention modules also include dropout and layer normalization and can handle a pair of sequences with different temporal lengths. As done in classification tasks, the attention outputs are averaged along the temporal dimension, yielding a single vector. Since the OCR and ASR sentiment features are already single vectors, each modality is represented by a single vector after the attention modules. We then concatenate all five feature vectors along the channel dimension, resulting in a single vector representing the entire video. Finally, this vector is fed into a linear layer followed by a softmax layer, which outputs a probability for each emotion. 

\begin{figure}[t]
	\centerline{\includegraphics[width=0.6\textwidth]{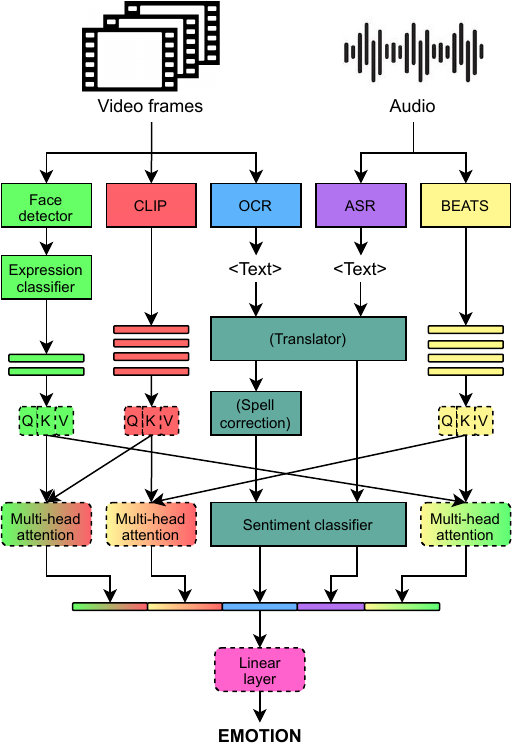}}
	\caption{Video emotion classification pipeline. }
	\label{fig:emotion_classifier}
\end{figure}

Our model is shown in Figure~\ref{fig:emotion_classifier}. Blocks with rounded and dashed outlines represent trained modules. The models with parentheses are used conditionally. Other blocks are pretrained feature extractors and are used in inference mode. Q, K, and V represent query, key, and value projections.

\subsubsection{Implementation details and hyperparameters}

We initially extract video frames at $1$ frame per second, using $n=16$ video frames and corresponding audio chunks as input to our model. Due to the limited size of the Ekman-6 dataset, we adopt this small, fixed input length to reduce the risk of overfitting. In the subsequent genre classification task, where more data is available, we lift this constraint.

During training, we select the $n$ video frames and audio chunks from random locations for data augmentation. During testing and inference, we extract them at equidistant intervals to ensure comprehensive temporal representation. We report classification performance using the provided training and testing splits, which included $819$ and $818$ videos, respectively.

We use cross-entropy loss, a batch size of $32$, a dropout rate of $0.5$, and Adam optimizer with a learning rate of 1e-5 \parencite{adam}. Attention modules have $4$ heads and a dimensionality of $512$. The model has around 11M trainable parameters. We used $10\%$ of the training split as the validation split and stopped training when validation accuracy started to drop. 


\subsubsection{Dataset cleaning}
\label{cleaning}
Our initial examination of the Ekman- 6 dataset revealed several problematic samples. While we report classification results on the unedited Ekman-6 dataset, we also clean it to train the model used in our video-based music generator later in Chapter~\ref{chap:video_music}. The Ekman-6 dataset was created by scraping the web for videos using search keywords that matched not only the categorized emotion but also related terms. We view each video to detect the problematic samples. After inspecting their file names, we identify the following problematic search keywords for each emotion class, which are \underline{underlined}.

\textbf{Anger}: A single person being \underline{annoy}ing, with no other person present to be annoyed or angry.

\textbf{Disgust}: Flashing lights or rapid camera movement, presumably to induce \underline{dizzy}ness or \underline{nausea}. It also includes videos related to \underline{boredom} and \underline{loathing}.

\textbf{Fear}: Counter-\underline{terror}ism, \underline{underwater} footage, 9/11 \underline{terror}ism attack aftermath, and suspect \underline{apprehension}.

\textbf{Joy}: \underline{Joy}ride (driving a car), the music "Ode to \underline{Joy}", and people named \underline{Joy}.

\textbf{Sadness}: \underline{pensive}

\textbf{Surprise}: \underline{distraction}, and people performing impressive feats labeled as \underline{astonishing}.

We identify and remove $128$ and $130$ problematic videos from the training and testing splits, respectively. Using the cleaned data, the classification accuracy increases by $2.6\%$. However, we exclude this result from our comparison with the state-of-the-art because data cleaning alters the test split's content, affecting the comparison's fairness.

\subsubsection{Inference web application}

As a complementary work, we present an open-source web application for performing inference on user-provided videos, designed to support both the general public and the research community\footnote{\url{https://www.serkansulun.com/app}}. Hosted on Google Colab, it offers free GPU runtime. Users can upload their own videos, provide a YouTube link, or use sample videos provided within the application. The application is self-contained and ready to use, requiring no setup from the user. The process is streamlined into 5 steps, with only 5 mouse clicks needed to obtain the results. After connecting to a GPU runtime, users should follow these steps:

\textbf{Step 1:}
Automatically download and extract the codebase, and install the required Python libraries. This step takes approximately $2$ minutes.

\textbf{Step 2:}
Download and build the feature extractor models and the classifier model. As these models are deep neural networks, this step takes about $3$ minutes. Note that Steps 1 and 2 only need to be completed once, even if classifying multiple videos.

\textbf{Step 3:}
Select how to load the input video. The options are "Sample video", "YouTube link", and "Upload video".

\textbf{Step 4:}
Depending on the choice from step 3, the user then selects the specific video. Steps 3 and 4 take only a few seconds to complete.

\textbf{Step 5:}
Extract the frames and audio from the input video, run the pretrained feature extractors, and finally run the emotion classifier. The outputs include text from automatic speech recognition (ASR) with its sentiment, text from optical character recognition (OCR) with its sentiment, and predictions from the BEATs audio classifier. Additionally, a sample frame is displayed showing detected faces with predicted emotional expressions, detected OCR boxes, and a caption generated by CLIPCap \parencite{clipcap}. Note that this sample frame is for demonstration purposes, while all $n$ frames are used for the final emotion classification. For a $60$-second video, this step takes approximately $30$ seconds.

\subsubsection{Results and discussion}

\begin{table}[h]
	\centering
	\caption{Classification accuracies compared to the state-of-the-art on the Ekman-6 dataset.}
	\begin{tabular}{l|c}
		Method  & Accuracy (\%)       \\ \hline
		ITE \parencite{ekman6}     & 51.20          \\
		CFN \parencite{cfn}    & 51.80          \\
		MART \parencite{mart}   & 53.17          \\
		VAANet \parencite{vaanet}  & 55.30          \\
		CTEN \parencite{cten}   & 58.20          \\
		KeyFrame \parencite{keyframe} & 59.51 \\
		LRCANet \parencite{lrcanet} & 59.78          \\
		FAEIL \parencite{faeil} & 60.44          \\
		TAM \parencite{tam} & 61.00          \\
		\textbf{VEMOCLAP (Ours)} & \textbf{65.28}
	\end{tabular}
	\label{table:emotion_results}

\end{table}

In Table \ref{table:emotion_results}, we present the quantitative performance of our model on the Ekman-6 dataset using the provided training and testing splits, showing that our method outperforms the state-of-the-art by $4.3\%$. 

\begin{figure}[h]
	\centerline{\includegraphics[width=0.7\textwidth]{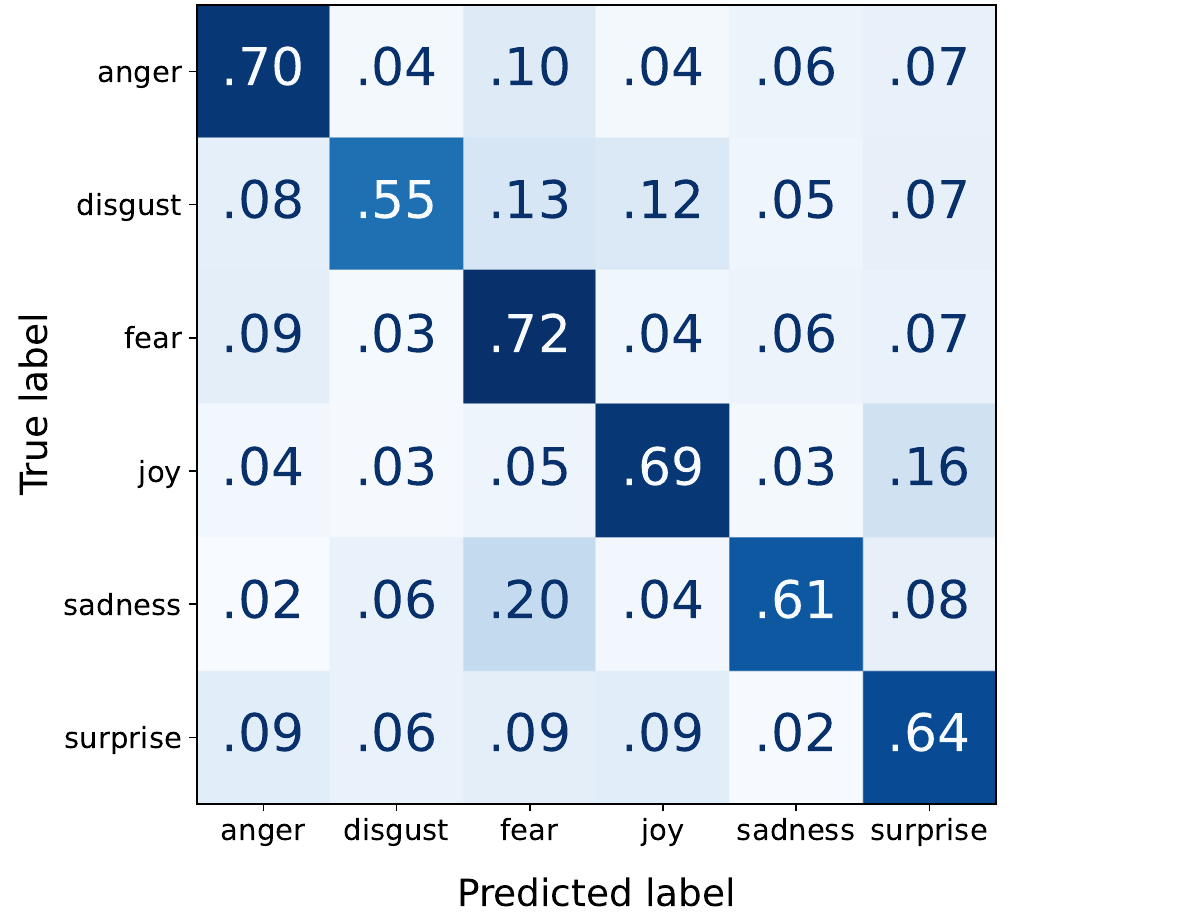}}
	\caption{Confusion matrix with values normalized over true labels on the test split of Ekman-6 dataset.}
	\label{fig:matrix}
\end{figure}

Figure \ref{fig:matrix} shows the confusion matrix for our classification results on the test split. Disgust has the lowest classification accuracy among the six emotions. This may be because, unlike the others, disgust is highly influenced by social conditioning and shaped by cultural moral values~\parencite{disgust}. 20\% of videos labeled as sadness are misclassified as fear, likely due to the shared negative emotional tone of both categories. Similarly, 16\% of videos labeled as joy are misclassified as surprise. Based on our manual inspection, this may be because most videos in the surprise category convey a positive affect---often depicting individuals who are not only surprised but also joyful.

To the best of our knowledge, our inference web application is the first click-and-run software for video emotion classification. The trained VEMOCLAP is also incorporated into our video-based music generation pipeline as we will explain in Chapter \ref{chap:video_music}.

\subsection{Genre classification}
\label{sec:genre_classification}

We continue with another semantic task, namely genre classification. In this task, we exploit a large-scale cinematic dataset to challenge the limitations on number of input frames.

\subsubsection{Dataset and feature extraction}

We use the MovieNet cinematic dataset, which includes metadata for 375k movies and YouTube trailer links for 33k of them, making it the largest labeled trailer dataset available. Importantly, this dataset is multi-label, indicating that an individual video can be associated with multiple genres.

\begin{figure}[htbp]
	\centering
	\includegraphics[width=0.5\columnwidth]{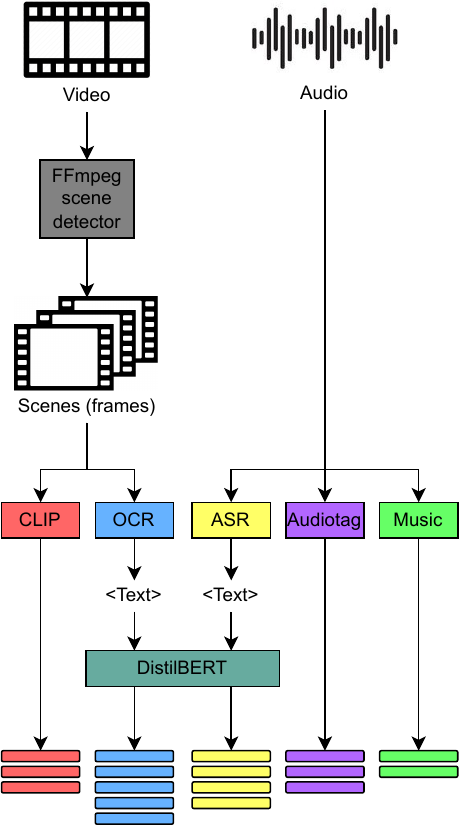} %
	\caption{Video feature extraction pipeline for genre classification.}
	\label{fig:genre_features}
\end{figure}

We extract CLIP, OCR, ASR, Audiotag, and Music features as illustrated in Figure~\ref{fig:genre_features}. Unlike the emotion classification pipeline, we process the text features using the general-purpose language model DistilBERT, which produces a sequence of feature vectors corresponding to each token (word or subword).

\subsubsection{Classification methods}

We build multiple models to take the previously extracted features of a given video as input and predict the genre of the video. Combining different features is challenging, especially since the encoded vectors have different lengths in both channel and temporal dimensions. For example, considering the audio event and music features, the lengths of each vector are $128$ and $64$, respectively. Furthermore, since the audio event and music networks operate on chunks of audio with lengths of $3$ and $22$ seconds respectively, for the same video, there are more embedded vectors for the former. And finally, the number of vectors for the same feature differs between different videos. We address these issues using different solutions and different classification models.

We build and train three distinct models to fuse and classify pretrained features: \textit{a multi-layer perceptron (MLP)}; the \textit{single-transformer} model that integrates features across all modalities; and the \textit{multi-transformer} model, where individual transformers handle features from specific modalities. As a baseline, we also implement a simpler model that does not utilize diverse pretrained features and instead uses two branches to independently process raw video frames and audio.

The final layer of all our models is a fully-connected (FC) layer with a size of $21$, outputting probabilities belonging to $21$ different genres. This layer is followed by a sigmoid layer to make sure each output is a probability between $0$ and $1$.

\paragraph{MLP}

\begin{figure}[htbp]
	\centering
	\includegraphics[width=0.75\columnwidth]{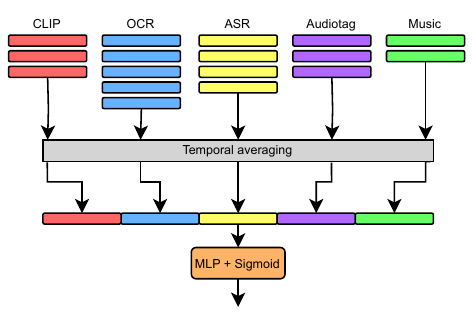} %
	\caption{Our MLP model.}
	\label{fig:mlp}
\end{figure}

We first implement a simple MLP classifier, mirroring the state-of-the-art approach used by \textcite{movienet}. While their method processes raw video and audio from a few short segments, we instead use pretrained features extracted from the entire video. Since MLPs require fixed-length input, we follow a similar strategy by averaging the feature vectors from each modality over time. These averaged vectors are then concatenated and passed to the MLP. Our architecture is illustrated in Figure~\ref{fig:mlp}.

In our two other models namely single-transformer and multi-transformer, we optionally use the temporal averaging for text features such as OCR and ASR. While non-textual features, namely CLIP, Audiotag, and Music are obtained from the activations \textit{before} the final prediction layer, the OCR and ASR features stem from running the DistilBERT model on the predicted text. Here, any error in the predicted text is propagated into the DistilBERT model, potentially corrupting the output features. In order to reduce the resulting noise, we experiment with averaging the textual features, namely OCR and ASR, along the temporal dimension.

\paragraph{Single-transformer}

To mitigate the significant information loss caused by averaging features along the temporal dimension, we employ a transformer model to capture both short- and long-term correspondences within the video sequence. We use the transformer as a sequence classifier by prepending the sequences with a special learnable vector, known as the \texttt{<CLS>} vector. While the transformer generates output vectors for each element in the input sequence, in classification tasks, only the output vector corresponding to the \texttt{<CLS>} vector is passed to the next layer, and the others are discarded \parencite{bert}.

\begin{figure}[htbp]
	\centering
	\includegraphics[width=0.75\columnwidth]{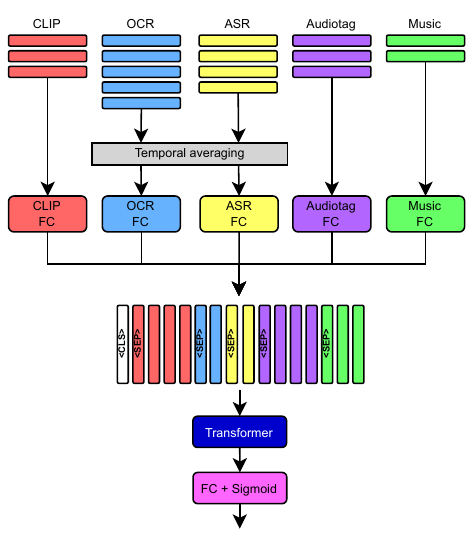} %
	\caption{Our single-transformer model.}
	\label{fig:single}
\end{figure}

While the transformer can handle sequences with varying lengths, the vectors in the sequence still need to have equal sizes along the channel dimension. To ensure that, we use fully-connected layers to transform feature vectors from different modalities into the same size. We note that this layer is applied to each vector in the sequence individually, hence not changing the number of vectors in its input sequence. Next, we concatenate all the vectors along the time dimension before feeding them into a single transformer. This approach aims at exploiting the correspondences between different features at the temporal level by processing all of them as a single sequence. However, a trade-off is the complexity of handling different modalities with a single transformer block using a single set of weights.

To make sure the transformer can distinguish feature vectors from different modalities, we use separately learned positional embeddings for sequences from each modality~\parencite{learned_position}. We additionally make use of separator (\texttt{<SEP>}) vectors, namely, an encoded version of the \texttt{<SEP>} token~\parencite{bert}. These are learned vectors that are specific for each modality, prepended to each sequence. Finally, as in almost all classifier transformers, we prepend the input sequence to the transformer with a learnable classifying (\texttt{<CLS>}) vector, namely, an encoded version of the \texttt{<CLS>} token \parencite{bert}. The output vector corresponding to this classifying vector, i.e. the first vector, is considered as the prediction of the transformer and fed into the final fully-connected layer, while the remaining vectors of the transformer's output sequence are discarded. The overall model is shown in Figure \ref{fig:single}.

\paragraph{Multi-transformer}

\begin{figure}[htbp]
	\centering
	\includegraphics[width=0.75\columnwidth]{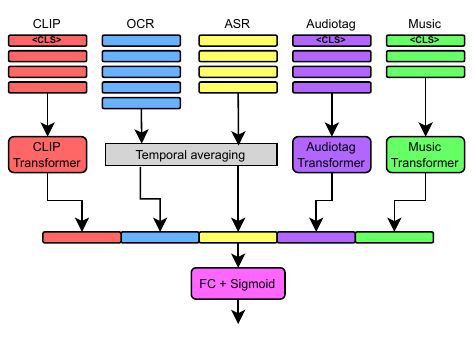} %
	\caption{Our multi-transformer model.}
	\label{fig:multi}
\end{figure}

Given the hypothesis that using a single transformer to process features from multiple modalities can be inefficient due to the increased complexity of the input, we devise a final model that incorporates separate transformer models for each modality, as shown in Figure \ref{fig:multi}. The inputs to all transformers are prepended with \texttt{<CLS>} token vectors, and the corresponding output vectors are concatenated along the channel dimension before being fed into a fully connected layer to obtain the final probabilities. This approach leverages potential correspondences between different features at the global level, though not at the temporal level.

\paragraph{Vision transformer baseline model}

To evaluate our approach of using multimodal pretrained features, we also implement a strong baseline for comparison. Unlike our main models, this baseline doesn't incorporate a wide range of pretrained features, but it still leverages powerful pretrained networks along with additional trainable layers. It operates on sequences of 2-dimensional raw visual frames and audio spectrograms, in contrast to the previously introduced models that work with sequences of 1-dimensional pretrained features. 

Our baseline model uses the Vision Transformer (ViT) to divide each video frame into patches, encode them, and process the resulting sequence of vectors with a standard transformer architecture~\parencite{vit}. For audio, it employs the Audio Spectrogram Transformer (AST), a variant of ViT designed to operate directly on audio spectrograms~\parencite{ast}. Despite being purely attention-based, without any convolutional layers, ViT outperforms convolutional neural networks, achieving state-of-the-art results in image \parencite{vit}, video \parencite{vivit}, and audio \parencite{ast} classification. 

While our primary objective is video classification, directly applying the video vision transformer (ViViT) \parencite{vivit} to our task is not a suitable approach for two main reasons. First, ViViT does not incorporate audio, which is a crucial aspect of our task. Second, our dataset contains discontinuous frames that depict different scenes with no visual continuity between them. As a result, it is more appropriate to handle these frames individually rather than concatenating patches from different frames, as ViViT does.

The ViT and AST models are pretrained on the ImageNet~\parencite{imagenet} and AudioSet~\parencite{audioset} datasets, which are large-scale benchmarks for image and audio classification, respectively. In our preliminary experiments, we attempted to train the entire model, including fine-tuning ViT and AST, but observed severe overfitting that could not be alleviated by standard regularization techniques such as dropout~\parencite{dropout}. To avoid this, we keep the parameters of the ViT and AST frozen and replace their output layers with trainable MLPs, naming them image-MLP and audio-MLP.

The pretrained AST is designed to work with audio spectrograms organized into 10-second segments, which we refer to as "spectrogram frames." To distinguish them, we use the term "image frames" to describe the visual scenes in our context. Since the trailers in our dataset are longer than $10$ seconds, we split the full audio spectrogram into 10-second chunks, with a $50\%$ overlap. Similarly, the ViT works with individual image frames, while our trailers consist of multiple frames. This leads to an output sequence of vectors, where each vector corresponds to an individual image or audio frame. To process these sequences and obtain the final predictions, we produce a fusion module. To this end, we use standard transformer classifiers for both image and audio sequences, yielding a single vector for each modality. Finally, we concatenate these two vectors and process them using a linear layer and a sigmoid layer to obtain the final label predictions.

\begin{figure}[htbp]
	\centering
	\includegraphics[width=0.6\columnwidth]{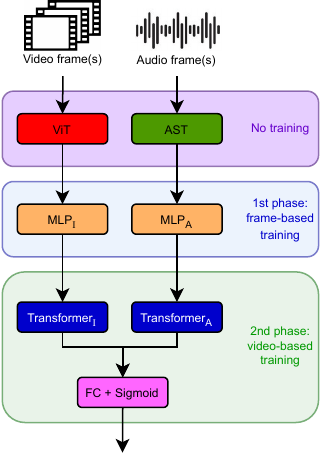} %
	\caption{Our baseline model. The subscripts "I" and "A" refer to image and audio, respectively.}
	\label{fig:baseline}
\end{figure}

Even when the parameters of the pretrained ViT and AST models are kept frozen, training the full model in an end-to-end fashion leads to overfitting. This is due to the relatively small size of our training dataset, which contains only 33k samples, while the original authors of ViT and AST train their models on extensive datasets such as Imagenet with 21M samples \parencite{imagenet} and AudioSet with 2M samples \parencite{audioset}, respectively. To address this overfitting issue we employ a two-stage training approach. First, we train the image- and audio-MLPs separately on individual frames. During this phase, the target is the label of the video to which the frames belong. Next, we freeze the parameters of the MLPs and train the fusion module using complete videos. The baseline model and its training scheme are illustrated in Figure \ref{fig:baseline}.

\subsubsection{Implementation details and hyperparameters}
We apply a filtering process to exclude extremely long and short videos from our dataset. Specifically, we compute the quartiles of video durations and exclude samples that were outside the inner fence, i.e., durations shorter than $Q1 - 1.5 \times IQR$ (corresponding to $19.6$ seconds) and longer than $Q3 + 1.5 \times IQR$ (corresponding to $214.4$ seconds). This filtering yields 26412 videos from the original set of 32647. Moreover, following the approach in the MovieNet paper \parencite{movienet}, we limit the labels to the 21 most frequent genres, namely, action, adventure, animation, biography, comedy, crime, documentary, drama, family, fantasy, history, horror, music, musical, mystery, romance, sci-fi, sport, thriller, war, and western.

To ensure an unbiased split of the dataset, we arranged the videos alphabetically by their YouTube IDs---given that these IDs are generated randomly. Following the methodology in the original MovieNet paper, we divide the dataset into training, validation, and testing splits with ratios of $0.7$, $0.1$, and $0.2$, corresponding to 18488, 2641, and 5283 samples respectively \parencite{movienet}. Hyperparameters are determined using a grid search, optimizing for the highest mean average precision on the validation split. The test split is utilized solely for reporting the final results.

While a straightforward way to balance precision and recall is to change the decision threshold, this method does not allow for comparison against the works in the literature where most report their results while setting a fixed decision threshold at $0.5$ \parencite{movienet}. Alternatively, during training, we can balance precision and recall by applying a constant weight to the loss associated with positive labels. For a single sample, the weighted binary cross-entropy loss becomes:

\begin{align*}
\textit{Loss} = -\frac{1}{C} \sum_{c=1}^{C} w \cdot y_c \cdot \log(p_c) + (1 - y_c) \cdot \log(1 - p_c)
\end{align*}

Here, $w$ represents the weight for positive labels and $C$ is the total number of classes. To align precision and recall values with those observed in MovieNet, we apply a weight of $0.25$.

Both the transformer architecture and the averaging module of the MLP architecture can handle sequences of varying lengths. However, during training, we use fixed-length inputs to facilitate minibatch processing. Longer sequences are truncated to the desired length, while shorter ones are padded with zero vectors. The length of feature vectors for each pretrained modality is determined through exploratory data analysis, specifically using box plots to analyze sequence lengths across all samples. The maximum sequence length is set to the upper adjacent value (upper whisker of the box plot), which is 1.5 times the interquartile range above the third quartile. As a result, the sequence lengths are defined as 216 for CLIP, 64 for OCR, 86 for ASR, 140 for Audiotag, and 18 for Musicnet. When using a single-transformer model, the concatenated sequence length totals 524. It is important to note that during inference on individual samples, our models can handle inputs of varying lengths without the need for padding or truncating.

For the MLP model, there is 1 layer, a model dimension of 256, and a total of 57k parameters. The single-transformer model consists of 2 layers, 8 attention heads, and a model dimension of 256, with a total of 8.56M parameters. The multi-transformer model features individual transformers, each with 1 layer, 8 attention heads, and a model dimension of 128, totaling 6.98M parameters. Considering the baseline model, both the image-MLP and audio-MLP have 2 layers and a dimensionality of 768. The fusing transformers each have 1 layer, 4 attention heads, and a dimensionality of 768. All transformers are trained with gradient clipping at a norm of 1. Models are trained using the Adam optimizer \parencite{adam}, with a learning rate of 1e-5, a dropout rate of 0.5, and a batch size of 32. The models are implemented using the PyTorch library \parencite{pytorch} and trained on a single NVIDIA Quadro RTX 6000 GPU with 24 GB of memory.

\subsubsection{Results and discussion}

We first compare the performance of our models against the baseline model and the models presented in the MovieNet paper \parencite{movienet}. Inference is performed on the test set one sample (video) at a time, using the full duration of each video without padding or truncating the sequences. Following the literature, we report macro-averaged precision, recall, and mean average precision (mAP) values, averaged over individual labels (genres).


Table \ref{table:overall} shows the overall performance of our models compared to models in the literature. In this table, P, R, and mAP denote precision, recall, and mean average precision respectively. For precision and recall, $0.5$ is used as the decision threshold. The results for the state-of-the-art models are taken from the original MovieNet paper~\parencite{movienet}. For clarity, the full names of the abbreviated models are: TSN (Temporal Segment Network) \parencite{tsn}, I3D (Two-Stream Inflated 3D ConvNets)~\parencite{i3d}, and TRN (Temporal Relation Network)~\parencite{trn}.

Our baseline model outperforms the models in the literature in terms of recall and mean average precision. Models using pretrained features outperform all other models across all metrics. Our best-performing model, the multi-transformer, significantly improves the state of the art. It outperforms all other models in every metric except precision, where it performs marginally worse than our MLP model.

\begin{table}[htbp]
	\centering
	
	\caption{Performance of our models against the state of the art.}
	\label{table:overall}
		\begin{tabular}{l|ccc}
			Models & P@0.5          & R@0.5          & mAP            \\ \hline \hline
			TSN   & 78.31 & 17.95 & 43.70 \\
			I3D   & 69.58 & 16.54 & 35.79 \\
			TRN   & 77.63 & 21.74 & 45.23 \\
			MovieNet & 79.74 & 24.97 & 46.88 \\ \hline
			Baseline & 73.78 & 32.17 & 59.73 \\ \hline
			MLP   & \textbf{82.05} & 33.51 & 63.16 \\
			Single-transformer & 81.00 & 37.11 & 65.09 \\
			Multi-transformer & 82.00 & \textbf{38.33} & \textbf{66.02} 
		\end{tabular}
	
\end{table}

In Table \ref{table:genres} we present a detailed analysis of the results across all genres. For conciseness, we only show the results for the best model in the literature MovieNet (MN)~\parencite{movienet}, our baseline (BL) model, and our best performing model multi-transformer (MT). P, R, and mAP denote precision, recall, and mean average precision respectively. For precision and recall, $0.5$ is used as the decision threshold.

\begin{table}[htbp]
	\centering
	
	\caption{Results per genre and the macro averages. MN: MovieNet~\parencite{movienet}, BL: baseline (ours), MT: multi-transformer (ours).}
	
	\label{table:genres}
	
		
		\begin{tabular}{l||ccc|ccc|ccc}
			& \multicolumn{3}{c|}{P@0.5}               & \multicolumn{3}{c|}{R@0.5}                     & \multicolumn{3}{c}{mAP}     \\ \hline
			& MN            & BL & MT          & MN             & BL      & MT          & MN            & BL & MT          \\ \hline \hline
			Action & 73.96 & 86.59 & \textbf{87.37} & 22.21 & 40.48 & \textbf{48.12} & 54.60 & 75.64 & \textbf{79.20} \\
			Adventure & 75.24 & 74.71 & \textbf{77.51} & 24.72 & 22.48 & \textbf{28.67} & 53.06 & 56.76 & \textbf{60.93} \\
			Animation & 93.16 & \textbf{96.38} & 96.19 & 74.09 & 86.59 & \textbf{92.28} & 86.45 & 94.65 & \textbf{96.18} \\
			Biography & \textbf{100.00} & 53.85 & 70.00 & 0.04  & 2.86  & \textbf{5.71} & 9.13  & 24.29 & \textbf{36.68} \\
			Comedy & 68.61 & 88.55 & \textbf{90.39} & 48.65 & \textbf{57.94} & 57.05 & 68.81 & 85.47 & \textbf{87.84} \\
			Crime & 74.12 & 72.00 & \textbf{80.22} & \textbf{39.30} & 8.33  & 22.53 & 49.25 & 50.81 & \textbf{60.25} \\
			Documentary & 85.49 & 91.30 & \textbf{92.37} & 4.79  & 70.87 & \textbf{81.71} & 21.03 & 90.29 & \textbf{94.64} \\
			Drama & 71.16 & 85.70 & \textbf{89.40} & \textbf{79.42} & 49.16 & 55.06 & 79.95 & 83.08 & \textbf{86.77} \\
			Family & 82.55 & 83.08 & \textbf{86.58} & 27.11 & 40.14 & \textbf{48.08} & 52.19 & 67.09 & \textbf{73.60} \\
			Fantasy & 69.83 & 79.37 & \textbf{87.88} & 13.51 & 12.02 & \textbf{20.91} & 39.12 & 48.78 & \textbf{59.59} \\
			History & 82.90 & 63.16 & \textbf{90.00} & \textbf{12.52} & 6.49  & 9.73  & 34.41 & 33.63 & \textbf{41.05} \\
			Horror & 70.03 & 86.49 & \textbf{88.38} & 8.76  & 51.04 & \textbf{55.82} & 35.51 & 79.48 & \textbf{84.24} \\
			Music & \textbf{89.04} & 66.67 & 68.06 & 27.24 & 20.48 & \textbf{29.52} & 47.13 & 42.17 & \textbf{48.36} \\
			Musical & 73.58 & 70.59 & \textbf{75.00} & 4.45  & 11.21 & \textbf{14.02} & 22.88 & 32.86 & \textbf{41.27} \\
			Mystery & \textbf{76.42} & 0.00  & 57.14 & \textbf{7.76} & 0.00  & 1.20  & \textbf{39.70} & 25.28 & 30.68 \\
			Romance & 71.93 & 76.47 & \textbf{82.32} & 14.02 & 13.65 & \textbf{15.75} & 49.27 & 51.16 & \textbf{58.59} \\
			Sci-Fi & 81.35 & 72.09 & \textbf{83.60} & 14.51 & 29.67 & \textbf{37.80} & 44.14 & 54.11 & \textbf{68.00} \\
			Sport & \textbf{94.97} & 69.51 & 75.86 & 21.99 & 42.86 & \textbf{49.62} & 39.59 & 60.44 & \textbf{66.80} \\
			Thriller & 64.98 & \textbf{79.35} & 78.17 & 14.50 & 28.79 & \textbf{37.82} & 49.80 & 69.54 & \textbf{71.76} \\
			War   & \textbf{86.27} & 64.71 & 75.86 & 12.80 & 20.89 & \textbf{27.85} & 34.41 & 47.66 & \textbf{56.40} \\
			Western & 88.89 & 88.89 & \textbf{89.80} & 51.93 & 59.70 & \textbf{65.67} & 73.99 & 81.14 & \textbf{83.53} \\ \hline
			AVERAGE & 79.74 & 73.78 & \textbf{82.00} & 24.97 & 32.17 & \textbf{38.33} & 46.88 & 59.73 & \textbf{66.02} \\ 
			
		\end{tabular}
	
\end{table}

Out of all $21$ genres, our model outperforms the other models in $14$ of them in terms of precision, $16$ of them in terms of recall, and $20$ of them in terms of mean average precision. Outlier values such as very low recall for genres such as Biography, History, and Mystery can be attributed to the inherent imbalance within the MovieNet dataset~\parencite{movienet}. We deliberately avoided balancing the data before training to maintain a fair comparison with the classification model presented in the MovieNet paper. We also believe that this dataset closely reflects the real-world distribution of cinematic genres, accurately portraying the relative rarity of genres such as Biography, History, and Mystery.

In Table \ref{table:features} we present an ablation study using our best-performing model multi-transformer, demonstrating the gain in terms of mean average precision along with the increase in runtime, due to the addition of each feature while comparing against our baseline model. The first row shows the performance of the baseline model. Asterisk ($\ast$) indicates averaging the features over the sequence (temporal) dimension. Runtime is defined as the average duration in seconds, to process a single video, both extracting its pretrained features and classifying it, during inference. 

\begin{table}[htbp]
	\centering
	\caption{The effect of inclusion of pretrained features on mean average precision and runtime.}
	
	\begin{tabular}{c|c|c|c|c||c|c}
		
		CLIP & Music & Audiotag & OCR & ASR & mAP & Runtime (s.)          \\ \hline \hline
		&               &               &               &               & 59.73 & 7.08 \\ \hline \hline
		$\checkmark$  &               &               &               &               & 64.73 & 5.32     \\ \hline
		$\checkmark$  & $\checkmark$  &               &               &               & 65.17 & 5.76      \\ \hline
		$\checkmark$  & $\checkmark$  & $\checkmark$  &               &               & 65.31 & 5.95      \\ \hline
		$\checkmark$  & $\checkmark$  & $\checkmark$  & $\checkmark$  &               & 63.33 & 25.85      \\ \hline
		$\checkmark$  & $\checkmark$  & $\checkmark$  & $\ast$        &               & 65.46 & 31.52      \\ \hline
		$\checkmark$  & $\checkmark$  & $\checkmark$  & $\checkmark$  & $\checkmark$  & 64.66 & 31.44      \\ \hline
		$\checkmark$  & $\checkmark$  & $\checkmark$  & $\ast$        & $\ast$        & \textbf{66.02} & 33.59
	\end{tabular}
	
	\label{table:features}
\end{table}

Although the inclusion of textual features such as OCR and ASR initially seems to hurt the performance, possibly due to the text prediction errors, incorporating their averaged version along the temporal dimension reduces the noise and does improve mean average precision. Incorporating textual features also significantly increases runtime. This is primarily because models such as CLIP, Music, and Audiotag are designed to predict a single embedding vector or label, while OCR and ASR models predict text sequences, which can be viewed as predicting many labels corresponding to each word and subword, in an autoregressive way. But most importantly, the settings in which the textual features are excluded yield a better classification performance and faster runtime compared to the baseline. Furthermore, our classification models can seamlessly incorporate features that result from newer and potentially more efficient ASR and OCR models that can be developed in the future, further reducing the overall runtime.

\begin{figure}[htbp]
	\centering
	
	\includegraphics[width=0.6\columnwidth]{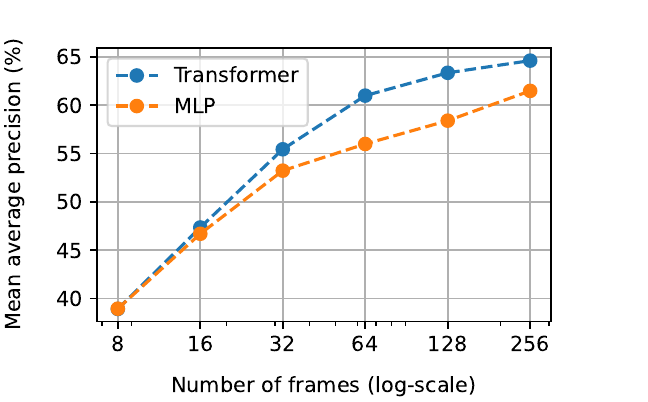}
	\caption{Number of frames vs. mean average precision, only using CLIP features.}
	\label{fig:frames}
\end{figure}

Finally in Figure \ref{fig:frames} we attempt to further explain the reasons behind the success of our models. The models in the literature extract a small and fixed number of frames at random locations from each video for classification. The MovieNet baseline \parencite{movienet} only uses $8$ scenes. They furthermore use MLPs, which cannot process sequences, hence the features from these small number of frames need further averaging along the temporal dimension, causing additional loss of information. Using a simplified experiment, we show the importance of the number of frames fed into the model. Using only the CLIP features, we employ $8$, $16$, $32$, $64$, $128$ and $256$ frames obtained from random locations in each video. We only use the CLIP features since the number of feature vectors differs amongst the pretrained features, and finding an exact number of features that would correspond to the specific number of CLIP features is challenging. 

As we use a single input modality, both our single-transformer and multi-transformer architectures defaults to a standard transfomer classifier. We therefore process the CLIP feature sequences using an MLP and a transformer. Regarding the MLP, the features from different frames are averaged along the temporal dimension as in \parencite{movienet}. On the contrary, the transformer model is capable of processing sequences hence we feed the features from each frame \textit{as is}.

In Figure \ref{fig:frames}, we can see that using a higher number of frames consistently improves the performance. Furthermore, as the number of frames increases, the superiority of the transformer over the MLP becomes apparent, due to its ability to process long sequences seamlessly.

In summary, the success of our genre classification models is due to combining diverse pretrained features, leveraging all video scenes and the full audio track, and avoiding temporal averaging. Instead, we feed the complete feature sequences into a transformer, which excels at handling long sequences. These design choices are feasible thanks to the large scale of the MovieNet dataset.

Our work on cinematic genre classification highlights the potential for enhancing video emotion classification, especially as large-scale video datasets with emotion labels become available in the future. For now, due to the limited size of existing video emotion datasets, the emotion classifier in our video-based music generation system uses a fixed, smaller set of input frames.

We are now ready to bring all components together to realize our video-based music generator. This system combines modules for video emotion analysis and temporal analysis (via scene cut detection) with a music generator that incorporates both emotion and temporal conditioning. The final MIDI output is synthesized into audio using the Fluidsynth software. (As a reminder, Figure~\ref{fig:flowchart}, introduced in the Introduction, illustrates the full system architecture.) We detail the integration process in the following chapter.

\chapter{Video-based music generation} \label{chap:video_music}

In this chapter, we explain how we integrate all the components to create our video-based music generator. Specifically, we combine the four modules of our two-stage, two-branch pipeline: the temporal boundary conditioner for music generation (Section \ref{sec:boundary_conditioning}), the emotion conditioner for music generation (Section \ref{sec:emotion_based_midi_generation}), the video scenecut (boundary) extractor (Section \ref{sec:scenecut}), and the the video emotion classifier (Section \ref{sec:emotion_classifier}). We name our model \textit{EMSYNC}, as it aligns music with video by matching their \textbf{em}otions and \textbf{sync}hronizing their temporal boundaries.

Before we proceed, we address the issue of incompatible emotion representations between the video emotion classifier (VEMOCLAP) and emotion-based music conditioning. While \protect{VEMOCLAP} outputs probability distributions for discrete emotion categories, the music generator is conditioned on emotions represented as continuous valence and arousal values, ranging from -1 to 1.

\section{Emotion mapping}

We establish a correspondence between the emotion representations in our video analysis and music generator models. Specifically, we map discrete emotions to continuous valence-arousal values.

\textcite{mapping} conducted user studies to determine the corresponding valence and arousal values for discrete emotion categories. They presented their findings in a table containing the mean and standard deviation values of valence and arousal for each categorical emotion. We extract and use the values corresponding to Ekman's six basic emotions, as shown in Table \ref{tab:mapping}~\parencite{ekman}.

\begin{table}[htbp]
	\centering
		\caption{Means and standard deviations of valence and arousal values corresponding to categorical emotions of~\protect\textcite{ekman}, obtained from the user studies of \textcite{mapping}.}
		\begin{tabular}{c|cc|cc}

			& \multicolumn{2}{c|}{Valence} & \multicolumn{2}{c}{Arousal} \\ 

			Emotion & Mean & Std       & Mean & Std         \\ \hline 

			anger         & -0.51             & 0.20   & \phantom{-}0.59          & 0.29                \\ 
			disgust       & -0.60             & 0.20   & \phantom{-}0.35          & 0.41                \\ 
			fear          & -0.64             & 0.20   & \phantom{-}0.60           & 0.32                \\ 
			joy           & \phantom{-}0.76   & 0.22   & \phantom{-}0.48          & 0.26                \\ 
			sadness       & -0.63             & 0.23   & -0.27                    & 0.34                \\ 
			surprise      & \phantom{-}0.40   & 0.30   & \phantom{-}0.67          & 0.27                \\ 

		\end{tabular}

	\label{tab:mapping}
\end{table}

Using the output probabilities of the video classifier along with the means and standard deviations for each emotion, we construct a mixture of Gaussian distributions. Users can either sample from the mixture---by first selecting an emotion category based on the output probabilities and then sampling from the corresponding Gaussian distribution---or use the mean of the mixture, which is calculated as the weighted average of the emotion category means, with weights determined by the output probabilities. 

Additionally, we introduce a parameter that represents the maximum absolute value of means across all emotion categories. Using this parameter, we compute shifting and scaling parameters that are applied to all valence-arousal distributions. This approach enables users to adjust the conditioning of the music generator, choosing between a wider or narrower range of emotions. The following equations show how we readjust the ranges of valence and arousal values.

\begin{align*}
\mathbf{v}_{\text{means}}^{\prime} &= \frac{2r \cdot \left(\mathbf{v}_{\text{means}} - \min(\mathbf{v}_{\text{means}})\right)}{\max(\mathbf{v}_{\text{means}}) - \min(\mathbf{v}_{\text{means}})} - r \\
\mathbf{v}_{\text{stds}}^{\prime} &= \frac{2r \cdot \mathbf{v}_{\text{stds}}}{\max(\mathbf{v}_{\text{means}}) - \min(\mathbf{v}_{\text{means}})} \\
\mathbf{a}_{\text{means}}^{\prime} &= \frac{2r \cdot \left(\mathbf{a}_{\text{means}} - \min(\mathbf{a}_{\text{means}})\right)}{\max(\mathbf{a}_{\text{means}}) - \min(\mathbf{a}_{\text{means}})} - r \\
\mathbf{a}_{\text{stds}}^{\prime} &= \frac{2r \cdot \mathbf{a}_{\text{stds}}}{\max(\mathbf{a}_{\text{means}}) - \min(\mathbf{a}_{\text{means}})}
\end{align*}

We select $r$ to represent the new maximum absolute value of means across all emotion categories, and hence, the overall range becomes $2r$. The prime superscript signify new values. Bold letters represent the vectors that contains the mean and standard deviation values of valence and arousal of all emotion categories. While we shift and scale the means, we only scale the standard deviations as standard deviation only represents the spread of the data and not the positioning. The minimum and the maximum values can be found directly from Table~\ref{tab:mapping} as:

\begin{align*}
\min(\mathbf{v}_{\text{means}}) &= -0.64, \quad \max(\mathbf{v}_{\text{means}}) = 0.76, \quad \min(\mathbf{a}_{\text{means}}) &= -0.27, \quad \max(\mathbf{a}_{\text{means}}) = 0.67.
\end{align*}

\section{Implementation details}

\begin{figure}[t] 
	\centering
	\includegraphics[width=0.5\columnwidth]{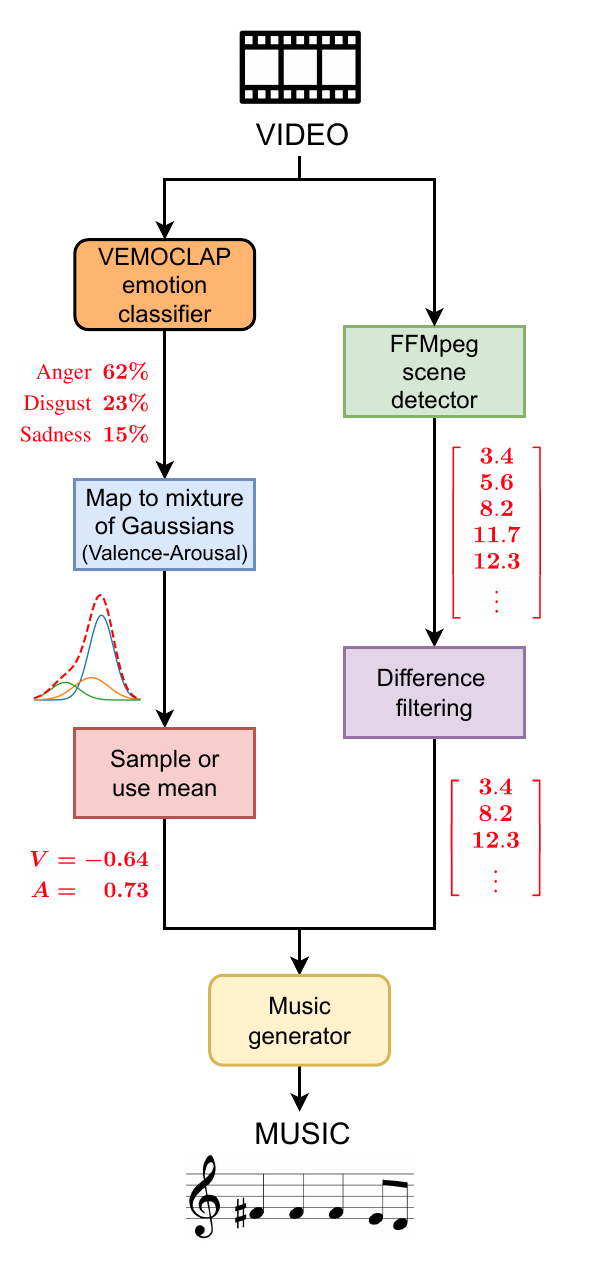}
	\caption{Our video-based music generation pipeline. The values next to the arrows are sample values.}
	\label{fig:video_midi}
\end{figure}

The video-based music generation pipeline is shown in Figure~\ref{fig:video_midi}. In the left branch, we perform emotion-conditioning. We begin by setting the maximum absolute value of the means across all emotion categories to 0.8 and adjust the valence-arousal distributions accordingly. The VEMOCLAP model classifies the video's emotions and outputs a probability distribution. This distribution is then mapped to a mixture of Gaussians, with the probability distribution serving as the weights for the valence-arousal distributions of each emotion category. From this, we can either sample from the distribution or use the mean of the mixture. To ensure more robust results, we choose to use the mean of the mixture. The final output consists of two values for valence and arousal, which become the emotion conditions for the music generator.

The right branch focuses on temporal conditioning. Here, the FFMpeg scene detector identifies scene cut positions by detecting when the average pixel difference between consecutive frames exceeds 27\% of the pixel value range (0-255). A difference filter is then applied to remove positions that are too close together, preventing a cacophony of overlapping strong chords placed too near each other. In our experiments, we discard scenes that are less than 4 seconds apart. The resulting timestamps serve as the temporal conditions for the music generator.

The music generator produces tokens until the generated music matches the video’s duration. After the MIDI is synthesized into an audio waveform, a 3-second fade-out is applied to avoid an abrupt ending. After training the video emotion classifier and the music generator---along with its temporal and emotional conditioning mechanisms---we do not perform any additional training or fine-tuning on the combined model. All hyperparameters are determined through trial and error, as well as manual inspection of the outputs.

\begin{figure}[t]  
	\centering
	\includegraphics[width=0.9\textwidth]{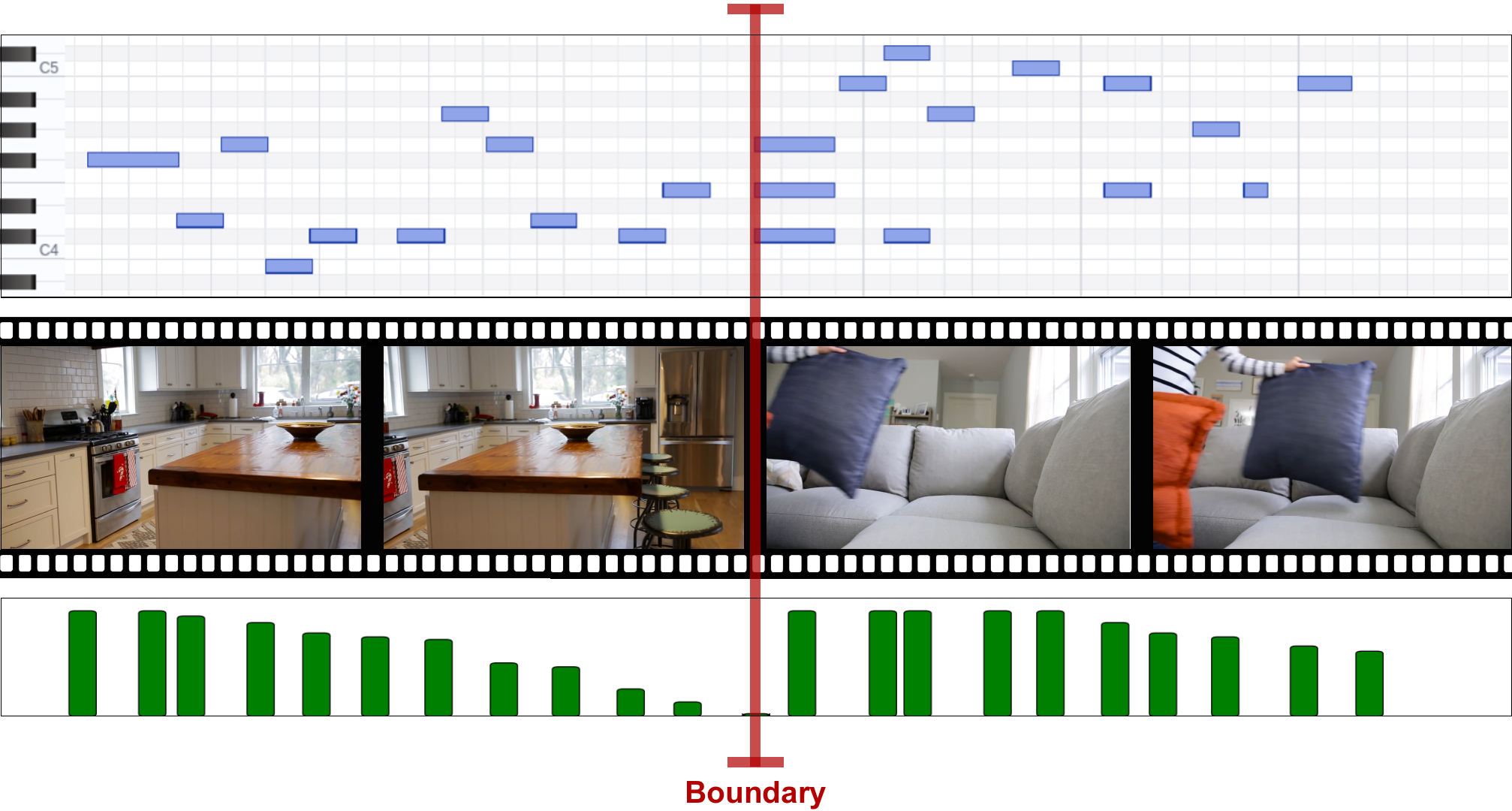}  
	\caption{Graphical illustration of boundaries.}
	\label{fig:boundaries}
\end{figure}

Figure \ref{fig:boundaries} provides a graphical illustration of how we align musical boundaries (strong chords) with video boundaries (scene cuts). During the training of our boundary conditioning mechanism, we use the timestamps of long-duration chords as target boundaries. In video-based inference, these target boundaries are replaced with the timestamps of scene cuts extracted from the video.

In the top row, we present a piece of symbolic music, where a chord consisting of three or more simultaneous notes with a duration exceeding a set threshold of two beats defines a musical boundary. These musical boundaries are only used during training, as explained in Section~\ref{sec:boundary_conditioning}. The middle row shows a video, where scene cuts serve as video boundaries and are used during inference. The bottom row illustrates the boundary offsets, which are calculated based on the input boundaries.

It is important to note that these figures are for illustrative purposes, and in this example, the offsets do not perfectly align with the music, except at the boundaries. Additionally, during video-based inference, we do not expect the music and video boundaries to match perfectly. As previously mentioned, our music generator creates strong chords in the vicinity of the input boundaries, ensuring rhythmic coherence with the rest of the generated music without disrupting its flow.

For emotion-based MIDI generation, we use our continuous-token model described in Section \ref{sec:music_models}. We select this model because it utilizes continuous-valued conditions, avoiding the loss of information associated with discretization. Additionally, while its performance is nearly on par with our top-performing continuous-concatenated model, it does not increase the model's dimensionality. However, all of our emotion conditioning modules are compatible for integration into the overall model.

\begin{figure}[t] 
	\centering
	\includegraphics[width=0.55\columnwidth]{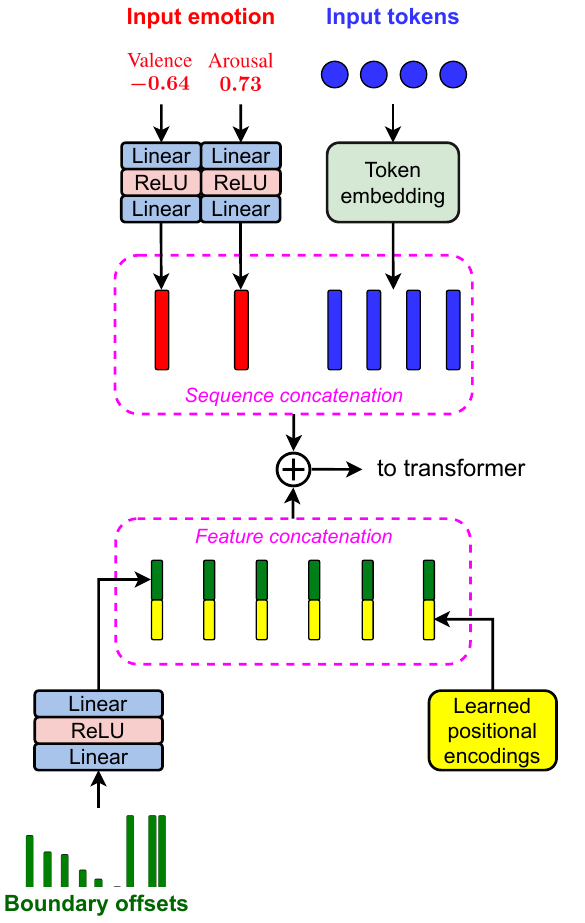}
	\caption{Our music generator. Numbers underneath valence and arousal are sample values.}
	\label{fig:music_generator}
\end{figure}

Figure \ref{fig:music_generator} illustrates the overall conditional music generation pipeline, integrating both emotional and temporal conditioning. Emotion conditioning is achieved by concatenating conditioning vectors with token embeddings along the sequence dimension. As outlined in Section \ref{sec:boundary_conditioning}, temporal boundary conditioning is implemented by concatenating boundary offsets with positional encodings along the feature dimension. The transformer's input is represented as:

\begin{align}
\mathbf{X}_{\text{trans}} &= \text{Concat}_s\left( \text{FFN}_v(x_v), \text{FFN}_a(x_a), \text{Embedding}(\mathbf{x}_t) \right) + \text{Concat}_f(\text{FFN}_b(\mathbf{b}), \mathbf{W_{pe}})
\end{align}

where \( x_v \) and \( x_a \) are the valence and arousal inputs, \( \mathbf{x}_t \) is the input token sequence, \( \mathbf{b} \) represents the input boundary offsets, and \( \mathbf{W_{pe}} \) is the learned positional encoding. The feed-forward networks \( \text{FFN}_v \), \( \text{FFN}_a \), and \( \text{FFN}_b \) process the valence, arousal, and boundary offset inputs, respectively. The operations \( \text{Concat}_s \) and \( \text{Concat}_f \) denote concatenation along the sequence and feature dimensions, respectively. These two conditioning mechanisms are designed to be easily integrated without interfering with each other. Furthermore, neither of the conditioning mechanisms modifies the transformer's body, making it easy to fine-tune our trained models for other types of music generation tasks by simply removing or replacing the conditioning modules.

We choose the size of our music generator to ensure a fair comparison against the state-of-the-art methods which we detail later. Our music generator consists of 11 layers, 8 attention heads, and a dimensionality of 512, totaling 37M parameters. We train it with a context length of 1216 tokens and a batch size of 64. Training is performed using cross-entropy loss with a learning rate of 2e-4 for the first 300k steps, followed by 5e-5 for the next 300k steps. We use the Adam optimizer with gradient clipping to a norm of 1~\parencite{adam}. We do not apply any regularization methods and do not observe overfitting, as the model is trained on large-scale, densely labeled data where it predicts each token of its input sequence.

\section{Evaluation}
\label{sec:video_music_eval}
We conduct objective and subjective evaluations of our model, comparing it with two open-source state-of-the-art methods for generating MIDI from arbitrary videos: Video2Music~\parencite{kang} and CMT~\parencite{di}, both introduced earlier in Section~\ref{sec:intro_video_based_music}.

Video2Music shares similarities with our approach in leveraging both low-level video features and high-level emotional conditioning. However, a major difference lies in the music generation process: their model generates only chord sequences, which are then transformed into piano melodies using fixed, hand-crafted arpeggiation patterns. These patterns all rely on a coarse time grid, limiting their ability to capture the expressive timing of human performance. In contrast, our model directly generates multi-instrument music in a direct and learned manner, using an event-based representation with fine temporal precision. Moreover, Video2Music requires the user to provide the musical key and initial chords, while our method is fully automatic.

The Controllable Music Transformer (CMT) generates multi-instrument music directly and is trained on the Lakh Pianoroll Dataset, similar to our approach. However, unlike our model, it relies solely on low-level video features such as timing, motion saliency, and motion speed, without incorporating high-level features like emotion. Additionally, similar to Video2Music, it restricts note placement to beat subdivisions, limiting its ability to capture expressive timing nuances.

For our final evaluation, where we compare our results to Video2Music and CMT, we set our model size to be comparable to theirs: our model has 37M parameters, compared to 39M in CMT and 33M in Video2Music. For all methods, we synthesize output MIDI files into audio waveforms using Fluidsynth and apply peak normalization up to -3 dB.

\subsection{Datasets}

We train our music generator on the Lakh Pianoroll-5 dataset~\parencite{lpd}. For objective and subjective evaluation, we first use the \mbox{EmoMV-C} dataset, which consists of music videos containing music audio tracks to be used as ground truth for objective evaluation~\parencite{emomv}. We exclude auditory features during the emotion classification to prevent data leakage. To assess the model's generalization across different video types, we also include the Pittsburgh Advertisements (Ads) dataset, which contains advertisement videos that may or may not feature music~\parencite{ads}. As advertisements often rely on music to maximize viewer engagement, the Ads dataset provides a real-world scenario for evaluation~\parencite{ads_music}.

We use the full EmoMV-C validation split, which contains 48 thirty-second videos. We filter the Ads dataset to match the total duration of EmoMV-C. We select videos associated with the four basic emotions commonly used in music emotion classification~\parencite{music_emotion_classification}, covering the quadrants of Russell's valence-arousal model~\parencite{valence_arousal}: \textit{cheerful}, \textit{calm}, \textit{angry}, and \textit{sad}, providing a broad coverage of the emotional space for evaluation. While some of the chosen emotions do overlap with Ekman's categories~\parencite{ekman}, we avoid explicitly using Ekman's emotions to select the evaluation videos. This prevents biasing the evaluation toward our method, since our video emotion classifier is trained to output Ekman's categories. We exclude videos shorter than 1 minute. To ensure an unbiased selection, we use YouTube IDs, which are generated randomly by YouTube. We sort these IDs alphabetically and select the first six videos from each emotion category, resulting in 24 evaluation videos. Finally, we trim the videos to a uniform duration of 1 minute. The resulting evaluation videos are then fed into the compared models.

\subsection{Objective evaluation}

In objective evaluation, we first evaluate music similarity using the ground-truth audio from the EmoMV-C dataset. Although the ground truth is not in MIDI, all methods generate MIDI outputs that are synthesized to audio, enabling a fair comparison in the audio domain. We evaluate the Kullback-Leibler Divergence (KL) between the labels of generated and ground-truth music, calculated using a pretrained music tagging model~\parencite{harmoniccnn}. This measures how closely the distribution of semantic tags in the generated music matches that of the ground-truth. 
We avoid the commonly used Fréchet Audio Distance, as it compares overall audio distributions suited for unconditional generation~\parencite{fad}, whereas video-based music generation requires per-sample evaluation to assess alignment with each video's unique content.

We additionally evaluate temporal audio-video alignment (AV-Align) score, which is the Intersection-over-Union between video and audio peaks~\parencite{avalign}, using both EmoMV-C and Ads datasets. Video peaks are detected from the mean optical flow magnitude~\parencite{flow}, while audio peaks correspond to audio onsets~\parencite{onsets}. We calculate AV-Align scores at two temporal resolutions (windows): one video frame (33.3 ms at 30 FPS) and one second. For each window, local peaks in both video and audio are detected and matched within the same temporal range to assess alignment. Finally, we report the average runtime for each model when generating music for a one-minute video, including model initialization.

\subsection{Subjective evaluation}

Using the evaluation videos from the EmoMV-C and the Ads datasets, we generate accompanying music with the three models and create survey pages, each containing two videos. For each video, the three music versions, generated by the compared methods, are presented side by side with anonymized method names, with each methods's output appearing an equal number of times in the left, center, and right positions. 

We enrolled 153 remote participants and asked them to rank the three models based on the standard criteria used in previous works~\parencite{di,kang}:

\begin{itemize}
	\item \textit{Music Richness (MR)} — the richness and diversity of the music, independent of the video content.
	\item \textit{Music Quality (MQ)} — the overall quality of the music, independent of the video content.
	\item \textit{Emotion Match (EM)} — how well the music matches the video in terms of emotion.
	\item \textit{Timing Match (TM)} — how well the music synchronizes with the video in terms of rhythm and timing.
	\item \textit{Overall Match (OM)} — how well the music matches the video as a whole.
\end{itemize}

\begin{figure}[htbp] 
	\centering
	\includegraphics[width=1\columnwidth]{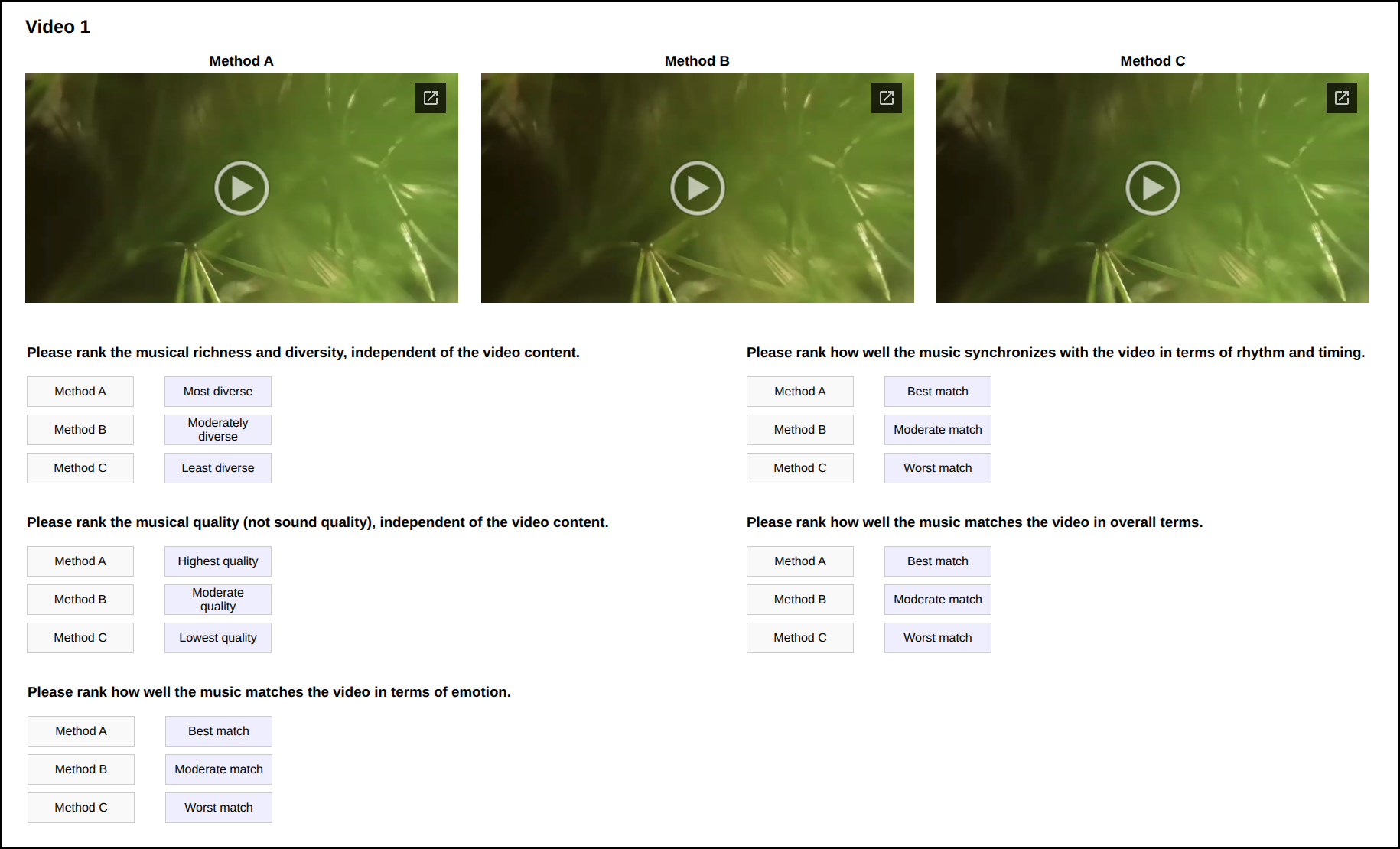}
	\caption{A screenshot of our video-based music generation survey.}
	\label{fig:survey}
\end{figure}

Each model is assigned a unique ranking of 1 (best), 2, or 3 (worst). Figure \ref{fig:survey} shows a sample view of the user survey. The model names are anonymized as Method A, Method B, and Method C. Users are asked to rank the methods by dragging the method names from the left-hand side to the right-hand side, ensuring that missing a question does not result in an error. For readers, we also provide output samples from eight videos online, with method names deanonymized and each method's output appearing in the same position for clarity.\footnote{\url{https://serkansulun.com/emsync}}

\section{Results and discussion}
\label{sec:results_video_music}

We compare CMT~\parencite{di}, Video2Music~\parencite{kang}, and our method EMSYNC across all subjective criteria using average rankings. Table~\ref{tab:group1}, Table~\ref{tab:group2}, and Table~\ref{tab:all_participants} present the results for Group 1, Group 2, and all participants, respectively. Rankings are first averaged per video and then averaged across all videos. A lower score indicates a better ranking (1 being best, 3 worst). Standard deviations across videos are shown in parentheses.

\begin{table}[htbp]
	\centering
	\caption{Results for participants with a self-reported understanding of music theory. Lower values indicate better rankings (1 = best, 3 = worst). Values are reported as mean (standard deviation).}
	\begin{tabular}{l|ccccc}
		& \multicolumn{1}{c}{Music} & \multicolumn{1}{c}{Music} & \multicolumn{1}{c}{Emotion} & \multicolumn{1}{c}{Timing} & \multicolumn{1}{c}{Overall} \\
		& \multicolumn{1}{c}{Richness} & \multicolumn{1}{c}{Quality} & \multicolumn{1}{c}{Match} & \multicolumn{1}{c}{Match} & \multicolumn{1}{c}{Match} \\
		\hline \hline
		\makecell{CMT\\\parencite{di}}           & 1.85 (0.56) & 2.44 (0.60) & 2.00 (0.53) & 2.12 (0.76) & 2.08 (0.64) \\ \hline
		\makecell{Video2Music\\\parencite{kang}}  & 2.52 (0.71) & 1.85 (0.67) & 2.12 (0.86) & 1.96 (0.76) & 2.17 (0.83) \\ \hline
		\makecell{EMSYNC\\(Ours)} & \textbf{1.62} (0.71) & \textbf{1.71} (0.76) & \textbf{1.88} (0.85) & \textbf{1.92} (0.75) & \textbf{1.75} (0.79) \\
	\end{tabular}
	\label{tab:group1}
\end{table}

\begin{table}[htbp]
	\centering
	\caption{Results for participants without a self-reported understanding of music theory. Lower values indicate better rankings (1 = best, 3 = worst). Values are reported as mean (standard deviation).}
	\begin{tabular}{l|ccccc}
		& \multicolumn{1}{c}{Music} & \multicolumn{1}{c}{Music} & \multicolumn{1}{c}{Emotion} & \multicolumn{1}{c}{Timing} & \multicolumn{1}{c}{Overall} \\
		& \multicolumn{1}{c}{Richness} & \multicolumn{1}{c}{Quality} & \multicolumn{1}{c}{Match} & \multicolumn{1}{c}{Match} & \multicolumn{1}{c}{Match} \\
		\hline \hline
		\makecell{CMT\\\parencite{di}}         & 1.79 (0.64) & 2.56 (0.61) & 2.46 (0.57) & 2.31 (0.75) & 2.35 (0.58) \\ \hline
		\makecell{Video2Music\\\parencite{kang}}   & 2.71 (0.36) & 1.77 (0.53) & 1.85 (0.73) & 1.98 (0.65) & 2.08 (0.73) \\ \hline
		\makecell{EMSYNC\\(Ours)} & \textbf{1.50} (0.49) & \textbf{1.67} (0.67) & \textbf{1.69} (0.64) & \textbf{1.71} (0.69) & \textbf{1.56} (0.66) \\
	\end{tabular}
	\label{tab:group2}
\end{table}

\begin{table}[htbp]
	\centering
	\caption{Results for all participants. Lower values indicate better rankings (1 = best, 3 = worst). Values are reported as mean (standard deviation).}
	\begin{tabular}{l|ccccc}
		& \multicolumn{1}{c}{Music} & \multicolumn{1}{c}{Music} & \multicolumn{1}{c}{Emotion} & \multicolumn{1}{c}{Timing} & \multicolumn{1}{c}{Overall} \\
		& \multicolumn{1}{c}{Richness} & \multicolumn{1}{c}{Quality} & \multicolumn{1}{c}{Match} & \multicolumn{1}{c}{Match} & \multicolumn{1}{c}{Match} \\
		\hline \hline
		\makecell{CMT\\\parencite{di}}          & 1.82 (0.60) & 2.50 (0.60) & 2.23 (0.59) & 2.22 (0.75) & 2.22 (0.62) \\ \hline
		\makecell{Video2Music\\\parencite{kang}}   & 2.61 (0.57) & 1.81 (0.60) & 1.99 (0.80) & 1.97 (0.70) & 2.12 (0.78) \\ \hline
		\makecell{EMSYNC\\(Ours)} & \textbf{1.56} (0.61) & \textbf{1.69} (0.71) & \textbf{1.78} (0.75) & \textbf{1.81} (0.72) & \textbf{1.66} (0.73) \\
	\end{tabular}
	\label{tab:all_participants}
\end{table}

\begin{figure}[htbp] 
	\centering
	\includegraphics[width=1\columnwidth]{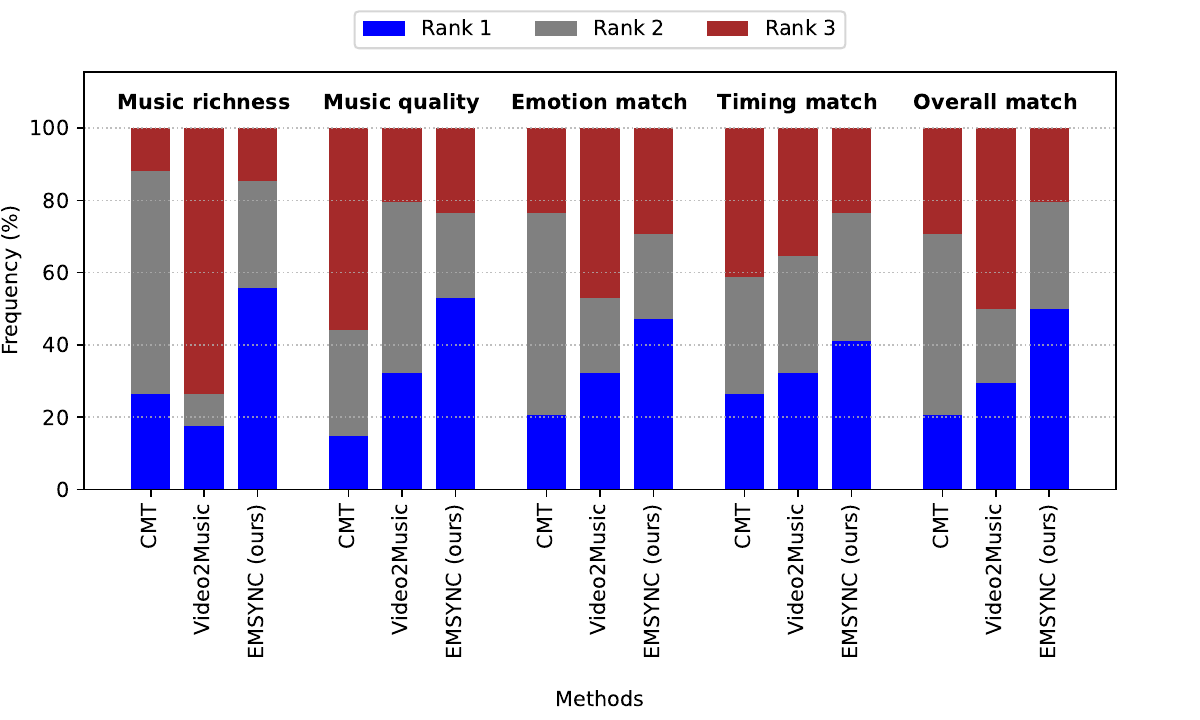}
	\caption{Stacked bar chart results for participants with a self-reported understanding of music theory.}
	\label{fig:stacked_musician}
\end{figure}

\begin{figure}[htbp] 
	\centering
	\includegraphics[width=1\columnwidth]{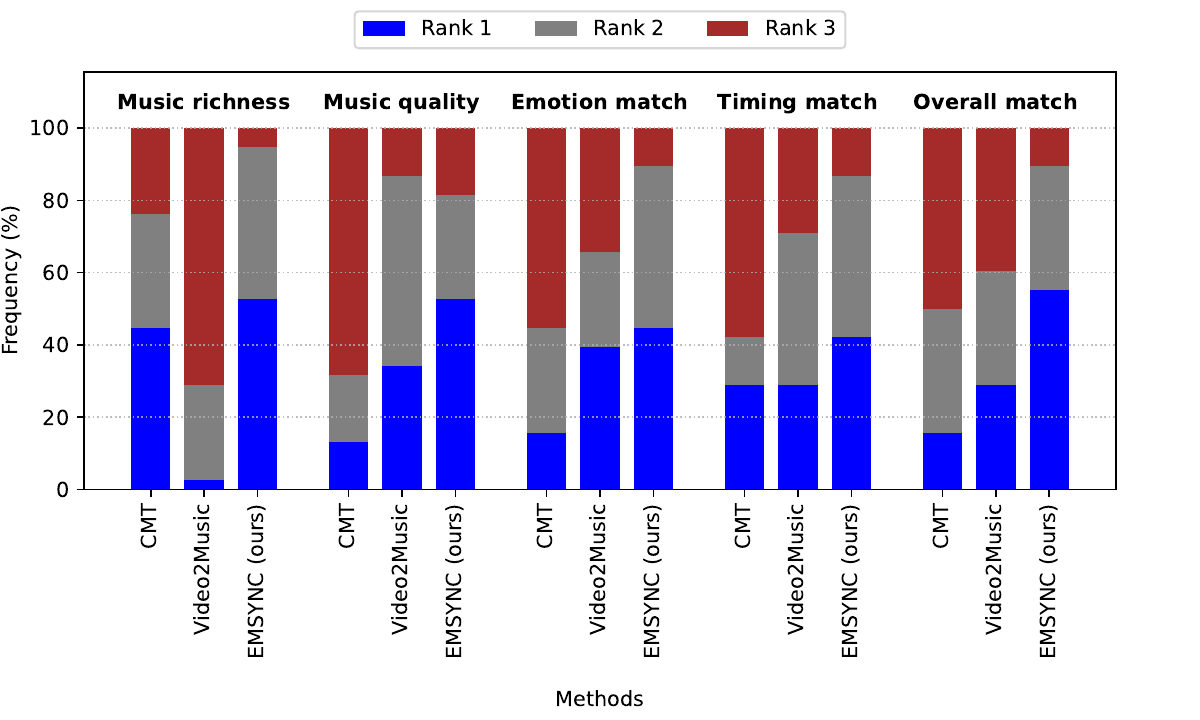}
	\caption{Stacked bar chart results for participants without a self-reported understanding of music theory.}
	\label{fig:stacked_nonmusician}
\end{figure}

\begin{figure}[htbp] 
	\centering
	\includegraphics[width=1\columnwidth]{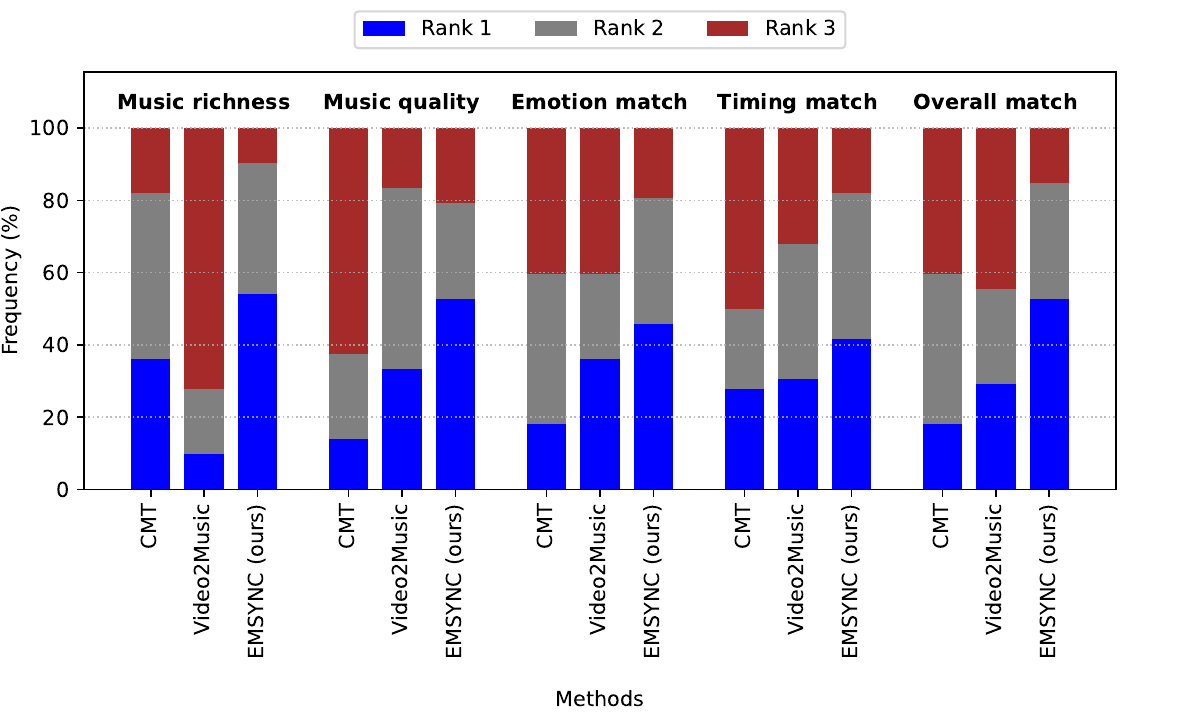}
	\caption{Stacked bar chart results for all participants.}
	\label{fig:stacked_all}
\end{figure}

Figures~\ref{fig:stacked_musician}, \ref{fig:stacked_nonmusician}, and \ref{fig:stacked_all} present stacked bar charts showing the frequencies of participant responses from Group 1, Group 2, and all participants, respectively. Each chart displays the frequency of each rank assigned, grouped by evaluation metric and method. 

Finally, we report the average runtime for each model when generating music for a one-minute video, including model initialization: EMSYNC takes 1.42 minutes, CMT 3.55 minutes, and Video2Music 1.61 minutes.

In terms of music richness, Video2Music receives the worst average ranking across both participant groups. This may be attributed to its reliance on fixed arpeggiation patterns or its limitation to piano-only output. Participants with music theory knowledge assign the best rank to CMT less frequently. A possible explanation, based on our preliminary review of model outputs, is that CMT adjusts note density using sparse constraints such as video motion, which can lead to rhythmically incoherent results. Participants without music theory knowledge may perceive this as adding variety and contributing to musical richness, while others may not. Additionally, participants without music theory knowledge seldom rank Video2Music first, possibly because it is readily apparent, even without any music theory knowledge, that the output lacks variety due to its restriction to a single instrument, whereas the other methods produce multi-instrumental music.

In terms of music quality, CMT receives the worst average ranking from both participant groups. This may again be due to its continual modulation of note density. Regarding emotion match, in both groups, CMT is the least frequently assigned the best rank, which is expected given that it relies solely on low-level features and lacks high-level conditioning such as emotion. However, a surprising observation is that participants with music theory knowledge are also the least likely to assign CMT the worst rank. As a result, its average ranking in this category is better than that of Video2Music. This may highlight the inherent difficulty of high-level video emotion classification and suggests that there is still room for improvement in the emotion analysis components of both our model and Video2Music.

In terms of timing match, our method is particularly favored by participants without music theory knowledge. This may support our hypothesis that alignment with sparse temporal cues, such as scene cuts, is more easily perceived by the general public. In terms of overall match, the effectiveness of our method becomes even more evident. It achieves the best performance across all participant groups, while the average rankings for the other methods remain worse than 2.

EMSYNC consistently outperforms other methods across all metrics for all groups. These results confirm that EMSYNC produces the most diverse and high-quality music with the best emotional alignment, temporal synchronization, and overall video compatibility while also being the most computationally efficient model.

This chapter concludes the presentation of our video-based music generation system in its entirety. Readers may recall our initial overview of the model pipeline in Figure~\ref{fig:flowchart}, and note that we have now discussed each component in detail—except for the audio generator. To build a working prototype capable of producing listenable outputs, we used preset sound libraries together with the Fluidsynth software, which maps these sounds to individual MIDI notes. The main limitation of this approach is the lack of timbral diversity: because the sound library is fixed, the resulting audio lacks nuance and realism, often sounding overly mechanical compared to performances by human musicians.

While several existing methods generate music directly in the audio domain, they are typically conditioned only on text and operate as black-box models~\parencite{musicgen1,musicgen2,musicgen3,musicgen4}. Since their outputs exist purely as audio, they are not editable or interpretable. As such, it may be more accurate to describe their outputs as musical sounds rather than musical compositions. This black-box approach contrasts with the human process of music creation, which typically involves two distinct stages: composition (defining chords, notes, melodies, and structure) followed by production (generating audio through performance). We argue that a promising direction for automatic music generation is to employ DNNs separately for both stages---composition via MIDI generation, and production via MIDI-conditioned audio generation. Although research into MIDI-based audio generation is ongoing, current efforts are often limited to single-track piano music, largely due to the scarcity of paired multi-instrument MIDI and audio datasets~\parencite{maestro,midi2audio2}. In the next chapter, we present a preliminary analysis of audio generation and its challenges, particularly when relying on synthetic data to compensate for the lack of real-world audio recordings.

\chapter{Audio bandwidth extension} \label{chap:audio}

As our video-based music generator creates soundtracks in the symbolic music format, they are synthesized into audio using computer software and a soundfont library, which includes fixed sounds for particular pitches of particular instruments. However, the end result is very mechanical and unrealistic, compared to the music produced by human musicians. We therefore investigate audio enhancement to take the first step towards improving the quality of audio synthesized from symbolic music.

There are currently no multi-instrument music datasets with paired audio (performed by humans) and MIDI, which would be required to train an end-to-end model capable of generating audio that sounds like human performance from synthesized MIDI. Even if such a dataset were available, creating human-like audio from MIDI would still pose an ill-defined problem, as an infinite number of human performances can be derived from the same music composition (i.e., MIDI file). In the absence of a multi-instrumental dataset with MIDI-audio pairs, one potential approach is to train DNNs using synthetic audio generated from MIDI via sound libraries. However, the quality of the output from these trained models would be limited by the quality of the synthesized audio, rendering the DNN useless.

We investigate another ill-defined audio enhancement problem, namely audio bandwidth extension. This task deals with creating full-band audio from band-limited audio. The information from the limited bands are missing, and there can be multiple candidates for this missing information, making this task ill-defined. We also explore this problem due to the use of synthetic data. Overall, studying audio bandwidth extension provides valuable insights that are applicable to other ill-defined audio generation tasks that utilize synthetic data, such as MIDI-to-audio generation. A key advantage of investigating audio bandwidth extension is the abundance of data available, as one can generate an unlimited amount of input data by processing any full-band audio with various low-pass filters.

When considering audio bandwidth extension for the enhancement of real-world recordings, no full-bandwidth version exists and as such there is no "ground-truth" to act as a target for DNNs. To this end, training data is typically obtained by low-pass filtering full-bandwidth recordings. However, since real-world band-limited samples are not the result of some hypothetical universal digital low-pass filter, it can be challenging to develop robust techniques for bandwidth extension which rely on a loose approximation of the bandwidth reduction process, and in turn to generalize to unseen recordings. While the trained DNNs can perform well on training data created with one type of low-pass filter, they may fail to generalize to audio content subjected to different types of low-pass filter. Throughout our work, we label this as \textit{filter overfitting} which can be understood as a lack of \textit{filter generalization}.

While the use of low-pass filtering is widespread among existing work on audio bandwidth extension using DNNs, to the best of our knowledge, no work to date as thoroughly investigated the topic of filter generalization. We argue that the lack of generalization to various types of signal deterioration is an important challenge in creating audio enhancement models for real-world deployment. In our work, we present a rigorous analysis of filter generalization, evaluating generalization to different filters used to preprocess input data, on the task of bandwidth extension of complex music signals, using two popular DNN architectures. 

To evaluate sample overfitting, we use a testing data split different from the training data split, which is \textit{unseen data} for the trained models. Additionally, to evaluate filter generalization, we preprocess the testing input data, with a filter that does not match the filters that preprocess the training input data, i.e., an \textit{unseen filter} and compare it to the test setting where the filters preprocessing the training and testing data do match, i.e, \textit{seen filters}. We argue that testing with the unseen filter can be thought of as a representation of real-world signal degradation, in which the true underlying degradation function is unknown. 

We evaluate two different regularization methods that are used in the literature to increase generalization. In particular, we compare the usage of data augmentation, batch normalization, and dropout, against the baseline of not using any regularization methods. We introduce a novel data augmentation technique of using a set of different low-pass filters to preprocess the input data, in which the unseen test filter is never present. We examine the training process by tracking the model's performance throughout training iterations, by performing validation using both seen and unseen filters.

One of the DNN models we employ is the \textit{U-Net}, which was first used for biomedical image segmentation \parencite{unet}, and later in audio signal processing tasks such singing voice separation \parencite{audiounet}, and eventually for audio enhancement \parencite{kuleshov, bwe_gan, tfnet, waveunet}. In addition to the U-Net, we also use the deep residual network model (\textit{ResNet}) \parencite{resnet} since it is one of the most widely used DNN architectures in signal processing tasks. Even though the U-Net is a popular architecture in the recent audio processing literature, to the best of our knowledge, no work in the domain of audio processing compares the U-Net against the well-established baseline of the ResNet. A small number of comparative studies exist in the fields of image processing and medical imaging, in which either the number of parameters of the compared models is not stated \parencite{celltracing}, or in which the ResNet has significantly fewer parameters than the U-Net \parencite{mr,hdr,cancer}. In all these works, the ResNet outperforms the U-Net by a small margin. We also present a comparison between the U-Net and ResNet, where each has a similar number of parameters.

Our main findings indicate that filter overfitting occurs for both the U-Net and ResNet, although to different degrees, and that the use of multi-filter data augmentation, as opposed to more traditional regularization techniques, is a promising means to mitigate this overfitting problem and thus improve filter generalization for bandwidth extension.

We continue by presenting our methodology in detail, starting with the models we employ.

\section{Models}
\label{sec:models}
In this section, we define the two baseline models: U-Net and ResNet. For both models, we follow the approach of \textcite{kuleshov} and use raw audio as the input rather than time-frequency transforms (e.g., as in \textcite{matthew}). As such we remove any need for phase reconstruction in the output. However, since we address bandwidth extension and not audio super-resolution per se, our inputs are not subsampled. Hence the sizes of the input and the output are equal for all our models. 
\subsection{U-Net}

\label{sec:unet}

The U-Net architecture \parencite{unet}, like the its predecessor the auto-encoder, consists of two main groups. The first group contains downsampling layers and followed in the second group by upsampling layers, as shown in Figure \ref{fig:audio_models} (left). In the U-Net, individual downsampling and upsampling layers at the same scale are connected through stacking connections, e.g., the output of the first downsampling convolutional block is stacked with the input of the last upsampling convolutional block. As is common in enhancement models, an additive connection from the input to the output is also used, so that the network only needs to model the \textit{difference} between the input and the target signals, rather than creating the target signal from scratch. 

In the downsampling group, one-dimensional convolutional layers with stride $2$ are used, effectively halving the activation length. Borrowing from image processing terminology, the upsampling group includes "sub-pixel" layers (also known as the pixel shuffler) \parencite{subpixel} to double the activation length. Sub-pixel layers work by weaving the samples in the spatial dimension, taken from alternate channels, effectively halving the channel length. The number of parameters is selected to replicate the original work using U-Net for audio super-resolution \parencite{kuleshov, github}, which we denote as \textit{Audio-SR-U-Net} throughout. This resulted in a network with $56.4$ million parameters. 

\subsection{ResNet}
\label{sec:resnet}

\begin{figure}[htbp]
	\centering
	\begin{subfigure}{.45\textwidth}
		\centering
		\includegraphics[width=.9\textwidth]{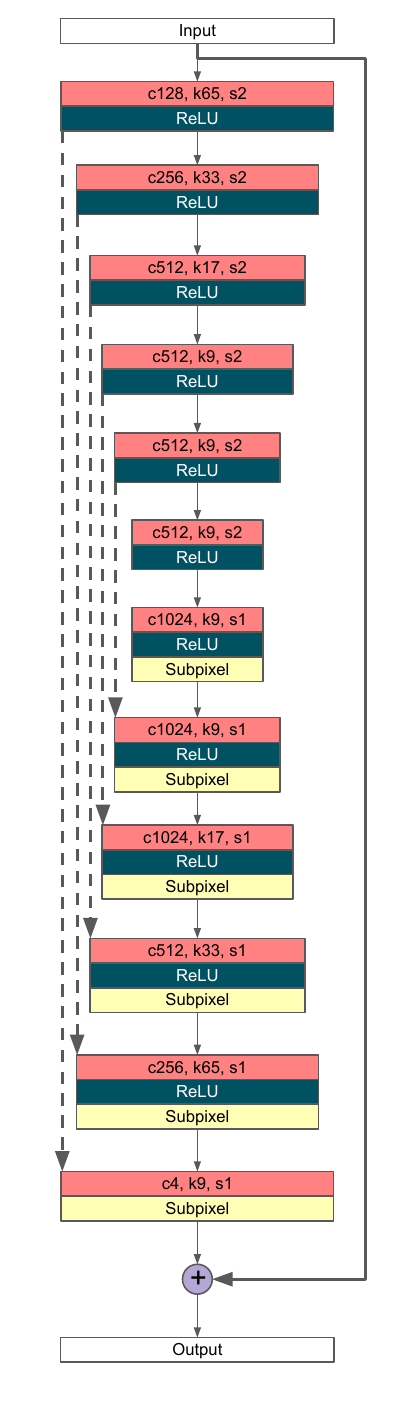}  
	\end{subfigure}
	\hfill
	\begin{subfigure}{.45\textwidth}
		\centering
		\includegraphics[width=.9\textwidth]{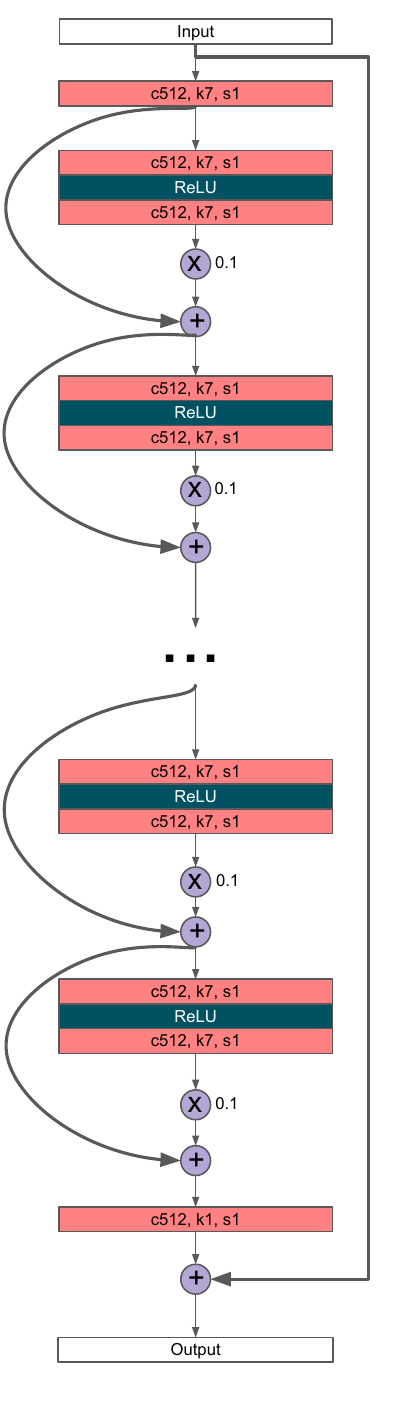}  
	\end{subfigure}
	\caption{Audio bandwidth extension models used. (Left) U-Net model, where dashed lines indicate stacking connections. (Right) ResNet model with 15 residual blocks. c, k, and s indicate channel size, kernel size, and stride of the convolutional layers, respectively.}
	\label{fig:audio_models}
\end{figure}

A common issue with training vanilla feed-forward neural networks with many layers is the "vanishing gradient" problem, in which the gradient back-propagated to the earliest layers approaches zero, due to repeated multiplications. Residual networks \parencite{resnet} eliminated this problem by using \textit{residual blocks}, which only model a fraction of the difference between their inputs and outputs. Commonly, each residual block includes two convolutional layers and a nonlinear function in between them. Very deep models include \textit{residual scaling} in which the output of each residual block is multiplied by a small number, e.g., $0.1$, and then summed with its input, to further stabilize training. Our ResNet model is represented in Figure \ref{fig:audio_models} (right).

Unlike the U-Net, the ResNet activation lengths stay constant throughout the network. In this way we can avoid any loss of temporal information since our goal is to create a high-resolution output of equal length to the input. Note that we use a simple design where all convolutional layers except the last one have the same number of parameters. Similar to the U-Net implementation, it has $55.1$ million parameters.

In all our models, all convolutions apply the appropriate zero padding to keep the activation sizes constant. This is even true for downsampling convolutions since the downsampling effect is achieved using strided convolutions.

To analyze generalization, we present ablation studies, in which we incorporate common methods to avoid overfitting, defined as \textit{regularization methods}.

\section{Regularization methods}
\label{sec:regularization}

We now detail the standard regularization methods such as dropout and batch normalization, as well as our novel data augmentation strategy.

\subsection{Dropout}

One of the simplest and most common methods to prevent overfitting is dropout, where activation units are dropped based on a fixed probability \parencite{dropout}. This introduces noise within the hidden layers and prevents the neurons from excessive co-adaptation.

Even though dropout has been largely superceded in the literature by batch normalization, especially in residual networks, new state-of-the-art residual models, namely wide residual networks \parencite{wide_resnet} do employ it. Furthermore, \textit{Audio-SR-U-Net}'s open-source implementation\footnote{\url{https://github.com/kuleshov/audio-super-res}} uses a dropout layer instead of batch normalization and thus, we followed this implementation in our U-Net model and used dropout layers after each upsampling convolutional layer. In our ResNet model, we placed dropout layers between the two convolutional layers of each residual block. For all experiments, we set the dropout rate at $0.5$.

\subsection{Batch normalization}

While training DNNs, updating the parameters of the model effectively changes the distribution of the inputs for the next layers. This is defined as \textit{internal covariate shift} and batch normalization addresses this problem by normalizing the layer inputs \parencite{batchnorm}. Even though batch normalization is mainly proposed to speed up training, it provides regularization as well. Because the parameters for the normalization are learned based on each batch, they can only provide a noisy estimate of the true mean and variance. Normalization using these estimated parameters introduces noise within the hidden layers and reduces overfitting. 

For the U-Net, we followed \textit{Audio-SR-U-Net} model \parencite{kuleshov} and inserted batch normalization layers after each downsampling convolutional layer. For the ResNet, batch normalization is used after each convolutional layer.

\subsection{Data augmentation}

To increase sample generalization of DNNs, data augmentation is used, where the input data samples are transformed before being fed into the DNN, effectively increasing the number and diversity of training samples. Data augmentation is very common in image-based tasks, and mostly utilizes geometric transformations such as rotating, flipping, or cropping \parencite{dataaugmentation}. Geometric transformations of this kind when applied to music signals typically do not produce realistic samples and while some work has been conducted on data augmentation for musical signals \parencite{mcfee2015_augmentation} it primarily targets robustness for classification tasks such as instrument identification via lossy transformations including time-stretching and pitch shifting, which cannot be applied in this context. 

\begin{figure}[t]
	\begin{center}
		\includegraphics[width=0.7\linewidth]{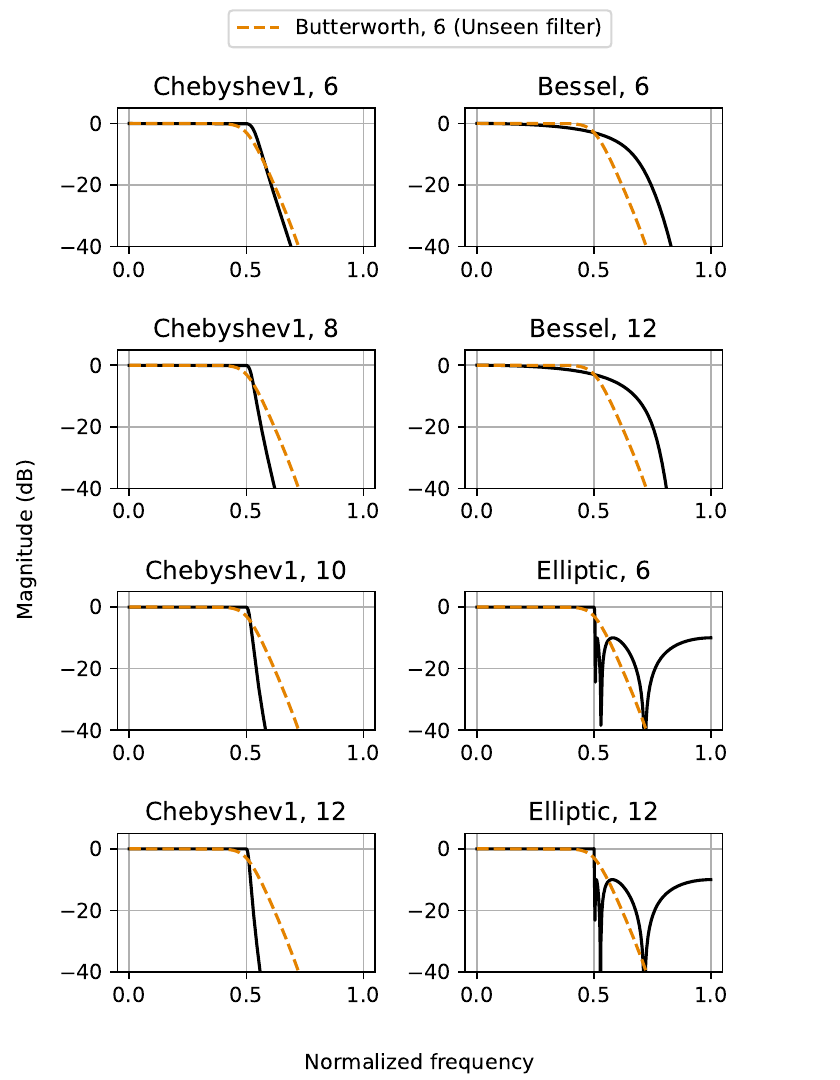}
	\end{center}
	\caption{Frequency responses of the training filters. The frequency response of the unseen filter, $6$th order Butterworth is superimposed on each plot.}
	\label{fig:filters}
\end{figure}

Since our main goal is to explore and then improve filter generalization, we propose a data augmentation method where many different types of filters are used during training. Our baseline methods without data augmentation, uses a \textit{single-filter} training setting, specifically a $6$th order Chebyshev Type $1$, denoted "Chebyshev-$1$, $6$". Whereas using data augmentation, in a \textit{multi-filter} training setting, we adopt a set of eight different filters, picked randomly for each input sample during training. These eight filters consist of \textit{Chebyshev-$1$}, \textit{Bessel}, and \textit{Elliptic} filters of different orders. To evaluate filter generalization, we reserve the $6$th order \textit{Butterworth} filter as the unseen filter. The filters are summarized in Table \ref{table:filters}, and their usage during evaluation is detailed in Section \ref{sec:eval}. A graphical overview of their different frequency magnitude responses is shown in Figure \ref{fig:filters}.

\begin{table}[htbp]
	\centering
	\caption{The types and orders of the low-pass filters used, under two different training settings, \textit{single-filter} (no data augmentation) and \textit{multi-filter} (data augmentation).}
	\begin{tabular}{c||c|c}
		& \begin{tabular}[c]{@{}c@{}}Single-filter\\ (No data augmentation) \vspace{1mm}\end{tabular}  & \begin{tabular}[c]{@{}c@{}}Multi-filter\\ (Data augmentation) \vspace{1mm}\end{tabular}    \\ \hline \hline
		\begin{tabular}[c]{@{}c@{}} \vspace{1mm} \\ Training \vspace{4mm}  \\ \hline \\ \vspace{-2mm} \\ Validation with\\ seen filter(s) \\ \vspace{-2mm} \end{tabular}                & Chebyshev-$1$, $6$  & \begin{tabular}[c]{@{}c@{}} \vspace{-2mm} \\ Chebyshev-$1$, $6$\\ Chebyshev-$1$, $8$\\ Chebyshev-$1$, $10$\\ Chebyshev-$1$, $12$\\ Bessel, $6$\\ Bessel, $12$\\ Elliptic, $6$\\ Elliptic, $12$ \vspace{1mm}\end{tabular} \\ \hline
		\begin{tabular}[c]{@{}c@{}} \vspace{-2mm} \\ Validation with\\ unseen filter \\ \vspace{-2mm} \\ \hline \\ \vspace{-5mm} \\ Testing with\\ unseen filter \\ \vspace{-2mm} \end{tabular} & Butterworth, $6$ & Butterworth, $6$                                                                                                                                                \\ \hline
		\begin{tabular}[c]{@{}c@{}}\vspace{-2mm} \\ Testing with\\ seen filter \\ \vspace{-2mm}\end{tabular}                                        & Chebyshev-$1$, $6$  & Chebyshev-$1$, $6$                                                                                                                                                
	\end{tabular}

	\label{table:filters}
\end{table}

Combining the three regularization methods with the two baseline architectures we obtain the following eight models which are used in our experiments: baseline U-Net (\textit{U-Net}), U-Net with data augmentation (\textit{U-Net DA}), U-Net with batch normalization (\textit{U-Net BN}), U-Net with dropout (\textit{U-Net DO}), baseline ResNet (\textit{ResNet}), ResNet with data augmentation (\textit{ResNet DA}), ResNet with batch normalization (\textit{ResNet BN}), and ResNet with dropout (\textit{ResNet DO}).

\section{Dataset}

Machine learning approaches to bandwidth extension formulate the problem via the use of datasets that contain both full-bandwidth (high-quality) and band-limited (low-quality) versions of each audio signal. A straightforward way to construct these pairs of samples is to obtain a high-quality dataset and then to low-pass filter it. Even though there are many musical audio datasets, especially within the music information retrieval community, many of them are collated from diverse sources (including researchers' personal audio collections) and often contain audio content has been compressed (e.g. via lossy MP3/AAC encoding), hence they are not strictly full-bandwidth. 

Other than the need for full-bandwidth musical audio content, our proposed approach is intended to be agnostic to musical style. To this end, any uncompressed full-bandwidth musical content could be used as training material, however in order to allow reproducibility, we select the following two publicly available datasets, which contain full-bandwidth, stereo and multi-track musical audio: MedleyDB (version 2.0) \parencite{medleydb} and DSD100 \parencite{dsd100}. In each dataset, the audio content is sampled at $44100$\,Hz (i.e., CD quality), and hence its bandwidth is $22050$\,Hz.

MedleyDB consists of $121$ songs, while DSD100 has two splits for training and testing, each containing $50$ songs. Given the inclusion of isolated multi-track stems, both datasets have found high uptake in music mixing and audio source separation research. However, we seek to address bandwidth extension for multi-instrument music as opposed to isolated single instruments, and thus we retain only the stereo mixes of each song. To create band-limited input samples, we apply a low-pass filter with a fixed cut-off of $11025$\,Hz, i.e., half the bandwidth of the original.

The DSD100 test split is used for testing, the last $8$ songs of DSD100 training split are used for validation, with all remaining songs of DSD100 training split plus the entire MedleyDB dataset used for training.

\section{Evaluation}
\label{sec:eval}

We present our evaluation methods for validation, which reveal overfitting trends during training, and for testing, which indicate final performance, along with an explanation of the metric used.

\subsection{Metric}

Since our investigation focuses more on the question of generalization rather than performance, we evaluate our experiments using a single well-established metric, the signal-to-noise ratio (SNR): 
\begin{equation}
\textrm{SNR}(x,\hat{x}) = 10 \log_{10} \frac{||x||_2^2}{||x-\hat{x}||_2^2} 
\end{equation}
where $x$ is the reference signal and $\hat{x}$ is its approximation. While calculating the $2$-norms, the signals are used in their stereo forms. In the specific context of our work, we consider SNR to be an appropriate choice to investigate overfitting since our models are trained with the mean-squared loss, and minimizing it corresponds to maximizing SNR.

To provide additional insight into performance, we evaluate the perceptual quality of the output audio samples, using the VGG distance, as used recently \textcite{denoising} for the evaluation of music enhancement. The VGG distance between two audio samples is defined as the distance between their embeddings created by the \textit{VGGish} network pretrained on audio classification \parencite{vggish}. 
\textcite{perceptual_metric} shows that the distance between deep embeddings correlates better to human evaluation, compared to hand-crafted metrics such as Perceptual Evaluation of Speech Quality (PESQ) \parencite{pesq} and the Virtual Speech Quality Objective Listener (ViSQOL) \parencite{visqol}, across various audio enhancement tasks including bandwidth extension. The \textit{VGGish} embeddings are also used in measuring the Fréchet Audio Distance (FAD), a state-of-the-art evaluation method to assess the perceptual quality of a collection of output samples \parencite{fad}. However, because FAD is used to compare two collections rather than individual audio signals, it is not applicable in our case.

To obtain the VGG embeddings, we used the \textit{VGGish} network's open-source implementation\footnote{\url{https://github.com/tensorflow/models/tree/master/research/audioset/vggish}}. We used the default parameters except setting the sampling frequency to $44100$\,Hz and the maximum frequency to $22050$\,Hz. In contrast to the SNR calculation, the reference implementation downmixes the stereo signals to mono before calculating the VGG embeddings. After post-processing, the embeddings take values from $0$ to $255$.  Similar to \textcite{perceptual_metric}, we employ the mean absolute distance to define the VGG distance as:

\begin{equation}
\textrm{VGG}(x,\hat{x}) = \frac{1}{n} \sum_{i=1}^n |y_i - \hat{y}_i |
\end{equation}

where $x$ is the reference signal, $\hat{x}$ is its approximation; $y$ and $\hat{y}$ are their embeddings, respectively. $n$ is the size of the embedding tensors, which depends on the length of $x$.

\subsection{Testing}

To assess the overall performance of our models, we perform testing once, at the end of the training. The test split of the DSD100 dataset is reserved for our testing stage. Due to GPU memory limitations, our networks cannot process full-length songs in a single forward pass, hence they process non-overlapping chunks of audio and the outputs are later concatenated to create full-length output songs. For both VGG distance and SNR, we calculate them at the song level first, based on these full-length songs, and then take the mean over the data split to obtain the final test values.

To evaluate filter generalization, we perform two tests for each model, using seen and unseen filters. As summarized in Table \ref{table:filters}, the $6$th order Butterworth filter is selected as the unseen filter, as it is not used in any training setting. The seen filter only includes $6$th order Chebyshev-$1$, as this is the only filter common to both single and multi-filter training settings.

\subsection{Validation}
\label{sssec:validation}

To observe generalization or overfitting throughout training, we perform validation repeatedly, once every $2500$ training iterations. We perform validation on 8-second audio excerpts, starting from the $8$th second of each song, for only eight songs. These eight songs are the last $8$ of the DSD100 training split. Since the validation is performed repeatedly throughout training, we keep the validation set sample size small. We believe that this small sample size is sufficient, because validation is only used to observe the progression of training, and the final performance evaluation is done in the testing stage. Final SNR values are calculated first by calculating the SNR over each $8$\,s excerpt, and then averaging over the validation songs.

Similar to testing, the validation is also performed twice, using seen and unseen filters. Validation with the unseen filter uses the $6$th order Butterworth filter, as in testing. Because validation with the seen filter(s) is done to observe the training progression of each model and not to compare different models, the filters employed are the same as those in the corresponding training setting. As seen in Table \ref{table:filters}, in the single-filter setting, validation with the seen filter only has the Chebyshev-$1$ filter, and in the multi-filter setting, it uses all eight training filters, with each assigned to processing a different song in the validation data split. 

\section{Implementation details}

We build and train our models using the Pytorch framework \parencite{pytorch} and a single Nvidia GeForce GTX 1080 Ti GPU. The model weights are initialized randomly with values drawn from the normal distribution with zero mean and unit variance. The batch size is $8$. We use the Adam optimizer \parencite{adam} with an initial learning rate of $5$e-$4$, and with beta values 0.9 and 0.999. The learning rate is halved when the training loss reaches a plateau. We record the average training loss every $2500$ iterations, and consider a plateau to correspond to no decrease in loss for $5$ such consecutive measurements. Training samples are created by first randomly picking an audio file from the training dataset and then, at a random location in the audio file, extracting a chunk of stereo audio, with a length of $8192$ samples, corresponding to $186$ milliseconds. However, since all our models are fully-convolutional, they can process audio signals with arbitrary lengths. We train our models until convergence and for testing we use the model weights taken from the conclusion of the training.

\section{Results}
\label{sec:results_audio}

\begin{figure*}[htbp]
	\begin{center}
		\includegraphics[width=1\textwidth]{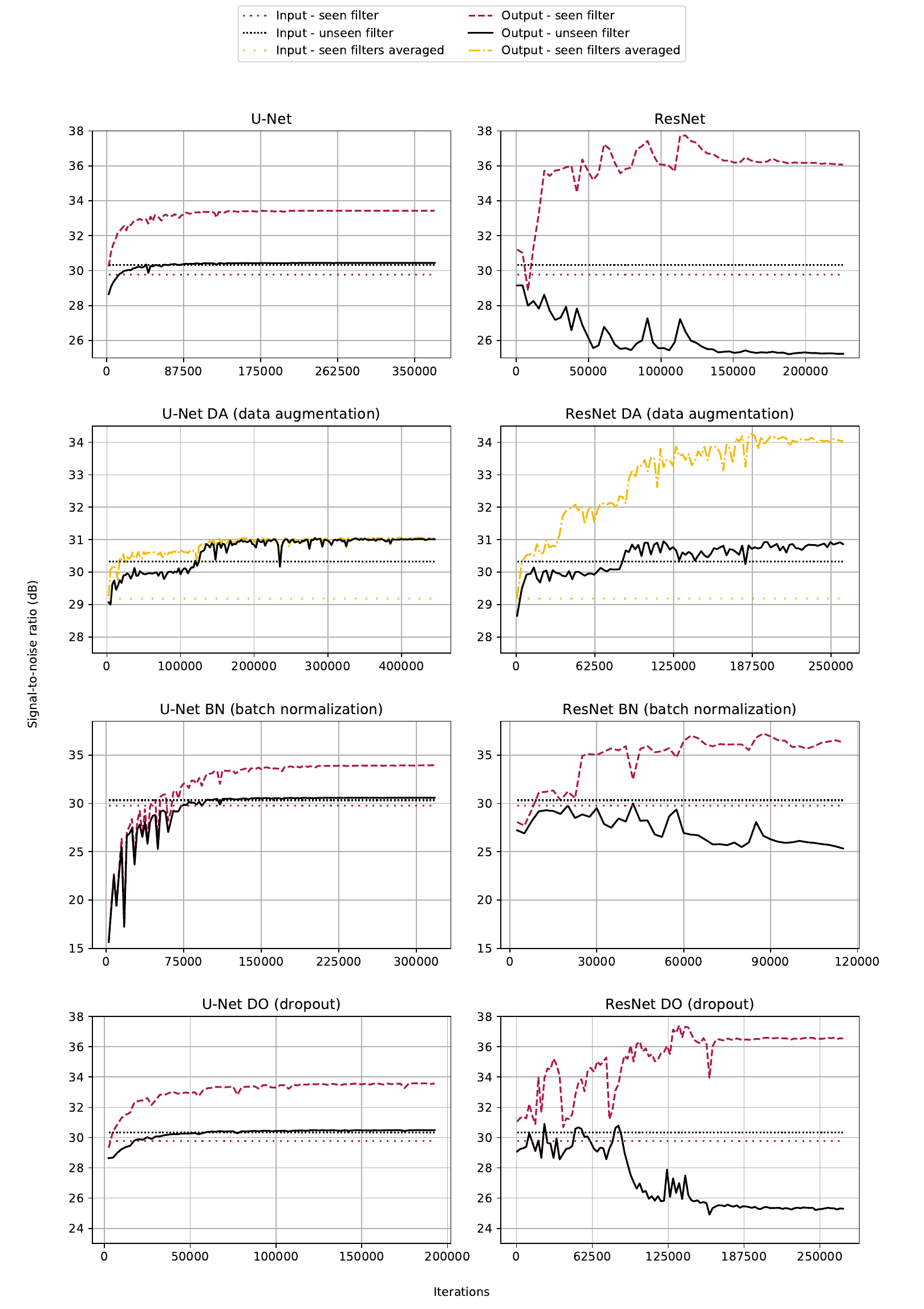}
	\end{center}
	\caption{Validation performance of our models throughout training.}
	\label{fig:all}
\end{figure*}

\subsection{Validation Data}
Figure \ref{fig:all} provides a high-level overview of the performance of all of the different models and training schemes, with the SNR as a function of the training iterations. While the horizontal dashed lines indicate a baseline of input SNR levels, the rest of the lines denote output SNRs for both validation settings. The SNR levels of the input validation with seen filter(s) are different for the experiments with data augmentation since a different number of training filters are used as summarized in Table \ref{table:filters}, and as shown in Figure \ref{fig:filters} their differing frequency responses naturally lead to different baseline SNRs.

Examining the first row of Figure \ref{fig:all} we see for both networks, when the input filter is known, then large improvements in SNR over the baseline are possible. However, contrasting the U-Net with the ResNet we see performance with the unseen filter is markedly different. For the U-Net the output SNR converges to the baseline, but for the ResNet performance degrades as training continues. In this way, we see quite clear evidence of a lack of filter generalization in both models.

Moving to the second row, where training includes data augmentation, we can observe a different pattern, where both networks can improve upon the baseline SNR for the unseen filter. Contrasting the U-Net with the ResNet, we can see the ResNet offers a greater improvement upon the set of seen filters than the U-Net, albeit for approximately the same number of parameters. 

Inspection of the third and fourth rows which include the two regularization techniques, we can observe a largely similar pattern to the first row, whereby the U-Net again converges to the input SNR, and the ResNet results in a lower SNR than the input. In summary, we see that for both networks, it is only training with data augmentation that we are able to find any improvement in SNR over the input. 



\subsection{Testing Data}
As described in Section \ref{sssec:validation}, the validation dataset is small, and the results shown in Figure \ref{fig:all} are calculated and averaged across short excerpts of $8$\,s in duration. In Table \ref{table:audio_results}, we present the performance of our models on the testing data, which includes the measurement of the SNR and the VGG distance as a perceptual measure, across the entire duration of the test dataset. $\Delta$SNR and $-\Delta$VGG represent the improvements with respect to the inputs. For SNR, $\Delta$SNR and $-\Delta$VGG higher is better and for VGG lower is better. DA, BN, and DO correspond to data augmentation, batch normalization, and dropout, respectively. The value range of the VGG embeddings and the VGG distances is $0$ to $255$. 

When testing with the seen filter, the ResNet models without data augmentation outperform all variants of U-Net by at least $4$\,dB, achieving more than a $7$\,dB improvement over the input SNR. The best performing model is ResNet with dropout, improving upon the input SNR by $7.4$\,dB. We also observe that the inclusion of data augmentation reduces performance when evaluated using the seen filter.

\begin{table}[htbp]
	\centering
		\caption{Output signal-to-noise ratio (SNR) and absolute VGG distances (VGG) on the test dataset, and their improvements with respect to the inputs.}
	
	\begin{tabular}{cccccc}
		Filter & Experiment & SNR  & $\Delta$SNR  & VGG & $-\Delta$VGG  \tabularnewline   \hline
		\multirow{9}{*}{\begin{tabular}[c]{@{}c@{}}Chebyshev1$-6$ \\(seen filter) \end{tabular}}    
		
		& Input       & $27.86$ &         & $46.55$ &         \\
		& U-Net       & $30.34$ & $+2.47$ & $41.04$ & $+5.51$ \\
		& U-Net DA    & $29.78$ & $+1.91$ & $44.29$ & $+2.26$ \\
		& U-Net BN    & $30.90$ & $+3.03$ & $41.52$ & $+5.03$ \\
		& U-Net DO    & $30.49$ & $+2.62$ & $41.51$ & $+5.04$ \\
		& ResNet      & $34.94$ & $+7.08$ & $39.02$ & $+7.53$ \\
		& ResNet DA   & $30.48$ & $+2.62$ & $40.11$ & $+6.43$ \\
		& ResNet BN   & $34.37$ & $+6.50$ & $39.41$ & $+7.14$ \\
		& ResNet DO   & $\mathbf{35.27}$ & $\mathbf{+7.41}$ & $\mathbf{37.23}$ & $\mathbf{+9.32}$ \\
		
		\hline
		\multirow{9}{*}{\begin{tabular}[c]{@{}c@{}}Butterworth$-6$ \\(unseen filter) \end{tabular}}
		
		& Input       & $27.37$ &         & $47.11$ &         \\
		& U-Net       & $28.55$ & $+1.18$ & $41.90$ & $+5.21$ \\
		& U-Net DA    & $29.00$ & $+1.63$ & $44.80$ & $+2.31$ \\
		& U-Net BN    & $28.77$ & $+1.40$ & $42.06$ & $+5.06$ \\
		& U-Net DO    & $28.62$ & $+1.24$ & $42.34$ & $+4.78$ \\
		& ResNet      & $21.96$ & $-5.41$ & $47.12$ & $-0.01$ \\
		& ResNet DA   & $\mathbf{29.16}$ & $\mathbf{+1.78}$ & $\mathbf{40.52}$ & $\mathbf{+6.59}$ \\
		& ResNet BN  
		& $23.23$ & $-4.14$ & $46.38$ & $+0.73$ \\
		& ResNet DO  
		& $22.10$ & $-5.27$ & $46.15$ & $+0.96$ \\
		
	\end{tabular}

	\label{table:audio_results}
\end{table}

When testing with the unseen filter, the two best performing models use our proposed data augmentation method. Here, the ResNet variants without data augmentation produce output SNR levels which are well below those of the input. Their output SNRs are around $5.3$\,dB lower than the input, and around $6.6$\,dB lower than their U-Net based counterparts. The addition of data augmentation improves the performance of both the baseline U-Net and ResNet. Although this improvement is marginal for the U-Net, at $0.45$\,dB, for the ResNet, we observe a much larger improvement of $7.2$\,dB. In testing with the unseen filter, the best performing model is the ResNet with data augmentation, which improves upon the input SNR by $1.8$\,dB.

Considering the VGG distances, the results of the U-Net variants do not change much across different filters. Compared to the seen filter setting, the ResNet variants without data augmentation exhibit worse results with the unseen filter, however, these values are very close to the input value, hence the filter overfitting in terms of the VGG distance is not as severe as the SNR. For the unseen filter setting, while the incorporation of data augmentation worsens the VGG distance by $2.8$ for U-Net, it produces a much larger improvement of $6.6$ for ResNet, making ResNet with data augmentation the best performing model in terms of VGG distance and SNR.

Quantitative results for each test song, along with three audio excerpts can be found online\footnote{\url{https://serkansulun.com/bwe}}.

\subsection{Sample Overfitting}
In Table \ref{table:nooverfit} we present the performance of our baseline models, without any regularization method, on the training and testing data splits separately, and evaluated on all samples in the data splits, across their full duration. Inputs are created using the low-pass filter which was also used during training (the seen filter, $6$th order Chebyshev-$1$). To provide insight into whether any sample overfitting is occurring (i.e., that the networks are somehow memorizing the audio content of the training data) we use the seen filter, the 6th order Chebyshev-$1$, during testing. For both the baseline U-Net and ResNet, between training and testing data splits, the SNR improvement over the input suggests no overfitting to the audio samples themselves.

\begin{table}[ht]
	\centering
	\caption{Output signal-to-noise ratio (SNR) of our baseline models, without any regularization method, on training and testing data splits separately, and their comparison to the input SNR.}
	\begin{tabular}{c|ccc}
		\begin{tabular}[c]{@{}c@{}}Data\\ split\end{tabular} & Experiment  & SNR (dB)   & \begin{tabular}[c]{@{}c@{}}Improvement\\ over input (dB)\end{tabular} \\ \hline \hline
		\multirow{3}{*}{Training}                               & Input  & 25.99 &                                                                  \\
		& U-Net  & 28.34 & +2.35                                                            \\
		& ResNet & 33.00 & +7.01                                                            \\ \hline
		\multirow{3}{*}{Testing}                                & Input  & 27.86 &                                                                  \\
		& U-Net  & 30.34 & +2.48                                                            \\
		& ResNet & 34.94 & +7.08                                                           
	\end{tabular}

	\label{table:nooverfit}
\end{table}

\subsection{U-Net vs ResNet: Model Comparison}
As stipulated in Section \ref{sec:models}, we allow both the U-Net and ResNet to have a similar number of parameters. However, we informally observed a distinct difference in training time. In Table \ref{table:specs}, we show several objective properties of these networks, namely the number of parameters, number of multiply-accumulate operations (MACs), and runtimes of our baseline models. Number of MACs roughly correspond to half of the number of floating point operations (FLOP). Runtime rate is the time spent in seconds, to process a signal with length of one second, during testing, i.e., a forward pass where no gradients are calculated. While both models have roughly the same number of parameters, we see that the U-Net has a much lower runtime and fewer MACs. This is due to its autoencoder-like shape, in which the convolutional layers with more channels are near the bottleneck of the network, where the spatial activation size is the smallest, effectively reducing the number of MACs and the runtime. Looking again at Figure \ref{fig:all}, we can speculate that the the ResNet has greater representation power than the U-Net, as shown by its ability to better model multiple known filters than the U-Net, albeit at the cost of slower training and inference. 

\begin{table}[h]
	\centering
	\caption{Number of parameters, number of multiply-accumulate operations (MACs) and runtimes of our models.}
	\begin{tabular}{c|ccc}
		Model       & \begin{tabular}[c]{@{}c@{}}Number of \\ parameters\end{tabular} & \begin{tabular}[c]{@{}c@{}}Number of \\ MACs\end{tabular} & \begin{tabular}[c]{@{}c@{}}Runtime \\ rate\end{tabular} \\ \hline
		U-Net  & 56.4M                                                           & 415.3G                                                    & 0.14                                                    \\
		ResNet & 55.1M                                                           & 3609.4G                                                   & 1.06                                                   
	\end{tabular}
	
	\label{table:specs}
\end{table}

\subsection{Visualization of bandwidth extension}

\begin{figure}[htbp]
	\begin{center}
		\includegraphics[width=0.99\linewidth]{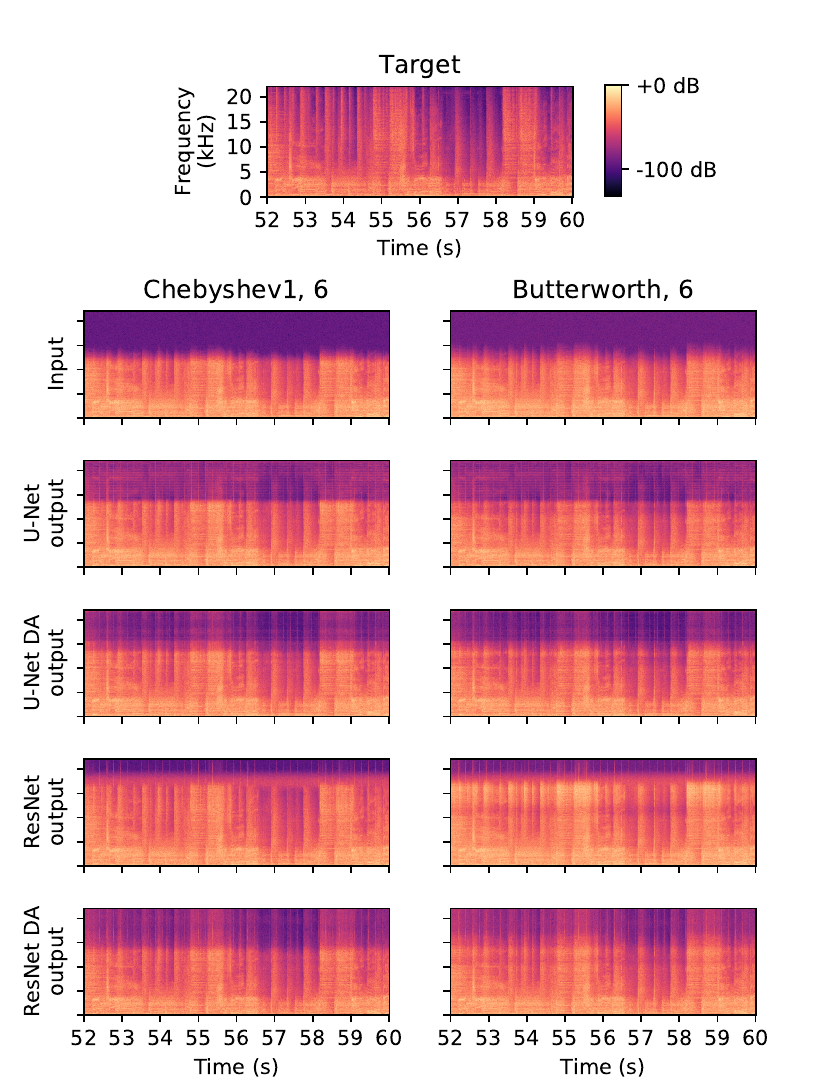}
	\end{center}
	\caption{Spectograms of sample audio segments.}
	\label{fig:spectrograms}
\end{figure}

\begin{figure}[htbp]
	\begin{center}
		\includegraphics[width=0.99\linewidth]{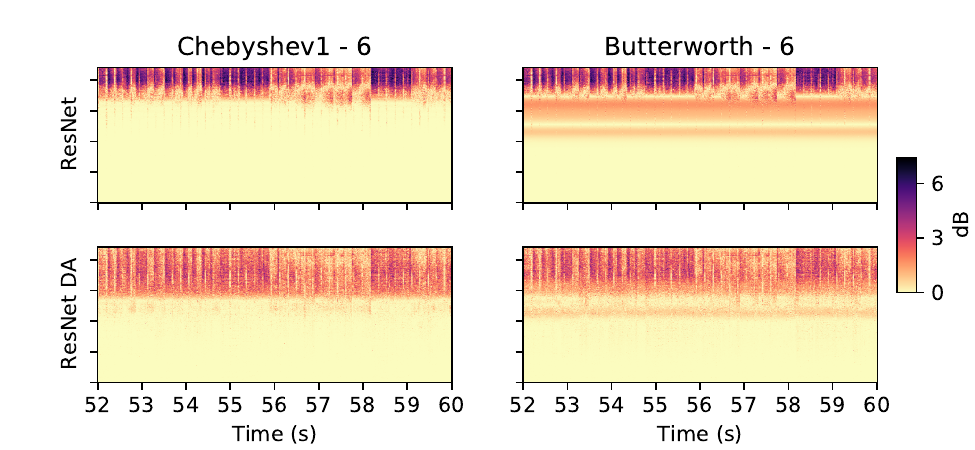}
	\end{center}
	\caption{Absolute difference with respect to the target spectrogram. The colormap is inverted for better visibility.}
	\label{fig:diff_spectrograms}
\end{figure}

While our proposed method operates entirely in the time-domain, we provide a graphical overview of the outputs of the two networks contrasting the baseline versions with the inclusion of data augmentation for both seen and unseen filters. To this end, we illustrate the spectrograms of one audio excerpt from the test set under each of these conditions in Figure \ref{fig:spectrograms}. The inspection of the figure reveals quite different behavior of the U-Net compared to the ResNet. In general, we can observe more prominent high-frequency information in the output of the ResNet. Of particular note, is the frequency region between approximately $12$-$17$\,kHz for baseline ResNet, and the unseen Butterworth filter, which, contrasting with the target, appears to have "over-enhanced" this region. By contrast, once the data augmentation is included, this high-frequency boosting is no longer evident. To emphasize this phenomenon further, in Figure \ref{fig:diff_spectrograms} we display the absolute difference with respect to the target spectrogram, for the baseline ResNet and ResNet with data augmentation. For the unseen Butterworth filter, in the upper half of the spectrogram, the absolute difference of the ResNet with data augmentation is much smoother compared to the baseline ResNet. In this visual representation, we can clearly observe that under all conditions the lower part of the absolute difference spectrogram is essentially unchanged, which reflects the direct additive connection of the input to the output in the network architectures.


\section{Discussion}

We raise the issue of filter generalization for deep neural networks applied to musical audio bandwidth extension. Contrary to many problems for which deep learning is used, we do not find any evidence of overfitting to audio samples themselves (i.e. the training data), but rather, we observe a clear trend for state of the art DNNs to overfit to filter shapes. When these DNNs are presented with audio excerpts that have been pre-processed with low-pass filters not included in the training, then no meaningful extension of the bandwidth can be obtained. Furthermore, the use of widely adopted regularization layers such as batch normalization and dropout fall short in alleviating this problem. Looking to the wider context and long term goal of musical audio bandwidth extension for audio enhancement, we believe that filter overfitting is a critical issue worthy of continued focus.


To address the filter overfitting issue we propose a novel data augmentation approach, which uses multiple filters at the time of training. Our results demonstrate that without data augmentation, filter overfitting increases as training progresses, whereas including data augmentation is a promising step towards achieving filter generalization. While the improvement in generalization for the U-Net is quite small, a more pronounced effect can be observed for the ResNet, which retains high performance across multiple seen training filters.
It is particularly noteworthy that the ResNet variants without data augmentation produce very poor results when tested with an unseen filter, with output quality well below that of the input. In this way, the incorporation of data augmentation was the only means to achieve SNR levels that are above the input. 

In addition to the primary findings concerning filter generalization, this is, to the best of our knowledge, the first comparison between U-Nets and ResNets in the field of audio processing, and perhaps the first ever comparison of these approaches given a similar number of parameters. Examining the results of testing with the seen filter, we observe that baseline ResNet outperforms baseline U-Net by a large margin. However, when testing with the unseen filter, baseline ResNet performs worst.

We argue that the ResNet has more representation power than the U-Net because while the U-Net reduces the spatial activation sizes in its downsampling blocks, the ResNet keeps the spatial activation sizes constant, starting from its input until its output, thus minimizing the loss of information. Even though the networks have the same number of parameters, we can quantify this higher representation power by comparing the number of multiply-accumulate operations. This higher representation power results in the ResNet performing much better in tests with the seen filter while demonstrating much higher levels of filter overfitting when there is no data augmentation. We show that using the proposed data augmentation method, this powerful network can be successfully regularized, and achieves the best SNR when testing with the unseen filter. While we chose to keep the number of parameters within the two models roughly equal, we note that compared to the ResNet, the U-Net is $7.5$ times faster and does nearly $9$ times fewer multiply-accumulate operations (MACs). In this way, the U-Net may be a preferred architecture for real-time streaming applications. 

We now move on to the conclusion, where we summarize the key findings and implications of our work and highlight potential future directions.

\chapter{Conclusion} \label{chap:conclusion}

In this thesis, we have presented a novel approach for video-based music generation, creating EMSYNC—an innovative framework that synchronizes music with video by aligning both emotional content and temporal boundaries. By integrating video analysis, emotion-based conditioning, and temporal boundary conditioning into a single music generation pipeline, EMSYNC provides a solution to key challenges in generating emotionally and rhythmically aligned music for video.

\section{Contributions}

One of the major contributions of this work is the creation of models for emotion classification of arbitrary videos and genre classification of cinematic trailers. These models are key to the success of our video-based music generation system, as they provide the emotional context and genre-specific features necessary for music generation. Our emotion classifier, VEMOCLAP, is capable of identifying the emotional tone of arbitrary videos, making it a versatile tool for various applications beyond music generation. Similarly, our genre classifier for cinematic trailers leverages pretrained models for feature extraction and allows for efficient, high-quality genre classification. These pretrained models are valuable resources for the community and have wide-ranging potential in multimedia content analysis, including automatic content categorization, sentiment analysis, and recommendation systems.

Another significant contribution is our music generator, which is the first to be conditioned on continuous-valued conditions—specifically, valence and arousal values—rather than discrete emotion categories. This breakthrough enables much finer control over the generated music, as it allows for the precise adjustment of emotional expression and musical dynamics. By avoiding the discretization of emotion features, we preserve the continuous nature of emotional expression, resulting in music that is more nuanced and reflective of the emotional complexity in the video content. This continuous-conditioning approach offers broader flexibility for fine-tuning music generation, allowing for more expressive and adaptive outputs that better align with varying emotional tones across different videos. Furthermore, this approach has implications for other sequence generation models, particularly in areas like natural language processing, where fine-grained control over generated content could enhance coherence and emotional depth in generated sequences.

We also created the first large-scale, emotion-labeled MIDI dataset specifically designed for music generation. By combining emotional annotations derived from the Spotify Developers API and song lyrics, this dataset provides a powerful resource for training emotion-based music generators. The impact of this emotion-labeled dataset is profound, as it bridges a gap in the field of symbolic music generation by providing emotion-rich training data that allows the generation of music explicitly aligned with emotional states. The dataset opens up new opportunities for research in affective music generation, enabling the creation of systems capable of producing music that responds dynamically to emotional inputs in various contexts such as films, games, and interactive media.

Furthermore, the emotion-labeled MIDI datasets created as part of this work have broader implications for the field of artificial intelligence and music. The dataset is a unique resource for training systems that not only generate music but also understand and adapt to the emotional qualities of a piece. By providing both continuous emotion labels and symbolic music data, we have set the stage for the investigation of more advanced music generation systems that can respond to emotional cues in a natural and seamless manner.

Our work also makes an important contribution by introducing a system for synchronizing video content with music through temporal boundary conditioning. By aligning musical events with video scene cuts, our approach ensures that the music is not only emotionally aligned but also rhythmically synchronized with the video’s key moments. This synchronization enhances the overall multimedia experience, particularly in applications such as film scoring, advertisements, and video games, where precise timing is crucial.

While MIDI events are not inherently aligned to a fixed time grid, our approach nonetheless enables effective temporal conditioning of the generator. This has broader implications for sequence processing tasks that involve transformers, as it demonstrates a new way of integrating temporal boundaries into generative models. This methodology can extend to various applications, such as text generation, speech synthesis, and other sequence-based tasks, where precise temporal alignment is crucial for maintaining coherence and enhancing the overall output.

In summary, EMSYNC represents a significant step forward in the field of video-based music generation. It is the first system to integrate emotion-based and boundary-based conditioning for music generation, enabling the creation of music that is both emotionally rich and rhythmically synchronized with video. The pretrained emotion and genre classifiers, the continuous-conditioning approach to music generation, and the emotion-labeled MIDI dataset are all major innovations that pave the way for more advanced and versatile music generation systems.

Moreover, EMSYNC outperforms the two existing methods in user studies across all subjective metrics, for both participant groups---with and without self-reported knowledge of music theory. Our method consistently ranks best in music richness, quality, emotional alignment, and timing synchronization, as confirmed by participants with varying levels of familiarity with music theory. This demonstrates not only the technical efficiency of EMSYNC but also its potential to reshape video-based music generation, offering a more expressive and adaptive solution to synchronizing music with video content. The results of these studies further validate the practical effectiveness of EMSYNC, establishing it as a notable advancement in this field.

Finally, we provided a preliminary analysis of audio generation through the audio bandwidth extension task. We highlighted a significant issue with the use of synthetic data, which is common in many works in the field of audio generation. Specifically, we demonstrated that models trained with one data synthesis method do not generalize well to data created by other synthesis methods. To address this, we took the first step towards mitigating this problem by diversifying the data synthesis methods to better approximate real-world scenarios. We believe that the insights from this work are valuable for other audio generation tasks, including MIDI-to-audio.

We make every stage of our research accessible to the community through open-source code, datasets, pretrained models, and interactive demos. By releasing these resources, we aim to foster collaboration and innovation not only within the research community but also for the general public. This openness allows other researchers to build upon our work, test new ideas, and advance the field of video-based music generation. Additionally, providing access to these tools empowers practitioners, developers, and enthusiasts to explore and apply our methods in diverse real-world contexts, from film production to game development, and beyond. Through this contribution, we hope to inspire further exploration and progress in both the academic and creative sectors.

Our major contributions in this thesis are summarized below.

\begin{itemize}
	\item We provide emotion labels in the form of valence-arousal values for 34,791 songs in the Lakh MIDI dataset \parencite{lmd}, creating the largest existing MIDI dataset with emotion labels.
	
	\item We present the first MIDI generator that can be conditioned on continuous valence-arousal values.
	
	\item We create the first MIDI generator that can be conditioned on temporal boundaries in a direct manner.
	
	\item Our cinematic trailer genre classifier outperforms all existing methods on MovieNet, the largest cinematic trailer dataset \parencite{movienet}.
	
	\item Our video emotion classifier outperforms all existing methods on Ekman-6, the largest user-generated (arbitrary) video dataset with emotion labels \parencite{ekman6}.
	
	\item Our video-based music generator outperforms all existing methods in user studies, across all metrics and demographics.
	
	\item We demonstrate that training audio generation models with synthetic data leads to generalization issues, and we take initial steps to address this challenge.
	
\end{itemize}

We list the papers we published throughout our research below.

\begin{itemize}
	
\item \textbf{S. Sulun}, M. E. P. Davies, and P. Viana, "Symbolic Music Generation Conditioned on Continuous-Valued Emotions," IEEE Access, vol. 10, pp. 44617–44626, 2022.

\item \sloppypar \textbf{S. Sulun}, P. Oliveira, and P. Viana, "Emotion4MIDI: A Lyrics-Based Emotion-Labeled Symbolic Music Dataset," in Progress in Artificial Intelligence, Cham: Springer Nature Switzerland, 2023, pp. 77–89.

\item \textbf{S. Sulun}, P. Viana, and M. E. P. Davies, "Video Soundtrack Generation by Aligning Emotions and Temporal Boundaries," IEEE Transactions on Multimedia, 2026. [in print].
	
\item \textbf{S. Sulun}, P. Viana, and M. E. P. Davies, "VEMOCLAP - A video emotion classification web application," IEEE International Symposium on Multimedia (ISM), 2024.

\item \textbf{S. Sulun}, P. Viana, and M. E. P. Davies, "Movie trailer genre classification using multimodal pretrained features," Expert Systems with Applications, vol. 258, p. 125209, 2024.

\item \textbf{S. Sulun} and M. E. P. Davies, "On filter generalization for music bandwidth extension using deep neural networks," IEEE Journal of Selected Topics in Signal Processing, vol. 15, no. 1, pp. 132–142, 2020.

\end{itemize}

\section{Future directions}

Looking ahead, we aim to offer insight and inspiration to future researchers in the fields of machine learning and multimedia. Throughout our research, we favored using large and relatively simple models trained on large datasets, rather than complex models trained on small datasets. The qualitative and quantitative superiority of our approach aligns with the current trend that training larger models on larger datasets yields better results~\parencite{larger_models,larger1,larger2}. This approach has long been adopted by the industry, with text generation models like ChatGPT\footnote{\url{https://www.chatgpt.com}}, LLaMA\footnote{\url{https://www.llama.com}}, and Gemini\footnote{\url{https://gemini.google.com/app}}, as well as image generation models like DALL-E\footnote{\url{https://openai.com/index/dall-e-3}} and Imagen\footnote{\url{https://deepmind.google/technologies/imagen-3}}, and audio generation models like Suno\footnote{\url{https://suno.com}}.

While the hardware capabilities of major tech companies give them a clear advantage, another overlooked factor is their access to large-scale proprietary datasets. Therefore, we emphasize the importance of creating newer, larger open-source, and peer-reviewed datasets, rather than focusing solely on building complex models on small datasets for incremental improvements over the existing state-of-the-art.

One effective approach to creating better datasets is by labeling existing ones. The success of image processing models like CLIP stems from the contrastive pretraining method, which pairs images with tags and other textual descriptions scraped from the internet~\parencite{clip}. Similar methods could be applied to other types of multimedia. However, a key challenge with this approach is the intentional or accidental use of copyrighted material. To keep pace with industry research advances, it seems likely that academic labs will need to collaborate with legal teams or allocate research funds for legal advice, just as the industry does.

Many datasets rely on content from video-sharing platforms~\parencite{movienet,ekman6}. The rise of large language models has made text data particularly valuable. Newer datasets should leverage the tags and descriptions of user-uploaded videos. Contrastive learning strategies can be especially useful for extracting meaningful features from pairs of video and text~\parencite{contrastive}.

Video emotion classification becomes more challenging when the analyzed video has a complex plot, with semantic variations and subtleties. Extracting this complex semantic information directly from raw pixels and audio would likely be inefficient, if not impossible. Our approach of using pretrained features is a step toward efficiently understanding this semantic complexity. As a next step, videos could be transcribed in detail into text using similar pretrained features, and large language models (LLMs) with reasoning capabilities~\parencite{reasoning} could process the resulting text to detect potential plot twists and nuances.

When data is scarce, using high-dimensional representations can lead to overfitting, often referred to as the \textit{curse of dimensionality}. Our method of using pretrained features effectively serves as a feature reduction technique, condensing large raw frames into smaller, more meaningful features. Other features can certainly be added. For instance, emotion classification of speech was one aspect we planned to include but struggled to find a reasonably performing open-source model. However, even though pretrained feature extractors work in inference mode, using many of them would significantly increase runtime. To mitigate this, employing knowledge distillation to reduce the sizes of these pretrained networks would be advantageous \parencite{distillation_2015}. Furthermore, combining multiple architectures into a single unified model would be a challenging yet rewarding research task. Finally, utilizing multiple datasets from different classifications through continual learning could help address data scarcity~\parencite{continual}.

The data scarcity problem is also evident with labeled MIDI datasets. Our work utilizes features from the Spotify Developer API to label MIDI files, which we refer to as "weak labels" since these features were originally created for audio, not MIDI. However, our work demonstrates that even when the labels are weak, having a large dataset with these labels is still beneficial. Future researchers can gather textual features, such as descriptions or tags associated with musical audio from the internet, assign them to their respective MIDI files, and apply LLMs and/or contrastive learning methods in a similar manner.

Generating multi-instrument audio from MIDI, or vice versa, using deep neural networks (DNNs) is a relatively underexplored area. Effective audio-to-MIDI models could facilitate the synthetic creation of large MIDI datasets. On the other hand, MIDI-to-audio models with realistic outputs would enable the production of high-quality sound from symbolic music compositions. As diffusion models have already demonstrated their ability to generate images of unprecedented quality, they could also be a promising candidate for high-quality, realistic audio generation \parencite{diffusion}.

Finally, the advent of Suno certainly represents a disruptive breakthrough in the field of music generation. While their methods are proprietary, it is plausible to assume that large private datasets of music in audio format contribute significantly to their success. The presentation of an open-source, peer-reviewed method that achieves similar levels of success would mark an incredible advancement in music generation research.



\PrintBib



@inproceedings{diffusion,
  author       = {Jascha Sohl{-}Dickstein and
Eric A. Weiss and
Niru Maheswaranathan and
Surya Ganguli},
title        = {Deep Unsupervised Learning using Nonequilibrium Thermodynamics},
booktitle    = {Proceedings of the 32nd International Conference on Machine Learning,
{ICML} 2015, Lille, France, 6-11 July 2015},
series       = {{JMLR} Workshop and Conference Proceedings},
volume       = {37},
pages        = {2256--2265},
publisher    = {JMLR.org},
year         = {2015},
url          = {http://proceedings.mlr.press/v37/sohl-dickstein15.html},
bibsource    = {dblp computer science bibliography, https://dblp.org}
}

@inproceedings{epia,
	author       = {Serkan Sulun and
	Pedro Oliveira and
	Paula Viana},
	title        = {Emotion4{MIDI}: {A} Lyrics-Based Emotion-Labeled Symbolic Music Dataset},
	booktitle    = {Progress in Artificial Intelligence - 22nd {EPIA} Conference on Artificial
	Intelligence, {EPIA} 2023, Faial Island, Azores, September 5-8, 2023,
	Proceedings, Part {II}},
	series       = {Lecture Notes in Computer Science},
	volume       = {14116},
	pages        = {77--89},
	publisher    = {Springer},
	year         = {2023},
	url          = {https://doi.org/10.1007/978-3-031-49011-8\_7},
	doi          = {10.1007/978-3-031-49011-8\_7},
	bibsource    = {dblp computer science bibliography, https://dblp.org}
}

@article{emsync,
	author       = {Serkan Sulun and
	Paula Viana and
	Matthew E. P. Davies},
	title        = {Video Soundtrack Generation by Aligning Emotions and Temporal Boundaries},
  journal   = {IEEE Transactions on Multimedia},
  year      = {2026},
  note      = {In print},
  url       = {https://arxiv.org/abs/2502.10154}
}

@inproceedings{sentiment,
  author       = {Jos{\'{e}} Camacho{-}Collados and
Kiamehr Rezaee and
Talayeh Riahi and
Asahi Ushio and
Daniel Loureiro and
Dimosthenis Antypas and
Joanne Boisson and
Luis Espinosa Anke and
Fangyu Liu and
Eugenio Mart{\'{\i}}nez C{\'{a}}mara},
title        = {TweetNLP: Cutting-Edge Natural Language Processing for Social Media},
booktitle    = {Proceedings of the The 2022 Conference on Empirical Methods in Natural
Language Processing, {EMNLP} 2022 - System Demonstrations, Abu Dhabi,
UAE, December 7-11, 2022},
pages        = {38--49},
publisher    = {Association for Computational Linguistics},
year         = {2022},
url          = {https://doi.org/10.18653/v1/2022.emnlp-demos.5},
doi          = {10.18653/V1/2022.EMNLP-DEMOS.5},
bibsource    = {dblp computer science bibliography, https://dblp.org}
}

@article{valence_arousal,
  title = {A Circumplex Model of Affect.},
  author = {Russell, James A.},
  year = {1980},
  journal = {Journal of personality and social psychology},
  volume = {39},
  number = {6},
  pages = {1161},
  publisher = {American Psychological Association},
  doi={10.1037/h0077714}
}

@article{mapping,
title = {Evidence for a three-factor theory of emotions},
journal = {Journal of Research in Personality},
volume = {11},
number = {3},
pages = {273-294},
year = {1977},
issn = {0092-6566},
doi = {https://doi.org/10.1016/0092-6566(77)90037-X},
url = {https://www.sciencedirect.com/science/article/pii/009265667790037X},
author = {James A Russell and Albert Mehrabian},
}

@article{ekman,
  title = {Universals and Cultural Differences in Facial Expressions of Emotion},
  author = {Ekman, Paul},
  year = {1971},
  journal = {Nebraska Symposium on Motivation},
  volume = {19},
  pages = {207--283},
  publisher = {University of Nebraska Press},
  address = {US},
  issn = {0146-7875},
  url={https://psycnet.apa.org/record/1973-11154-001},
}

@inproceedings{musictransformer,
  author       = {Cheng{-}Zhi Anna Huang and
Ashish Vaswani and
Jakob Uszkoreit and
Ian Simon and
Curtis Hawthorne and
Noam Shazeer and
Andrew M. Dai and
Matthew D. Hoffman and
Monica Dinculescu and
Douglas Eck},
title        = {Music Transformer: Generating Music with Long-Term Structure},
booktitle    = {7th International Conference on Learning Representations, {ICLR} 2019,
New Orleans, LA, USA, May 6-9, 2019},
publisher    = {OpenReview.net},
year         = {2019},
url          = {https://openreview.net/forum?id=rJe4ShAcF7},
bibsource    = {dblp computer science bibliography, https://dblp.org}
}

@article{event_encoding,
  author       = {Sageev Oore and
Ian Simon and
Sander Dieleman and
Douglas Eck and
Karen Simonyan},
title        = {This time with feeling: learning expressive musical performance},
journal      = {Neural Computing and Applications},
volume       = {32},
number       = {4},
pages        = {955--967},
year         = {2020},
url          = {https://doi.org/10.1007/s00521-018-3758-9},
doi          = {10.1007/S00521-018-3758-9},
bibsource    = {dblp computer science bibliography, https://dblp.org}
}

@inproceedings{lpd,
title={MuseGAN: Multi-track Sequential Generative Adversarial Networks for Symbolic Music Generation and Accompaniment}, volume={32}, url={https://ojs.aaai.org/index.php/AAAI/article/view/11312}, DOI={10.1609/aaai.v32i1.11312}, number={1}, journal={Proceedings of the AAAI Conference on Artificial Intelligence}, author={Dong, Hao-Wen and Hsiao, Wen-Yi and Yang, Li-Chia and Yang, Yi-Hsuan}, year={2018}, month={4} }

@phdthesis{lmd,
  author       = {Colin Raffel},
title        = {Learning-Based Methods for Comparing Sequences, with Applications
to Audio-to-MIDI Alignment and Matching},
school       = {Columbia University, {USA}},
year         = {2016},
url          = {https://doi.org/10.7916/D8N58MHV},
doi          = {10.7916/D8N58MHV},
bibsource    = {dblp computer science bibliography, https://dblp.org}
}

@inproceedings{di,
  author       = {Shangzhe Di and
Zeren Jiang and
Si Liu and
Zhaokai Wang and
Leyan Zhu and
Zexin He and
Hongming Liu and
Shuicheng Yan},
title        = {Video Background Music Generation with Controllable Music Transformer},
booktitle    = {{MM} '21: {ACM} Multimedia Conference, Virtual Event, China, October
20 - 24, 2021},
pages        = {2037--2045},
publisher    = {{ACM}},
year         = {2021},
url          = {https://doi.org/10.1145/3474085.3475195},
doi          = {10.1145/3474085.3475195},
bibsource    = {dblp computer science bibliography, https://dblp.org}
}

@inproceedings{foley,
  title = {Foley Music: Learning to Generate Music from Videos},
  shorttitle = {Foley Music},
  booktitle = {Computer Vision - ECCV 2020 - 16th European Conference},
  author = {Gan, Chuang and Huang, Deng and Chen, Peihao and Tenenbaum, Joshua B. and Torralba, Antonio},
  year = {2020},
  volume = {12356},
  pages = {758--775},
  publisher = {Springer},
  doi = {10.1007/978-3-030-58621-8_44},
  urldate = {2025-01-24}
}

@inproceedings{sighttosound,
  title = {Sight to Sound: An End-to-End Approach for Visual Piano Transcription},
  shorttitle = {Sight to Sound},
  booktitle = {IEEE International Conference on Acoustics, Speech and Signal Processing, ICASSP 2020},
  author = {Koepke, A. Sophia and Wiles, Olivia and Moses, Yael and Zisserman, Andrew},
  year = {2020},
  pages = {1838--1842},
  publisher = {IEEE},
  doi = {10.1109/ICASSP40776.2020.9053115}
}

@inproceedings{audeo,
  author       = {Kun Su and
Xiulong Liu and
Eli Shlizerman},
title        = {Audeo: Audio Generation for a Silent Performance Video},
booktitle    = {Advances in Neural Information Processing Systems 33: Annual Conference
on Neural Information Processing Systems 2020, NeurIPS 2020, December
6-12, 2020, virtual},
year         = {2020},
url          = {https://proceedings.neurips.cc/paper/2020/hash/227f6afd3b7f89b96c4bb91f95d50f6d-Abstract.html},
bibsource    = {dblp computer science bibliography, https://dblp.org}
}

@inproceedings{rhythmicnet,
  author       = {Kun Su and
Xiulong Liu and
Eli Shlizerman},
title        = {How Does it Sound?},
booktitle    = {Advances in Neural Information Processing Systems 34: Annual Conference
on Neural Information Processing Systems 2021, NeurIPS 2021, December
6-14, 2021, virtual},
pages        = {29258--29273},
year         = {2021},
url          = {https://proceedings.neurips.cc/paper/2021/hash/f4e369c0a468d3aeeda0593ba90b5e55-Abstract.html},
bibsource    = {dblp computer science bibliography, https://dblp.org}
}

@inproceedings{zhuo,
  title = {Video Background Music Generation: Dataset, Method and Evaluation},
  shorttitle = {Video Background Music Generation},
  booktitle = {{IEEE/CVF} International Conference on Computer Vision, ICCV 2023},
  author = {Zhuo, Le and Wang, Zhaokai and Wang, Baisen and Liao, Yue and Bao, Chenxi and Peng, Stanley and Han, Songhao and Zhang, Aixi and Fang, Fei and Liu, Si},
  year = {2023},
  pages = {15591--15601},
  publisher = {{IEEE}},
  doi = {10.1109/ICCV51070.2023.01433}
}

@article{kang,
  title = {Video2Music: Suitable Music Generation from Videos Using an Affective Multimodal Transformer Model},
  shorttitle = {Video2Music},
  author = {Kang, Jaeyong and Poria, Soujanya and Herremans, Dorien},
  year = {2024},
  month = sep,
  journal = {Expert Systems with Applications},
  volume = {249},
  pages = {123640},
  issn = {0957-4174},
  doi = {10.1016/j.eswa.2024.123640},
  urldate = {2024-04-10}
}

@inproceedings{music_dataset,
  author       = {Hang Zhao and
Chuang Gan and
Andrew Rouditchenko and
Carl Vondrick and
Josh H. McDermott and
Antonio Torralba},
title        = {The Sound of Pixels},
booktitle    = {Computer Vision - {ECCV} 2018 - 15th European Conference, Munich,
Germany, September 8-14, 2018, Proceedings, Part {I}},
series       = {Lecture Notes in Computer Science},
volume       = {11205},
pages        = {587--604},
publisher    = {Springer},
year         = {2018},
url          = {https://doi.org/10.1007/978-3-030-01246-5\_35},
doi          = {10.1007/978-3-030-01246-5\_35},
bibsource    = {dblp computer science bibliography, https://dblp.org}
}

@article{video_midi_dataset,
  author       = {Bochen Li and
Xinzhao Liu and
Karthik Dinesh and
Zhiyao Duan and
Gaurav Sharma},
title        = {Creating a Multitrack Classical Music Performance Dataset for Multimodal
Music Analysis: Challenges, Insights, and Applications},
journal      = {{IEEE} Transactions on Multimedia},
volume       = {21},
number       = {2},
pages        = {522--535},
year         = {2019},
url          = {https://doi.org/10.1109/TMM.2018.2856090},
doi          = {10.1109/TMM.2018.2856090},
bibsource    = {dblp computer science bibliography, https://dblp.org}
}

@inproceedings{gcn,
  title = {Semi-Supervised Classification with Graph Convolutional Networks},
  booktitle = {5th International Conference on Learning Representations, ICLR 2017},
  author = {Kipf, Thomas N. and Welling, Max},
  year = {2017},
  url = {https://openreview.net/forum?id=SJU4ayYgl},
}

@inproceedings{resnet,
  title = {Deep Residual Learning for Image Recognition},
  booktitle = {IEEE/CVF Conference on Computer Vision and Pattern Recognition, CVPR 2016},
  author = {He, Kaiming and Zhang, Xiangyu and Ren, Shaoqing and Sun, Jian},
  year = {2016},
  pages = {770--778},
  doi = {10.1109/CVPR.2016.90}
}

@inproceedings{compound_words,
  title = {Compound Word Transformer: Learning to Compose Full-Song Music over Dynamic Directed Hypergraphs},
  shorttitle = {Compound Word Transformer},
  booktitle = {Proceedings of the AAAI Conference on Artificial Intelligence},
  author = {Hsiao, Wen-Yi and Liu, Jen-Yu and Yeh, Yin-Cheng and Yang, Yi-Hsuan},
  year = {2021},
  pages = {178--186},
  doi = {10.1609/AAAI.V35I1.16091},
  urldate = {2025-01-24}
}

@inproceedings{onset_and_frames,
  title = {Onsets and Frames: Dual-Objective Piano Transcription},
  shorttitle = {Onsets and Frames},
  booktitle = {Proceedings of the 19th International Society for Music Information Retrieval Conference},
  author = {Hawthorne, Curtis and Elsen, Erich and Song, Jialin and Roberts, Adam and Simon, Ian and Raffel, Colin and Engel, Jesse H. and Oore, Sageev and Eck, Douglas},
  year = {2018},
  pages = {50--57},
  url = {http://ismir2018.ircam.fr/doc/pdfs/19\_Paper.pdf},
}

@inproceedings{ndb,
  author       = {Jesse H. Engel and
Kumar Krishna Agrawal and
Shuo Chen and
Ishaan Gulrajani and
Chris Donahue and
Adam Roberts},
title        = {GANSynth: Adversarial Neural Audio Synthesis},
booktitle    = {7th International Conference on Learning Representations, {ICLR} 2019,
New Orleans, LA, USA, May 6-9, 2019},
publisher    = {OpenReview.net},
year         = {2019},
url          = {https://openreview.net/forum?id=H1xQVn09FX},
bibsource    = {dblp computer science bibliography, https://dblp.org}
}

@inproceedings{metrics_music,
  author       = {Shih{-}Lun Wu and
                  Yi{-}Hsuan Yang},
  title        = {The Jazz Transformer on the Front Line: Exploring the Shortcomings
                  of AI-composed Music through Quantitative Measures},
  booktitle    = {Proceedings of the 21th International Society for Music Information
                  Retrieval Conference, {ISMIR} 2020},
  pages        = {142--149},
  year         = {2020},
  url          = {http://archives.ismir.net/ismir2020/paper/000339.pdf},
  bibsource    = {dblp computer science bibliography, https://dblp.org}
}

@inproceedings{contrastive,
  author       = {Jean{-}Baptiste Alayrac and
Adri{\`{a}} Recasens and
Rosalia Schneider and
Relja Arandjelovic and
Jason Ramapuram and
Jeffrey De Fauw and
Lucas Smaira and
Sander Dieleman and
Andrew Zisserman},
title        = {Self-Supervised MultiModal Versatile Networks},
booktitle    = {Advances in Neural Information Processing Systems},
year         = {2020},
url          = {https://proceedings.neurips.cc/paper/2020/hash/0060ef47b12160b9198302ebdb144dcf-Abstract.html},
bibsource    = {dblp computer science bibliography, https://dblp.org}
}

@inproceedings{musicgen1,
	author       = {Max W. Y. Lam and
	Qiao Tian and
	Tang Li and
	Zongyu Yin and
	Siyuan Feng and
	Ming Tu and
	Yuliang Ji and
	Rui Xia and
	Mingbo Ma and
	Xuchen Song and
	Jitong Chen and
	Yuping Wang and
	Yuxuan Wang},
	title        = {Efficient Neural Music Generation},
	booktitle    = {Advances in Neural Information Processing Systems 36: Annual Conference
	on Neural Information Processing Systems 2023, NeurIPS 2023, New Orleans,
	LA, USA, December 10 - 16, 2023},
	year         = {2023},
	url          = {http://papers.nips.cc/paper\_files/paper/2023/hash/38b23e2328096520e9c889ae03e372c9-Abstract-Conference.html},
	bibsource    = {dblp computer science bibliography, https://dblp.org}
}

@inproceedings{musicgen2,
	author       = {Jade Copet and
	Felix Kreuk and
	Itai Gat and
	Tal Remez and
	David Kant and
	Gabriel Synnaeve and
	Yossi Adi and
	Alexandre D{\'{e}}fossez},
	title        = {Simple and Controllable Music Generation},
	booktitle    = {Advances in Neural Information Processing Systems 36: Annual Conference
	on Neural Information Processing Systems 2023, NeurIPS 2023, New Orleans,
	LA, USA, December 10 - 16, 2023},
	year         = {2023},
	url          = {http://papers.nips.cc/paper\_files/paper/2023/hash/94b472a1842cd7c56dcb125fb2765fbd-Abstract-Conference.html},
	bibsource    = {dblp computer science bibliography, https://dblp.org}
}

@article{musicgen3,
	title = {AudioLDM 2: Learning Holistic Audio Generation With Self-Supervised Pretraining},
	shorttitle = {AudioLDM 2},
	author = {Liu, Haohe and Yuan, Yi and Liu, Xubo and Mei, Xinhao and Kong, Qiuqiang and Tian, Qiao and Wang, Yuping and Wang, Wenwu and Wang, Yuxuan and Plumbley, Mark D.},
	year = {2024},
	journal = {{IEEE/ACM} Transactions on Audio, Speech, and Language Processing},
	volume = {32},
	pages = {2871--2883},
	doi = {10.1109/TASLP.2024.3399607}
}

@inproceedings{musicgen4,
	title = {Fast Timing-Conditioned Latent Audio Diffusion},
	booktitle = {Forty-First International Conference on Machine Learning, ICML 2024, Vienna, Austria, July 21-27, 2024},
	author = {Evans, Zach and Carr, C. J. and Taylor, Josiah and Hawley, Scott H. and Pons, Jordi},
	year = {2024},
	publisher = {OpenReview.net},
	url = {https://openreview.net/forum?id=jOlO8t1xdx},
}

@inproceedings{swin,
  title = {Video Swin Transformer},
  booktitle = {IEEE/CVF Conference on Computer Vision and Pattern Recognition, CVPR 2022},
  author = {Liu, Ze and Ning, Jia and Cao, Yue and Wei, Yixuan and Zhang, Zheng and Lin, Stephen and Hu, Han},
  year = {2022},
  pages = {3192--3201},
  publisher = {IEEE},
  doi = {10.1109/CVPR52688.2022.00320}
}

@inproceedings{vivit,
  author       = {Anurag Arnab and
Mostafa Dehghani and
Georg Heigold and
Chen Sun and
Mario Lucic and
Cordelia Schmid},
title        = {ViViT: {A} Video Vision Transformer},
booktitle    = {2021 {IEEE/CVF} International Conference on Computer Vision, {ICCV}
2021, Montreal, QC, Canada, October 10-17, 2021},
pages        = {6816--6826},
publisher    = {{IEEE}},
year         = {2021},
url          = {https://doi.org/10.1109/ICCV48922.2021.00676},
doi          = {10.1109/ICCV48922.2021.00676},
bibsource    = {dblp computer science bibliography, https://dblp.org}
}

@inproceedings{denoising,
	author       = {Yunpeng Li and
	Marco Tagliasacchi and
	Beat Gfeller and
	Dominik Roblek},
	title        = {Learning to Denoise Historical Music},
	booktitle    = {Proceedings of the 21th International Society for Music Information
	Retrieval Conference, {ISMIR} 2020, Montreal, Canada, October 11-16,
	2020},
	pages        = {504--511},
	year         = {2020},
	url          = {http://archives.ismir.net/ismir2020/paper/000057.pdf},
	bibsource    = {dblp computer science bibliography, https://dblp.org}
}

@inproceedings{vggish,
	author       = {Shawn Hershey and
	Sourish Chaudhuri and
	Daniel P. W. Ellis and
	Jort F. Gemmeke and
	Aren Jansen and
	R. Channing Moore and
	Manoj Plakal and
	Devin Platt and
	Rif A. Saurous and
	Bryan Seybold and
	Malcolm Slaney and
	Ron J. Weiss and
	Kevin W. Wilson},
	title        = {{CNN} architectures for large-scale audio classification},
	booktitle    = {2017 {IEEE} International Conference on Acoustics, Speech and Signal
	Processing, {ICASSP} 2017, New Orleans, LA, USA, March 5-9, 2017},
	pages        = {131--135},
	publisher    = {{IEEE}},
	year         = {2017},
	url          = {https://doi.org/10.1109/ICASSP.2017.7952132},
	doi          = {10.1109/ICASSP.2017.7952132},
	bibsource    = {dblp computer science bibliography, https://dblp.org}
}

@inproceedings{visqol,
	author       = {Andrew Hines and
	Jan Skoglund and
	Anil C. Kokaram and
	Naomi Harte},
	title        = {ViSQOL: The Virtual Speech Quality Objective Listener},
	booktitle    = {{IWAENC} 2012 - International Workshop on Acoustic Signal Enhancement,
	Proceedings, {RWTH} Aachen University, Germany, September 4th - 6th,
	2012},
	publisher    = {VDE-Verlag},
	year         = {2012},
	url          = {http://www.vde-verlag.de/proceedings-de/453451035.html},
	bibsource    = {dblp computer science bibliography, https://dblp.org}
}

@inproceedings{fad,
	author       = {Kevin Kilgour and
	Mauricio Zuluaga and
	Dominik Roblek and
	Matthew Sharifi},
	editor       = {Gernot Kubin and
	Zdravko Kacic},
	title        = {Fr{\'{e}}chet Audio Distance: {A} Reference-Free Metric for Evaluating
	Music Enhancement Algorithms},
	booktitle    = {20th Annual Conference of the International Speech Communication Association,
	Interspeech 2019, Graz, Austria, September 15-19, 2019},
	pages        = {2350--2354},
	publisher    = {{ISCA}},
	year         = {2019},
	url          = {https://doi.org/10.21437/Interspeech.2019-2219},
	doi          = {10.21437/INTERSPEECH.2019-2219},
	bibsource    = {dblp computer science bibliography, https://dblp.org}
}

@inproceedings{perceptual_metric,
	author       = {Pranay Manocha and
	Adam Finkelstein and
	Richard Zhang and
	Nicholas J. Bryan and
	Gautham J. Mysore and
	Zeyu Jin},
	title        = {A Differentiable Perceptual Audio Metric Learned from Just Noticeable
	Differences},
	booktitle    = {21st Annual Conference of the International Speech Communication Association,
	Interspeech 2020, Virtual Event, Shanghai, China, October 25-29, 2020},
	pages        = {2852--2856},
	publisher    = {{ISCA}},
	year         = {2020},
	url          = {https://doi.org/10.21437/Interspeech.2020-1191},
	doi          = {10.21437/INTERSPEECH.2020-1191},
	bibsource    = {dblp computer science bibliography, https://dblp.org}
}

@inproceedings{pesq,
	author       = {Antony W. Rix and
	John G. Beerends and
	Michael P. Hollier and
	Andries P. Hekstra},
	title        = {Perceptual evaluation of speech quality (PESQ)-a new method for speech
	quality assessment of telephone networks and codecs},
	booktitle    = {{IEEE} International Conference on Acoustics, Speech, and Signal Processing,
	{ICASSP} 2001, 7-11 May, 2001, Salt Palace Convention Center, Salt
	Lake City, Utah, USA, Proceedings},
	pages        = {749--752},
	publisher    = {{IEEE}},
	year         = {2001},
	url          = {https://doi.org/10.1109/ICASSP.2001.941023},
	doi          = {10.1109/ICASSP.2001.941023},
	bibsource    = {dblp computer science bibliography, https://dblp.org}
}

@article{frame_prediction_sulun,
	author       = {Serkan Sulun and
	A. Murat Tekalp},
	title        = {Can learned frame prediction compete with block motion compensation
	for video coding?},
	journal      = {Signal Image Video Process.},
	volume       = {15},
	number       = {2},
	pages        = {401--410},
	year         = {2021},
	url          = {https://doi.org/10.1007/s11760-020-01751-y},
	doi          = {10.1007/S11760-020-01751-Y},
	bibsource    = {dblp computer science bibliography, https://dblp.org}
}

@article{masters,
  title = {Deep Learned Frame Prediction for Video Compression},
  author = {Sulun, Serkan},
  year = {2018},
  journal = {arXiv preprint arXiv:1811.10946},
  eprint = {1811.10946},
  primaryclass = {cs, eess},
  url = {http://arxiv.org/abs/1811.10946},
  archiveprefix = {arXiv}
}

@inproceedings{vemoclap,
	title = {{VEMOCLAP}: A Video Emotion Classification Web Application},
	shorttitle = {VEMOCLAP},
	booktitle = {2024 International Symposium on Multimedia (ISM)},
	author = {Sulun, Serkan and Viana, Paula and Davies, Matthew E. P.},
	year = {2024},
	month = dec,
	pages = {137--140},
	publisher = {IEEE Computer Society},
	doi = {10.1109/ISM63611.2024.00029},
	urldate = {2025-05-17},
	isbn = {9798331511111},
	langid = {english}
}

@mastersthesis{cbet,
	title={Emotion mining from text},
	author={Shahraki, Ameneh Gholipour},
	year={2015},
	school={University of Alberta},
	url={https://era.library.ualberta.ca/items/27ae961f-d9a6-4a5a-9b6f-f180478ea573}
}

@article{chords,
title = {The representation of harmonic structure in music: Hierarchies of stability as a function of context},
journal = {Cognition},
volume = {13},
number = {1},
pages = {63-102},
year = {1983},
issn = {0010-0277},
doi = {10.1016/0010-0277(83)90003-3},
author = {Jamshed Bharucha and Carol L. Krumhansl},
}

@inproceedings{position1,
  title = {Transformer Language Models without Positional Encodings Still Learn Positional Information},
  booktitle = {Findings of the Association for Computational Linguistics: EMNLP 2022},
  author = {Haviv, Adi and Ram, Ori and Press, Ofir and Izsak, Peter and Levy, Omer},
  year = {2022},
  pages = {1382--1390},
  publisher = {Association for Computational Linguistics},
  doi = {10.18653/V1/2022.FINDINGS-EMNLP.99},
  urldate = {2025-01-31}
}

@article{position2,
  title = {The Impact of Positional Encoding on Length Generalization in Transformers},
  author = {Kazemnejad, Amirhossein and Padhi, Inkit and Natesan Ramamurthy, Karthikeyan and Das, Payel and Reddy, Siva},
  year = {2024},
  journal = {Advances in Neural Information Processing Systems},
  volume = {36},
  url = {https://proceedings.neurips.cc/paper_files/paper/2023/hash/4e85362c02172c0c6567ce593122d31c-Abstract-Conference.html},
}

@inproceedings{ads,
  author       = {Zaeem Hussain and
Mingda Zhang and
Xiaozhong Zhang and
Keren Ye and
Christopher Thomas and
Zuha Agha and
Nathan Ong and
Adriana Kovashka},
title        = {Automatic Understanding of Image and Video Advertisements},
booktitle    = {2017 {IEEE} Conference on Computer Vision and Pattern Recognition,
{CVPR} 2017, Honolulu, HI, USA, July 21-26, 2017},
pages        = {1100--1110},
publisher    = {{IEEE} Computer Society},
year         = {2017},
url          = {https://doi.org/10.1109/CVPR.2017.123},
doi          = {10.1109/CVPR.2017.123},
bibsource    = {dblp computer science bibliography, https://dblp.org}
}

@article{ads_music,
  title = {Musical Influences in Advertising: How Music Modifies First Impressions of Product Endorsers and Brands},
  shorttitle = {Musical Influences in Advertising},
  author = {Zander, Mark F.},
  year = {2006},
  month = oct,
  journal = {Psychology of Music},
  volume = {34},
  number = {4},
  pages = {465--480},
  issn = {0305-7356, 1741-3087},
  doi = {10.1177/0305735606067158},
  urldate = {2025-02-01},
  copyright = {https://journals.sagepub.com/page/policies/text-and-data-mining-license},
  langid = {english}
}

@incollection{music_emotion_classification,
  author       = {Yi{-}Hsuan Yang and
                  Yu{-}Ching Lin and
                  Heng Tze Cheng and
                  I{-}Bin Liao and
                  Yeh{-}Chin Ho and
                  Homer H. Chen},
  title        = {Toward Multi-modal Music Emotion Classification},
  booktitle    = {Advances in Multimedia Information Processing - {PCM} 2008},
  series       = {Lecture Notes in Computer Science},
  volume       = {5353},
  pages        = {70--79},
  publisher    = {Springer},
  year         = {2008},
  url          = {https://doi.org/10.1007/978-3-540-89796-5\_8},
  doi          = {10.1007/978-3-540-89796-5\_8},
  bibsource    = {dblp computer science bibliography, https://dblp.org}
}

@inproceedings{music_emotion_features,
	title = {REMUPP: An Interface for Evaluation of Relations between Musical Parameters and Perceived Properties},
	shorttitle = {REMUPP},
	booktitle = {Proceedings of the International Conference on Advances in Computer Entertainment Technology, ACE 2005, Valencia, Spain, June 15-15, 2005},
	author = {Wingstedt, Johnny and Berg, Jan and Liljedahl, Mats and Lindberg, Stefan},
	year = {2005},
	pages = {346--349},
	publisher = {ACM},
	doi = {10.1145/1178477.1178543}
}

@inbook{music_and_emotion2,
	author="Hunter, Patrick G.
	and Schellenberg, E. Glenn",
	title="Music and Emotion",
	bookTitle="Music Perception",
	year="2010",
	publisher="Springer New York",
	address="New York, NY",
	pages="129--164",
	isbn="978-1-4419-6114-3",
	doi="10.1007/978-1-4419-6114-3_5",
	url="https://doi.org/10.1007/978-1-4419-6114-3_5"
}

@inproceedings{music_emotion_features2,
	title = {Relations between Selected Musical Parameters and Expressed Emotions: Extending the Potential of Computer Entertainment},
	shorttitle = {Relations between Selected Musical Parameters and Expressed Emotions},
	booktitle = {Proceedings of the International Conference on Advances in Computer Entertainment Technology, ACE 2005, Valencia, Spain, June 15-15, 2005},
	author = {Berg, Jan and Wingstedt, Johnny},
	year = {2005},
	pages = {164--171},
	publisher = {ACM},
	doi = {10.1145/1178477.1178499}
}

@article{multimedia,
  title = {Current Trends in Consumption of Multimedia Content Using Online Streaming Platforms: A User-Centric Survey},
  shorttitle = {Current Trends in Consumption of Multimedia Content Using Online Streaming Platforms},
  author = {{Falkowski-Gilski}, Przemyslaw and Uhl, Tadeus},
  year = {2020},
  journal = {Computer Science Review},
  volume = {37},
  pages = {100268},
  issn = {1574-0137},
  doi = {10.1016/j.cosrev.2020.100268},
  urldate = {2025-02-10}
}

@book{soundtrack,
  title={Theories of the Soundtrack},
  author={Buhler, James},
  year={2018},
  publisher={Oxford University Press},
  url={https://books.google.com/books?id=n_NwDwAAQBAJ&lpg=PP1&ots=RKwzh_dMet&lr=&pg=PP1&redir_esc=y#v=onepage&q&f=false}
}

@misc{yolo,
	title = {{YOLOv4}: Optimal Speed and Accuracy of Object Detection},
	shorttitle = {YOLOv4},
	author = {Bochkovskiy, Alexey and Wang, Chien-Yao and Liao, Hong-Yuan Mark},
	year = {2020},
	month = apr,
	number = {arXiv:2004.10934},
	eprint = {2004.10934},
	primaryclass = {cs, eess},
	publisher = {arXiv},
	doi = {10.48550/arXiv.2004.10934},
	urldate = {2024-08-31},
	archiveprefix = {arXiv}
}

@inproceedings{vit,
	title = {An Image Is Worth 16x16 Words: Transformers for Image Recognition at Scale},
	shorttitle = {An Image Is Worth 16x16 Words},
	booktitle = {9th International Conference on Learning Representations, ICLR 2021, Virtual Event, Austria, May 3-7, 2021},
	author = {Dosovitskiy, Alexey and Beyer, Lucas and Kolesnikov, Alexander and Weissenborn, Dirk and Zhai, Xiaohua and Unterthiner, Thomas and Dehghani, Mostafa and Minderer, Matthias and Heigold, Georg and Gelly, Sylvain and Uszkoreit, Jakob and Houlsby, Neil},
	year = {2021},
	publisher = {OpenReview.net},
	url = {https://openreview.net/forum?id=YicbFdNTTy},
}

@inproceedings{relative_position,
	title = {Self-Attention with Relative Position Representations},
	booktitle = {Proceedings of the 2018 Conference of the North American Chapter of the Association for Computational Linguistics: Human Language Technologies, NAACL-HLT, New Orleans, Louisiana, USA, June 1-6, 2018, Volume 2 (Short Papers)},
	author = {Shaw, Peter and Uszkoreit, Jakob and Vaswani, Ashish},
	year = {2018},
	pages = {464--468},
	publisher = {Association for Computational Linguistics},
	doi = {10.18653/V1/N18-2074},
	urldate = {2025-02-24}
}

@article{clipcap,
  author       = {Ron Mokady and
Amir Hertz and
Amit H. Bermano},
title        = {ClipCap: {CLIP} Prefix for Image Captioning},
journal      = {CoRR},
volume       = {abs/2111.09734},
year         = {2021},
url          = {https://arxiv.org/abs/2111.09734},
eprinttype    = {arXiv},
eprint       = {2111.09734},
bibsource    = {dblp computer science bibliography, https://dblp.org}
}

@inproceedings{beats,
  author       = {Sanyuan Chen and
Yu Wu and
Chengyi Wang and
Shujie Liu and
Daniel Tompkins and
Zhuo Chen and
Wanxiang Che and
Xiangzhan Yu and
Furu Wei},
title        = {BEATs: Audio Pre-Training with Acoustic Tokenizers},
booktitle    = {International Conference on Machine Learning, {ICML} 2023, 23-29 July
2023, Honolulu, Hawaii, {USA}},
series       = {Proceedings of Machine Learning Research},
volume       = {202},
pages        = {5178--5193},
publisher    = {{PMLR}},
year         = {2023},
url          = {https://proceedings.mlr.press/v202/chen23ag.html},
bibsource    = {dblp computer science bibliography, https://dblp.org}
}

@inproceedings{dynamic,
  author       = {Amjad Almahairi and
Nicolas Ballas and
Tim Cooijmans and
Yin Zheng and
Hugo Larochelle and
Aaron C. Courville},
title        = {Dynamic Capacity Networks},
booktitle    = {Proceedings of the 33nd International Conference on Machine Learning,
{ICML} 2016, New York City, NY, USA, June 19-24, 2016},
series       = {{JMLR} Workshop and Conference Proceedings},
volume       = {48},
pages        = {2549--2558},
publisher    = {JMLR.org},
year         = {2016},
url          = {http://proceedings.mlr.press/v48/almahairi16.html},
bibsource    = {dblp computer science bibliography, https://dblp.org}
}

@inproceedings{xlmroberta,
  author       = {Alexis Conneau and
Kartikay Khandelwal and
Naman Goyal and
Vishrav Chaudhary and
Guillaume Wenzek and
Francisco Guzm{\'{a}}n and
Edouard Grave and
Myle Ott and
Luke Zettlemoyer and
Veselin Stoyanov},
title        = {Unsupervised Cross-lingual Representation Learning at Scale},
booktitle    = {Proceedings of the 58th Annual Meeting of the Association for Computational
Linguistics, {ACL} 2020, Online, July 5-10, 2020},
pages        = {8440--8451},
publisher    = {Association for Computational Linguistics},
year         = {2020},
url          = {https://doi.org/10.18653/v1/2020.acl-main.747},
doi          = {10.18653/V1/2020.ACL-MAIN.747},
bibsource    = {dblp computer science bibliography, https://dblp.org}
}

@article{t5,
author = {Raffel, Colin and Shazeer, Noam and Roberts, Adam and Lee, Katherine and Narang, Sharan and Matena, Michael and Zhou, Yanqi and Li, Wei and Liu, Peter J.},
title = {Exploring the limits of transfer learning with a unified text-to-text transformer},
year = {2020},
issue_date = {January 2020},
publisher = {JMLR.org},
volume = {21},
number = {1},
issn = {1532-4435},
journal = {Journal of Machine Learning Research},
month = jan,
articleno = {140},
numpages = {67},
}

@inproceedings{cfn,
	title = {Emotion in Context: Deep Semantic Feature Fusion for Video Emotion Recognition},
	shorttitle = {Emotion in Context},
	booktitle = {ACM-MM 2016},
	author = {Chen, Chen and Wu, Zuxuan and Jiang, Yu-Gang},
	year = {2016},
	pages = {127--131},
	publisher = {ACM},
	doi = {10.1145/2964284.2967196}
}

@inproceedings{mart,
  author       = {Zhicheng Zhang and
Pancheng Zhao and
Eunil Park and
Jufeng Yang},
title        = {{MART:} Masked Affective RepresenTation Learning via Masked Temporal
Distribution Distillation},
booktitle    = {{IEEE/CVF} Conference on Computer Vision and Pattern Recognition,
{CVPR} 2024, Seattle, WA, USA, June 16-22, 2024},
pages        = {12830--12840},
publisher    = {{IEEE}},
year         = {2024},
url          = {https://doi.org/10.1109/CVPR52733.2024.01219},
doi          = {10.1109/CVPR52733.2024.01219},
bibsource    = {dblp computer science bibliography, https://dblp.org}
}

@inproceedings{vaanet,
	title = {An End-to-End Visual-Audio Attention Network for Emotion Recognition in User-Generated Videos},
	booktitle = {AAAI 2020},
	author = {Zhao, Sicheng and Ma, Yunsheng and Gu, Yang and Yang, Jufeng and Xing, Tengfei and Xu, Pengfei and Hu, Runbo and Chai, Hua and Keutzer, Kurt},
	year = {2020},
	pages = {303--311},
	publisher = {AAAI Press},
	doi = {10.1609/AAAI.V34I01.5364},
	urldate = {2024-08-31}
}

@inproceedings{cten,
  author       = {Zhicheng Zhang and
Lijuan Wang and
Jufeng Yang},
title        = {Weakly Supervised Video Emotion Detection and Prediction via Cross-Modal
Temporal Erasing Network},
booktitle    = {{IEEE/CVF} Conference on Computer Vision and Pattern Recognition,
{CVPR} 2023, Vancouver, BC, Canada, June 17-24, 2023},
pages        = {18888--18897},
publisher    = {{IEEE}},
year         = {2023},
url          = {https://doi.org/10.1109/CVPR52729.2023.01811},
doi          = {10.1109/CVPR52729.2023.01811},
bibsource    = {dblp computer science bibliography, https://dblp.org}
}

@article{lrcanet,
  author       = {Yun Yi and
Jin Zhou and
Hanli Wang and
Pengjie Tang and
Min Wang},
title        = {Emotion recognition in user-generated videos with long-range correlation-aware
network},
journal      = {{IET} Image Processing},
volume       = {18},
number       = {12},
pages        = {3288--3301},
year         = {2024},
url          = {https://doi.org/10.1049/ipr2.13174},
doi          = {10.1049/IPR2.13174},
bibsource    = {dblp computer science bibliography, https://dblp.org}
}

@article{eswa,
	title = {Movie Trailer Genre Classification Using Multimodal Pretrained Features},
	author = {Sulun, Serkan and Viana, Paula and Davies, Matthew E. P.},
	year = {2024},
	journal = {Expert Systems with Applications},
	pages = {125209},
	issn = {0957-4174},
	doi = {10.1016/j.eswa.2024.125209}
}

@inproceedings{tam,
  author       = {Jicai Pan and
Shangfei Wang and
Lin Fang},
title        = {Representation Learning through Multimodal Attention and Time-Sync
Comments for Affective Video Content Analysis},
booktitle    = {{MM} '22: The 30th {ACM} International Conference on Multimedia, Lisboa,
Portugal, October 10 - 14, 2022},
pages        = {42--50},
publisher    = {{ACM}},
year         = {2022},
url          = {https://doi.org/10.1145/3503161.3548018},
doi          = {10.1145/3503161.3548018},
bibsource    = {dblp computer science bibliography, https://dblp.org}
}

@article{faeil,
  author={Zhang, Haimin and Xu, Min},
journal={IEEE Transactions on Multimedia}, 
title={Recognition of Emotions in User-Generated Videos through Frame-Level Adaptation and Emotion Intensity Learning}, 
year={2023},
volume={25},
number={},
pages={881-891},
doi={10.1109/TMM.2021.3134167}}

@article{keyframe,
	title = {User-generated video emotion recognition based on key frames},
	author = {Wei, Jie and Yang, Xinyu and Dong, Yizhuo},
	year = {2021},
	month = apr,
	journal = {Multimedia Tools and Applications},
  volume       = {80},
number       = {9},
pages        = {14343--14361},
year         = {2021},
url          = {https://doi.org/10.1007/s11042-020-10203-1},
doi          = {10.1007/S11042-020-10203-1},
}

@misc{vgg,
  author       = {Karen Simonyan and
Andrew Zisserman},
title        = {Very Deep Convolutional Networks for Large-Scale Image Recognition},
booktitle    = {3rd International Conference on Learning Representations, {ICLR} 2015,
San Diego, CA, USA, May 7-9, 2015, Conference Track Proceedings},
year         = {2015},
url          = {http://arxiv.org/abs/1409.1556},
bibsource    = {dblp computer science bibliography, https://dblp.org}
}

@inproceedings{clip,
	author       = {Alec Radford and
	Jong Wook Kim and
	Chris Hallacy and
	Aditya Ramesh and
	Gabriel Goh and
	Sandhini Agarwal and
	Girish Sastry and
	Amanda Askell and
	Pamela Mishkin and
	Jack Clark and
	Gretchen Krueger and
	Ilya Sutskever},
	title        = {Learning Transferable Visual Models From Natural Language Supervision},
	booktitle    = {Proceedings of the 38th International Conference on Machine Learning,
	{ICML} 2021, 18-24 July 2021, Virtual Event},
	series       = {Proceedings of Machine Learning Research},
	volume       = {139},
	pages        = {8748--8763},
	publisher    = {{PMLR}},
	year         = {2021},
	url          = {http://proceedings.mlr.press/v139/radford21a.html},
	bibsource    = {dblp computer science bibliography, https://dblp.org}
}

@article{audiotag,
	author       = {Qiuqiang Kong and
	Yin Cao and
	Turab Iqbal and
	Yuxuan Wang and
	Wenwu Wang and
	Mark D. Plumbley},
	title        = {{PANNs}: Large-Scale Pretrained Audio Neural Networks for Audio Pattern
	Recognition},
	journal      = {{IEEE} {ACM} Transactions on Audio, Speech, and Language Processing},
	volume       = {28},
	pages        = {2880--2894},
	year         = {2020},
	
	doi          = {10.1109/TASLP.2020.3030497},
	bibsource    = {dblp computer science bibliography, https://dblp.org}
}

@article{distilbert,
	title={{DistilBERT}, a distilled version of {BERT}: smaller, faster, cheaper and lighter},
	author={Sanh, Victor and Debut, Lysandre and Chaumond, Julien and Wolf, Thomas},
	journal={arXiv preprint arXiv:1910.01108},
	year={2019},
	doi={10.48550/arXiv.1910.01108}
}

@inproceedings{transformer,
  author       = {Ashish Vaswani and
Noam Shazeer and
Niki Parmar and
Jakob Uszkoreit and
Llion Jones and
Aidan N. Gomez and
Lukasz Kaiser and
Illia Polosukhin},
title        = {Attention is All you Need},
booktitle    = {Advances in Neural Information Processing Systems 30: Annual Conference
on Neural Information Processing Systems 2017, December 4-9, 2017,
Long Beach, CA, {USA}},
pages        = {5998--6008},
year         = {2017},
url          = {https://proceedings.neurips.cc/paper/2017/hash/3f5ee243547dee91fbd053c1c4a845aa-Abstract.html},
bibsource    = {dblp computer science bibliography, https://dblp.org}
}

@article{moviescope,
	title = {Moviescope: Large-Scale Analysis of Movies Using Multiple Modalities},
	author = {{Cascante-Bonilla}, Paola and Sitaraman, Kalpathy and Luo, Mengjia and Ordonez, Vicente},
	year = {2019},
	journal = {arXiv preprint arXiv:1908.03180},
	eprint = {1908.03180},
	doi = {10.48550/arXiv.1908.03180},
	archiveprefix = {arXiv}
}

@inproceedings{mmtf,
	author       = {Yashar Deldjoo and
	Mihai Gabriel Constantin and
	Bogdan Ionescu and
	Markus Schedl and
	Paolo Cremonesi},
	title        = {{MMTF-14K:} a multifaceted movie trailer feature dataset for recommendation
	and retrieval},
	booktitle    = {Proceedings of the 9th {ACM} Multimedia Systems Conference, MMSys
	2018, Amsterdam, The Netherlands, June 12-15, 2018},
	pages        = {450--455},
	publisher    = {{ACM}},
	year         = {2018},
	doi          = {10.1145/3204949.3208141},
	bibsource    = {dblp computer science bibliography, https://dblp.org}
}

@inproceedings{lmtd,
	author       = {Gabriel S. Simoes and
	Jonatas Wehrmann and
	Rodrigo C. Barros and
	Duncan D. Ruiz},
	title        = {Movie genre classification with Convolutional Neural Networks},
	booktitle    = {2016 International Joint Conference on Neural Networks, {IJCNN} 2016,
	Vancouver, BC, Canada, July 24-29, 2016},
	pages        = {259--266},
	publisher    = {{IEEE}},
	year         = {2016},
	doi          = {10.1109/IJCNN.2016.7727207},
	bibsource    = {dblp computer science bibliography, https://dblp.org}
}

@inproceedings{zhou,
	author       = {Howard Zhou and
	Tucker Hermans and
	Asmita V. Karandikar and
	James M. Rehg},
	title        = {Movie genre classification via scene categorization},
	booktitle    = {Proceedings of the 18th International Conference on Multimedia 2010,
	Firenze, Italy, October 25-29, 2010},
	pages        = {747--750},
	publisher    = {{ACM}},
	year         = {2010},
	doi          = {10.1145/1873951.1874068},
	bibsource    = {dblp computer science bibliography, https://dblp.org}
}

@article{wehrmann,
	author       = {Jonatas Wehrmann and
	Rodrigo C. Barros},
	title        = {Movie genre classification: {A} multi-label approach based on convolutions
	through time},
	journal      = {Applied Soft Computing},
	volume       = {61},
	pages        = {973--982},
	issn = {1568-4946},
	year         = {2017},
	doi          = {10.1016/j.asoc.2017.08.029},
	bibsource    = {dblp computer science bibliography, https://dblp.org}
}

@article{places,
	author       = {Bolei Zhou and
	Agata Lapedriza and
	Aditya Khosla and
	Aude Oliva and
	Antonio Torralba},
	title        = {Places: {A} 10 Million Image Database for Scene Recognition},
	journal      = {{IEEE} Transactions on Pattern Analysis and Machine Intelligence},
	volume       = {40},
	number       = {6},
	pages        = {1452--1464},
	year         = {2018},
	doi          = {10.1109/TPAMI.2017.2723009},
	bibsource    = {dblp computer science bibliography, https://dblp.org}
}

@article{yadav,
	author       = {Ashima Yadav and
	Dinesh Kumar Vishwakarma},
	title        = {A unified framework of deep networks for genre classification using
	movie trailer},
	journal      = {Applied Soft Computing},
	volume       = {96},
	pages        = {106624},
	year         = {2020},
	doi          = {10.1016/j.asoc.2020.106624},
	issn = {1568-4946},
	bibsource    = {dblp computer science bibliography, https://dblp.org}
}

@inproceedings{movieclip,
	author       = {Digbalay Bose and
	Rajat Hebbar and
	Krishna Somandepalli and
	Haoyang Zhang and
	Yin Cui and
	Kree Cole{-}McLaughlin and
	Huisheng Wang and
	Shrikanth Narayanan},
	title        = {{MovieCLIP}: Visual Scene Recognition in Movies},
	booktitle    = {{IEEE/CVF} Winter Conference on Applications of Computer Vision, {WACV}
	2023, Waikoloa, HI, USA, January 2-7, 2023},
	pages        = {2082--2091},
	publisher    = {{IEEE}},
	year         = {2023},
	doi          = {10.1109/WACV56688.2023.00212},
	bibsource    = {dblp computer science bibliography, https://dblp.org}
}

@inproceedings{tsn,
	title = {Temporal Segment Networks: Towards Good Practices for Deep Action Recognition},
	shorttitle = {Temporal Segment Networks},
	booktitle = {Computer Vision - ECCV 2016 - 14th European Conference, Amsterdam, The Netherlands, October 11-14, 2016, Proceedings, Part VIII},
	author = {Wang, Limin and Xiong, Yuanjun and Wang, Zhe and Qiao, Yu and Lin, Dahua and Tang, Xiaoou and Gool, Luc Van},
	year = {2016},
	series = {Lecture Notes in Computer Science},
	volume = {9912},
	pages = {20--36},
	publisher = {Springer},
	doi = {10.1007/978-3-319-46484-8_2},
	urldate = {2023-07-31}
}

@inproceedings{i3d,
	ids = {carreiraQuoVadisAction2017},
	title = {Quo Vadis, Action Recognition? A New Model and the Kinetics Dataset},
	shorttitle = {Quo Vadis, Action Recognition?},
	booktitle = {2017 IEEE Conference on Computer Vision and Pattern Recognition, CVPR 2017, Honolulu, HI, USA, July 21-26, 2017},
	author = {Carreira, Joao and Zisserman, Andrew},
	year = {2017},
	pages = {4724--4733},
	publisher = {IEEE Computer Society},
	doi = {10.1109/CVPR.2017.502}
}

@inproceedings{trn,
	title = {Temporal Relational Reasoning in Videos},
	booktitle = {Computer Vision - ECCV 2018 - 15th European Conference, Munich, Germany, September 8-14, 2018, Proceedings, Part I},
	author = {Zhou, Bolei and Andonian, Alex and Oliva, Aude and Torralba, Antonio},
	year = {2018},
	series = {Lecture Notes in Computer Science},
	volume = {11205},
	pages = {831--846},
	publisher = {Springer},
	doi = {10.1007/978-3-030-01246-5_49},
	urldate = {2022-01-10}
}

@inproceedings{adam,
	author       = {Diederik P. Kingma and
	Jimmy Ba},
	title        = {Adam: {A} Method for Stochastic Optimization},
	booktitle    = {3rd International Conference on Learning Representations, {ICLR} 2015,
	San Diego, CA, USA, May 7-9, 2015, Conference Track Proceedings},
	year         = {2015},
	url          = {http://arxiv.org/abs/1412.6980},
	bibsource    = {dblp computer science bibliography, https://dblp.org}
}

@inproceedings{pytorch,
	title = {{PyTorch}: An Imperative Style, High-Performance Deep Learning Library},
	booktitle = {Advances in Neural Information Processing Systems 32: Annual Conference on Neural Information Processing Systems 2019, NeurIPS 2019, 8-14 December 2019, Vancouver, BC, Canada},
	author = {Paszke, Adam and Gross, Sam and Massa, Francisco and Lerer, Adam and Bradbury, James and Chanan, Gregory and Killeen, Trevor and Lin, Zeming and Gimelshein, Natalia and Antiga, Luca and Desmaison, Alban and Kopf, Andreas and Yang, Edward and DeVito, Zachary and Raison, Martin and Tejani, Alykhan and Chilamkurthy, Sasank and Steiner, Benoit and Fang, Lu and Bai, Junjie and Chintala, Soumith},
	year = {2019},
	pages = {8024--8035},
	url = {https://proceedings.neurips.cc/paper_files/paper/2019/file/bdbca288fee7f92f2bfa9f7012727740-Paper.pdf},
}

@inproceedings{ast,
	title = {{AST}: Audio Spectrogram Transformer},
	shorttitle = {AST},
	booktitle = {Interspeech 2021, 22nd Annual Conference of the International Speech Communication Association, Brno, Czechia, 30 August - 3 September 2021},
	author = {Gong, Yuan and Chung, Yu-An and Glass, James R.},
	year = {2021},
	pages = {571--575},
	publisher = {ISCA},
	doi = {10.21437/Interspeech.2021-698}
}

@inproceedings{lakhnes,
  author       = {Chris Donahue and
Huanru Henry Mao and
Yiting Ethan Li and
Garrison W. Cottrell and
Julian J. McAuley},
title        = {LakhNES: Improving Multi-instrumental Music Generation with Cross-domain
Pre-training},
booktitle    = {Proceedings of the 20th International Society for Music Information
Retrieval Conference, {ISMIR} 2019, Delft, The Netherlands, November
4-8, 2019},
pages        = {685--692},
year         = {2019},
url          = {http://archives.ismir.net/ismir2019/paper/000083.pdf},
bibsource    = {dblp computer science bibliography, https://dblp.org}
}

@article{musenet,
	title={MuseNet},
	author={Payne, Christine},
	journal={OpenAI Blog},
	volume={3},
	year={2019},
	url = {https://openai.com/blog/musenet},
}

@article{ctrl,
  author       = {Nitish Shirish Keskar and
Bryan McCann and
Lav R. Varshney and
Caiming Xiong and
Richard Socher},
title        = {{CTRL:} {A} Conditional Transformer Language Model for Controllable
Generation},
journal      = {CoRR},
volume       = {abs/1909.05858},
year         = {2019},
url          = {http://arxiv.org/abs/1909.05858},
eprinttype    = {arXiv},
eprint       = {1909.05858},
bibsource    = {dblp computer science bibliography, https://dblp.org}
}

@inproceedings{vgmidi,
	title = {Learning to Generate Music with Sentiment},
	booktitle = {Proceedings of the 20th International Society for Music Information Retrieval Conference, ISMIR 2019, Delft, the Netherlands, November 4-8, 2019},
	author = {Ferreira, Lucas and Whitehead, Jim},
	year = {2019},
	pages = {384--390},
	url={http://archives.ismir.net/ismir2019/paper/000045.pdf},
}

@inproceedings{conditionalLM,
  author       = {Tom{\'{a}}s Mikolov and
Geoffrey Zweig},
title        = {Context dependent recurrent neural network language model},
booktitle    = {2012 {IEEE} Spoken Language Technology Workshop (SLT), Miami, FL,
USA, December 2-5, 2012},
pages        = {234--239},
publisher    = {{IEEE}},
year         = {2012},
url          = {https://doi.org/10.1109/SLT.2012.6424228},
doi          = {10.1109/SLT.2012.6424228},
bibsource    = {dblp computer science bibliography, https://dblp.org}
}

@inproceedings{conditional_translator,
	title = {Controlling Politeness in Neural Machine Translation via Side Constraints},
	booktitle = {NAACL HLT 2016, the 2016 Conference of the North American Chapter of the Association for Computational Linguistics: Human Language Technologies, San Diego California, USA, June 12-17, 2016},
	author = {Sennrich, Rico and Haddow, Barry and Birch, Alexandra},
	year = {2016},
	pages = {35--40},
	doi = {10.18653/v1/n16-1005}
}

@article{image_caption_rnn,
  author       = {Oriol Vinyals and
Alexander Toshev and
Samy Bengio and
Dumitru Erhan},
title        = {Show and Tell: Lessons Learned from the 2015 {MSCOCO} Image Captioning
Challenge},
journal      = {{IEEE} Transactions on Pattern Analysis Machine Intelligence},
volume       = {39},
number       = {4},
pages        = {652--663},
year         = {2017},
url          = {https://doi.org/10.1109/TPAMI.2016.2587640},
doi          = {10.1109/TPAMI.2016.2587640},
bibsource    = {dblp computer science bibliography, https://dblp.org}
}

@article{image_caption_transformer,
AUTHOR = {Zhu, Xinxin and Li, Lixiang and Liu, Jing and Peng, Haipeng and Niu, Xinxin},
TITLE = {Captioning Transformer with Stacked Attention Modules},
JOURNAL = {Applied Sciences},
VOLUME = {8},
YEAR = {2018},
NUMBER = {5},
ARTICLE-NUMBER = {739},
URL = {https://www.mdpi.com/2076-3417/8/5/739},
ISSN = {2076-3417},
DOI = {10.3390/app8050739}
}

@inproceedings{image_caption_attention,
  author       = {Kelvin Xu and
Jimmy Ba and
Ryan Kiros and
Kyunghyun Cho and
Aaron C. Courville and
Ruslan Salakhutdinov and
Richard S. Zemel and
Yoshua Bengio},
title        = {Show, Attend and Tell: Neural Image Caption Generation with Visual
Attention},
booktitle    = {Proceedings of the 32nd International Conference on Machine Learning,
{ICML} 2015, Lille, France, 6-11 July 2015},
series       = {{JMLR} Workshop and Conference Proceedings},
volume       = {37},
pages        = {2048--2057},
publisher    = {JMLR.org},
year         = {2015},
url          = {http://proceedings.mlr.press/v37/xuc15.html},
bibsource    = {dblp computer science bibliography, https://dblp.org}
}

@inproceedings{midinet,
  author       = {Li{-}Chia Yang and
Szu{-}Yu Chou and
Yi{-}Hsuan Yang},
title        = {MidiNet: {A} Convolutional Generative Adversarial Network for Symbolic-Domain
Music Generation},
booktitle    = {Proceedings of the 18th International Society for Music Information
Retrieval Conference, {ISMIR} 2017, Suzhou, China, October 23-27,
2017},
pages        = {324--331},
year         = {2017},
url          = {https://ismir2017.smcnus.org/wp-content/uploads/2017/10/226\_Paper.pdf},
bibsource    = {dblp computer science bibliography, https://dblp.org}
}

@inproceedings{coconet,
  author       = {Cheng{-}Zhi Anna Huang and
Tim Cooijmans and
Adam Roberts and
Aaron C. Courville and
Douglas Eck},
title        = {Counterpoint by Convolution},
booktitle    = {Proceedings of the 18th International Society for Music Information
Retrieval Conference, {ISMIR} 2017, Suzhou, China, October 23-27,
2017},
pages        = {211--218},
year         = {2017},
url          = {https://ismir2017.smcnus.org/wp-content/uploads/2017/10/187\_Paper.pdf},
bibsource    = {dblp computer science bibliography, https://dblp.org}
}

@inproceedings{musicvae,
  author       = {Adam Roberts and
Jesse H. Engel and
Colin Raffel and
Curtis Hawthorne and
Douglas Eck},
title        = {A Hierarchical Latent Vector Model for Learning Long-Term Structure
in Music},
booktitle    = {Proceedings of the 35th International Conference on Machine Learning,
{ICML} 2018, Stockholmsm{\"{a}}ssan, Stockholm, Sweden, July
10-15, 2018},
series       = {Proceedings of Machine Learning Research},
volume       = {80},
pages        = {4361--4370},
publisher    = {{PMLR}},
year         = {2018},
url          = {http://proceedings.mlr.press/v80/roberts18a.html},
bibsource    = {dblp computer science bibliography, https://dblp.org}
}

@inproceedings{sketchnet,
  author       = {Ke Chen and
Cheng{-}i Wang and
Taylor Berg{-}Kirkpatrick and
Shlomo Dubnov},
title        = {Music SketchNet: Controllable Music Generation via Factorized Representations
of Pitch and Rhythm},
booktitle    = {Proceedings of the 21th International Society for Music Information
Retrieval Conference, {ISMIR} 2020, Montreal, Canada, October 11-16,
2020},
pages        = {77--84},
year         = {2020},
url          = {http://archives.ismir.net/ismir2020/paper/000146.pdf},
bibsource    = {dblp computer science bibliography, https://dblp.org}
}

@inproceedings{musical_style,
  author       = {Kristy Choi and
Curtis Hawthorne and
Ian Simon and
Monica Dinculescu and
Jesse H. Engel},
title        = {Encoding Musical Style with Transformer Autoencoders},
booktitle    = {Proceedings of the 37th International Conference on Machine Learning,
{ICML} 2020, 13-18 July 2020, Virtual Event},
series       = {Proceedings of Machine Learning Research},
volume       = {119},
pages        = {1899--1908},
publisher    = {{PMLR}},
year         = {2020},
url          = {http://proceedings.mlr.press/v119/choi20b.html},
bibsource    = {dblp computer science bibliography, https://dblp.org}
}

@inproceedings{music_machine,
	title = {Flexible Generation with the Multi-Track Music Machine},
	booktitle = {Proceedings of the 21st International Society for Music Information Retrieval Conference, ISMIR 2020},
	author = {Ens, Jeffrey and Pasquier, Philippe},
	year = {2020},
	month = oct,
	url={https://ddmal.music.mcgill.ca/ISMIR-Conf/static/lbd/ISMIR2020-LBD-423-abstract.pdf}
}

@inproceedings{tonal_tension,
	title = {A Variational Autoencoder for Music Generation Controlled by Tonal Tension},
	booktitle = {Proceedings of the 2020 Joint Conference on AI Music Creativity},
	author = {Guo, Rui and Simpson, Ivor and Magnusson, Thor and Kiefer, Chris and Herremans, Dorien},
	year = {2020},
	url = {https://arxiv.org/abs/2010.06230},
}

@inproceedings{pati,
  author       = {Kun{-}Youl Park and
Hyung Soon Kim},
title        = {Narrowband to wideband conversion of speech using {GMM} based transformation},
booktitle    = {{IEEE} International Conference on Acoustics, Speech, and Signal Processing.
{ICASSP} 2000, 5-9 June, 2000, Hilton Hotel and Convention Center,
Istanbul, Turkey},
pages        = {1843--1846},
publisher    = {{IEEE}},
year         = {2000},
url          = {https://doi.org/10.1109/ICASSP.2000.862114},
doi          = {10.1109/ICASSP.2000.862114},
bibsource    = {dblp computer science bibliography, https://dblp.org}
}

@inproceedings{yang,
  author       = {Ruihan Yang and
Dingsu Wang and
Ziyu Wang and
Tianyao Chen and
Junyan Jiang and
Gus Xia},
title        = {Deep Music Analogy Via Latent Representation Disentanglement},
booktitle    = {Proceedings of the 20th International Society for Music Information
Retrieval Conference, {ISMIR} 2019, Delft, The Netherlands, November
4-8, 2019},
pages        = {596--603},
year         = {2019},
url          = {http://archives.ismir.net/ismir2019/paper/000072.pdf},
bibsource    = {dblp computer science bibliography, https://dblp.org}
}

@inproceedings{fadernets,
  author       = {Hao Hao Tan and
Dorien Herremans},
title        = {Music FaderNets: Controllable Music Generation Based On High-Level
Features via Low-Level Feature Modelling},
booktitle    = {Proceedings of the 21th International Society for Music Information
Retrieval Conference, {ISMIR} 2020, Montreal, Canada, October 11-16,
2020},
pages        = {109--116},
year         = {2020},
url          = {http://archives.ismir.net/ismir2020/paper/000222.pdf},
bibsource    = {dblp computer science bibliography, https://dblp.org}
}

@inproceedings{maestro,
  author       = {Curtis Hawthorne and
Andriy Stasyuk and
Adam Roberts and
Ian Simon and
Cheng{-}Zhi Anna Huang and
Sander Dieleman and
Erich Elsen and
Jesse H. Engel and
Douglas Eck},
title        = {Enabling Factorized Piano Music Modeling and Generation with the {MAESTRO}
Dataset},
booktitle    = {7th International Conference on Learning Representations, {ICLR} 2019,
New Orleans, LA, USA, May 6-9, 2019},
publisher    = {OpenReview.net},
year         = {2019},
url          = {https://openreview.net/forum?id=r1lYRjC9F7},
bibsource    = {dblp computer science bibliography, https://dblp.org}
}

@inproceedings{gru,
  author       = {Kyunghyun Cho and
Bart van Merrienboer and
Dzmitry Bahdanau and
Yoshua Bengio},
title        = {On the Properties of Neural Machine Translation: Encoder-Decoder Approaches},
booktitle    = {Proceedings of SSST@EMNLP 2014, Eighth Workshop on Syntax, Semantics
and Structure in Statistical Translation, Doha, Qatar, 25 October
2014},
pages        = {103--111},
publisher    = {Association for Computational Linguistics},
year         = {2014},
url          = {https://aclanthology.org/W14-4012/},
doi          = {10.3115/V1/W14-4012},
bibsource    = {dblp computer science bibliography, https://dblp.org}
}

@inproceedings{attention,
  author       = {Dzmitry Bahdanau and
Kyunghyun Cho and
Yoshua Bengio},
title        = {Neural Machine Translation by Jointly Learning to Align and Translate},
booktitle    = {3rd International Conference on Learning Representations, {ICLR} 2015, San Diego, CA, USA, May 7-9, 2015, Conference Track Proceedings},
year         = {2015},
url          = {http://arxiv.org/abs/1409.0473},
bibsource    = {dblp computer science bibliography, https://dblp.org}
}

@inproceedings{encoder_decoder_rnn,
	title = {Learning Phrase Representations Using RNN Encoder-Decoder for Statistical Machine Translation},
	booktitle = {Proceedings of the 2014 Conference on Empirical Methods in Natural Language Processing, EMNLP 2014, October 25-29, 2014, Doha, Qatar, A Meeting of SIGDAT, a Special Interest Group of the ACL},
	author = {Cho, Kyunghyun and {van Merrienboer}, Bart and G{\"u}l{\c c}ehre, {\c C}aglar and Bahdanau, Dzmitry and Bougares, Fethi and Schwenk, Holger and Bengio, Yoshua},
	year = {2014},
	pages = {1724--1734},
	doi = {10.3115/v1/d14-1179}
}

@inproceedings{gpt3,
  author       = {Tom B. Brown and
Benjamin Mann and
Nick Ryder and
Melanie Subbiah and
Jared Kaplan and
Prafulla Dhariwal and
Arvind Neelakantan and
Pranav Shyam and
Girish Sastry and
Amanda Askell and
Sandhini Agarwal and
Ariel Herbert{-}Voss and
Gretchen Krueger and
Tom Henighan and
Rewon Child and
Aditya Ramesh and
Daniel M. Ziegler and
Jeffrey Wu and
Clemens Winter and
Christopher Hesse and
Mark Chen and
Eric Sigler and
Mateusz Litwin and
Scott Gray and
Benjamin Chess and
Jack Clark and
Christopher Berner and
Sam McCandlish and
Alec Radford and
Ilya Sutskever and
Dario Amodei},
title        = {Language Models are Few-Shot Learners},
booktitle    = {Advances in Neural Information Processing Systems 33: Annual Conference
on Neural Information Processing Systems 2020, NeurIPS 2020, December
6-12, 2020, virtual},
year         = {2020},
url          = {https://proceedings.neurips.cc/paper/2020/hash/1457c0d6bfcb4967418bfb8ac142f64a-Abstract.html},
bibsource    = {dblp computer science bibliography, https://dblp.org}
}

@inproceedings{paraphrase,
  author       = {Kalpesh Krishna and
John Wieting and
Mohit Iyyer},
title        = {Reformulating Unsupervised Style Transfer as Paraphrase Generation},
booktitle    = {Proceedings of the 2020 Conference on Empirical Methods in Natural
Language Processing, {EMNLP} 2020, Online, November 16-20, 2020},
pages        = {737--762},
publisher    = {Association for Computational Linguistics},
year         = {2020},
url          = {https://doi.org/10.18653/v1/2020.emnlp-main.55},
doi          = {10.18653/V1/2020.EMNLP-MAIN.55},
bibsource    = {dblp computer science bibliography, https://dblp.org}
}

@inproceedings{bias,
	title = {Towards Controllable Biases in Language Generation},
	booktitle = {Proceedings of the 2020 Conference on Empirical Methods in Natural Language Processing: Findings, EMNLP 2020, Online Event, 16-20 November 2020},
	author = {Sheng, Emily and Chang, Kai-Wei and Natarajan, Prem and Peng, Nanyun},
	year = {2020},
	series = {Findings of ACL},
	volume = {EMNLP 2020},
	pages = {3239--3254},
	doi = {10.18653/v1/2020.findings-emnlp.291}
}

@article{dialogue,
  author       = {Eric Michael Smith and
Diana Gonzalez{-}Rico and
Emily Dinan and
Y{-}Lan Boureau},
title        = {Controlling Style in Generated Dialogue},
journal      = {CoRR},
volume       = {abs/2009.10855},
year         = {2020},
url          = {https://arxiv.org/abs/2009.10855},
eprinttype    = {arXiv},
eprint       = {2009.10855},
bibsource    = {dblp computer science bibliography, https://dblp.org}
}

@inproceedings{retrieve,
	title = {Retrieve and Refine: Improved Sequence Generation Models for Dialogue},
	booktitle = {Proceedings of the 2nd International Workshop on Search-Oriented Conversational AI, SCAI@EMNLP 2018, Brussels, Belgium, October 31, 2018},
	author = {Weston, Jason and Dinan, Emily and Miller, Alexander H.},
	year = {2018},
	pages = {87--92},
	doi = {10.18653/v1/w18-5713}
}

@inproceedings{iterative,
  author       = {Sumanth Dathathri and
Andrea Madotto and
Janice Lan and
Jane Hung and
Eric Frank and
Piero Molino and
Jason Yosinski and
Rosanne Liu},
title        = {Plug and Play Language Models: {A} Simple Approach to Controlled Text
Generation},
booktitle    = {8th International Conference on Learning Representations, {ICLR} 2020,
Addis Ababa, Ethiopia, April 26-30, 2020},
publisher    = {OpenReview.net},
year         = {2020},
url          = {https://openreview.net/forum?id=H1edEyBKDS},
bibsource    = {dblp computer science bibliography, https://dblp.org}
}

@inproceedings{wang,
  author       = {Ziyu Wang and
Dingsu Wang and
Yixiao Zhang and
Gus Xia},

title        = {Learning Interpretable Representation for Controllable Polyphonic
Music Generation},
booktitle    = {Proceedings of the 21th International Society for Music Information
Retrieval Conference, {ISMIR} 2020, Montreal, Canada, October 11-16,
2020},
pages        = {662--669},
year         = {2020},
url          = {http://archives.ismir.net/ismir2020/paper/000094.pdf},
bibsource    = {dblp computer science bibliography, https://dblp.org}
}

@inproceedings{balstm,
	author    = {Johnson, Daniel D.},
	title     = {Generating Polyphonic Music Using Tied Parallel Networks},
	booktitle = {Computational Intelligence in Music, Sound, Art and Design},
	series    = {Lecture Notes in Computer Science},
	volume    = {10198},
	publisher = {Springer},
	location = {Amsterdam, The Netherlands},
	pages     = {128--143},
	year      = {2017},
	url       = {https://doi.org/10.1007/978-3-319-55750-2\_9},
	doi       = {10.1007/978-3-319-55750-2\_9},
	bibsource = {dblp computer science bibliography, https://dblp.org}
}

@INPROCEEDINGS{emo_lstm,
	author={Zhao, Kun and Li, Siqi and Cai, Juanjuan and Wang, Hui and Wang, Jingling},
	booktitle={2019 IEEE 3rd Information Technology, Networking, Electronic and Automation Control Conference (ITNEC)}, 
	title={An Emotional Symbolic Music Generation System based on LSTM Networks}, 
	year={2019},
	volume={},
	number={},
	pages={2039-2043},
	doi={10.1109/ITNEC.2019.8729266}}

@book{discrete_emotions,
	author = {Ekman, Paul},
	title = {Unmasking the face; a guide to recognizing emotions from facial clues},
	publisher = {Prentice-Hall},
	year = {1975},
	address = {Englewood Cliffs, N.J},
	isbn = {1883536367},
	url={https://psycnet.apa.org/record/1975-31746-000}
}

@inproceedings{emotion_survey_continuous,
  author       = {Hatice Gunes and
Bj{\"{o}}rn W. Schuller and
Maja Pantic and
Roddy Cowie},
title        = {Emotion representation, analysis and synthesis in continuous space:
{A} survey},
booktitle    = {Ninth {IEEE} International Conference on Automatic Face and Gesture
Recognition {(FG} 2011), Santa Barbara, CA, USA, 21-25 March 2011},
pages        = {827--834},
publisher    = {{IEEE} Computer Society},
year         = {2011},
url          = {https://doi.org/10.1109/FG.2011.5771357},
doi          = {10.1109/FG.2011.5771357},
bibsource    = {dblp computer science bibliography, https://dblp.org}
}

@inproceedings{game_soundtrack,
	title = {Dynamic Game Soundtrack Generation in Response to a Continuously Varying Emotional Trajectory},
	booktitle = {Audio Engineering Society Conference: 56th International Conference: Audio for Games},
	author = {Williams, Duncan and Kirke, Alexis and Eaton, Joel and Miranda, Eduardo and Daly, Ian and Hallowell, James and Roesch, Etienne and Hwang, Faustina and Nasuto, Slawomir J.},
	year = {2015},
	url={https://aes2.org/publications/elibrary-page/?id=17593}
}

@article{medicinal,
	title = {The Effects of Neurofeedback Training with Background Music on EEG Patterns of ADD and ADHD Children},
	author = {Pratt, Rosalie Rebollo and Abel, Hans-Henning and Skidmore, Jon},
	year = {1995},
	journal = {International Journal of Arts Medicine},
	volume = {4},
	number = {1},
	pages = {24--31},
	address = {US},
	issn = {1057-4263},
	url={https://psycnet.apa.org/record/1996-02780-004}
}

@article{bci,
	title = {Brain-Computer Music Interfacing (BCMI): From Basic Research to the Real World of Special Needs},
	shorttitle = {Brain-Computer Music Interfacing (BCMI)},
	author = {Miranda, Eduardo R. and Magee, Wendy L. and Wilson, John J. and Eaton, Joel and Palaniappan, Ramaswamy},
	year = {2011},
	journal = {Music \& Medicine},
	volume = {3},
	number = {3},
	pages = {134--140},
	doi={10.47513/mmd.v3i3.370}
}

@book{music_and_emotion,
    author = {Juslin, Patrik N and Sloboda, John A},
title = {Music And Emotion: Theory and research},
publisher = {Oxford University Press},
year = {2001},
month = {08},
isbn = {9780192631886},
doi = {10.1093/oso/9780192631886.001.0001},
url = {https://doi.org/10.1093/oso/9780192631886.001.0001},
}

@article{emotion_models_for_music,
author = {Tuomas Eerola and Jonna K. Vuoskoski},
title ={A comparison of the discrete and dimensional models of emotion in music},

journal = {Psychology of Music},
volume = {39},
number = {1},
pages = {18-49},
year = {2011},
doi = {10.1177/0305735610362821},

URL = { 

https://doi.org/10.1177/0305735610362821



},
eprint = { 

https://doi.org/10.1177/0305735610362821



}
}

@inproceedings{nucleus,
  author       = {Ari Holtzman and
Jan Buys and
Li Du and
Maxwell Forbes and
Yejin Choi},
title        = {The Curious Case of Neural Text Degeneration},
booktitle    = {8th International Conference on Learning Representations, {ICLR} 2020,
Addis Ababa, Ethiopia, April 26-30, 2020},
publisher    = {OpenReview.net},
year         = {2020},
url          = {https://openreview.net/forum?id=rygGQyrFvH},
bibsource    = {dblp computer science bibliography, https://dblp.org}
}

@inproceedings{kuleshov,
	author    = {Volodymyr Kuleshov and
	S. Zayd Enam and
	Stefano Ermon},
	title     = {Audio Super-Resolution using Neural Networks},
	booktitle = {5th International Conference on Learning Representations, {ICLR} 2017,
	Toulon, France, April 24-26, 2017, Workshop Track Proceedings},
	publisher = {OpenReview.net},
	year      = {2017},
	url       = {https://openreview.net/forum?id=S1gNakBFx},
	bibsource = {dblp computer science bibliography, https://dblp.org}
}

@article{sulun_audio,
  author={Sulun, Serkan and Davies, Matthew E. P.},
journal={IEEE Journal of Selected Topics in Signal Processing}, 
title={On Filter Generalization for Music Bandwidth Extension Using Deep Neural Networks}, 
year={2021},
volume={15},
number={1},
pages={132-142},
doi={10.1109/JSTSP.2020.3037485}}

@inproceedings{customer,
  author       = {Minqing Hu and
Bing Liu},
title        = {Mining and summarizing customer reviews},
booktitle    = {Proceedings of the Tenth {ACM} {SIGKDD} International Conference on
Knowledge Discovery and Data Mining, Seattle, Washington, USA, August
22-25, 2004},
pages        = {168--177},
publisher    = {{ACM}},
year         = {2004},
url          = {https://doi.org/10.1145/1014052.1014073},
doi          = {10.1145/1014052.1014073},
bibsource    = {dblp computer science bibliography, https://dblp.org}
}

@article{finance,
	title={Sentiment analysis on social media for stock movement prediction},
	author={Nguyen, Thien Hai and Shirai, Kiyoaki and Velcin, Julien},
	journal={Expert Systems with Applications},
	volume={42},
	number={24},
	pages={9603--9611},
	year={2015},
	publisher={Elsevier},
	url          = {https://doi.org/10.1016/j.eswa.2015.07.052},
	doi          = {10.1016/J.ESWA.2015.07.052},
}

@inproceedings{politics,
  author       = {Mohit Iyyer and
Peter Enns and
Jordan L. Boyd{-}Graber and
Philip Resnik},
title        = {Political Ideology Detection Using Recursive Neural Networks},
booktitle    = {Proceedings of the 52nd Annual Meeting of the Association for Computational
Linguistics, {ACL} 2014, June 22-27, 2014, Baltimore, MD, USA, Volume
1: Long Papers},
pages        = {1113--1122},
publisher    = {The Association for Computer Linguistics},
year         = {2014},
url          = {https://doi.org/10.3115/v1/p14-1105},
doi          = {10.3115/V1/P14-1105},
bibsource    = {dblp computer science bibliography, https://dblp.org}
}

@Article{emotionpv,
AUTHOR = {Almeida, João and Vilaça, Luís and Teixeira, Inês N. and Viana, Paula},
TITLE = {Emotion Identification in Movies through Facial Expression Recognition},
JOURNAL = {Applied Sciences},
VOLUME = {11},
YEAR = {2021},
NUMBER = {15},
ARTICLE-NUMBER = {6827},
URL = {https://www.mdpi.com/2076-3417/11/15/6827},
ISSN = {2076-3417},
DOI = {10.3390/app11156827}
}

@inproceedings{ngrams,
  author       = {Bo Pang and
Lillian Lee and
Shivakumar Vaithyanathan},
title        = {Thumbs up? Sentiment Classification using Machine Learning Techniques},
booktitle    = {Proceedings of the 2002 Conference on Empirical Methods in Natural
Language Processing, {EMNLP} 2002, Philadelphia, PA, USA, July 6-7,
2002},
pages        = {79--86},
year         = {2002},
url          = {https://aclanthology.org/W02-1011/},
doi          = {10.3115/1118693.1118704},
bibsource    = {dblp computer science bibliography, https://dblp.org}
}

@inproceedings{appraisal,
  author       = {Casey Whitelaw and
Navendu Garg and
Shlomo Argamon},
title        = {Using appraisal groups for sentiment analysis},
booktitle    = {Proceedings of the 2005 {ACM} {CIKM} International Conference on Information
and Knowledge Management, Bremen, Germany, October 31 - November 5,
2005},
pages        = {625--631},
publisher    = {{ACM}},
year         = {2005},
url          = {https://doi.org/10.1145/1099554.1099714},
doi          = {10.1145/1099554.1099714},
bibsource    = {dblp computer science bibliography, https://dblp.org}
}

@book{gpt,
	title = {Improving Language Understanding by Generative Pre-Training},
	author = {Radford, Alec and Narasimhan, Karthik and Salimans, Tim and Sutskever, Ilya},
	year = {2018},
	publisher = {OpenAI},
	url={http://www.nlpir.org/wordpress/wp-content/uploads/2019/06/Improving-language-understanding-by-generative-pre-training.pdf}
}

@article{emotion_review,
  author       = {Sheetal Kusal and
Shruti Patil and
Jyoti Choudrie and
Ketan Kotecha and
Deepali Rahul Vora and
Ilias O. Pappas},
title        = {A Review on Text-Based Emotion Detection - Techniques, Applications,
Datasets, and Future Directions},
journal      = {CoRR},
volume       = {abs/2205.03235},
year         = {2022},
url          = {https://doi.org/10.48550/arXiv.2205.03235},
doi          = {10.48550/ARXIV.2205.03235},
eprinttype    = {arXiv},
eprint       = {2205.03235},
bibsource    = {dblp computer science bibliography, https://dblp.org}
}

@book{music_survey,
  author       = {Jean{-}Pierre Briot and
Ga{\"{e}}tan Hadjeres and
Fran{\c{c}}ois{-}David Pachet},
title        = {Deep Learning Techniques for Music Generation},
publisher    = {Springer},
year         = {2020},
url          = {https://doi.org/10.1007/978-3-319-70163-9},
doi          = {10.1007/978-3-319-70163-9},
isbn         = {978-3-319-70162-2},
bibsource    = {dblp computer science bibliography, https://dblp.org}
}

@incollection{sentiment_survey,
  author       = {Bing Liu and
Lei Zhang},
title        = {A Survey of Opinion Mining and Sentiment Analysis},
booktitle    = {Mining Text Data},
pages        = {415--463},
publisher    = {Springer},
year         = {2012},
url          = {https://doi.org/10.1007/978-1-4614-3223-4\_13},
doi          = {10.1007/978-1-4614-3223-4\_13},
bibsource    = {dblp computer science bibliography, https://dblp.org}
}

@inproceedings{overfitting,
  author       = {Zhou Yu and
Jun Yu and
Jianping Fan and
Dacheng Tao},
title        = {Multi-modal Factorized Bilinear Pooling with Co-attention Learning
for Visual Question Answering},
booktitle    = {{IEEE} International Conference on Computer Vision, {ICCV} 2017, Venice,
Italy, October 22-29, 2017},
pages        = {1839--1848},
publisher    = {{IEEE} Computer Society},
year         = {2017},
url          = {https://doi.org/10.1109/ICCV.2017.202},
doi          = {10.1109/ICCV.2017.202},
bibsource    = {dblp computer science bibliography, https://dblp.org}
}

@article{aac,
author = {Duncan Williams and Alexis Kirke and Eduardo R Miranda and Etienne Roesch and Ian Daly and Slawomir Nasuto},
title ={Investigating affect in algorithmic composition systems},

journal = {Psychology of Music},
volume = {43},
number = {6},
pages = {831-854},
year = {2015},
doi = {10.1177/0305735614543282},
}

@inproceedings{mirexlike,
title      = {{Multi-modal Music Emotion Recognition: A New Dataset, Methodology and Comparative Analysis}},
shorttitle = {Multi-modal Music Emotion Recognition},
booktitle  = {{International Symposium on Computer Music Multidisciplinary Research}},
author     = {Panda, Renato and Malheiro, Ricardo and Rocha, Bruno and Oliveira, António and Paiva, Rui Pedro},
year       = {2013},
url        = {https://hdl.handle.net/10316/94095}
}

@inproceedings{emopia,
  author       = {Hsiao{-}Tzu Hung and
Joann Ching and
Seungheon Doh and
Nabin Kim and
Juhan Nam and
Yi{-}Hsuan Yang},
title        = {{EMOPIA:} {A} Multi-Modal Pop Piano Dataset For Emotion Recognition
and Emotion-based Music Generation},
booktitle    = {Proceedings of the 22nd International Society for Music Information
Retrieval Conference, {ISMIR} 2021, Online, November 7-12, 2021},
pages        = {318--325},
year         = {2021},
url          = {https://archives.ismir.net/ismir2021/paper/000039.pdf},
bibsource    = {dblp computer science bibliography, https://dblp.org}
}

@article{neuroscience,
	title={Brain correlates of music-evoked emotions},
	author={Koelsch, Stefan},
	journal={Nature Reviews Neuroscience},
	volume={15},
	number={3},
	pages={170--180},
	year={2014},
	publisher={Nature Publishing Group UK London},
	doi={10.1038/nrn3666}
}

@article{psychology,
author = {Carol L. Krumhansl},
title ={Music: A Link Between Cognition and Emotion},

journal = {Current Directions in Psychological Science},
volume = {11},
number = {2},
pages = {45-50},
year = {2002},
doi = {10.1111/1467-8721.00165},

URL = { 

https://doi.org/10.1111/1467-8721.00165



},
eprint = { 

https://doi.org/10.1111/1467-8721.00165



}

}

@book{composing_w_emotion,
	title={Emotion and meaning in music},
	author={Meyer, Leonard B},
	year={2008},
	publisher={University of Chicago Press},
	url={https://press.uchicago.edu/ucp/books/book/chicago/E/bo28551887.html}
}

@article{blues_lstm,
	title   = {A first look at music composition using {LSTM} recurrent neural networks},
	volume  = {103},
	journal = {Istituto Dalle Molle Di Studi Sull Intelligenza Artificiale},
	author  = {Eck, Douglas and Schmidhuber, Juergen},
	year    = {2002},
	pages   = {48},
	url = {https://people.idsia.ch/~juergen/blues/IDSIA-07-02.pdf}
}

@inproceedings{wnli,
  author       = {Hector J. Levesque and
Ernest Davis and
Leora Morgenstern},
title        = {The Winograd Schema Challenge},
booktitle    = {Principles of Knowledge Representation and Reasoning: Proceedings
of the Thirteenth International Conference, {KR} 2012, Rome, Italy,
June 10-14, 2012},
publisher    = {{AAAI} Press},
year         = {2012},
url          = {http://www.aaai.org/ocs/index.php/KR/KR12/paper/view/4492},
bibsource    = {dblp computer science bibliography, https://dblp.org}
}

@inproceedings{goemotions,
	author       = {Dorottya Demszky and
	Dana Movshovitz{-}Attias and
	Jeongwoo Ko and
	Alan S. Cowen and
	Gaurav Nemade and
	Sujith Ravi},
	title        = {GoEmotions: {A} Dataset of Fine-Grained Emotions},
	booktitle    = {Proceedings of the 58th Annual Meeting of the Association for Computational
	Linguistics, {ACL} 2020, Online, July 5-10, 2020},
	pages        = {4040--4054},
	publisher    = {Association for Computational Linguistics},
	year         = {2020},
	url          = {https://doi.org/10.18653/v1/2020.acl-main.372},
	doi          = {10.18653/V1/2020.ACL-MAIN.372},
	bibsource    = {dblp computer science bibliography, https://dblp.org}
}

@inproceedings{msd,
  author       = {Thierry Bertin{-}Mahieux and
Daniel P. W. Ellis and
Brian Whitman and
Paul Lamere},
title        = {The Million Song Dataset},
booktitle    = {Proceedings of the 12th International Society for Music Information
Retrieval Conference, {ISMIR} 2011, Miami, Florida, USA, October 24-28,
2011},
pages        = {591--596},
publisher    = {University of Miami},
year         = {2011},
url          = {http://ismir2011.ismir.net/papers/OS6-1.pdf},
bibsource    = {dblp computer science bibliography, https://dblp.org}
}

@article{sulun_midi,
  author       = {Serkan Sulun and
Matthew E. P. Davies and
Paula Viana},
title        = {Symbolic Music Generation Conditioned on Continuous-Valued Emotions},
journal      = {{IEEE} Access},
volume       = {10},
pages        = {44617--44626},
year         = {2022},
url          = {https://doi.org/10.1109/ACCESS.2022.3169744},
doi          = {10.1109/ACCESS.2022.3169744},
bibsource    = {dblp computer science bibliography, https://dblp.org}
}

@article{huggingface,
  author       = {Thomas Wolf and
Lysandre Debut and
Victor Sanh and
Julien Chaumond and
Clement Delangue and
Anthony Moi and
Pierric Cistac and
Tim Rault and
R{\'{e}}mi Louf and
Morgan Funtowicz and
Jamie Brew},
title        = {HuggingFace's Transformers: State-of-the-art Natural Language Processing},
journal      = {CoRR},
volume       = {abs/1910.03771},
year         = {2019},
url          = {http://arxiv.org/abs/1910.03771},
eprinttype    = {arXiv},
eprint       = {1910.03771},
bibsource    = {dblp computer science bibliography, https://dblp.org}
}

@incollection{restoration,
author="Godsill, Simon
and Rayner, Peter
and Capp{\'e}, Olivier",
title="Digital Audio Restoration",
bookTitle="Applications of Digital Signal Processing to Audio and Acoustics",
year="2002",
publisher="Springer US",
address="Boston, MA",
pages="133--194",
isbn="978-0-306-47042-4",
doi="10.1007/0-306-47042-X_4",
url="https://doi.org/10.1007/0-306-47042-X_4"
}

@article{audioinpainting,
  author       = {Amir Adler and
Valentin Emiya and
Maria G. Jafari and
Michael Elad and
R{\'{e}}mi Gribonval and
Mark D. Plumbley},
title        = {Audio Inpainting},
journal      = {{IEEE} Transactions on Speech Audio Processing},
volume       = {20},
number       = {3},
pages        = {922--932},
year         = {2012},
url          = {https://doi.org/10.1109/TASL.2011.2168211},
doi          = {10.1109/TASL.2011.2168211},
bibsource    = {dblp computer science bibliography, https://dblp.org}
}

@inproceedings{bansal2005,
  author       = {Dhananjay Bansal and
Bhiksha Raj and
Paris Smaragdis},
title        = {Bandwidth expansion of narrowband speech using non-negative matrix
factorization},
booktitle    = {9th European Conference on Speech Communication and Technology, INTERSPEECH-Eurospeech
2005, Lisbon, Portugal, September 4-8, 2005},
pages        = {1505--1508},
publisher    = {{ISCA}},
year         = {2005},
url          = {https://doi.org/10.21437/Interspeech.2005-528},
doi          = {10.21437/INTERSPEECH.2005-528},
bibsource    = {dblp computer science bibliography, https://dblp.org}
}

@article{medleydb,
	title={MedleyDB 2.0: New data and a system for sustainable data collection},
	author={Bittner, Rachel M and Wilkins, Julia and Yip, Hanna and Bello, Juan P},
	journal={ISMIR Late Breaking and Demo Papers},
	year={2016},
	url={https://bpb-us-e1.wpmucdn.com/wp.nyu.edu/dist/2/2294/files/2016/08/bittner-medleydb.pdf?bid=2294}
}

@inproceedings{
	dsd100,
  author       = {Antoine Liutkus and
Fabian{-}Robert St{\"{o}}ter and
Zafar Rafii and
Daichi Kitamura and
Bertrand Rivet and
Nobutaka Ito and
Nobutaka Ono and
Julie Fontecave},
title        = {The 2016 Signal Separation Evaluation Campaign},
booktitle    = {Latent Variable Analysis and Signal Separation - 13th International
Conference, {LVA/ICA} 2017, Grenoble, France, February 21-23, 2017,
Proceedings},
series       = {Lecture Notes in Computer Science},
volume       = {10169},
pages        = {323--332},
year         = {2017},
url          = {https://doi.org/10.1007/978-3-319-53547-0\_31},
doi          = {10.1007/978-3-319-53547-0\_31},
bibsource    = {dblp computer science bibliography, https://dblp.org}
}

@inproceedings{matthew,
	title={High frequency magnitude spectrogram reconstruction for music mixtures using convolutional autoencoders},
	author={Miron, Marius and Davies, Matthew E. P.},
	booktitle={21st International Conference on Digital Audio Effects (DAFx2018)},
	pages={173--180},
	year={2018},
	url={https://www.researchgate.net/publication/329372321_High_frequency_magnitude_spectrogram_reconstruction_for_music_mixtures_using_convolutional_autoencoders}
}

@inproceedings{alexnet,
  author       = {Alex Krizhevsky and
Ilya Sutskever and
Geoffrey E. Hinton},
title        = {ImageNet Classification with Deep Convolutional Neural Networks},
booktitle    = {Advances in Neural Information Processing Systems 25: 26th Annual
Conference on Neural Information Processing Systems 2012. Proceedings
of a meeting held December 3-6, 2012, Lake Tahoe, Nevada, United States},
pages        = {1106--1114},
year         = {2012},
url          = {https://proceedings.neurips.cc/paper/2012/hash/c399862d3b9d6b76c8436e924a68c45b-Abstract.html},
bibsource    = {dblp computer science bibliography, https://dblp.org}
}

@inproceedings{batchnorm,
  author       = {Sergey Ioffe and
Christian Szegedy},
title        = {Batch Normalization: Accelerating Deep Network Training by Reducing
Internal Covariate Shift},
booktitle    = {Proceedings of the 32nd International Conference on Machine Learning,
{ICML} 2015, Lille, France, 6-11 July 2015},
series       = {{JMLR} Workshop and Conference Proceedings},
volume       = {37},
pages        = {448--456},
publisher    = {JMLR.org},
year         = {2015},
url          = {http://proceedings.mlr.press/v37/ioffe15.html},
bibsource    = {dblp computer science bibliography, https://dblp.org}
}

@article{lstm,
  author       = {Sepp Hochreiter and
J{\"{u}}rgen Schmidhuber},
title        = {Long Short-Term Memory},
journal      = {Neural Comput.},
volume       = {9},
number       = {8},
pages        = {1735--1780},
year         = {1997},
url          = {https://doi.org/10.1162/neco.1997.9.8.1735},
doi          = {10.1162/NECO.1997.9.8.1735},
bibsource    = {dblp computer science bibliography, https://dblp.org}
}

@inproceedings{gan,
author = {Goodfellow, Ian and Pouget-Abadie, Jean and Mirza, Mehdi and Xu, Bing and Warde-Farley, David and Ozair, Sherjil and Courville, Aaron and Bengio, Yoshua},
title = {Generative adversarial networks},
year = {2020},
issue_date = {November 2020},
publisher = {Association for Computing Machinery},
address = {New York, NY, USA},
volume = {63},
number = {11},
issn = {0001-0782},
url = {https://doi.org/10.1145/3422622},
doi = {10.1145/3422622},
journal = {Communications of the ACM},
month = oct,
pages = {139–144},
numpages = {6}
}

@inproceedings{imagesr2,
  author       = {Bee Lim and
Sanghyun Son and
Heewon Kim and
Seungjun Nah and
Kyoung Mu Lee},
title        = {Enhanced Deep Residual Networks for Single Image Super-Resolution},
booktitle    = {2017 {IEEE} Conference on Computer Vision and Pattern Recognition
Workshops, {CVPR} Workshops 2017, Honolulu, HI, USA, July 21-26, 2017},
pages        = {1132--1140},
publisher    = {{IEEE} Computer Society},
year         = {2017},
url          = {https://doi.org/10.1109/CVPRW.2017.151},
doi          = {10.1109/CVPRW.2017.151},
bibsource    = {dblp computer science bibliography, https://dblp.org}
}

@article{gan_overfitting1,
  author       = {Ben Adlam and
Charles Weill and
Amol Kapoor},
title        = {Investigating Under and Overfitting in Wasserstein Generative Adversarial
Networks},
journal      = {CoRR},
volume       = {abs/1910.14137},
year         = {2019},
url          = {http://arxiv.org/abs/1910.14137},
eprinttype    = {arXiv},
eprint       = {1910.14137},
bibsource    = {dblp computer science bibliography, https://dblp.org}
}

@article{gan_overfitting2,
  author       = {Qiantong Xu and
Gao Huang and
Yang Yuan and
Chuan Guo and
Yu Sun and
Felix Wu and
Kilian Q. Weinberger},
title        = {An empirical study on evaluation metrics of generative adversarial
networks},
journal      = {CoRR},
volume       = {abs/1806.07755},
year         = {2018},
url          = {http://arxiv.org/abs/1806.07755},
eprinttype    = {arXiv},
eprint       = {1806.07755},
bibsource    = {dblp computer science bibliography, https://dblp.org}
}

@inproceedings{subpixel,
  author       = {Wenzhe Shi and
Jose Caballero and
Ferenc Huszar and
Johannes Totz and
Andrew P. Aitken and
Rob Bishop and
Daniel Rueckert and
Zehan Wang},
title        = {Real-Time Single Image and Video Super-Resolution Using an Efficient
Sub-Pixel Convolutional Neural Network},
booktitle    = {2016 {IEEE} Conference on Computer Vision and Pattern Recognition,
{CVPR} 2016, Las Vegas, NV, USA, June 27-30, 2016},
pages        = {1874--1883},
publisher    = {{IEEE} Computer Society},
year         = {2016},
url          = {https://doi.org/10.1109/CVPR.2016.207},
doi          = {10.1109/CVPR.2016.207},
bibsource    = {dblp computer science bibliography, https://dblp.org}
}

@article{waveunet,
  author       = {Craig Macartney and
Tillman Weyde},
title        = {Improved Speech Enhancement with the Wave-U-Net},
journal      = {CoRR},
volume       = {abs/1811.11307},
year         = {2018},
url          = {http://arxiv.org/abs/1811.11307},
eprinttype    = {arXiv},
eprint       = {1811.11307},
bibsource    = {dblp computer science bibliography, https://dblp.org}
}

@inproceedings{tfnet,
  author       = {Teck{-}Yian Lim and
Raymond A. Yeh and
Yijia Xu and
Minh N. Do and
Mark Hasegawa{-}Johnson},
title        = {Time-Frequency Networks for Audio Super-Resolution},
booktitle    = {2018 {IEEE} International Conference on Acoustics, Speech and Signal
Processing, {ICASSP} 2018, Calgary, AB, Canada, April 15-20, 2018},
pages        = {646--650},
publisher    = {{IEEE}},
year         = {2018},
url          = {https://doi.org/10.1109/ICASSP.2018.8462049},
doi          = {10.1109/ICASSP.2018.8462049},
bibsource    = {dblp computer science bibliography, https://dblp.org}
}

@inproceedings{audiounet,
  author       = {Andreas Jansson and
Eric J. Humphrey and
Nicola Montecchio and
Rachel M. Bittner and
Aparna Kumar and
Tillman Weyde},
title        = {Singing Voice Separation with Deep U-Net Convolutional Networks},
booktitle    = {Proceedings of the 18th International Society for Music Information
Retrieval Conference, {ISMIR} 2017, Suzhou, China, October 23-27,
2017},
pages        = {745--751},
year         = {2017},
url          = {https://ismir2017.smcnus.org/wp-content/uploads/2017/10/171\_Paper.pdf},
bibsource    = {dblp computer science bibliography, https://dblp.org}
}

@misc{github,
	title = {kuleshov/audio-super-res},
	copyright = {MIT},
	url = {https://github.com/kuleshov/audio-super-res},
	author = {Kuleshov, Volodymyr},
	month = apr,
	year = {2020}
}

@inproceedings{unet,
  author       = {Olaf Ronneberger and
Philipp Fischer and
Thomas Brox},
title        = {U-Net: Convolutional Networks for Biomedical Image Segmentation},
booktitle    = {Medical Image Computing and Computer-Assisted Intervention - {MICCAI}
2015 - 18th International Conference Munich, Germany, October 5 -
9, 2015, Proceedings, Part {III}},
series       = {Lecture Notes in Computer Science},
volume       = {9351},
pages        = {234--241},
publisher    = {Springer},
year         = {2015},
url          = {https://doi.org/10.1007/978-3-319-24574-4\_28},
doi          = {10.1007/978-3-319-24574-4\_28},
bibsource    = {dblp computer science bibliography, https://dblp.org}
}

@inproceedings{wide_resnet,
  author       = {Sergey Zagoruyko and
Nikos Komodakis},
title        = {Wide Residual Networks},
booktitle    = {Proceedings of the British Machine Vision Conference 2016, {BMVC}
2016, York, UK, September 19-22, 2016},
publisher    = {{BMVA} Press},
year         = {2016},
url          = {https://bmva-archive.org.uk/bmvc/2016/papers/paper087/index.html},
bibsource    = {dblp computer science bibliography, https://dblp.org}
}

@inproceedings{bwe_gan2,
  author       = {Jonas Sautter and
Friedrich Faubel and
Markus Buck and
Gerhard Schmidt},
title        = {Artificial Bandwidth Extension Using a Conditional Generative Adversarial
Network with Discriminative Training},
booktitle    = {{IEEE} International Conference on Acoustics, Speech and Signal Processing,
{ICASSP} 2019, Brighton, United Kingdom, May 12-17, 2019},
pages        = {7005--7009},
publisher    = {{IEEE}},
year         = {2019},
url          = {https://doi.org/10.1109/ICASSP.2019.8682649},
doi          = {10.1109/ICASSP.2019.8682649},
bibsource    = {dblp computer science bibliography, https://dblp.org}
}

@article{bwe_gan,
  author       = {Sung Kim and
Visvesh Sathe},
title        = {Bandwidth Extension on Raw Audio via Generative Adversarial Networks},
journal      = {CoRR},
volume       = {abs/1903.09027},
year         = {2019},
url          = {http://arxiv.org/abs/1903.09027},
eprinttype    = {arXiv},
eprint       = {1903.09027},
bibsource    = {dblp computer science bibliography, https://dblp.org}
}

@inproceedings{li,
  author       = {Kehuang Li and
Zhen Huang and
Yong Xu and
Chin{-}Hui Lee},
title        = {DNN-based speech bandwidth expansion and its application to adding
high-frequency missing features for automatic speech recognition of
narrowband speech},
booktitle    = {16th Annual Conference of the International Speech Communication Association,
{INTERSPEECH} 2015, Dresden, Germany, September 6-10, 2015},
pages        = {2578--2582},
publisher    = {{ISCA}},
year         = {2015},
url          = {https://doi.org/10.21437/Interspeech.2015-555},
doi          = {10.21437/INTERSPEECH.2015-555},
bibsource    = {dblp computer science bibliography, https://dblp.org}
}

@inproceedings{imagesr1,
  author       = {Yuchen Fan and
Honghui Shi and
Jiahui Yu and
Ding Liu and
Wei Han and
Haichao Yu and
Zhangyang Wang and
Xinchao Wang and
Thomas S. Huang},
title        = {Balanced Two-Stage Residual Networks for Image Super-Resolution},
booktitle    = {2017 {IEEE} Conference on Computer Vision and Pattern Recognition
Workshops, {CVPR} Workshops 2017, Honolulu, HI, USA, July 21-26, 2017},
pages        = {1157--1164},
publisher    = {{IEEE} Computer Society},
year         = {2017},
url          = {https://doi.org/10.1109/CVPRW.2017.154},
doi          = {10.1109/CVPRW.2017.154},
bibsource    = {dblp computer science bibliography, https://dblp.org}
}

@inproceedings{imagesr3,
  author       = {Yulun Zhang and
Yapeng Tian and
Yu Kong and
Bineng Zhong and
Yun Fu},
title        = {Residual Dense Network for Image Super-Resolution},
booktitle    = {2018 {IEEE} Conference on Computer Vision and Pattern Recognition,
{CVPR} 2018, Salt Lake City, UT, USA, June 18-22, 2018},
pages        = {2472--2481},
publisher    = {Computer Vision Foundation / {IEEE} Computer Society},
year         = {2018},
url          = {http://openaccess.thecvf.com/content\_cvpr\_2018/html/Zhang\_Residual\_Dense\_Network\_CVPR\_2018\_paper.html},
doi          = {10.1109/CVPR.2018.00262},
bibsource    = {dblp computer science bibliography, https://dblp.org}
}

@inproceedings{celltracing,
  author       = {Markus Rempfler and
Sanjeev Kumar and
Valentin Stierle and
Philipp Paulitschke and
Bjoern Andres and
Bjoern H. Menze},
title        = {Cell Lineage Tracing in Lens-Free Microscopy Videos},
booktitle    = {Medical Image Computing and Computer Assisted Intervention - {MICCAI}
2017 - 20th International Conference, Quebec City, QC, Canada, September
11-13, 2017, Proceedings, Part {II}},
series       = {Lecture Notes in Computer Science},
volume       = {10434},
pages        = {3--11},
publisher    = {Springer},
year         = {2017},
url          = {https://doi.org/10.1007/978-3-319-66185-8\_1},
doi          = {10.1007/978-3-319-66185-8\_1},
bibsource    = {dblp computer science bibliography, https://dblp.org}
}

@article{mr,
	title = {{MR} image reconstruction using deep learning: evaluation of network structure and loss functions},
	volume = {9},
	issn = {2223-4292},
	shorttitle = {{MR} image reconstruction using deep learning},
	doi = {10.21037/qims.2019.08.10},
	number = {9},
	urldate = {2020-04-12},
	journal = {Quantitative Imaging in Medicine and Surgery},
	author = {Ghodrati, Vahid and Shao, Jiaxin and Bydder, Mark and Zhou, Ziwu and Yin, Wotao and Nguyen, Kim-Lien and Yang, Yingli and Hu, Peng},
	month = sep,
	year = {2019},
	pmid = {31667138},
	pmcid = {PMC6785508},
	pages = {1516--1527}
}

@inproceedings{hdr,
  author       = {Shangzhe Wu and
Jiarui Xu and
Yu{-}Wing Tai and
Chi{-}Keung Tang},
title        = {Deep High Dynamic Range Imaging with Large Foreground Motions},
booktitle    = {Computer Vision - {ECCV} 2018 - 15th European Conference, Munich,
Germany, September 8-14, 2018, Proceedings, Part {II}},
series       = {Lecture Notes in Computer Science},
volume       = {11206},
pages        = {120--135},
publisher    = {Springer},
year         = {2018},
url          = {https://doi.org/10.1007/978-3-030-01216-8\_8},
doi          = {10.1007/978-3-030-01216-8\_8},
bibsource    = {dblp computer science bibliography, https://dblp.org}
}

@inproceedings{cancer,
  author       = {Eleni Chiou and
Francesco Giganti and
Elisenda Bonet{-}Carne and
Shonit Punwani and
Iasonas Kokkinos and
Eleftheria Panagiotaki},
title        = {Prostate Cancer Classification on {VERDICT} {DW-MRI} Using Convolutional
Neural Networks},
booktitle    = {Machine Learning in Medical Imaging - 9th International Workshop,
{MLMI} 2018, Held in Conjunction with {MICCAI} 2018, Granada, Spain,
September 16, 2018, Proceedings},
series       = {Lecture Notes in Computer Science},
volume       = {11046},
pages        = {319--327},
publisher    = {Springer},
year         = {2018},
url          = {https://doi.org/10.1007/978-3-030-00919-9\_37},
doi          = {10.1007/978-3-030-00919-9\_37},
bibsource    = {dblp computer science bibliography, https://dblp.org}
}

@article{cheng1994,
  author       = {Yan Ming Cheng and
Douglas D. O'Shaughnessy and
Paul Mermelstein},
title        = {Statistical recovery of wideband speech from narrowband speech},
journal      = {{IEEE} Transactions on Speech Audio Processing},
volume       = {2},
number       = {4},
pages        = {544--548},
year         = {1994},
url          = {https://doi.org/10.1109/89.326637},
doi          = {10.1109/89.326637},
bibsource    = {dblp computer science bibliography, https://dblp.org}
}

@inproceedings{yoshida1994,
  author       = {Yuki Yoshida and
Masanobu Abe},
title        = {An algorithm to reconstruct wideband speech from narrowband speech
based on codebook mapping},
booktitle    = {The 3rd International Conference on Spoken Language Processing, {ICSLP}
1994, Yokohama, Japan, September 18-22, 1994},
pages        = {1591--1594},
publisher    = {{ISCA}},
year         = {1994},
url          = {https://doi.org/10.21437/ICSLP.1994-412},
doi          = {10.21437/ICSLP.1994-412},
bibsource    = {dblp computer science bibliography, https://dblp.org}
}

@inproceedings{epps1999,
	title = {A new technique for wideband enhancement of coded narrowband speech},
	booktitle = {1999 {IEEE} {Workshop} on {Speech} {Coding} {Proceedings}. {Model}, {Coders}, and {Error} {Criteria} ({Cat}. {No}. {99EX351})},
	publisher = {IEEE},
	author = {Epps, Julien and Holmes, W. Harvey},
	year = {1999},
	pages = {174--176},
	url={https://ieeexplore.ieee.org/iel5/6345/16956/00781522.pdf}
}

@inproceedings{nakatoh1997,
  author       = {Yoshihisa Nakatoh and
Mineo Tsushima and
Takeshi Norimatsu},
title        = {Generation of broadband speech from narrowband speech using piecewise
linear mapping},
booktitle    = {Fifth European Conference on Speech Communication and Technology,
{EUROSPEECH} 1997, Rhodes, Greece, September 22-25, 1997},
pages        = {1643--1646},
publisher    = {{ISCA}},
year         = {1997},
url          = {https://doi.org/10.21437/Eurospeech.1997-469},
doi          = {10.21437/EUROSPEECH.1997-469},
bibsource    = {dblp computer science bibliography, https://dblp.org}
}

@inproceedings{chennoukh2001,
  author       = {Samir Chennoukh and
Andreas Johannes Gerrits and
G. Miet and
Robert Johannes Sluijter},
title        = {Speech enhancement via frequency bandwidth extension using line spectral
frequencies},
booktitle    = {{IEEE} International Conference on Acoustics, Speech, and Signal Processing,
{ICASSP} 2001, 7-11 May, 2001, Salt Palace Convention Center, Salt
Lake City, Utah, USA, Proceedings},
pages        = {665--668},
publisher    = {{IEEE}},
year         = {2001},
url          = {https://doi.org/10.1109/ICASSP.2001.940919},
doi          = {10.1109/ICASSP.2001.940919},
bibsource    = {dblp computer science bibliography, https://dblp.org}
}

@inproceedings{park2000,
  author       = {Kun{-}Youl Park and
Hyung Soon Kim},
title        = {Narrowband to wideband conversion of speech using {GMM} based transformation},
booktitle    = {{IEEE} International Conference on Acoustics, Speech, and Signal Processing.
{ICASSP} 2000, 5-9 June, 2000, Hilton Hotel and Convention Center,
Istanbul, Turkey},
pages        = {1843--1846},
publisher    = {{IEEE}},
year         = {2000},
url          = {https://doi.org/10.1109/ICASSP.2000.862114},
doi          = {10.1109/ICASSP.2000.862114},
bibsource    = {dblp computer science bibliography, https://dblp.org}
}

@inproceedings{nour2008,
  author       = {Amr H. Nour{-}Eldin and
Peter Kabal},
title        = {Mel-frequency cepstral coefficient-based bandwidth extension of narrowband
speech},
booktitle    = {9th Annual Conference of the International Speech Communication Association,
{INTERSPEECH} 2008, Brisbane, Australia, September 22-26, 2008},
pages        = {53--56},
publisher    = {{ISCA}},
year         = {2008},
url          = {https://doi.org/10.21437/Interspeech.2008-11},
doi          = {10.21437/INTERSPEECH.2008-11},
bibsource    = {dblp computer science bibliography, https://dblp.org}
}

@article{jax2003,
	title = {On artificial bandwidth extension of telephone speech},
	volume = {83},
	issn = {0165-1684},
	doi = {10.1016/S0165-1684(03)00082-3},
	language = {en},
	number = {8},
	urldate = {2020-04-28},
	journal = {Signal Processing},
	author = {Jax, Peter and Vary, Peter},
	month = aug,
	year = {2003},
	pages = {1707--1719}
}

@inproceedings{bauer2008,
  author       = {Patrick Bauer and
Tim Fingscheidt},
title        = {An HMM-based artificial bandwidth extension evaluated by cross-language
training and test},
booktitle    = {Proceedings of the {IEEE} International Conference on Acoustics, Speech,
and Signal Processing, {ICASSP} 2008, March 30 - April 4, 2008, Caesars
Palace, Las Vegas, Nevada, {USA}},
pages        = {4589--4592},
publisher    = {{IEEE}},
year         = {2008},
url          = {https://doi.org/10.1109/ICASSP.2008.4518678},
doi          = {10.1109/ICASSP.2008.4518678},
bibsource    = {dblp computer science bibliography, https://dblp.org}
}

@inproceedings{mmimdb,
	title = {Gated Multimodal Units for Information Fusion},
	booktitle = {5th International Conference on Learning Representations, ICLR 2017, Toulon, France, April 24-26, 2017, Workshop Track Proceedings},
	author = {Ovalle, John Edison Arevalo and Solorio, Thamar and {Montes-y-G{\'o}mez}, Manuel and Gonz{\'a}lez, Fabio A.},
	year = {2017},
	publisher = {OpenReview.net},
	url = {https://openreview.net/forum?id=S12\_nquOe},
	urldate = {2025-02-24}
}

@article{song2009,
	title = {A study of {HMM}-based bandwidth extension of speech signals},
	volume = {89},
	issn = {01651684},
	doi = {10.1016/j.sigpro.2009.03.037},
	language = {en},
	number = {10},
	urldate = {2020-04-28},
	journal = {Signal Processing},
	author = {Song, Geun-Bae and Martynovich, Pavel},
	month = oct,
	year = {2009},
	pages = {2036--2044}
}

@inproceedings{sun2013,
  author       = {Dennis L. Sun and
Rahul Mazumder},
title        = {Non-negative matrix completion for bandwidth extension: {A} convex
optimization approach},
booktitle    = {{IEEE} International Workshop on Machine Learning for Signal Processing,
{MLSP} 2013, Southampton, United Kingdom, September 22-25, 2013},
pages        = {1--6},
publisher    = {{IEEE}},
year         = {2013},
url          = {https://doi.org/10.1109/MLSP.2013.6661924},
doi          = {10.1109/MLSP.2013.6661924},
bibsource    = {dblp computer science bibliography, https://dblp.org}
}

@inproceedings{iser2003,
  author       = {Bernd Iser and
Gerhard Schmidt},
title        = {Neural networks versus codebooks in an application for bandwidth extension
of speech signals},
booktitle    = {8th European Conference on Speech Communication and Technology, {EUROSPEECH}
2003 - {INTERSPEECH} 2003, Geneva, Switzerland, September 1-4, 2003},
pages        = {565--568},
publisher    = {{ISCA}},
year         = {2003},
url          = {https://doi.org/10.21437/Eurospeech.2003-227},
doi          = {10.21437/EUROSPEECH.2003-227},
bibsource    = {dblp computer science bibliography, https://dblp.org}
}

@article{kontio2007,
	title = {Neural {Network}-{Based} {Artificial} {Bandwidth} {Expansion} of {Speech}},
	volume = {15},
	issn = {1558-7916},
	doi = {10.1109/TASL.2006.885934},
	number = {3},
	urldate = {2020-04-28},
	journal = {IEEE Transactions on Audio, Speech and Language Processing},
	author = {Kontio, Juho and Laaksonen, Laura and Alku, Paavo},
	month = mar,
	year = {2007},
	pages = {873--881}
}

@inproceedings{dataaugmentation,
  author       = {Luke Taylor and
Geoff Nitschke},
title        = {Improving Deep Learning with Generic Data Augmentation},
booktitle    = {{IEEE} Symposium Series on Computational Intelligence, {SSCI} 2018,
Bangalore, India, November 18-21, 2018},
pages        = {1542--1547},
publisher    = {{IEEE}},
year         = {2018},
url          = {https://doi.org/10.1109/SSCI.2018.8628742},
doi          = {10.1109/SSCI.2018.8628742},
bibsource    = {dblp computer science bibliography, https://dblp.org}
}

@inproceedings{mcfee2015_augmentation,
  author       = {Brian McFee and
Eric J. Humphrey and
Juan Pablo Bello},

title        = {A Software Framework for Musical Data Augmentation},
booktitle    = {Proceedings of the 16th International Society for Music Information
Retrieval Conference, {ISMIR} 2015, M{\'{a}}laga, Spain, October
26-30, 2015},
pages        = {248--254},
year         = {2015},
url          = {http://ismir2015.uma.es/articles/228\_Paper.pdf},
bibsource    = {dblp computer science bibliography, https://dblp.org}
}

@article{perraudin2018inpainting,
	title={Inpainting of long audio segments with similarity graphs},
	author={Perraudin, Nathanael and Holighaus, Nicki and Majdak, Piotr and Balazs, Peter},
	journal={IEEE/ACM Transactions on Audio, Speech, and Language Processing},
	volume={26},
	number={6},
  pages        = {1079--1090},
year         = {2018},
url          = {https://doi.org/10.1109/TASLP.2018.2809864},
doi          = {10.1109/TASLP.2018.2809864},
bibsource    = {dblp computer science bibliography, https://dblp.org}
}

@inproceedings{larger_models,
  author       = {Jaime Sevilla and
Lennart Heim and
Anson Ho and
Tamay Besiroglu and
Marius Hobbhahn and
Pablo Villalobos},
title        = {Compute Trends Across Three Eras of Machine Learning},
booktitle    = {International Joint Conference on Neural Networks, {IJCNN} 2022, Padua,
Italy, July 18-23, 2022},
pages        = {1--8},
publisher    = {{IEEE}},
year         = {2022},
url          = {https://doi.org/10.1109/IJCNN55064.2022.9891914},
doi          = {10.1109/IJCNN55064.2022.9891914},
bibsource    = {dblp computer science bibliography, https://dblp.org}
}

@article{canny,
  author       = {John F. Canny},
title        = {A Computational Approach to Edge Detection},
journal      = {{IEEE} Transactions on Pattern Analysis Machine Intelligence},
volume       = {8},
number       = {6},
pages        = {679--698},
year         = {1986},
url          = {https://doi.org/10.1109/TPAMI.1986.4767851},
doi          = {10.1109/TPAMI.1986.4767851},
bibsource    = {dblp computer science bibliography, https://dblp.org}
}

@article{flow,
	title = {Determining Optical Flow},
	author = {Horn, Berthold KP and Schunck, Brian G.},
	year = {1981},
	journal = {Artificial intelligence},
	volume = {17},
	number = {1-3},
	pages = {185--203},
	doi          = {10.1016/0004-3702(81)90024-2},
}

@article{texture,
	title = {A Comparative Study of Texture Measures with Classification Based on Featured Distributions},
	author = {Ojala, Timo and Pietik{\"a}inen, Matti and Harwood, David},
	year = {1996},
	journal = {Pattern recognition},
	volume = {29},
	number = {1},
	pages = {51--59},
	publisher = {Elsevier},
	doi          = {10.1016/0031-3203(95)00067-4}
}

@inproceedings{scenechange,
  author       = {Igor Bieda and
Anton Kisil and
Taras Panchenko},
title        = {An Approach to Scene Change Detection},
booktitle    = {2021 11th {IEEE} International Conference on Intelligent Data Acquisition
and Advanced Computing Systems: Technology and Applications (IDAACS),
Cracow, Poland, September 22-25, 2021},
pages        = {489--493},
publisher    = {{IEEE}},
year         = {2021},
url          = {https://doi.org/10.1109/IDAACS53288.2021.9660887},
doi          = {10.1109/IDAACS53288.2021.9660887},
bibsource    = {dblp computer science bibliography, https://dblp.org}
}

@inproceedings{action,
	title = {Towards Understanding Action Recognition},
	booktitle = {Proceedings of the IEEE International Conference on Computer Vision},
	author = {Jhuang, Hueihan and Gall, Juergen and Zuffi, Silvia and Schmid, Cordelia and Black, Michael J.},
	year = {2013},
	pages = {3192--3199},
	url = {http://openaccess.thecvf.com/content_iccv_2013/html/Jhuang_Towards_Understanding_Action_2013_ICCV_paper.html},
	urldate = {2025-02-20}
}

@inproceedings{object,
  author       = {Joseph Redmon and
Santosh Kumar Divvala and
Ross B. Girshick and
Ali Farhadi},
title        = {You Only Look Once: Unified, Real-Time Object Detection},
booktitle    = {2016 {IEEE} Conference on Computer Vision and Pattern Recognition,
{CVPR} 2016, Las Vegas, NV, USA, June 27-30, 2016},
pages        = {779--788},
publisher    = {{IEEE} Computer Society},
year         = {2016},
url          = {https://doi.org/10.1109/CVPR.2016.91},
doi          = {10.1109/CVPR.2016.91},
bibsource    = {dblp computer science bibliography, https://dblp.org}
}

@article{emotion_intro,
  author       = {Mohammad Soleymani and
Maja Pantic and
Thierry Pun},
title        = {Multimodal Emotion Recognition in Response to Videos},
journal      = {{IEEE} Transactions on Affective Computing},
volume       = {3},
number       = {2},
pages        = {211--223},
year         = {2012},
url          = {https://doi.org/10.1109/T-AFFC.2011.37},
doi          = {10.1109/T-AFFC.2011.37},
bibsource    = {dblp computer science bibliography, https://dblp.org}
}

@article{emotionrepresentation1,
	title = {Categorical and Dimensional Ratings of Emotional Speech: Behavioral Findings From the Morgan Emotional Speech Set},
	shorttitle = {Categorical and Dimensional Ratings of Emotional Speech},
	author = {Morgan, Shae D.},
	year = {2019},
	month = nov,
	journal = {Journal of Speech, Language, and Hearing Research},
	volume = {62},
	number = {11},
	pages = {4015--4029},
	issn = {1092-4388, 1558-9102},
	doi = {10.1044/2019_JSLHR-S-19-0144},
	urldate = {2025-02-21},
	langid = {english}
}

@article{emotionrepresentation2,
AUTHOR={Matsuda, Yoshi-Taka  and Fujimura, Tomomi  and Katahira, Kentaro  and Okada, Masato  and Ueno, Kenichi  and Cheng, Kang  and Okanoya, Kazuo },

TITLE={The implicit processing of categorical and dimensional strategies: an fMRI study of facial emotion perception},

JOURNAL={Frontiers in Human Neuroscience},

VOLUME={Volume 7 - 2013},

YEAR={2013},

URL={https://www.frontiersin.org/journals/human-neuroscience/articles/10.3389/fnhum.2013.00551},

DOI={10.3389/fnhum.2013.00551},

ISSN={1662-5161},
}

@article{emotionrepresentation3,
	title = {Emotions in text: dimensional and categorical models},
	author = {Calvo, Rafael A. and Mac Kim, Sunghwan},
	year = {2013},
	month = aug,
	journal = {Computational Intelligence},
	volume = {29},
	number = {3},
	pages = {527--543},
	issn = {0824-7935, 1467-8640},
	doi = {10.1111/j.1467-8640.2012.00456.x},
	urldate = {2025-02-21},
	copyright = {http://onlinelibrary.wiley.com/termsAndConditionstextbackslash{}#vor},
	langid = {english}
	}

@book{humanemotions,
	title = {Human Emotions},
	author = {Izard, Carroll E.},
	year = {2013},
	publisher = {Springer Science \& Business Media},
	url = {http://link.springer.com/10.1007/978-1-4899-2209-0},
}

@incollection{emotionwheel,
title = {The Nature of Emotions: Clinical Implications},
shorttitle = {The Nature of Emotions},
booktitle = {Emotions and Psychopathology},
author = {Plutchik, Robert},
year = {1988},
pages = {1--20},
publisher = {Springer US},
address = {Boston, MA},
doi = {10.1007/978-1-4757-1987-1_1},
urldate = {2025-02-21},
isbn = {978-1-4757-1987-1},
langid = {english}
}

@inproceedings{facialemotion,
  author       = {Andrey V. Savchenko},
title        = {EmotiEffNets for Facial Processing in Video-based Valence-Arousal
Prediction, Expression Classification and Action Unit Detection},
booktitle    = {{IEEE/CVF} Conference on Computer Vision and Pattern Recognition,
{CVPR} 2023 - Workshops, Vancouver, BC, Canada, June 17-24, 2023},
pages        = {5716--5724},
publisher    = {{IEEE}},
year         = {2023},
url          = {https://doi.org/10.1109/CVPRW59228.2023.00606},
doi          = {10.1109/CVPRW59228.2023.00606},
bibsource    = {dblp computer science bibliography, https://dblp.org}
}

@article{physiologicalemotion,
title = {Emotion Classification in Arousal Valence Model using MAHNOB-HCI Database},
journal = {International Journal of Advanced Computer Science and Applications},
doi = {10.14569/IJACSA.2017.080344},
url = {http://dx.doi.org/10.14569/IJACSA.2017.080344},
year = {2017},
publisher = {The Science and Information Organization},
volume = {8},
number = {3},
author = {Mimoun Ben Henia Wiem and Zied Lachiri}
}

@inproceedings{musicemotion,
	title = {Music Mood Classification Model Based on Arousal-Valence Values},
	booktitle = {13th International Conference on Advanced Communication Technology (Icact2011)},
	author = {Kim, JungHyun and Lee, Seungjae and Kim, SungMin and Yoo, Won Young},
	year = {2011},
	pages = {292--295},
	publisher = {IEEE},
	url = {https://ieeexplore.ieee.org/iel5/5740523/5745722/05745796.pdf}
}

@article{musdb18,
	title={The {MUSDB18} corpus for music separation},
	author={Rafii, Zafar and Liutkus, Antoine and St{\"o}ter, Fabian-Robert and Mimilakis, Stylianos Ioannis and Bittner, Rachel},
	year={2017},
	url={https://inria.hal.science/hal-02190845/document}
}

@misc{musdb18hq,
	title = {{MUSDB18-HQ} - an Uncompressed Version of {MUSDB18}},
	author = {Rafii, Zafar and Liutkus, Antoine and St{\"o}ter, Fabian-Robert and Mimilakis, Stylianos Ioannis and Bittner, Rachel},
	year = {2019},
	month = aug,
	publisher = {Zenodo},
	doi = {10.5281/zenodo.3338372},
	urldate = {2025-05-02}
}

@article{ekman6,
	title = {Heterogeneous Knowledge Transfer in Video Emotion Recognition, Attribution and Summarization},
	author = {Xu, Baohan and Fu, Yanwei and Jiang, Yu-Gang and Li, Boyang and Sigal, Leonid},
	year = {2018},
	journal={IEEE Transactions on Affective Computing},
	pages = {255--270},
	doi = {10.1109/TAFFC.2016.2622690}
}

@misc{translation,
  author       = {Marta R. Costa{-}juss{\`{a}} and
James Cross and
Onur {\c{C}}elebi and
Maha Elbayad and
Kenneth Heafield and
Kevin Heffernan and
Elahe Kalbassi and
Janice Lam and
Daniel Licht and
Jean Maillard and
Anna Y. Sun and
Skyler Wang and
Guillaume Wenzek and
Al Youngblood and
Bapi Akula and
Lo{\"{\i}}c Barrault and
Gabriel Mejia Gonzalez and
Prangthip Hansanti and
John Hoffman and
Semarley Jarrett and
Kaushik Ram Sadagopan and
Dirk Rowe and
Shannon Spruit and
Chau Tran and
Pierre Andrews and
Necip Fazil Ayan and
Shruti Bhosale and
Sergey Edunov and
Angela Fan and
Cynthia Gao and
Vedanuj Goswami and
Francisco Guzm{\'{a}}n and
Philipp Koehn and
Alexandre Mourachko and
Christophe Ropers and
Safiyyah Saleem and
Holger Schwenk and
Jeff Wang},
title        = {No Language Left Behind: Scaling Human-Centered Machine Translation},
journal      = {CoRR},
volume       = {abs/2207.04672},
year         = {2022},
url          = {https://doi.org/10.48550/arXiv.2207.04672},
doi          = {10.48550/ARXIV.2207.04672},
eprinttype    = {arXiv},
eprint       = {2207.04672},
bibsource    = {dblp computer science bibliography, https://dblp.org}
}

@article{textgeneration,
	title = {The Survey: Text Generation Models in Deep Learning},
	shorttitle = {The Survey},
	author = {Iqbal, Touseef and Qureshi, Shaima},
	year = {2022},
	month = jun,
	journal = {Journal of King Saud University - Computer and Information Sciences},
	volume = {34},
	number = {6, Part A},
	pages = {2515--2528},
	issn = {1319-1578},
	doi = {10.1016/j.jksuci.2020.04.001},
	urldate = {2025-02-21}
}

@article{questionanswering,
	title = {Conversational Question Answering: A Survey},
	shorttitle = {Conversational Question Answering},
	author = {Zaib, Munazza and Zhang, Wei Emma and Sheng, Quan Z. and Mahmood, Adnan and Zhang, Yang},
	year = {2022},
	month = dec,
	journal = {Knowledge and Information Systems},
	volume = {64},
	number = {12},
	pages = {3151--3195},
	issn = {0219-1377, 0219-3116},
	doi = {10.1007/s10115-022-01744-y},
	urldate = {2025-02-21},
	langid = {english}
}

@inproceedings{subwords,
	title = {Neural Machine Translation of Rare Words with Subword Units},
	booktitle = {Proceedings of the 54th Annual Meeting of the Association for Computational Linguistics, ACL 2016, August 7-12, 2016, Berlin, Germany, Volume 1: Long Papers},
	author = {Sennrich, Rico and Haddow, Barry and Birch, Alexandra},
	year = {2016},
	publisher = {The Association for Computer Linguistics},
	doi = {10.18653/V1/P16-1162},
	urldate = {2025-02-21}
}

@article{rnn,
	title = {Learning Representations by Back-Propagating Errors},
	author = {Rumelhart, David E. and Hinton, Geoffrey E. and Williams, Ronald J.},
	year = {1986},
	journal = {Nature},
	volume = {323},
	number = {6088},
	pages = {533--536},
	publisher = {Nature Publishing Group},
	doi={10.1038/323533a0}
}

@article{layernorm,
  author       = {Lei Jimmy Ba and
Jamie Ryan Kiros and
Geoffrey E. Hinton},
title        = {Layer Normalization},
journal      = {CoRR},
volume       = {abs/1607.06450},
year         = {2016},
url          = {http://arxiv.org/abs/1607.06450},
eprinttype    = {arXiv},
eprint       = {1607.06450},
bibsource    = {dblp computer science bibliography, https://dblp.org}
}

@inproceedings{relu,
	title = {Rectified Linear Units Improve Restricted Boltzmann Machines},
	booktitle = {Proceedings of the 27th International Conference on Machine Learning (ICML-10)},
	author = {Nair, Vinod and Hinton, Geoffrey E.},
	year = {2010},
	pages = {807--814},
	url = {https://www.cs.toronto.edu/~hinton/absps/reluICML.pdf},
	urldate = {2025-02-22}
}

@inproceedings{social_video_emotion,
	title = {Pairwise Emotional Relationship Recognition in Drama Videos: Dataset and Benchmark},
	shorttitle = {Pairwise Emotional Relationship Recognition in Drama Videos},
	booktitle = {Proceedings of the 29th ACM International Conference on Multimedia},
	author = {Gao, Xun and Zhao, Yin and Zhang, Jie and Cai, Longjun},
	year = {2021},
	month = oct,
	pages = {3380--3389},
	publisher = {ACM},
	address = {Virtual Event China},
	doi = {10.1145/3474085.3475493},
	urldate = {2025-02-22},
	isbn = {978-1-4503-8651-7},
	langid = {english}
}

@inproceedings{caer,
	title = {Context-Aware Emotion Recognition Networks},
	booktitle = {2019 IEEE/CVF International Conference on Computer Vision, ICCV 2019, Seoul, Korea (South), October 27 - November 2, 2019},
	author = {Lee, Jiyoung and Kim, Seungryong and Kim, Sunok and Park, Jungin and Sohn, Kwanghoon},
	year = {2019},
	pages = {10142--10151},
	publisher = {IEEE},
	doi = {10.1109/ICCV.2019.01024}
}

@article{musicvideodataset,
	title = {Deep Learning-Based Late Fusion of Multimodal Information for Emotion Classification of Music Video},
	author = {Pandeya, Yagya Raj and Lee, Joonwhoan},
	year = {2021},
	month = jan,
	journal = {Multimedia Tools and Applications},
	volume = {80},
	number = {2},
	pages = {2887--2905},
	issn = {1573-7721},
	doi = {10.1007/s11042-020-08836-3},
	urldate = {2022-06-28},
	langid = {english}
}

@inproceedings{fer,
	title = {Training Deep Networks for Facial Expression Recognition with Crowd-Sourced Label Distribution},
	booktitle = {Proceedings of the 18th ACM International Conference on Multimodal Interaction},
	author = {Barsoum, Emad and Zhang, Cha and Ferrer, Cristian Canton and Zhang, Zhengyou},
	year = {2016},
	month = oct,
	pages = {279--283},
	publisher = {ACM},
	address = {Tokyo Japan},
	doi = {10.1145/2993148.2993165},
	urldate = {2025-02-22},
	isbn = {978-1-4503-4556-9},
	langid = {english}
}

@article{videoemotion8,
	title = {Predicting Emotions in User-Generated Videos},
	author = {Jiang, Yu-Gang and Xu, Baohan and Xue, Xiangyang},
	year = {2014},
	month = jun,
	journal = {Proceedings of the AAAI Conference on Artificial Intelligence},
	volume = {28},
	number = {1},
	issn = {2374-3468},
	doi = {10.1609/aaai.v28i1.8724},
	urldate = {2023-11-28},
	copyright = {Copyright (c)},
	langid = {english}
}

@inproceedings{odaq,
	title = {Odaq: Open Dataset of Audio Quality},
	shorttitle = {Odaq},
	booktitle = {ICASSP 2024 - 2024 IEEE International Conference on Acoustics, Speech and Signal Processing (ICASSP)},
	author = {Torcoli, Matteo and Wu, Chih-Wei and Dick, Sascha and Williams, Phillip A. and Modar Halimeh, Mhd and Wolcott, William and Habets, Emanu{\"e}l A. P.},
	year = {2024},
	month = apr,
	pages = {836--840},
	issn = {2379-190X},
	doi = {10.1109/ICASSP48485.2024.10447634},
	urldate = {2025-02-23}
}

@article{aec,
  author       = {Ross Cutler and
Ando Saabas and
Tanel P{\"{a}}rnamaa and
Marju Purin and
Evgenii Indenbom and
Nicolae{-}Catalin Ristea and
Jegor Guzvin and
Hannes Gamper and
Sebastian Braun and
Robert Aichner},
title        = {{ICASSP} 2023 Acoustic Echo Cancellation Challenge},
journal      = {CoRR},
volume       = {abs/2309.12553},
year         = {2023},
url          = {https://doi.org/10.48550/arXiv.2309.12553},
doi          = {10.48550/ARXIV.2309.12553},
eprinttype    = {arXiv},
eprint       = {2309.12553},
bibsource    = {dblp computer science bibliography, https://dblp.org}
}

@article{spass,
	title = {Dataset for Polyphonic Sound Event Detection Tasks in Urban Soundscapes: The Synthetic Polyphonic Ambient Sound Source (SPASS) Dataset},
	shorttitle = {Dataset for Polyphonic Sound Event Detection Tasks in Urban Soundscapes},
	author = {{Viveros-Mu{\~n}oz}, Rhoddy and Huijse, Pablo and Vargas, Victor and Espejo, Diego and Poblete, Victor and Arenas, Jorge P. and Vernier, Matthieu and Vergara, Diego and Su{\'a}rez, Enrique},
	year = {2023},
	month = oct,
	journal = {Data in Brief},
	volume = {50},
	pages = {109552},
	issn = {2352-3409},
	doi = {10.1016/j.dib.2023.109552},
	urldate = {2025-02-23}
}

@inproceedings{language_identification1,
  author       = {Phillip Keung and
Yichao Lu and
Gy{\"{o}}rgy Szarvas and
Noah A. Smith},
title        = {The Multilingual Amazon Reviews Corpus},
booktitle    = {Proceedings of the 2020 Conference on Empirical Methods in Natural
Language Processing, {EMNLP} 2020, Online, November 16-20, 2020},
pages        = {4563--4568},
publisher    = {Association for Computational Linguistics},
year         = {2020},
url          = {https://doi.org/10.18653/v1/2020.emnlp-main.369},
doi          = {10.18653/V1/2020.EMNLP-MAIN.369},
bibsource    = {dblp computer science bibliography, https://dblp.org}
}

@InProceedings{language_identification2,
  author       = {Alexis Conneau and
Ruty Rinott and
Guillaume Lample and
Adina Williams and
Samuel R. Bowman and
Holger Schwenk and
Veselin Stoyanov},
title        = {{XNLI:} Evaluating Cross-lingual Sentence Representations},
booktitle    = {Proceedings of the 2018 Conference on Empirical Methods in Natural
Language Processing, Brussels, Belgium, October 31 - November 4, 2018},
pages        = {2475--2485},
publisher    = {Association for Computational Linguistics},
year         = {2018},
url          = {https://doi.org/10.18653/v1/d18-1269},
doi          = {10.18653/V1/D18-1269},
bibsource    = {dblp computer science bibliography, https://dblp.org}
}

@InProceedings{language_identification3,
	title = {Machine translated multilingual STS benchmark dataset.},
	author={Philip May},
	year={2021},
	url={https://github.com/PhilipMay/stsb-multi-mt}
}

@inproceedings{movienet,
	ids = {huangMovieNetHolisticDataset2020},
	title = {{MovieNet}: A Holistic Dataset for Movie Understanding},
	booktitle = {Computer Vision - ECCV 2020 - 16th European Conference, Glasgow, UK, August 23-28, 2020, Proceedings, Part IV},
	author = {Huang, Qingqiu and Xiong, Yu and Rao, Anyi and Wang, Jiaze and Lin, Dahua},
	year = {2020},
	series = {Lecture Notes in Computer Science},
	volume = {12349},
	pages = {709--727},
	publisher = {Springer},
	doi = {10.1007/978-3-030-58548-8_41}
}

@inproceedings{whisper,
	title = {Robust Speech Recognition via Large-Scale Weak Supervision},
	booktitle = {International Conference on Machine Learning, ICML 2023, 23-29 July 2023, Honolulu, Hawaii, USA},
	author = {Radford, Alec and Kim, Jong Wook and Xu, Tao and Brockman, Greg and McLeavey, Christine and Sutskever, Ilya},
	year = {2023},
	series = {Proceedings of Machine Learning Research},
	volume = {202},
	pages = {28492--28518},
	publisher = {PMLR},
	url = {https://proceedings.mlr.press/v202/radford23a.html},
}

@inproceedings{musicnet,
	author       = {Keunwoo Choi and
	Gy{\"{o}}rgy Fazekas and
	Mark B. Sandler},
	title        = {Automatic Tagging Using Deep Convolutional Neural Networks},
	booktitle    = {Proceedings of the 17th International Society for Music Information
	Retrieval Conference, {ISMIR} 2016, New York City, United States,
	August 7-11, 2016},
	pages        = {805--811},
	year         = {2016},
	url          = {https://wp.nyu.edu/ismir2016/wp-content/uploads/sites/2294/2016/07/009\_Paper.pdf},
	bibsource    = {dblp computer science bibliography, https://dblp.org}
}

@inproceedings{bert,
	author       = {Jacob Devlin and
	Ming{-}Wei Chang and
	Kenton Lee and
	Kristina Toutanova},
	title        = {{BERT:} Pre-training of Deep Bidirectional Transformers for Language
	Understanding},
	booktitle    = {Proceedings of the 2019 Conference of the North American Chapter of
	the Association for Computational Linguistics: Human Language Technologies,
	{NAACL-HLT} 2019, Minneapolis, MN, USA, June 2-7, 2019, Volume 1 (Long
	and Short Papers)},
	pages        = {4171--4186},
	publisher    = {Association for Computational Linguistics},
	year         = {2019},
	doi          = {10.18653/v1/n19-1423},
	bibsource    = {dblp computer science bibliography, https://dblp.org}
}

@inproceedings{distillation_2006,
	author       = {Cristian Bucila and
	Rich Caruana and
	Alexandru Niculescu{-}Mizil},
	title        = {Model compression},
	booktitle    = {Proceedings of the Twelfth {ACM} {SIGKDD} International Conference
	on Knowledge Discovery and Data Mining, Philadelphia, PA, USA, August
	20-23, 2006},
	pages        = {535--541},
	publisher    = {{ACM}},
	year         = {2006},
	doi          = {10.1145/1150402.1150464},
	bibsource    = {dblp computer science bibliography, https://dblp.org}
}

@article{distillation_2015,
	title = {Distilling the Knowledge in a Neural Network},
	author = {Hinton, Geoffrey and Vinyals, Oriol and Dean, Jeff},
	year = {2015},
	journal = {arXiv preprint arXiv:1503.02531},
	doi = {10.48550/arXiv.1503.02531},
	archiveprefix = {arXiv}
}

@inproceedings{learned_position,
	author       = {Benyou Wang and
	Donghao Zhao and
	Christina Lioma and
	Qiuchi Li and
	Peng Zhang and
	Jakob Grue Simonsen},
	title        = {Encoding word order in complex embeddings},
	booktitle    = {8th International Conference on Learning Representations, {ICLR} 2020,
	Addis Ababa, Ethiopia, April 26-30, 2020},
	publisher    = {OpenReview.net},
	year         = {2020},
	url          = {https://openreview.net/forum?id=Hke-WTVtwr},
	bibsource    = {dblp computer science bibliography, https://dblp.org}
}

@article{movielens,
	ids = {harperMovielensDatasetsHistory2015},
	title = {The {MovieLens} Datasets: History and Context},
	author = {Harper, F. Maxwell and Konstan, Joseph A.},
	year = {2016},
	journal = {{ACM} Transactions on Interactive Intelligent Systems
},
	volume = {5},
	number = {4},
	pages = {19:1--19:19},
	doi = {10.1145/2827872}
}

@inproceedings{condensed,
	author       = {Max Bain and
	Arsha Nagrani and
	Andrew Brown and
	Andrew Zisserman},
	title        = {Condensed Movies: Story Based Retrieval with Contextual Embeddings},
	booktitle    = {Computer Vision - {ACCV} 2020 - 15th Asian Conference on Computer
	Vision, Kyoto, Japan, November 30 - December 4, 2020, Revised Selected
	Papers, Part {V}},
	series       = {Lecture Notes in Computer Science},
	volume       = {12626},
	pages        = {460--479},
	publisher    = {Springer},
	year         = {2020},
	doi          = {10.1007/978-3-030-69541-5\_28},
	bibsource    = {dblp computer science bibliography, https://dblp.org}
}

@inproceedings{imagenet,
	author       = {Jia Deng and
	Wei Dong and
	Richard Socher and
	Li{-}Jia Li and
	Kai Li and
	Li Fei{-}Fei},
	title        = {{ImageNet}: {A} large-scale hierarchical image database},
	booktitle    = {2009 {IEEE} Computer Society Conference on Computer Vision and Pattern
	Recognition {(CVPR} 2009), 20-25 June 2009, Miami, Florida, {USA}},
	pages        = {248--255},
	publisher    = {{IEEE} Computer Society},
	year         = {2009},
	doi          = {10.1109/CVPR.2009.5206848},
	bibsource    = {dblp computer science bibliography, https://dblp.org}
}

@article{dropout,
	title = {Dropout: {A} Simple Way to Prevent Neural Networks from Overfitting},
	shorttitle = {Dropout},
	author = {Srivastava, Nitish and Hinton, Geoffrey and Krizhevsky, Alex and Sutskever, Ilya and Salakhutdinov, Ruslan},
	year = {2014},
	journal = {The journal of machine learning research},
	volume = {15},
	number = {1},
	pages = {1929--1958},
	publisher = {JMLR. org},
	doi          = {10.5555/2627435.2670313},
}

@inproceedings{audioset,
	title = {Audio Set: {An} Ontology and Human-Labeled Dataset for Audio Events},
	booktitle = {2017 IEEE International Conference on Acoustics, Speech and Signal Processing, ICASSP 2017, New Orleans, LA, USA, March 5-9, 2017},
	author = {Gemmeke, Jort F. and Ellis, Daniel P. W. and Freedman, Dylan and Jansen, Aren and Lawrence, Wade and Moore, R. Channing and Plakal, Manoj and Ritter, Marvin},
	year = {2017},
	pages = {776--780},
	publisher = {IEEE},
	doi = {10.1109/ICASSP.2017.7952261}
}

@inproceedings{trailer_poster_classification,
	title = {Leveraging Efficient Training and Feature Fusion in Transformers for Multimodal Classification},
	booktitle = {IEEE International Conference on Image Processing, ICIP 2023, Kuala Lumpur, Malaysia, October 8-11, 2023},
	author = {Ak, Kenan Emir and Lee, Gwang-Gook and Xu, Yan and Shen, Mingwei},
	year = {2023},
	pages = {1420--1424},
	publisher = {IEEE},
	doi = {10.1109/ICIP49359.2023.10223098}
}

@article{trailer_poster_classification_2,
	title = {Simple and Effective Multimodal Learning Based on Pre-Trained Transformer Models},
	author = {Miyazawa, Kazuki and Kyuragi, Yuta and Nagai, Takayuki},
	year = {2022},
	journal = {IEEE Access},
	volume = {10},
	pages = {29821--29833},
	doi = {10.1109/ACCESS.2022.3159346}
}

@inproceedings{trailer_metadata_classification,
	title = {Multimodal Weighted Fusion of Transformers for Movie Genre Classification},
	booktitle = {Proceedings of the Third Workshop on Multimodal Artificial Intelligence},
	author = {Rodriguez Bribiesca, Isaac and Lopez Monroy, Adrian Pastor and Montes-y-Gomez, Manuel},
	year = {2021},
	month = jun,
	pages = {1--5},
	publisher = {Association for Computational Linguistics},
	address = {Mexico City, Mexico},
	doi = {10.18653/v1/2021.maiworkshop-1.1}
}

@inproceedings{larger1,
	title = 	 {{E}fficient{N}et: Rethinking Model Scaling for Convolutional Neural Networks},
	author =       {Tan, Mingxing and Le, Quoc},
	booktitle = 	 {Proceedings of the 36th International Conference on Machine Learning},
	pages = 	 {6105--6114},
	year = 	 {2019},
	volume = 	 {97},
	series = 	 {Proceedings of Machine Learning Research},
	month = 	 {6},
	publisher =    {PMLR},
	url = 	 {https://proceedings.mlr.press/v97/tan19a.html}
}

@article{larger2,
	title = {Freely Scalable and Reconfigurable Optical Hardware for Deep Learning},
	author = {Bernstein, Liane and Sludds, Alexander and Hamerly, Ryan and Sze, Vivienne and Emer, Joel and Englund, Dirk},
	year = {2021},
	journal = {Scientific reports},
	volume = {11},
	number = {1},
	pages = {3144},
	publisher = {Nature Publishing Group UK London},
	doi = {10.1038/s41598-021-82543-3},
	urldate = {2024-07-17}
}

@incollection{transfer1,
	title = {A Survey on Deep Transfer Learning},
	booktitle = {Artificial Neural Networks and Machine Learning -- ICANN 2018},
	author = {Tan, Chuanqi and Sun, Fuchun and Kong, Tao and Zhang, Wenchang and Yang, Chao and Liu, Chunfang},
	year = {2018},
	volume = {11141},
	pages = {270--279},
	publisher = {Springer International Publishing},
	address = {Cham},
	doi = {10.1007/978-3-030-01424-7_27},
	urldate = {2024-07-17},
	isbn = {978-3-030-01423-0}
}

@article{transfer2,
	title = {A Decade Survey of Transfer Learning (2010--2020)},
	author = {Niu, Shuteng and Liu, Yongxin and Wang, Jian and Song, Houbing},
	year = {2020},
	journal = {IEEE Transactions on Artificial Intelligence},
	volume = {1},
	number = {2},
	pages = {151--166},
	publisher = {IEEE},
	doi={10.1109/TAI.2021.3054609},
	urldate = {2024-07-17}
}

@article{avoiding_overfitting,
	title = {Avoiding Overfitting in the Analysis of High-Dimensional Data with Artificial Neural Networks ({ANNs})},
	author = {Defernez, Marianne and Kemsley, E. Katherine},
	year = {1999},
	month = jan,
	journal = {Analyst},
	volume = {124},
	number = {11},
	pages = {1675--1681},
	publisher = {The Royal Society of Chemistry},
	issn = {1364-5528},
	doi = {10.1039/A905556H},
	url = {https://pubs.rsc.org/en/content/articlepdf/1999/an/a905556h},
	langid = {english}
}

@article{activity_e,
	title = {Transfer Learning and Its Extensive Appositeness in Human Activity Recognition: A Survey},
	shorttitle = {Transfer Learning and Its Extensive Appositeness in Human Activity Recognition},
	author = {Ray, Abhisek and Kolekar, Maheshkumar H.},
	year = {2024},
	month = apr,
	journal = {Expert Systems with Applications},
	volume = {240},
	pages = {122538},
	issn = {0957-4174},
	doi = {10.1016/j.eswa.2023.122538},
	urldate = {2024-08-21}
}

@article{recommendation_e,
	title = {The Complementarity of a Diverse Range of Deep Learning Features Extracted from Video Content for Video Recommendation},
	author = {Almeida, Adolfo and {de Villiers}, Johan Pieter and De Freitas, Allan and Velayudan, Mergandran},
	year = {2022},
	month = apr,
	journal = {Expert Systems with Applications},
	volume = {192},
	pages = {116335},
	issn = {0957-4174},
	doi = {10.1016/j.eswa.2021.116335},
	urldate = {2024-08-21}
}

@article{summarization_e,
	title = {Deep Multi-Scale Pyramidal Features Network for Supervised Video Summarization},
	author = {Khan, Habib and Hussain, Tanveer and Ullah Khan, Samee and Ahmad Khan, Zulfiqar and Baik, Sung Wook},
	year = {2024},
	month = mar,
	journal = {Expert Systems with Applications},
	volume = {237},
	pages = {121288},
	issn = {0957-4174},
	doi = {10.1016/j.eswa.2023.121288},
	urldate = {2024-08-21}
}

@article{reasoning,
  author       = {Jason Wei and
Xuezhi Wang and
Dale Schuurmans and
Maarten Bosma and
Brian Ichter and
Fei Xia and
Ed H. Chi and
Quoc V. Le and
Denny Zhou},
title        = {Chain-of-Thought Prompting Elicits Reasoning in Large Language Models},
booktitle    = {Advances in Neural Information Processing Systems 35: Annual Conference
on Neural Information Processing Systems 2022, NeurIPS 2022, New Orleans,
LA, USA, November 28 - December 9, 2022},
year         = {2022},
url          = {http://papers.nips.cc/paper\_files/paper/2022/hash/9d5609613524ecf4f15af0f7b31abca4-Abstract-Conference.html},
bibsource    = {dblp computer science bibliography, https://dblp.org}
}

@article{continual,
  author       = {Liyuan Wang and
Xingxing Zhang and
Hang Su and
Jun Zhu},
title        = {A Comprehensive Survey of Continual Learning: Theory, Method and Application},
journal      = {{IEEE} Transactions on Pattern Analysis Machine Intelligence},
volume       = {46},
number       = {8},
pages        = {5362--5383},
year         = {2024},
url          = {https://doi.org/10.1109/TPAMI.2024.3367329},
doi          = {10.1109/TPAMI.2024.3367329},
bibsource    = {dblp computer science bibliography, https://dblp.org}
}

@article{scaling,
  author       = {Jared Kaplan and
Sam McCandlish and
Tom Henighan and
Tom B. Brown and
Benjamin Chess and
Rewon Child and
Scott Gray and
Alec Radford and
Jeffrey Wu and
Dario Amodei},
title        = {Scaling Laws for Neural Language Models},
journal      = {CoRR},
volume       = {abs/2001.08361},
year         = {2020},
url          = {https://arxiv.org/abs/2001.08361},
eprinttype    = {arXiv},
eprint       = {2001.08361},
bibsource    = {dblp computer science bibliography, https://dblp.org}
}

@book{soundtrack_emotion1,
	title = {Music Composition for Film and Television},
	author = {Schifrin, Lalo},
	year = {2011},
	publisher = {Hal Leonard Corporation},
	url = {https://books.google.com/books/about/Music_Composition_for_Film_and_Televisio.html?id=gTwLAQAAQBAJ}
}

@article{soundtrack_emotion2,
	title = {The Soundtrack (Putting Music in Its Place)},
	author = {Deutsch, Stephen},
	year = {2007},
	journal = {The Soundtrack},
	volume = {1},
	number = {1},
	pages = {3--13},
	url = {http://eprints.bournemouth.ac.uk/1307/}
}

@book{soundtrack_scenes1,
	title = {Film Music: A Neglected Art: A Critical Study of Music in Films},
	shorttitle = {Film Music},
	author = {Prendergast, Roy M.},
	year = {1992},
	publisher = {WW Norton \& Company},
	url={https://books.google.com/books/about/Film_Music.html?id=7ozTus2HZbsC}
}

@article{soundtrack_scenes2,
	title = {Understanding Musical Soundtracks},
	author = {Cohen, Annabel J.},
	year = {1990},
	month = jul,
	journal = {Empirical Studies of the Arts},
	volume = {8},
	number = {2},
	pages = {111--124},
	issn = {0276-2374, 1541-4493},
	doi = {10.2190/8Y6G-KTM8-VDX4-UHRW},
	urldate = {2025-04-15},
	copyright = {https://journals.sagepub.com/page/policies/text-and-data-mining-license},
	langid = {english}
}

@inproceedings{midi2audio2,
	title = {Music Generation: A Simplified Approach},
	shorttitle = {Music Generation},
	booktitle = {2024 3rd International Conference for Innovation in Technology (INOCON)},
	author = {Sambaragi, Laxmi M and Pradeep Naik, Pratiksha and Kakatkar, Nikhil N and Chikkamath, Satish and S R, Nirmala and Budihal, Suneeta V},
	year = {2024},
	month = mar,
	pages = {1--6},
	doi = {10.1109/INOCON60754.2024.10511477},
	urldate = {2025-04-16}
}

@article{transfer_learning,
	title={A survey of transfer learning},
	author={Weiss, Karl and Khoshgoftaar, Taghi M and Wang, DingDing},
	journal={Journal of Big data},
	volume={3},
	pages={1--40},
	year={2016},
	publisher={Springer},
	url          = {https://doi.org/10.1186/s40537-016-0043-6},
	doi          = {10.1186/S40537-016-0043-6},
}

@article{finetuning,
  author       = {Kenneth Ward Church and
Zeyu Chen and
Yanjun Ma},
title        = {Emerging trends: {A} gentle introduction to fine-tuning},
journal      = {Nat. Lang. Eng.},
volume       = {27},
number       = {6},
pages        = {763--778},
year         = {2021},
url          = {https://doi.org/10.1017/S1351324921000322},
doi          = {10.1017/S1351324921000322},
bibsource    = {dblp computer science bibliography, https://dblp.org}
}

@inproceedings{pretrained_features,
  author       = {Tal Reiss and
Niv Cohen and
Liron Bergman and
Yedid Hoshen},
title        = {{PANDA:} Adapting Pretrained Features for Anomaly Detection and Segmentation},
booktitle    = {{IEEE} Conference on Computer Vision and Pattern Recognition, {CVPR}
2021, virtual, June 19-25, 2021},
pages        = {2806--2814},
publisher    = {Computer Vision Foundation / {IEEE}},
year         = {2021},
url          = {https://openaccess.thecvf.com/content/CVPR2021/html/Reiss\_PANDA\_Adapting\_Pretrained\_Features\_for\_Anomaly\_Detection\_and\_Segmentation\_CVPR\_2021\_paper.html},
doi          = {10.1109/CVPR46437.2021.00283},
bibsource    = {dblp computer science bibliography, https://dblp.org}
}

@book{fundamentals_music_processing,
	title = {Fundamentals of Music Processing: Audio, Analysis, Algorithms, Applications},
	shorttitle = {Fundamentals of Music Processing},
	author = {M{\"u}ller, Meinard},
	year = {2015},
	publisher = {Springer International Publishing},
	address = {Cham},
	doi = {10.1007/978-3-319-21945-5},
	urldate = {2025-04-17},
	copyright = {https://www.springernature.com/gp/researchers/text-and-data-mining},
	isbn = {978-3-319-21944-8},
	langid = {english}
}

@proceedings{semeval,
  editor       = {Atul Kr. Ojha and
A. Seza Do{\u{g}}ru{\"{o}}z and
Harish Tayyar Madabushi and
Giovanni Da San Martino and
Sara Rosenthal and
Aiala Ros{\'{a}}},
title        = {Proceedings of the 18th International Workshop on Semantic Evaluation,
SemEval@NAACL 2024, Mexico City, Mexico, June 20-21, 2024},
publisher    = {Association for Computational Linguistics},
year         = {2024},
url          = {https://aclanthology.org/volumes/2024.semeval-1/},
isbn         = {979-8-89176-107-0},
bibsource    = {dblp computer science bibliography, https://dblp.org}
}

@article{semeval25,
	author       = {Shamsuddeen Hassan Muhammad and
	Nedjma Ousidhoum and
	Idris Abdulmumin and
	Seid Muhie Yimam and
	Jan Philip Wahle and
	Terry Ruas and
	Meriem Beloucif and
	Christine de Kock and
	Tadesse Destaw Belay and
	Ibrahim Said Ahmad and
	Nirmal Surange and
	Daniela Teodorescu and
	David Ifeoluwa Adelani and
	Alham Fikri Aji and
	Felermino Ali and
	Vladimir Araujo and
	Abinew Ali Ayele and
	Oana Ignat and
	Alexander Panchenko and
	Yi Zhou and
	Saif M. Mohammad},
	title        = {SemEval-2025 Task 11: Bridging the Gap in Text-Based Emotion Detection},
	journal      = {CoRR},
	volume       = {abs/2503.07269},
	year         = {2025},
	url          = {https://doi.org/10.48550/arXiv.2503.07269},
	doi          = {10.48550/ARXIV.2503.07269},
	eprinttype    = {arXiv},
	eprint       = {2503.07269},
	bibsource    = {dblp computer science bibliography, https://dblp.org}
}

@inproceedings{semeval19,
	author       = {Ankush Chatterjee and
	Kedhar Nath Narahari and
	Meghana Joshi and
	Puneet Agrawal},
	title        = {SemEval-2019 Task 3: EmoContext Contextual Emotion Detection in Text},
	booktitle    = {Proceedings of the 13th International Workshop on Semantic Evaluation,
	SemEval@NAACL-HLT 2019, Minneapolis, MN, USA, June 6-7, 2019},
	pages        = {39--48},
	publisher    = {Association for Computational Linguistics},
	year         = {2019},
	url          = {https://doi.org/10.18653/v1/s19-2005},
	doi          = {10.18653/V1/S19-2005},
	bibsource    = {dblp computer science bibliography, https://dblp.org}
}

@inproceedings{semeval18,
	author       = {Saif M. Mohammad and
	Felipe Bravo{-}Marquez and
	Mohammad Salameh and
	Svetlana Kiritchenko},
	title        = {SemEval-2018 Task 1: Affect in Tweets},
	booktitle    = {Proceedings of The 12th International Workshop on Semantic Evaluation,
	SemEval@NAACL-HLT 2018, New Orleans, Louisiana, USA, June 5-6, 2018},
	pages        = {1--17},
	publisher    = {Association for Computational Linguistics},
	year         = {2018},
	url          = {https://doi.org/10.18653/v1/s18-1001},
	doi          = {10.18653/V1/S18-1001},
	bibsource    = {dblp computer science bibliography, https://dblp.org}
}

@inproceedings{emobank,
	author       = {Sven Buechel and
	Udo Hahn},
	title        = {EmoBank: Studying the Impact of Annotation Perspective and Representation
	Format on Dimensional Emotion Analysis},
	booktitle    = {Proceedings of the 15th Conference of the European Chapter of the
	Association for Computational Linguistics, {EACL} 2017, Valencia,
	Spain, April 3-7, 2017, Volume 2: Short Papers},
	pages        = {578--585},
	publisher    = {Association for Computational Linguistics},
	year         = {2017},
	url          = {https://doi.org/10.18653/v1/e17-2092},
	doi          = {10.18653/V1/E17-2092},
	bibsource    = {dblp computer science bibliography, https://dblp.org}
}

@inproceedings{meld,
	author       = {Soujanya Poria and
	Devamanyu Hazarika and
	Navonil Majumder and
	Gautam Naik and
	Erik Cambria and
	Rada Mihalcea},
	title        = {{MELD:} {A} Multimodal Multi-Party Dataset for Emotion Recognition
	in Conversations},
	booktitle    = {Proceedings of the 57th Conference of the Association for Computational
	Linguistics, {ACL} 2019, Florence, Italy, July 28- August 2, 2019,
	Volume 1: Long Papers},
	pages        = {527--536},
	publisher    = {Association for Computational Linguistics},
	year         = {2019},
	url          = {https://doi.org/10.18653/v1/p19-1050},
	doi          = {10.18653/V1/P19-1050},
	bibsource    = {dblp computer science bibliography, https://dblp.org}
}

@article{instruction_tuning,
  author       = {Shengyu Zhang and
Linfeng Dong and
Xiaoya Li and
Sen Zhang and
Xiaofei Sun and
Shuhe Wang and
Jiwei Li and
Runyi Hu and
Tianwei Zhang and
Fei Wu and
Guoyin Wang},
title        = {Instruction Tuning for Large Language Models: {A} Survey},
journal      = {CoRR},
volume       = {abs/2308.10792},
year         = {2023},
url          = {https://doi.org/10.48550/arXiv.2308.10792},
doi          = {10.48550/ARXIV.2308.10792},
eprinttype    = {arXiv},
eprint       = {2308.10792},
bibsource    = {dblp computer science bibliography, https://dblp.org}
}

@inproceedings{adapters,
  author       = {Neil Houlsby and
Andrei Giurgiu and
Stanislaw Jastrzebski and
Bruna Morrone and
Quentin de Laroussilhe and
Andrea Gesmundo and
Mona Attariyan and
Sylvain Gelly},
title        = {Parameter-Efficient Transfer Learning for {NLP}},
booktitle    = {Proceedings of the 36th International Conference on Machine Learning,
{ICML} 2019, 9-15 June 2019, Long Beach, California, {USA}},
series       = {Proceedings of Machine Learning Research},
volume       = {97},
pages        = {2790--2799},
publisher    = {{PMLR}},
year         = {2019},
url          = {http://proceedings.mlr.press/v97/houlsby19a.html},
bibsource    = {dblp computer science bibliography, https://dblp.org}
}

@inproceedings{lora,
	title = {LoRA: Low-Rank Adaptation of Large Language Models},
	shorttitle = {LoRA},
	booktitle = {The Tenth International Conference on Learning Representations, ICLR 2022, Virtual Event, April 25-29, 2022},
	author = {Hu, Edward J. and Shen, Yelong and Wallis, Phillip and {Allen-Zhu}, Zeyuan and Li, Yuanzhi and Wang, Shean and Wang, Lu and Chen, Weizhu},
	year = {2022},
	publisher = {OpenReview.net},
	url = {https://openreview.net/forum?id=nZeVKeeFYf9},
	urldate = {2025-02-15}
}

@misc{uob,
	title = {UoB-NLP at SemEval-2025 Task 11: Leveraging Adapters for Multilingual and Cross-Lingual Emotion Detection},
	shorttitle = {UoB-NLP at SemEval-2025 Task 11},
	author = {Leon, Frances Laureano De and Wang, Yixiao and Feng, Yue and Lee, Mark G.},
	year = {2025},
	month = apr,
	number = {arXiv:2504.08543},
	eprint = {2504.08543},
	primaryclass = {cs},
	publisher = {arXiv},
	doi = {10.48550/arXiv.2504.08543},
	urldate = {2025-04-24},
	archiveprefix = {arXiv}
}

@misc{lotus,
	title = {Lotus at SemEval-2025 Task 11: RoBERTa with Llama-3 Generated Explanations for Multi-Label Emotion Classification},
	shorttitle = {Lotus at SemEval-2025 Task 11},
	author = {Ranjbar, Niloofar and Baghbani, Hamed},
	year = {2025},
	month = apr,
	number = {arXiv:2502.19935},
	eprint = {2502.19935},
	primaryclass = {cs},
	publisher = {arXiv},
	doi = {10.48550/arXiv.2502.19935},
	urldate = {2025-04-24},
	archiveprefix = {arXiv}
}

@misc{roberta,
	title = {RoBERTa: A Robustly Optimized BERT Pretraining Approach},
	shorttitle = {RoBERTa},
	author = {Liu, Yinhan and Ott, Myle and Goyal, Naman and Du, Jingfei and Joshi, Mandar and Chen, Danqi and Levy, Omer and Lewis, Mike and Zettlemoyer, Luke and Stoyanov, Veselin},
	year = {2019},
	month = jul,
	number = {arXiv:1907.11692},
	eprint = {1907.11692},
	primaryclass = {cs},
	publisher = {arXiv},
	doi = {10.48550/arXiv.1907.11692},
	urldate = {2024-08-31},
	archiveprefix = {arXiv}
}

@misc{mmtbert,
	title = {MMT-BERT: Chord-Aware Symbolic Music Generation Based on Multitrack Music Transformer and MusicBERT},
	shorttitle = {MMT-BERT},
	author = {Zhu, Jinlong and Sakurai, Keigo and Togo, Ren and Ogawa, Takahiro and Haseyama, Miki},
	year = {2024},
	month = sep,
	number = {arXiv:2409.00919},
	eprint = {2409.00919},
	primaryclass = {cs},
	publisher = {arXiv},
	doi = {10.48550/arXiv.2409.00919},
	urldate = {2025-04-24},
	archiveprefix = {arXiv}
}

@article{music_lyrics_emotion,
	title = {Affective Impact of Music Vs. Lyrics},
	author = {Stratton, Valerie N. and Zalanowski, Annette H.},
	year = {1994},
	month = jul,
	journal = {Empirical Studies of the Arts},
	volume = {12},
	number = {2},
	pages = {173--184},
	publisher = {SAGE Publications Inc},
	issn = {0276-2374},
	doi = {10.2190/35T0-U4DT-N09Q-LQHW},
	urldate = {2025-05-02},
	langid = {english}
}

@article{scenecut1,
	title={Perceiving scenes in film and in the world},
	author={Cutting, James E},
	journal={Moving image theory: Ecological considerations},
	pages={9--27},
	year={2005},
	url={https://www.researchgate.net/profile/James-Cutting/publication/237009484_Perceiving_scenes_in_film_and_in_the_world/links/0046351ae10f2baa8c000000/Perceiving-scenes-in-film-and-in-the-world.pdf},
}

@inproceedings{scenecut2,
  author       = {Naoki Nitanda and
Miki Haseyama and
Hideo Kitajima},
title        = {Audio signal segmentation and classification for scene-cut detection},
booktitle    = {International Symposium on Circuits and Systems {(ISCAS} 2005), 23-26
May 2005, Kobe, Japan},
pages        = {4030--4033},
publisher    = {{IEEE}},
year         = {2005},
url          = {https://doi.org/10.1109/ISCAS.2005.1465515},
doi          = {10.1109/ISCAS.2005.1465515},
bibsource    = {dblp computer science bibliography, https://dblp.org}
}

@article{sound_separation,
	title={A survey on: Sound source separation methods},
	author={Pimpale, Ms Monali R and Therese, Shanthi and Shinde, Vinayak},
	journal={International Journal},
	volume={3},
	number={11},
	pages={580--584},
	year={2016},
	url={https://www.academia.edu/download/50739135/V3I1103.pdf}
}

@article{disgust,
	author = {Jonathan Haidt and Paul Rozin and Clark Mccauley and Sumio Imada},
	title ={Body, Psyche, and Culture: The Relationship between Disgust and Morality},
	journal = {Psychology and Developing Societies},
	volume = {9},
	number = {1},
	pages = {107-131},
	year = {1997},
	doi = {10.1177/097133369700900105},
}

@article{emomv,
	title = {EmoMV: Affective Music-Video Correspondence Learning Datasets for Classification and Retrieval},
	shorttitle = {EmoMV},
	author = {Thao, Ha Thi Phuong and Roig, Gemma and Herremans, Dorien},
	year = {2023},
	month = mar,
	journal = {Information Fusion},
	volume = {91},
	pages = {64--79},
	issn = {1566-2535},
	doi = {10.1016/j.inffus.2022.10.002},
	urldate = {2025-05-21}
}

@inproceedings{harmoniccnn,
	title = {Data-Driven Harmonic Filters for Audio Representation Learning},
	booktitle = {ICASSP 2020 - 2020 IEEE International Conference on Acoustics, Speech and Signal Processing (ICASSP)},
	author = {Won, Minz and Chun, Sanghyuk and Nieto, Oriol and Serrc, Xavier},
	year = {2020},
	month = may,
	pages = {536--540},
	issn = {2379-190X},
	doi = {10.1109/ICASSP40776.2020.9053669},
	urldate = {2025-07-16}
}

@inproceedings{avalign,
	title = {Diverse and Aligned Audio-to-Video Generation via Text-to-Video Model Adaptation},
	booktitle = {Proceedings of the AAAI Conference on Artificial Intelligence},
	author = {Yariv, Guy and Gat, Itai and Benaim, Sagie and Wolf, Lior and Schwartz, Idan and Adi, Yossi},
	year = {2024},
	volume = {38},
	pages = {6639--6647},
	url = {https://ojs.aaai.org/index.php/AAAI/article/view/28486},
	urldate = {2025-06-20}
}

@inproceedings{onsets,
	title={Maximum filter vibrato suppression for onset detection},
	author={B{\"o}ck, Sebastian and Widmer, Gerhard},
	booktitle={16th International Conference on Digital Audio Effects (DAFx)},
	volume={7},
	pages={4},
	year={2013},
	organization={Citeseer}
}
\end{document}